%% file: main.tex
\documentclass[10pt,twocolumn,letterpaper]{article}

\usepackage[pagenumbers]{wacv} % To force page numbers, e.g. for an arXiv version

\usepackage{multirow}
\usepackage{titletoc}

\definecolor{wacvblue}{rgb}{0.21,0.49,0.74}
\usepackage[pagebackref,breaklinks,colorlinks,allcolors=wacvblue]{hyperref}

\def\wacvPaperID{1558} % *** Enter the WACV Paper ID here
\def\confName{WACV}
\def\confYear{2027}

\title{Improved Robustness from Biologically Inspired Sparse Contrast Representations}

\author{Lorena Stracke$^{1}$
\quad
Lia Nimmermann$^{1}$
\quad
Shashank Agnihotri$^{2}$
\quad
Bhaskar Choubey$^{3}$
\\
Margret Keuper$^{2,4}$
\quad
Volker Blanz$^{1}$\vspace{1mm}
\\
{\normalsize$^{1}$Media Systems, University of Siegen, Germany} \\
{\normalsize$^{2}$Data and Web Science Group, University of Mannheim, Germany} \\
{\normalsize$^{3}$Analogue Circuits and Image Sensors, University of Siegen, Germany}\\
{\normalsize$^{4}$Max-Planck-Institute for Informatics, Saarland Informatics Campus, Germany} \\
{\tt\small \{lorena.stracke,blanz,bhaskar.choubey\}@uni-siegen.de,}\\{\tt\small \{shashank.agnihotri,keuper\}@uni-mannheim.de}
}

\begin{document}

\maketitle

\begin{abstract}
Deep neural networks surpass humans on many vision benchmarks, yet remain far less robust to distribution shifts such as illumination and weather changes. Existing approaches address this challenge by additional training data, extensive augmentation, architectural modifications, or test-time adaptation. In this work, we explore a complementary direction: inspired by the human retina, we propose a fixed, model-agnostic preprocessing module that extracts signals that are more stable with respect to variations of illumination. Our method combines color remapping with local contrast extraction, producing sparse representations that emphasize structural features. We study its impact on semantic segmentation by training on Cityscapes and evaluating generalization under adverse conditions on Dark Zurich and ACDC. Our results show that the biologically inspired preprocessing preserves in-distribution performance while consistently improving robustness in challenging lighting scenarios, such as nighttime, where annotated training data are scarce. Moreover, the segmentation accuracy remains stable even when the contrast-based representation is sparsified by up to~70\%.
These gains suggest that rethinking the input representation itself can improve robustness while also opening opportunities for lower-latency, transmission-aware imaging sensors when sparsity can be exploited close to acquisition.
    %\keywords{Biologically Inspired Vision \and Early Visual Processing \and Robust Visual Perception \and Sparse Representations \and Semantic Segmentation}

\end{abstract}

\section{Introduction}
\label{sec:intro}

Modern vision systems degrade substantially under distribution shifts caused by adverse illumination and weather, despite strong in-distribution performance \cite{Geirhos.2018,Hendrycks.2019,shen2025assessing}.
This gap is particularly apparent for nighttime street scenes, where the input statistics change drastically due to low signal-to-noise ratios, altered contrast, glare, and illumination-dependent color responses~\cite{Hendrycks.2019,Geirhos.2018}.
Most existing robustness approaches address this by adding data (specialized adverse-condition datasets, heavy augmentation, or domain adaptation)~\cite{Hendrycks.2019}, modifying architectures~\cite{xie2019feature}, or performing test-time adaptation~\cite{wang2020tent}.
While effective, these directions increase system complexity and often remain sensitive when deployment conditions deviate from the assumed shift~\cite{Hendrycks.2019,Geirhos.2018}.

\input{tex_for_figures/teaser_new}

In this work, we pursue a complementary direction: reshape the \emph{input representation} before it reaches the network.
Our goal is not to replicate the human visual system, but to design a principled preprocessing front-end inspired by computational models of early vision and retinal processing.
Early developments in learning were closely tied to neuroscience ideas~\cite{McCulloch.1943}, and modern models still sometimes reveal parallels only post-hoc~\cite{agrawal2014convolutional}.
Here we revisit a classical early-vision principle: rather than preserving absolute brightness and color values, we emphasize signals that are more stable across illumination changes, namely opponent color channels and spatial differences in color and brightness. This is inspired by the mappings from photo-receptors (rods and short-, medium- and long-wavelength cones) to retinal ganglion cells, producing the signal that is transmitted in the optic nerve. Ganglion cells have receptive fields with concentric excitatory and inhibitory regions (center-surround cells), and color is represented in opponent red-green and yellow-blue channels. This mapping is considered to increase robustness to variations in brightness and color of illumination, and it leads to well-known  perceptual phenomena such as simultaneous contrast and color constancy, where perceived brightness and color depend strongly on context rather than absolute physical quantities~\cite{Palmer.1999}.
Motivated by this, we propose a fixed, model-agnostic preprocessing module that maps RGB images into contrast-centered channels, followed by a controlled sparsification step.
The module is lightweight, has no learnable parameters, and can be used as a plug-and-play front-end to standard segmentation architectures (\cref{fig:teaser_new}).

Our preprocessing instantiates two important mechanisms from early vision: center-surround processing and opponent color encoding~\cite{Marr-Hildreth.1980,Hurvich.1957,Thoreson.2019}.
We translate these principles into three computational stages (\cref{fig:Overview_pipeline}).
First, we apply a color remapping that expresses the signal in representations aligned with luminance processing and opponent channels (black--white, red--green, blue--yellow)~\cite{Hurvich.1957,Thoreson.2019}.
Second, we compute a spatial contrast field by approximating Difference of Gaussian (DoG) responses using a practical implementation: subtracting a blurred version of the signal from itself across scales, which converts the image into a signed contrast representation with near-zero values in locally homogeneous regions and concentrated magnitude around boundaries~\cite{Marr-Hildreth.1980}.
Third, we apply an explicit sparsification that sets low-magnitude contrast responses to zero, making the representation naturally compressible.

A contrast-domain representation is inherently sparse because large parts of natural images are locally smooth.
This sparsity becomes even more pronounced in low-light scenes where textures vanish into noise and only strong transitions remain reliable.
Empirically, our preprocessing produces near-zero values across most spatial locations, and models trained on dense preprocessed inputs remain accurate even when evaluated on strongly sparsified versions of the same representation (up to roughly 70\% sparsity in our zero-shot experiments; \cref{fig:Sparsity_results_zero-shot}).

This simple approach has improved robustness as a direct benefit: suppressing weak, illumination-dependent responses biases the model toward stable structural cues, which are exactly the cues that should transfer across lighting changes \cite{Palmer.1999,Geirhos.2018,Hendrycks.2019}.

We evaluate the proposed preprocessing for the task of semantic segmentation because it is a demanding downstream task that requires pixel-wise predictions and therefore stress-tests whether the transformed representation preserves fine spatial structure.
Street-scene benchmarks provide a controlled in-distribution training domain and established adverse-condition test sets that isolate illumination and weather shifts.
Specifically, we train on Cityscapes~\cite{Cityscapes_dataset} and evaluate robustness on Dark Zurich~\cite{Dark_zurich_dataset} and ACDC~\cite{ACDC_dataset}.
This setup directly probes the main question of this paper: can a fixed contrast-based representation preserve in-distribution performance while improving generalization under real nighttime and adverse conditions?

Our main contributions are:
\begin{itemize}
    \item A fixed, biologically inspired preprocessing front-end that combines opponent color or luminance remapping with a practical DoG-style contrast approximation, designed as a drop-in module for standard vision models~\cite{Marr-Hildreth.1980,Hurvich.1957,Thoreson.2019}.
    \item An empirical study of luminance, opponent, and single-color contrast representations for semantic segmentation under real distribution shifts (nighttime and adverse weather), showing improved robustness while maintaining competitive in-distribution accuracy~\cite{Cityscapes_dataset,Dark_zurich_dataset,ACDC_dataset}.
    \item Evidence that contrast-domain preprocessing yields highly sparse inputs that can be aggressively zeroed (up to \(\sim\)70\% in our zero-shot setting) without degrading segmentation performance, enabling compression and efficiency opportunities.
\end{itemize}

\section{Related Work}
\label{sec:related work}
\noindent\textbf{Biological Vision As A Model For Robustness. }

Biologically inspired neural computation has a long tradition in machine learning~\cite{McCulloch.1943, Floreano.2023, Kriegeskorte.2015}. Here we give a short overview of existing research that successfully applies concepts from the early visual system. 
Kim et al.~\cite{Kim.2016} introduced ON and OFF ReLU units to mimic the behavior of the bipolar cells that invert the signals from the sensor cells on the retina. 
Babaiee et al.~\cite{Babaiee.2021} incorporate center-surround filters, using the resulting data as additional inputs at deeper layers, improving robustness on NORB~\cite{NORB.2004}. 
Hasani et al.~\cite{Hasani.2019} applied DoG filters directly to the first activation maps in CNNs, simulating retinal filtering as part of the model architecture. 
Tsitiridis et al.~\cite{Tsitiridis.2019} combined the center-surround design with color-opponency, effectively modelling double-opponent cells and reported positive effects in the case of face biometrics.
It has also been shown that neural activations emerging in trained CNNs resemble center-surround receptive fields ~\cite{Pan.2023, olshausen1996emergence}. 

Some research has studied the role of opponent color cells, center-surround cells and double-opponent cells found in the human visual system (HVS), showing that these are important for lighting estimation and color constancy~\cite{gao2013colorconstancy,gao2024primary}, for detecting boundaries while suppressing texture~\cite{yang2015boundary}, and for object detection and instance segmentation~\cite{xue2021colorconstancy}. In contrast to these realistic simulations of the HVS, we focus on benefits of a HVS-motivated preprocessing on sensor design and downstream tasks.

Building on this, our work integrates DoG-style preprocessing, combined with color remapping into the input pipeline for segmentation models, without changing the network or training process, targeting robustness to nighttime and weather-based lighting shifts.

\input{tex_for_figures/overview_pipeline}

\noindent\textbf{Grayscale Input In Neural Networks. }
Grayscale input has been explored in various contexts~\cite{sommerhoff2024task}. Bui et al.~\cite{Bui.2016} showed that grayscale images can outperform RGB in object recognition across different types of classifiers. For facial expression and identity recognition, grayscale performs comparably to RGB~\cite{Bhatta.2025, Yudin.2020}, although shallow models rely more on color~\cite{Bui.2016}. RGB remains superior in color-sensitive tasks such as plant disease classification~\cite{Phong.2024}. 

Given the dominance of luminance-based perception in the human visual system (HVS) under low-light conditions~\cite{Palmer.1999}, luminance variants serve as a promising robustness baseline in our study.

\noindent\textbf{Robustness To Lighting Variations. }
Several works have highlighted the vulnerability of neural networks to lighting changes~\cite{hoffmann2021towards}. Sivaraman et al.~\cite{Sivaraman.2018} demonstrated performance drops under illumination shifts for ImageNet-trained CNNs. Hu et al.~\cite{Hu.2023} showed that brightness and shadow variations significantly degrade model performance. The large-scale benchmark by Hendrycks et al.~\cite{Hendrycks.2019} further confirmed that models trained on clean data are sensitive to a wide range of corruptions. 

Our work complements these prior works by targeting robustness through preprocessing mechanisms that mimic biologically grounded visual filters without explicit training on difficult data. Our focus lies in building an implicit bias toward robustness against adverse conditions.

\noindent\textbf{Sparsity in Deep Learning. }
Sparsity has long been recognized as a desirable property in representation learning, both from biological and computational perspectives. Early work on sparse coding demonstrated that enforcing sparse activations yields efficient and interpretable representations closely aligned with receptive fields observed in the visual cortex~\cite{olshausen1996emergence}. 
In deep learning, sparsity has been explored through activation regularization, sparse coding layers, structured pruning, and dynamically sparse networks. Sparse representations have been shown to improve robustness and generalization by reducing reliance on redundant or noisy features and by encouraging disentangled feature representations~\cite{glorot2011deep, frankle2018lottery, evci2020rigging}. 
Recent studies further suggest that sparsity can enhance robustness to distribution shifts and corruptions by focusing computation on salient structures rather than dense pixel-level detail~\cite{geirhos2020shortcut,achille2018information,olshausen1996emergence,simoncelli2001natural}.

While most prior work introduces sparsity through architectural constraints~\cite{glorot2011deep,frankle2018lottery}, training objectives~\cite{achille2018information}, or post-hoc pruning~\cite{han2015deep,frankle2018lottery,evci2020rigging}, %our approach introduces sparsity directly at the input level of the machine learning pipeline as a side effect of our biologically inspired preprocessing. 
our approach incorporates sparsity as the final step of the biologically inspired preprocessing, selectively extracting relevant information while discarding redundancies. 

\section{Biologically Inspired Image Preprocessing}
\label{sec:preprocessing}

Our goal is to improve robustness under adverse illumination and weather by reshaping the \emph{input representation} rather than altering architectures or training procedures.
Motivated by computational models of early vision, we design a fixed preprocessing front-end that emphasizes local contrast and opponent color structure while discarding illumination-dependent absolute signals~\cite{Palmer.1999,Hurvich.1957,Thoreson.2019}.
As summarized in \cref{fig:Overview_pipeline}, the preprocessing consists of three stages: (1) a linear \emph{color remapping}, (2) a \emph{center-surround} contrast operator approximating a Difference-of-Gaussians (DoG), and (3) an explicit \emph{sparsification} that removes small-magnitude responses.
The output has the same spatial resolution as the input. Note that only the sparsification adds non-linearity to the processing.

\subsection{Notation and Design Choices}
Let $\mathbf{x} \in \mathbb{R}^{H \times W \times 3}$ denote an RGB image with channels $(R,G,B)$.
We apply the preprocessing independently at each spatial location and per channel group, producing either one channel ($C=1$) or three channels ($C=3$), depending on the remapping.
When the remapping yields a single channel, we optionally replicate it to three channels for compatibility with pretrained backbones and unchanged model definitions.

\subsection{Color Remapping}
\label{sec:implementation}

The first stage transforms RGB into a representation that either (i) discards chromatic information and retains only luminance, motivated by low-light (scotopic) vision where rod signals dominate~\cite{Palmer.1999,Thoreson.2019}, or (ii) expresses the signal in opponent channels, motivated by opponent processing in the retina~\cite{Hurvich.1957,Palmer.1999,Thoreson.2019}.
As a control, we include an identity remapping.
All three options correspond to linear combinations of the RGB channels as indicated in \cref{fig:Overview_pipeline} (1).

Formally, we define a remapping matrix $\mathbf{M} \in \mathbb{R}^{C \times 3}$ and compute
\begin{equation}
\mathbf{y}(u,v) = \mathbf{M}\,\mathbf{x}(u,v), 
\qquad \mathbf{x}(u,v) \in \mathbb{R}^{3},\ \mathbf{y}(u,v) \in \mathbb{R}^{C},
\label{eq:color_remap_general}
\end{equation}
for every pixel $(u,v)$.
The remapped image is $\mathbf{y} \in \mathbb{R}^{H \times W \times C}$.

\noindent\textbf{Luminance.}
The luminance variant produces a single channel ($C=1$) and is defined by
\begin{equation}
\mathbf{M}_{\text{lum}} =
\begin{bmatrix}
\frac{1}{3} & \frac{1}{3} & \frac{1}{3}
\end{bmatrix}
\in \mathbb{R}^{1 \times 3}.
\label{eq:matrix_luminance_simple}
\end{equation}
To preserve architectural compatibility with standard 3-channel pretrained backbones, we optionally replicate the luminance channel to obtain a 3-channel tensor:
\begin{equation}
\mathbf{y}_{3\text{ch}} = \mathrm{replicate}_3(\mathbf{y}) \in \mathbb{R}^{H \times W \times 3}.
\label{eq:luminance_replicate}
\end{equation}
In addition, we evaluate a commonly used grayscale conversion that emphasizes green, motivated by human brightness sensitivity and widely used in RGB-to-grayscale conversion~\cite{Gunes.2016,Palmer.1999}:
\begin{equation}
\mathbf{M}_{\text{lum-g}} =
\begin{bmatrix}
0.299 & 0.587 & 0.114
\end{bmatrix}
\in \mathbb{R}^{1 \times 3}.
\label{eq:matrix_luminance_green_bias_1x3}
\end{equation}

\noindent\textbf{Color opponency.}
To model opponent coding under well-lit conditions, we use a 3-channel remapping ($C=3$) that yields black--white (luminance-like), red--green, and blue--yellow channels~\cite{Hurvich.1957,Palmer.1999,Thoreson.2019}:
\begin{equation}
\mathbf{M}_{\text{co}} =
\begin{bmatrix}
\phantom{-}\frac{1}{3} & \phantom{-}\frac{1}{3} & \phantom{-}\frac{1}{3}  \vspace{1mm}\\
\phantom{-}\frac{1}{2} & -\frac{1}{2} & \phantom{-}0 \vspace{1mm}\\
-\frac{1}{6} & -\frac{1}{6} & \phantom{-}\frac{1}{3}
\end{bmatrix}.
\label{eq:matrix_color_opponency}
\end{equation}
This mapping preserves three channels (and is invertible up to scale), so no replication is needed.

\noindent\textbf{Single-color (identity).}
As a control, we use the identity mapping $\mathbf{M}_{\text{id}} = \mathbf{I}_3$, keeping channels as $(R,G,B)$.
This isolates the effect of subsequent stages from the choice of color space.

\subsection{Spatial Contrast Extraction via a DoG-Style Operator}
\label{sec:contrast_extraction}

The second stage computes a signed local-contrast representation inspired by center-surround receptive fields of retinal ganglion cells, which encode contrast while discarding absolute luminance~\cite{Palmer.1999}.
We implement a practical approximation of DoG, a classical model closely related to the Marr-Hildreth edge operator~\cite{Marr-Hildreth.1980}.
Intuitively (see \cref{fig:Overview_pipeline} (2)), we compare a signal at a fine scale to coarser, blurred versions of itself and retain the residual.

Let $\mathbf{y}$ be the remapped image from Sec.~\ref{sec:implementation}.
We construct progressively blurred versions using repeated application of a $3 \times 3$ box filter, which approximates Gaussian smoothing when applied multiple times~\cite{Getreuer.2013}.
Let $\mathcal{B}(\cdot)$ denote one application of the box blur and define
\begin{equation}
\mathbf{y}^{(0)} = \mathbf{y}, \qquad
\mathbf{y}^{(i)} = \mathcal{B}\!\left(\mathbf{y}^{(i-1)}\right), \ i=1,\dots,d,
\label{eq:progressive_blur}
\end{equation}
where $d$ is the blur depth (number of blur steps).

We then compute the contrast image as the difference between the original and the average of the blurred signals:
\begin{equation}
\mathbf{z} \;=\; \mathbf{y}^{(0)} \;-\; \frac{1}{d}\sum_{i=1}^{d}\mathbf{y}^{(i)},
\qquad d \ge 1.
\label{eq:contrast_definition}
\end{equation}
This yields $\mathbf{z} \in \mathbb{R}^{H \times W \times C}$ with both positive and negative values: sign encodes contrast direction, while magnitude encodes contrast strength.
For $d=0$, we define $\mathbf{z}=\mathbf{y}$, i.e. no contrast extraction, which we use as a control (and together with $\mathbf{M}_{\text{id}}$ reproduces the baseline in \cref{fig:Overview_pipeline}).

\subsection{Sparsification by Magnitude Thresholding}
\label{sec:sparsity}

The contrast image $\mathbf{z}$ produced by \cref{eq:contrast_definition} is typically close to zero across locally homogeneous regions and concentrated around boundaries, which makes it naturally sparse.
We explicitly enforce sparsity by removing weak contrast responses, guided by the magnitude $|\mathbf{z}|$.
This aligns with the intuition that low-amplitude responses contribute little structural information and can be discarded without substantially affecting downstream perception~\cite{WANDELL2025360}. 
We define a sparsification operator $\mathcal{S}_{p}$ that sets a fraction $p\%$ of the smallest-magnitude values to zero.
%Let $\tau_p$ be the $p$-th percentile of $|\mathbf{z}|$ over all spatial positions and channels in an image.
Then
\begin{equation}
\left(\mathcal{S}_{p}(\mathbf{z})\right)_{u,v,c}
=
\begin{cases}
0, & \text{if } |\mathbf{z}_{u,v,c}| < \tau_p,\\
\mathbf{z}_{u,v,c}, & \text{otherwise,}
\end{cases}
\label{eq:sparsify_percentile}
\end{equation}
where $\tau_p$ is the $p$-th percentile of $|\mathbf{z}|$ computed over all spatial positions and channels within an image.
The final preprocessed input to the network is $\tilde{\mathbf{x}} = \mathcal{S}_{p}(\mathbf{z})$.

\input{tex_for_figures/sparsity_50_combined_image}

\Cref{fig:Combined_image_for_sparsity_50} provides an interpretation of this operation at $p=70\%$.
Panel (a) shows the signed contrast tensor (values near zero appear gray), panel (b) highlights in pink the coefficients that are set to zero, and panels (c,d) map the same mask back onto the original RGB image.
This visualization indicates that sparsification primarily removes coefficients in color-consistent or locally smooth regions, while retaining strong structural transitions such as object boundaries and high-contrast edges, which are crucial for semantic segmentation.

\section{Experiments}
\label{sec:Experiments}
In the following, we discuss the experimental setup, semantic segmentation results, and ablation studies. 
Please refer to the appendix for additional results (including image classification) and other details.

\subsection{Experimental Setup}
\label{subsec:experimental_setup}
For evaluation, we consider semantic segmentation as downstream task because it requires correct pixel-wise predictions and therefore directly tests whether the proposed preprocessing preserves fine spatial structure while improving robustness on Out-of-Distribution (OOD) data. While semantic segmentation serves as the primary downstream task, results for image classification on CIFAR-100 are reported in the appendix.
Models are trained on Cityscapes~\cite{Cityscapes_dataset} and evaluated on Cityscapes (independent and identically distributed \ie i.i.d.\ data), and OOD data: Dark Zurich~\cite{Dark_zurich_dataset}, and ACDC~\cite{ACDC_dataset}.
We report mean IoU (mIoU), mean class accuracy (mAcc), and average pixel accuracy (aAcc), averaged over three random seeds.
Unless stated otherwise, preprocessing uses three channels and contrast depth $d{=}5$.

We consider three preprocessing variants (luminance, color-opponency, single-color) with contrast extraction as defined in \cref{sec:preprocessing}. Normalization is performed prior to all preprocessing variants as well as in the baseline.
We evaluate five representative segmentation architectures, DeepLabV3+~\cite{DeepLabv3+}, Mask2Former~\cite{Mask2Former} with a ResNet-50~\cite{he2016deep} backbone, UPerNet~\cite{UPerNet}, SegFormer~\cite{SegFormer}, and Mask2Former with an InternImage-Huge backbone~\cite{wang2023internimage} (denoted just as InternImage) to assess robustness across model families.
Further details (backbones, schedules, hyperparameters) are provided in the appendix.

\subsection{Results}
\label{sec:results}

We evaluate robustness under distribution shift and the effect of sparsity using models trained on Cityscapes with dense inputs (0\% sparsity).
The baseline corresponds to standard RGB training and evaluation, and serves to demonstrate that standard RGB images are not amenable to sparsification, whereas our biologically inspired preprocessing enables effective sparsification. For our method, we train the same architectures with the preprocessing from \cref{sec:preprocessing} (color remapping and contrast extraction) but without any training-time sparsity. Summarized results for Dark Zurich are visualized in \cref{fig:improvement_plot}.

\input{tex_for_figures/improvement_plot}
\input{tex_for_figures/sparsity_zero_shot_results}

The dense evaluation point (0\% sparsity in \cref{fig:Sparsity_results_zero-shot}) shows that preprocessing preserves in-distribution performance: on Cityscapes, all variants remain on par with the baseline, indicating that the transformation does not remove task-relevant information in standard conditions.
Under nighttime distribution shift (Dark Zurich and ACDC Night), the baseline exhibits a pronounced performance drop, whereas all preprocessing variants improve OOD performance, supporting the main premise from \cref{sec:intro} that contrast-centered representations improve generalization without modifying downstream architectures.

We next assess whether the same dense-trained models can be evaluated on \emph{sparsified} inputs at inference time, without retraining.
For our preprocessing variants, sparsity is applied \emph{after} preprocessing, that is, in the signed contrast domain defined in \cref{sec:preprocessing}, where low-magnitude coefficients largely correspond to locally homogeneous regions.
As a sanity check, we apply the \emph{same} sparsity operator to the baseline directly in RGB space (zeroing the values closest to zero), which tests whether sparsity alone is meaningful without a contrast-domain representation.
\Cref{fig:Sparsity_results_zero-shot} shows a clear separation: baseline performance degrades steadily as sparsity increases, confirming that naively dropping RGB values removes information the network relies on.
In contrast, models trained with preprocessing tolerate substantial sparsity at test time: performance remains stable up to high sparsity levels (approximately 60\%-70\% in our experiments, depending on the architecture) on both i.i.d.\ and OOD benchmarks, while retaining their robustness advantage.
We also note that InternImage attains the highest overall performance (i.i.d.\ and OOD) and exhibits particularly strong stability under sparsity.
\iffalse
TODO:CHANGE THIS EXPLANATION FOR INTERNIMAGE
We hypothesize Mask2Former remains most stable under sparsity because its pixel decoder provides multi-scale high-resolution embeddings and its decoder's masked cross-attention restricts feature aggregation to predicted mask regions, reducing reliance on dense global evidence~\cite{Mask2Former}.
We discuss the implications of this in \cref{sec:discussion}.
\fi
Together, these results demonstrate that the proposed preprocessing not only improves robustness under real nighttime distribution shift, but also yields a representation that can be aggressively sparsified at inference time with minimal performance loss.

\subsection{Ablation Studies}
\label{sec:ablations}

\subsubsection{Training With Sparsity}
\label{sec:sparsity_trained}
The zero-shot results above show that dense-trained models with biologically inspired preprocessing can tolerate substantial sparsity at inference time.
We now test whether explicitly training with sparsified inputs further stabilizes performance and whether the conclusions remain consistent.
\input{tex_for_figures/comparison_of_learned_and_zuro_shot_results_deeplab_mask2former}

We train DeepLabV3+, Mask2Former, and UPerNet architectures using sparsity levels between 40\% and 90\% and evaluate at the same sparsity levels.
\Cref{fig:Sparse results trained} confirms the same qualitative behavior as the zero-shot study.
Preprocessing models remain stable under moderate sparsity, with reliable performance up to around 60\% and gradual degradation beyond that.
Compared to the baseline under sparsity, preprocessing consistently yields stronger robustness, showing that the benefit is not an artifact of zero-shot evaluation.
The remaining fluctuation between up to 60\% and 70\% stable sparsity (trained vs. zero-shot) suggests that the precise tolerance depends on architecture and training dynamics, but the overall observation remains unchanged: sparsity becomes usable only after the representation is transformed into contrast space.

\subsubsection{Beyond Nighttime}
\label{sec:beyond_nighttime}
We motivate nighttime robustness, but the preprocessing is not inherently tied to low luminance only.
We therefore evaluate generalization to all other available adverse weather settings of ACDC (Fog, Rain, Snow, Mean).
\input{tex_for_figures/ablation_going_beyond_nighttime_heatmap}

\Cref{tab:Validation results - Going beyond nighttime} shows that the preprocessing improves robustness beyond nighttime in several settings.
For fog and snow, contrast-centered representations help across architectures, suggesting that suppressing low-information regions and emphasizing structural transitions transfers to weather-induced shifts as well.
For rain, results are largely at-par: the RGB baseline is only marginally higher than the best preprocessing variant (on the order of $\sim$1\% mIoU), and this gap is within the variability across random seeds.
This suggests that chromatic cues can be slightly beneficial in this regime, while the proposed preprocessing remains competitive rather than failing under rain.
Overall, these results indicate that the preprocessing provides a general robustness bias rather than a purely nighttime-specific trick, while also highlighting that the usefulness of color information depends on the type of shift.

\subsubsection{Integration Into AllWeatherNet}
The architecture-agnostic plug-and-play design of our preprocessing is a major technical benefit: the module can be placed in front of \emph{any} existing neural network that processes visual data, decoupling signal conditioning from network inference. To further demonstrate this flexibility, we combine our preprocessing with the AllWeatherNet~\cite{qian2024allweather} robustness method using Mask2Former. The results are visualized in \cref{fig:allweathernet_results_main_paper}. Our approach consistently outperforms the AllWeatherNet baseline and, in several settings, even further increases performance by combining both approaches.
\input{tex_for_figures/allweathernet_results_main_paper}

\iffalse
\subsubsection{Depth of Contrast Extraction. }
\label{sec:depth_ablation}
Finally, we ablate the contrast depth $d$, which controls the extent of center-surround aggregation in our DoG-style operator from \cref{sec:contrast_extraction}.
To enable a broader sweep under resource constraints, we use DeepLabV3+ with a smaller backbone and evaluate $d \in \{0,\dots,10\}$.

\input{tex_for_figures/ablation_depth_parameter_search}

\Cref{fig:Validation results - Depth Parameter Search} shows that depth has a substantial effect on both i.i.d. accuracy and robustness, with the optimal value depending mildly on the remapping variant and test domain.
Very small depths can under-emphasize contrast, while very large depths can over-smooth and remove details.
Balancing robustness under shift and performance on Cityscapes, $d{=}5$ offers a strong trade-off across variants, and we therefore use it as the default throughout the main experiments.
\fi

\section{Discussion}
\label{sec:discussion}

Our findings suggest that fixed transformations inspired by classical early-vision principles can serve as an effective front-end for modern segmentation models.
By emphasizing opponent color and luminance structure and converting inputs into a signed local-contrast representation, the preprocessing removes illumination-dependent components while preserving the spatial cues that segmentation relies on.
This is reflected in two consistent empirical trends: (i) i.i.d.\ performance on Cityscapes remains essentially unchanged, and (ii) robustness improves under real nighttime distribution shift.
Beyond robustness, the contrast-domain representation concentrates information into a small fraction of coefficients, which explains why sparsification is effective after preprocessing but harmful when applied directly in RGB space.

\subsection{Architectural Implications of Sparse Contrast Representations}
A notable observation is that Mask2Former (with both backbones) is very stable even under strong sparsity at inference time.
As hypothesized in \cref{sec:results}, this may be explained by the combination of its multi-scale pixel decoder and the decoder's masked cross-attention, which restricts feature aggregation to predicted mask regions and thus reduces reliance on dense global evidence~\cite{Mask2Former}.
Additionally,  the dynamic space convolution in the InternImage encoder with Mask2Former decoder might be helping it to adapt to the sparse representations much better.
This points to a broader implication: future architectures could be designed to exploit sparsity more directly, for example, by incorporating sparsity-aware mechanisms already in the encoder (e.g., sparse attention, masked convolutions, or routing conditioned on non-zero contrast responses), rather than benefiting from sparsity only at later decoding stages.
Such designs could further improve both robustness and efficiency when operating on contrast-centered, sparsified inputs.

\subsection{Implementation on Sensor: Transmission-Aware Preprocessing}
\label{sec:sensor_discussion}
The sparsity induced by our preprocessing has an immediate systems-level implication if the front-end is implemented 
directly on a CMOS image sensor chip. It would reduce the power consumption of data converters (ADCs) and data transmission, 
 which is particularly relevant for always-on sensing and long-range links where communication energy scales with transmitted data volume. 

Sparsification can be implemented efficiently by increasing the minimum threshold level of the data converter, which improves conversion efficiency significantly. As we discuss in the appendix, a 
limited range of analog values must be digitized, 
with each bit reduction approximately halving the required power. Moreover, sparsification reduces the number of pixel values to be transmitted. We calculated a potential saving of roughly 33\% of the interface power budget~(see appendix).

Taken together, the learning and systems perspectives align: the same transformation that improves robustness also produces a representation that is cheaper to transmit.
A promising next step is to formalize this trade-off end-to-end, including adaptive sparsification strategies that meet a target bandwidth or power budget while preserving segmentation accuracy under domain shift.

\section{Conclusion}
\label{sec:conclusion}

We study a route to robustness under adverse illumination and weather complementary to existing approaches: reshaping the \emph{input representation} rather than modifying architectures or training procedures.
Inspired by classical early-vision principles, we introduce a fixed preprocessing front-end that (i) remaps color into luminance or opponent channels and (ii) computes a signed local-contrast representation via a DoG-style operator, optionally followed by (iii) magnitude-based sparsification.
Across multiple segmentation architectures trained on Cityscapes, this front-end preserves i.i.d.\ accuracy while consistently improving robustness on real-world nighttime benchmarks (Dark Zurich and ACDC Night).
Crucially, the contrast-domain representation is not only more robust, but it also provides better options for data compression: models trained on dense data remain stable even if the preprocessed signal is sparsified aggressively at inference time, whereas applying the same sparsity operator directly in RGB space leads to substantial degradation.
This difference indicates that our contrast-based representation captures task-relevant information in a more effective way by its tendency to assign values close to zero to less relevant pixels. 

Beyond the machine learning perspective, these results suggest a practical systems implication.
A fixed, lightweight preprocessing module that produces sparse contrast signals can reduce the amount of information that must be stored or transmitted, while simultaneously improving robustness under domain shift.
Together, our findings identify contrast-centered, sparsifiable representations as a simple and effective interface between sensing and recognition for more reliable and more efficient vision systems.

\noindent\textbf{Future Work. }
While the proposed front-end substantially reduces the robustness gap, a gap between i.i.d.\ and OOD performance remains; combining preprocessing with stronger training strategies (for example, data augmentation) might narrow it further.
This work focused on exploring the proposed methods and their impact, free of other confounders such as test-time adaptation and strong data augmentation.
Motivated by the strong sparsity stability of the architectures, a second direction is to develop sparsity-aware architectures that operate directly on sparse contrast signals, for example, by dropping near-zero tokens and using sparse attention or utilizing dynamic sparse kernels as in InternImage.
Finally, near-sensor implementation enables end-to-end sensing-to-inference co-design, where sparsity can translate to reduced transmitted payload and energy.

\noindent\textbf{Limitations. }
\iffalse
A main limitation is the availability of pixel-wise OOD benchmarks that allow training on a clean i.i.d.\ dataset while stress-testing generalization under real nighttime conditions.
This restricts the breadth of evaluation and motivates expanding to additional settings as such datasets become available.
\fi
A key limitation is the availability of pixel-wise benchmarks that combine training on clean i.i.d.\ data with testing for generalization under OOD conditions, particularly real nighttime scenarios. This limits the scope of the evaluation and motivates future studies on additional settings as suitable datasets become available.

{
    \small
    \bibliographystyle{ieeenat_fullname}
    \bibliography{main}
}

\input{appendix}

\end{document}

%% file: tex_for_figures/teaser_new.tex
\begin{figure}[tb]
  \centering
  \includegraphics[width=\linewidth]{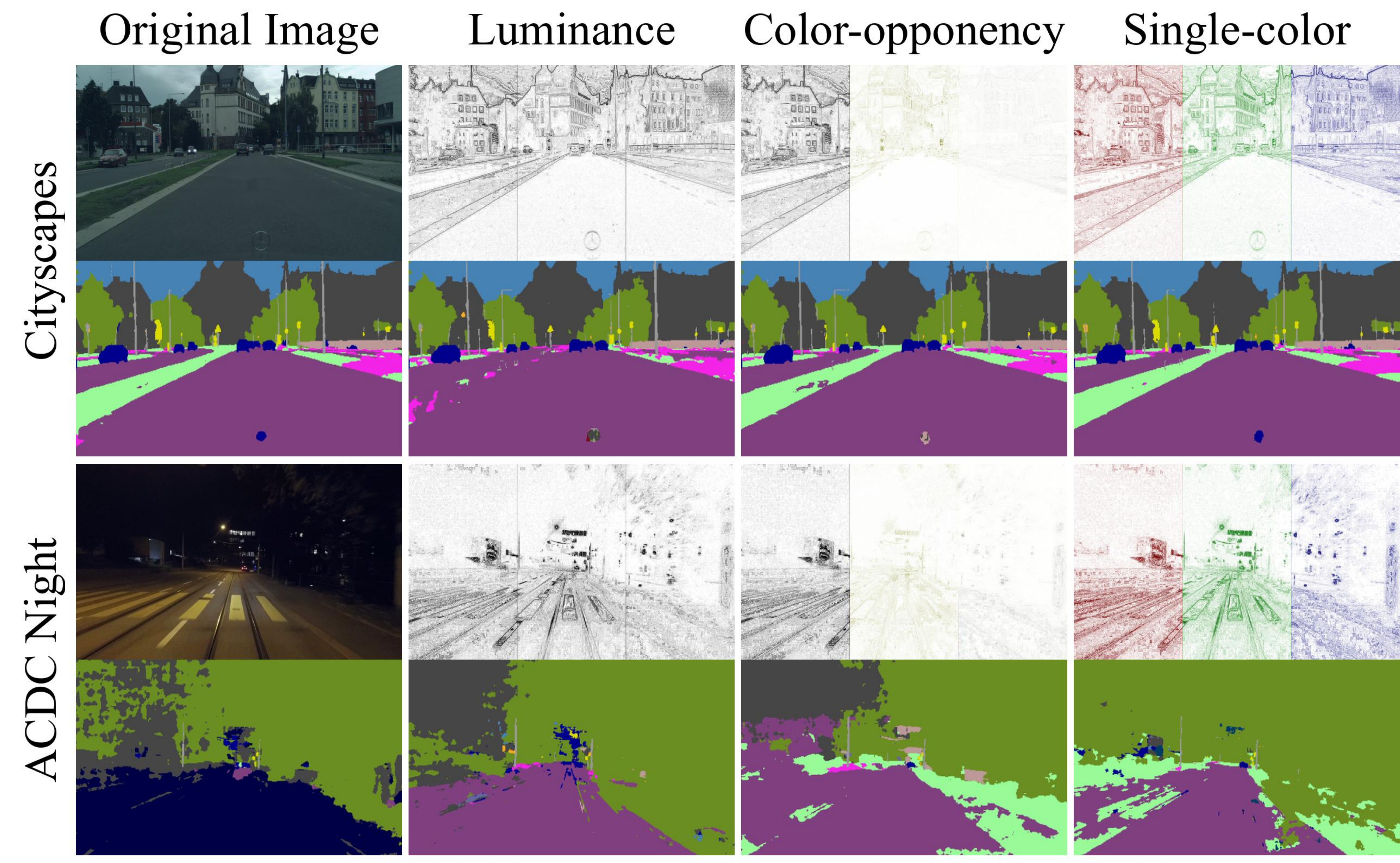}
  \iffalse
  \caption{Example images from the in-distribution dataset (Cityscapes~\cite{Cityscapes_dataset}), real-world robustness dataset ACDC Night~\cite{ACDC_dataset}), and preprocessed versions of the explored variants, showing a third of each image in the representation of one resulting channel. Below we plot the segmentation masks predicted by UPerNet trained with the respective preprocessing.}
  \fi
  %\caption{\textbf{Robustness can begin before the neural network.} We show examples from the in-distribution Cityscapes dataset~\cite{Cityscapes_dataset} and the real-world nighttime robustness benchmark ACDC Night~\cite{ACDC_dataset}, together with representative channels from our luminance, color-opponency, and single-color contrast preprocessing variants. The corresponding UPerNet predictions illustrate the central idea of this work: fixed, biologically inspired input transformations can preserve the structural cues needed for segmentation while reducing sensitivity to adverse illumination, suggesting a simple path toward more robust and potentially sensor-efficient vision systems.}
  %
  \caption{\textbf{Robustness can begin before the neural network.} In-distribution (above, Cityscapes~\cite{Cityscapes_dataset}) versus OOD low-light (below, ACDC Night~\cite{ACDC_dataset}) results for UPerNet. The leftmost column shows the baseline, which transfers poorly to ACDC Night, while our biologically inspired preprocessing in all considered variants (luminance, color-opponency, and single-color) significantly improves results across all labels by preserving robust structural cues.}
  
 %\includegraphics[width=\linewidth]{figures/teaser_combined.pdf}
 % \caption{Cropped example images from the in-distribution dataset (Cityscapes~\cite{Cityscapes_dataset}) and real-world robustness datasets (Dark Zurich~\cite{Dark_zurich_dataset} and ACDC~\cite{ACDC_dataset}). 
  %  Each row shows the original RGB image followed by the preprocessed variants explored in this work: luminance, color-opponency, and single-color. Single channels are also visualized using the color maps shown below including negative values, where white represents 0. Please refer to the appendix for more visualizations and segmentation masks.}
    \label{fig:teaser_new}
\end{figure}

%% file: tex_for_figures/overview_pipeline.tex
\begin{figure*}[tb]
  \centering
  \includegraphics[width=\linewidth]{figures/Simpler-pipeline-full.pdf} 
  \caption{Overview of the four evaluated pipelines. In comparison with a \textbf{baseline} (top row), we consider our preprocessing in three variants: \textbf{luminance}, \textbf{color-opponency} and \textbf{single-color}. Our preprocessing begins with a \textbf{color remapping (1)} from RGB values to three different color spaces. Then follows the \textbf{contrast extraction (2)}, where we simulate a convolution with a DoG-like kernel. Finally, we enforce \textbf{sparsity (3)} by setting low absolute values to zero, highlighted in pink. Each variant produces 3 channels that serve as a new input to different neural network architecures (right).}
  \label{fig:Overview_pipeline}
  %\vspace{-0.5em}
\end{figure*}

%% file: tex_for_figures/sparsity_50_combined_image.tex
\begin{figure}[tb]
    \centering
    \begin{subfigure}[t]{0.49\linewidth}
        \centering
        \includegraphics[width=\linewidth]{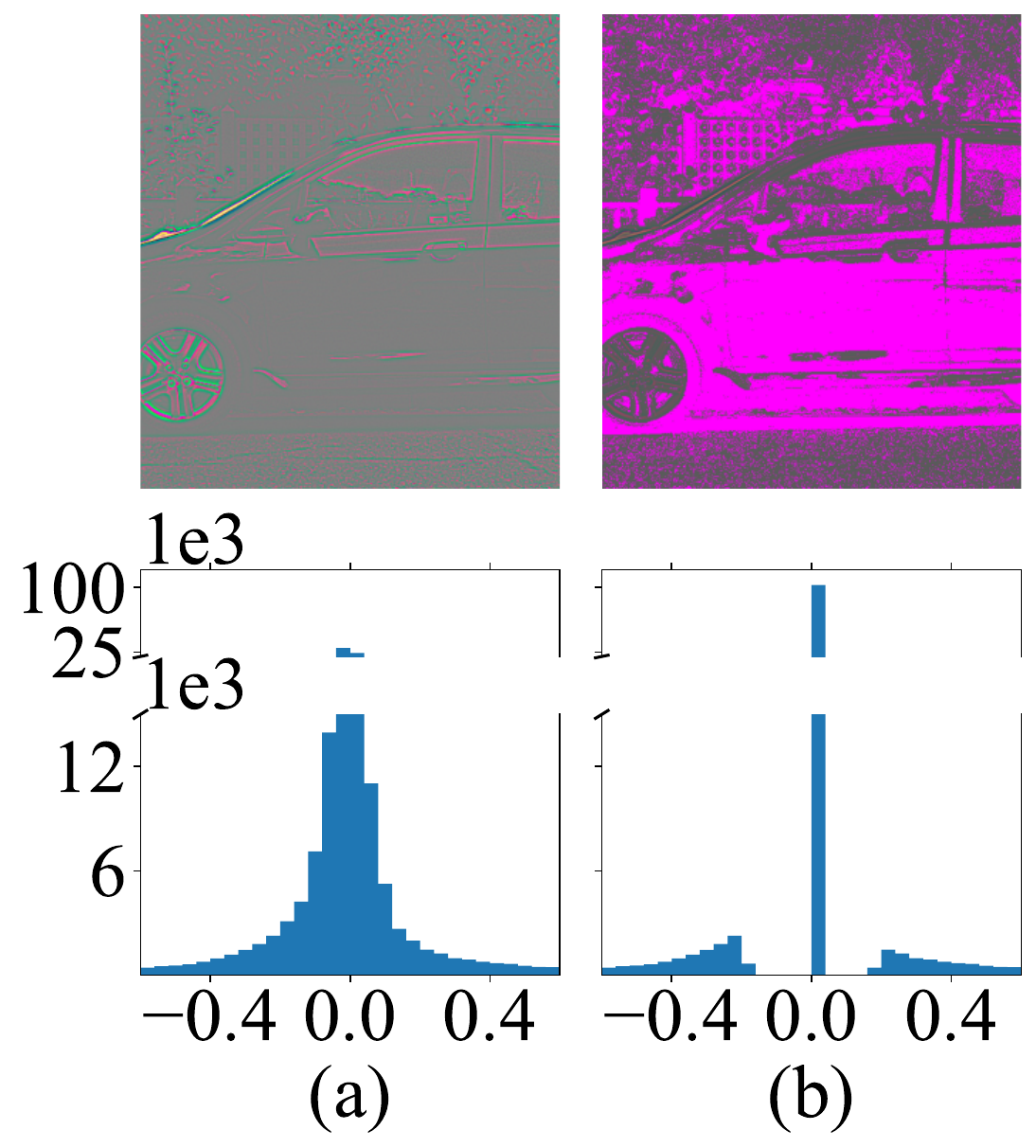}
        \label{fig:50_a}
    \end{subfigure}
    \hfill
    \begin{subfigure}[t]{0.49\linewidth}
        \centering
        \includegraphics[width=\linewidth]{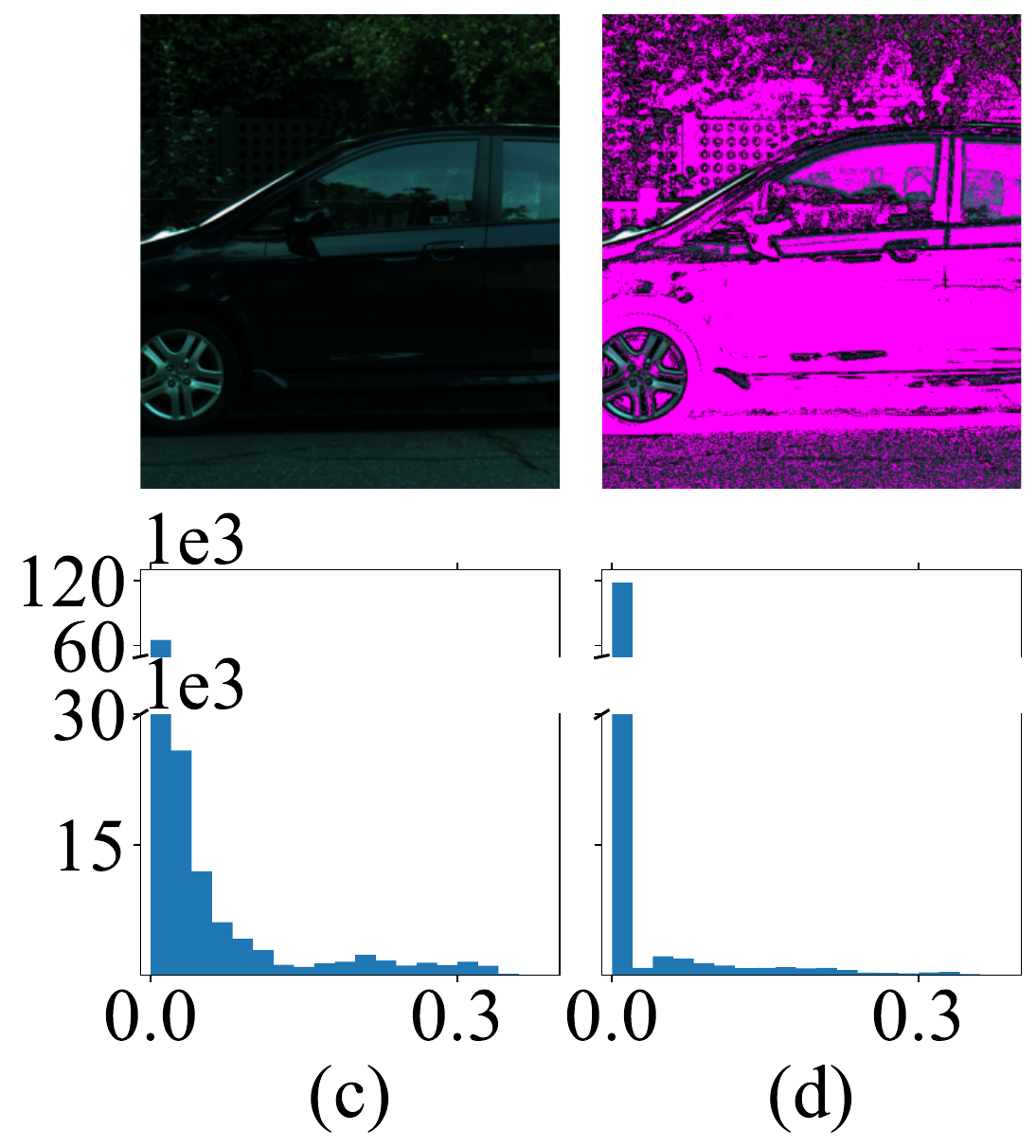}
        \label{fig:50_b}
    \end{subfigure}
    \vspace{-2em}
    \caption{Visualization and respective value distributions of an example image with our preprocessing and sparsification. (a) shows the contrast tensor obtained after our preprocessing, using the luminance variant (values around 0 appear gray); (b) the same contrast image after sparsification with p=70\%, with affected regions highlighted in pink; (c) shows the original RGB image, and (d) the original RGB image with the same sparsification mask applied as in (b). Important object structures and contours are preserved.}
    \label{fig:Combined_image_for_sparsity_50}
\end{figure}

%% file: tex_for_figures/improvement_plot.tex
\begin{figure}[ht]
  \centering
  \includegraphics[width=\linewidth]{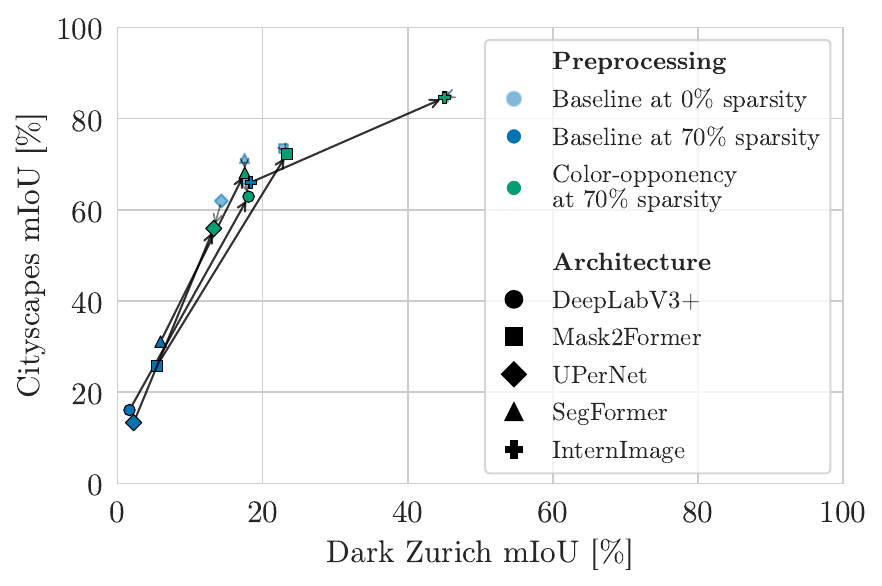}
  \caption{\textbf{Robustness improves without sacrificing i.i.d. performance.} This summary view illustrates the main trend: our preprocessing maintains high Cityscapes performance while improving nighttime generalization on Dark Zurich.}
  \label{fig:improvement_plot}
\end{figure}

%% file: tex_for_figures/sparsity_zero_shot_results.tex
\begin{figure*}[tb]
  \centering
  \includegraphics[width=1.0\textwidth]{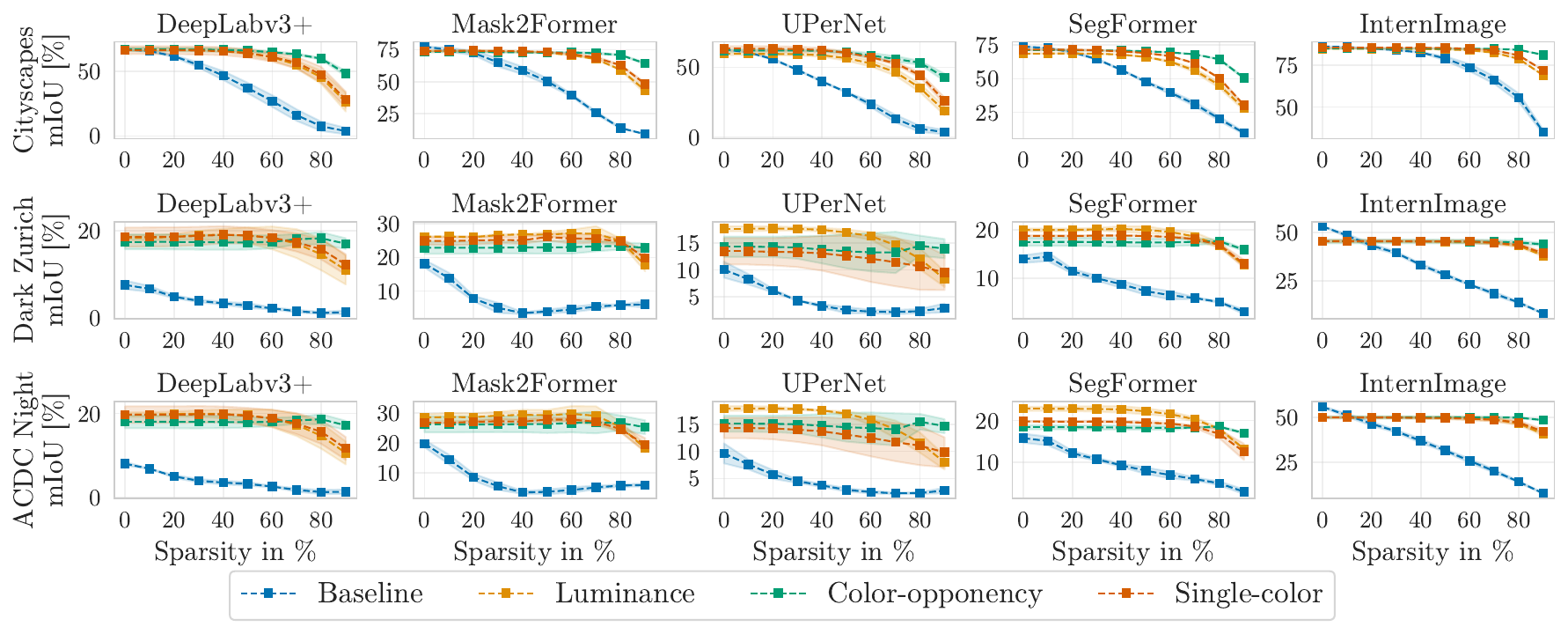}
  \caption{\textbf{Trained on Cityscapes without sparsity.} Sparsity-zero-shot mIoU results for DeepLabV3+, Mask2Former, UPerNet, SegFormer, and InternImage with all architectures trained on the Cityscapes dataset with our preprocessing and without sparsity. Performance is evaluated on Cityscapes, Dark Zurich, and ACDC Night datasets. The results cover sparsity levels from 0\% to 90\%, where 0\% denotes testing without sparsity. All preprocessing variations, luminance, color-opponency, and single-color, were trained including contrast extraction. For other metrics, please refer to the appendix.}
  \label{fig:Sparsity_results_zero-shot}
  %\vspace{-0.5em}
\end{figure*}

%\scriptsize \textcolor{red}{switch rows and columns here, so the figure fits the page}

%% file: tex_for_figures/comparison_of_learned_and_zuro_shot_results_deeplab_mask2former.tex
\begin{figure}[tb]
  \centering
  \includegraphics[width=1\linewidth]{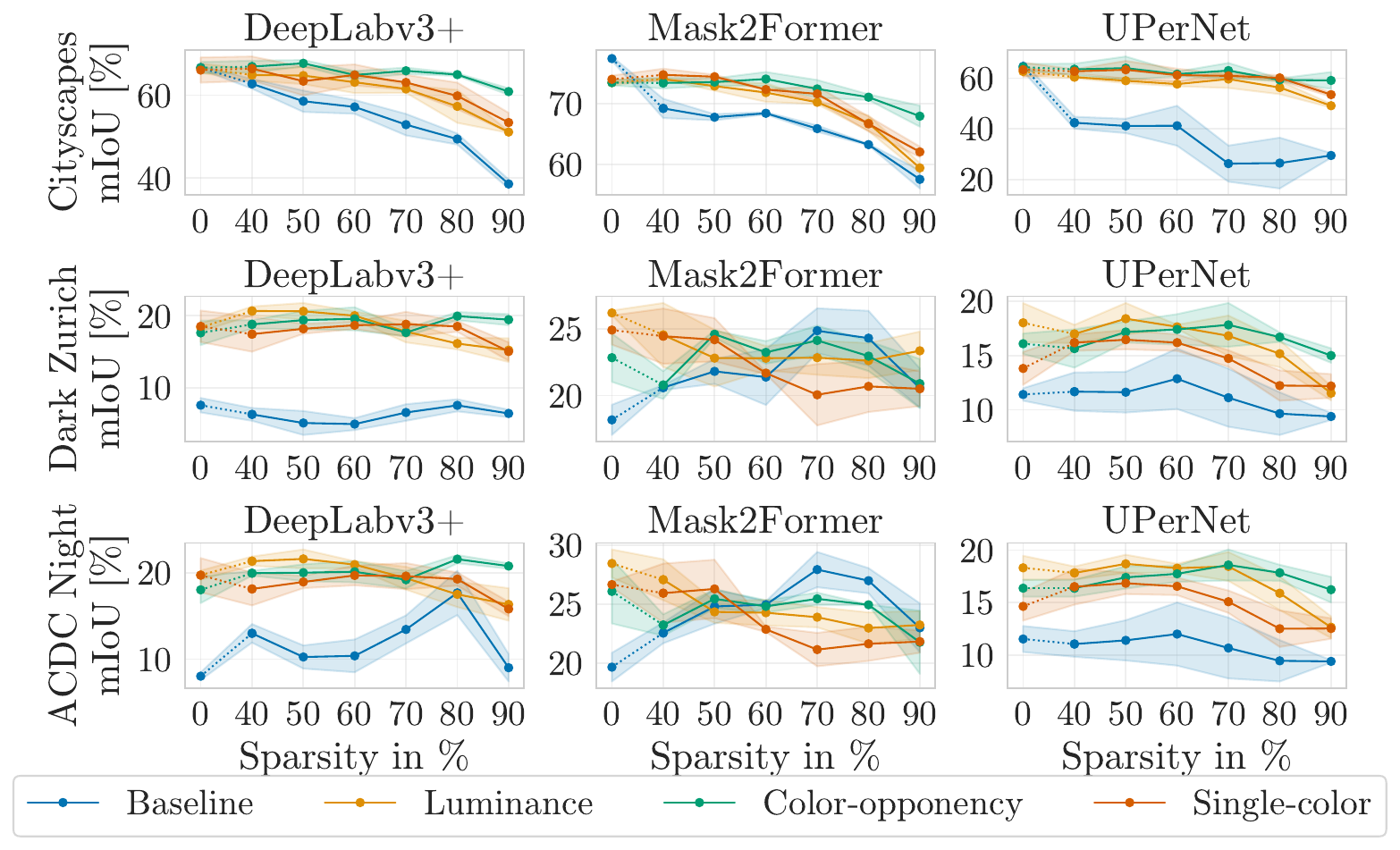}
  \caption{\textbf{Trained on Cityscapes with respective Sparsity.} mIoU results for DeepLabV3+, Mask2Former, and UPerNet, with all methods trained on the Cityscapes dataset. Performance is evaluated on Cityscapes, Dark Zurich, and ACDC Night datasets. Results are shown, covering sparsity levels from 40\% to 90\%. All preprocessing variations, luminance, color-opponency, and single-color, were trained, including contrast extraction. For other metrics, further discussions and evaluations, please refer the appendix.}
  \label{fig:Sparse results trained}
  %\vspace{-0.5em}
\end{figure}

%% file: tex_for_figures/ablation_going_beyond_nighttime_heatmap.tex
\begin{table}[tbh]
    \centering
    \caption{mIoU results of DeepLabv3+, Mask2Former, UPerNet, SegFormer, and InternImage on ACDC Fog, Rain, and Snow subsets, along with the ACDC Mean. All architectures were trained only on Cityscapes. All preprocessing variations, luminance, color-opponency, and single-color, were trained including contrast extraction. For other metrics, please refer to the appendix.}
    %\scriptsize
    \resizebox{\linewidth}{!}{
    \begin{tabular}{@{}l@{\hspace{1mm}}c@{\hspace{3mm}}r@{\hspace{3mm}}r@{\hspace{3mm}}r@{\hspace{3mm}}r@{}}
    \toprule
    Architecture & Preprocessing & ACDC Fog & ACDC Rain & ACDC Snow & ACDC Mean\\
    \midrule
    \multirow{4}{*}{DeepLabv3+} & - &  39.20~$\pm$~4.35 & 33.31~$\pm$~2.15 & 24.18~$\pm$~3.60 & 23.59~$\pm$~1.66\\
    & Luminance & 48.90~$\pm$~3.67 & 34.20~$\pm$~2.50 & 31.58~$\pm$~1.88 & 32.80~$\pm$~1.78\\
    & Color-opponency & \underline{49.46}~$\pm$~3.07 & \textbf{38.87}~$\pm$~2.24 & \textbf{32.97}~$\pm$~2.97& \underline{33.74}~$\pm$~1.99\\
    & Single-color & \textbf{50.66}~$\pm$~3.38 & \underline{38.60}~$\pm$~3.04 & \underline{32.60}~$\pm$~3.50& \textbf{34.63}~$\pm$~3.00\\
    \midrule
    \multirow{4}{*}{Mask2Former} & - &  \textbf{64.70}~$\pm$~3.04 & \textbf{49.91}~$\pm$~1.32 & \underline{48.20}~$\pm$~2.09 & 42.05~$\pm$~1.41\\
    & Luminance & 62.16~$\pm$~1.18 & \underline{47.20}~$\pm$~0.13 & \textbf{49.87}~$\pm$~2.06 & \textbf{46.83}~$\pm$~1.10\\
    & Color-opponency & 61.58~$\pm$~0.95 & 44.45~$\pm$~0.48 &  47.48~$\pm$~0.67& 44.93~$\pm$~0.66\\
    & Single-color & \underline{63.77}~$\pm$~3.25 & 44.86~$\pm$~3.38 & 46.97~$\pm$~3.98& \underline{45.23}~$\pm$~1.74\\
    \midrule
    \multirow{4}{*}{UPerNet} & - &   45.44~$\pm$~1.69 & \textbf{35.35}~$\pm$~2.12 & \underline{26.90}~$\pm$~1.92&25.65~$\pm$~0.91\\
    & Luminance & \textbf{46.52}~$\pm$~0.66 & 29.31~$\pm$~0.53 & 26.63~$\pm$~0.26& \underline{28.34}~$\pm$~0.81\\
    & Color-opponency & \underline{46.10}~$\pm$~0.60 & \underline{34.28}~$\pm$~1.23 & 26.74~$\pm$~1.63& \textbf{29.15}~$\pm$~1.62\\
    & Single-color & 45.70~$\pm$~2.04 & 33.35~$\pm$~1.69 & \textbf{27.84}~$\pm$~2.10& 28.22~$\pm$~0.89\\
    \midrule
    \multirow{4}{*}{SegFormer} & - &  \textbf{63.67}~$\pm$~0.94 & \textbf{45.50}~$\pm$~0.87 & \textbf{46.92}~$\pm$~0.51 & \textbf{41.11}~$\pm$~0.89\\
    & Luminance & 52.70~$\pm$~3.44 & 38.12~$\pm$~0.91 & 38.94~$\pm$~1.05 & 38.00~$\pm$~0.58\\
    & Color-opponency & \underline{57.79}~$\pm$~1.80 &  \underline{43.38}~$\pm$~2.29  & \underline{42.60}~$\pm$~0.63& \underline{38.87}~$\pm$~0.36\\
    & Single-color & 54.55~$\pm$~0.14 & 42.22~$\pm$~1.25 &  41.89~$\pm$~2.11& 38.74~$\pm$~0.98\\
    \midrule
    \multirow{4}{*}{InternImage} & - &  \textbf{84.12~$\pm$~0.25} & \textbf{74.16~$\pm$~1.03} & \textbf{74.18~$\pm$~0.57} & \textbf{72.02~$\pm$~0.29}\\
    & Luminance & 80.44~$\pm$~0.45 & 70.28~$\pm$~0.54 & \underline{71.16}~$\pm$~1.02 & 67.93~$\pm$~0.35\\
    & Color-opponency & 80.33~$\pm$~0.72 &  70.11~$\pm$~2.56  & 70.28~$\pm$~0.98 & 67.66~$\pm$~0.95\\
    & Single-color & \underline{81.38}~$\pm$~0.92 & \underline{71.90}~$\pm$~1.04 &  70.71~$\pm$~0.81 & \underline{68.44}~$\pm$~0.66\\
    \bottomrule
    \end{tabular}
    }
    \label{tab:Validation results - Going beyond nighttime}
\end{table}
%\fi

%% file: tex_for_figures/allweathernet_results_main_paper.tex
\begin{figure}[tb]
  \centering
  \includegraphics[width=\linewidth]{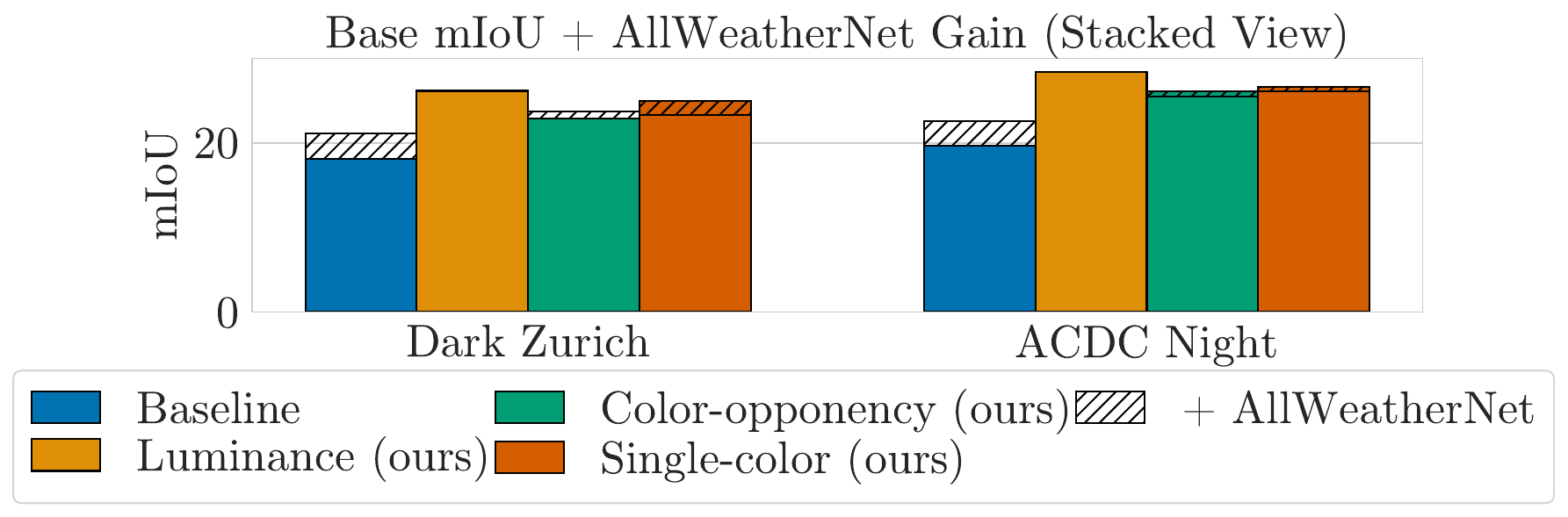}
  \caption{Evaluation of the Mask2Former trained on Cityscapes and validated on Dark Zurich and ACDC Night, including the deviation to results validated on the enhanced datasets using AllWeatherNet. For more datasets, please refer to the appendix.}
  \label{fig:allweathernet_results_main_paper}
\end{figure}

%% file: appendix.tex
\newpage
\appendix
\renewcommand{\theHsection}{\thesection}
\makeatletter
\renewcommand\subsubsection{\@startsection{subsubsection}{3}{\z@}%
  {-18\p@ \@plus -4\p@ \@minus -4\p@}
  {4\p@ \@plus 2\p@ \@minus 2\p@}
  {\normalfont\normalsize\bfseries}}
\makeatother
\onecolumn
\begin{center}
    {\LARGE \bfseries Improved Robustness from Biologically Inspired Sparse Contrast Representations \par}
    \vspace{1em}
    {\Large \bfseries Supplementary Material \par}
\end{center}
\section*{Contents}
\providecommand{\authcount}[1]{}
\setcounter{secnumdepth}{3}

\startcontents[sections]
\printcontents[sections]{}{1}{\setcounter{tocdepth}{3}}

\section{Implementation Details }
\label{sec:appendix:implementation_details}

\subsection{Detailed Description of the Preprocessing }
\label{sec:appendix:preprocessing_description}
Our preprocessing mimics aspects of the human visual system by applying color remapping and contrast extraction to the input image. As described in the main paper (see \cref{sec:preprocessing}), we investigate three preprocessing variants: luminance, color-opponency and single-color.

\noindent The preprocessing is applied after normalization before the input images are fed into the neural network. Depending on the configuration, it consists of two to four preprocessing steps, three of which are implemented with convolution layers.

\noindent In the first step, the input image is blurred multiple times with the intermediate results being concatenated along the channel dimension, resulting in a tensor with \(3(d+1)\) channels: the first three channels correspond to the RGB channels of the original image, followed by the RGB channels of the increasingly blurred images. 

\noindent For the second step another convolutional layer performs the color space conversion and the contrast computation by subtracting the increasingly blurred images from the original image. Depending on the configuration, this layer outputs either a single-channel or a three-channel representation. The calculations for the weights are provided in the main paper (see \cref{sec:implementation}). The weights are concatenated in accordance to the step before, resulting in a convolution with 3(d+1) input and 3 or 1 output channels.

\noindent To maintain compatibility with networks expecting a three-channel input, an optional third step duplicates result from the last step in case of a single channel output (luminance). We implemented this as a convolutional layer with one input channel, three output channels, a \(1\times 1\) kernel, stride 1, and no padding. It acts as a pointwise convolution, expanding the single channel to three identical channels.

\noindent As an optional final step after preprocessing, sparsity can be introduced to the contrast image. We first calculate a key value for which the given percentage of the absolute values are lower. Then we enforce sparsity by setting every pixel with an absolute value below that key value to zero~(\cref{sec:sparsity}). In the threshold variant of this step, we set the key value directly from the given threshold.

\subsection{Experimental Setup }
\label{sec:appendix:experimental_setup}
All experiments were conducted either on NVIDIA Tesla V100 GPUs (16 GB) or on NVIDIA H100 GPUs for the task of semantic segmentation, using a single GPU for training in all cases.

\noindent\textbf{Downstream Task. }
We consider semantic segmentation as the downstream task, since it requires correct pixel-wise predictions and is therefore well suited to analyze if our preprocessing preserves fine spatial structures while improving robustness to OOD data. 

\noindent\textbf{Datasets. } 
All models are trained on the Cityscapes dataset~\cite{Cityscapes_dataset}, while evaluation is carried out on Cityscapes, Dark Zurich~\cite{Dark_zurich_dataset}, and ACDC~\cite{ACDC_dataset}. 

\noindent\textbf{Evaluation Metrics. } 
Our evaluation includes the mean Intersection over Union (mIoU) computed between the predicted and ground truth masks, the mean class accuracy (mAcc), and the overall pixel accuracy (aAcc). All results are averaged over three seeds.

\noindent\textbf{Architectures. } 
For Deeplabv3+~\cite{DeepLabv3+}, Mask2Former~\cite{Mask2Former} with a ResNet-50~\cite{he2016deep} backbone, UPerNet~\cite{UPerNet} and SegFormer~\cite{SegFormer}, we utilize configuration files that are implmented in MMSegmentation~\cite{MMSegmentation}. We consider Deeplabv3+ (deeplabv3plus\linebreak\_r50-d8\_4xb2-80k\_cityscapes-512x1024.py), Mask2Former(mask2former\linebreak\_r50\_8xb2-90k\_cityscapes-512x1024.py), UPerNet  (upernet\_r50\_4xb2-\linebreak80k\_cityscapes-512x1024.py) and SegFormer (segformer\_mit-b0\_8xb1-\linebreak160k\_cityscapes-512x1024.py). We changed the scheduler from this SegFormer from (/\_base\_/schedules/schedule\_160k.py) to (/\_base\_/schedules/schedule\linebreak\_240k.py). Consequently, when we refer to SegFormer in this paper, it is trained for 240k iterations. For Mask2Former with an InternImage-Huge backbone~\cite{wang2023internimage} (denoted as InternImage), we adopt the implementation from OpenGVLab~\cite{wang2023internimage} and integrate our preprocessing methods, including sparsity. For the depth-search in \cref{sec:appendix:depth_of_contrast_extraction}, we utilize an alternative version of Deeplabv3+ (deeplabv3plus\_r18-d8\_4xb2-80k\_cityscapes-512x1024.py). We denote this as Deeplabv3+ (ResNetV1c-18)~\cite{ResNetV1c}.

\noindent\textbf{Training Regime. } 
The training was performed according to the configuration files provided by MMSegmentation or according to the implementation by OpenGVLab when utilizing InternImage, as described in the Architectures section, using pretrained weights unless explicitly stated otherwise.

\subsection{Implications on Sensor Design }
As mentioned in the main paper the induced sparsity can be used to reduce bandwidth and power consumption if implemented on a sensor. 
An implementation of the scheme directly on a CMOS image sensor chip would reduce the stress on both the data converters and the data transmission circuits, which are both known to account for a significant proportion of the power consumption in CMOS camera systems.
The proposed \emph{sparsification by magnitude threshold} can be implemented efficiently by increasing the minimum threshold level of the data converter. Data converters typically exhibit higher nonlinearity for analog input voltages close to the supply rails (0V and VDD). By avoiding the conversion of small signal values, the converter no longer needs to operate near these rails, thereby improving conversion efficiency and linearity.

As only a limited range of analog values must be digitized, the required ADC resolution can also be significantly reduced. The power consumed by an ADC is commonly approximated as $P = f_s \cdot \text{FOM}_w \cdot 2^{\text{ENOB}}$, where $f_s$ is the sampling frequency, $FOM_w$ is the Walden figure of merit, and $ENOB$ is the effective number of bits. Assuming a similar figure of merit, reducing the bit depth directly lowers ADC power consumption, with each bit reduction approximately halving the required power. Since typical image sensors employ one ADC per column, or per small group of columns, reducing the conversion resolution across the array can lead to substantial overall power savings.

The proposed approach also reduces transmission power requirements. After thresholding, a large proportion of pixel values become zero, which significantly decreases the amount of data that must be transmitted off-chip. For example, the CMV4000 image sensor consumes approximately 18 mW per channel when using LVDS (Low-Voltage Differential Signaling).
Reducing the number of output channels from 16 to 4 would therefore save approximately 216 mW, corresponding to roughly 33\% of the interface power budget~\cite{amsosram2025cmv4000}.

In our own designs, we have observed even higher reduction in the power requirement as we reduce the number of bits. As 60 to 70\% of pixels become zero in some of the images, we can even not use an LVD signaling and utilize lower power fixed-voltage line readouts, thereby further reducing the power requirement of these chips.

Approximating DoG kernels by box-filters (ie. identical weights)   simplifies the implementation as unity-gain amplifiers without coefficient storage or additional memory elements. These can be implemented within the sensor columns, where the available pitch is small due to small pixel dimensions, thereby requiring low-area circuits. Challenges that may require more research are potential additional noise sources and different quantization rates that need to be considered.

\section{Extended Zero-Shot Results}
\label{sec:appendix:experimental_results}
In this section, we provide comprehensive experimental results for all studies presented in the main paper, reporting the evaluation metrics mIoU, mAcc, and aAcc.

\noindent\textbf{Results.}
\label{sec:results_appendix}
We report detailed results for our main experiments (see \cref{sec:results}). For each architecture, including DeepLabv3+, Mask2Former, UPerNet, SegFormer, and InternImage, we show the baseline as well as the three preprocessing variants: luminance, color-opponency, and single-color.

Results at 0\% sparsity on Cityscapes, Dark Zurich, and ACDC Night are visualized in \cref{fig:validation_results_robustness_under_nighttime_conditions_appendix}, showing mIoU, mAcc, and aAcc for each configuration.
For each architecture, we also provide separate tables reporting validation results on Cityscapes, Dark Zurich, all ACDC subsets, and the ACDC Mean. Each table presents mIoU, mAcc, and aAcc for the baseline and the three preprocessing variants. Results are reported for sparsity levels from 0\% to 90\% in increments of 10\%. All experiments are conducted in a zero-shot setting, meaning the models are evaluated without sparsity-specific training.

Results for DeepLabv3+ are reported in \cref{tab:sparse_results_deeplabv3+_cityscapes_zero_shot} to \cref{tab:sparse_results_deeplabv3plus_acdc_full_zero_shot}, for Mask2Former in \cref{tab:sparse_results_mask2Former_cityscapes_zero_shot} to \cref{tab:sparse_results_mask2former_acdc_full}, for UPerNet in \cref{tab:sparse_results_upernet_cityscapes_zero_shot} to \cref{tab:sparse_results_upernet_acdc_full}, for SegFormer in \cref{tab:sparse_results_segformer_cityscapes_zero_shot} to \cref{tab:sparse_results_segformer_acdc_full}, and for InternImage in \cref{tab:sparse_results_internimage_cityscapes_val} to \cref{tab:sparse_results_internimage_acdc_4_mean}. For each architecture and dataset, one table is provided.

\input{tex_for_figures/appendix_main_results_mIoU_depth_5}

\subsection{DeepLabv3+}
\input{tables/sparsity_zero_shot_deeplab}

\subsection{Mask2Former}
\input{tables/sparsity_zero_shot_mask2former}
\subsection{UPerNet}
\input{tables/sparsity_zero_shot_upernet}
\subsection{SegFormer}
\input{tables/sparsity_zero_shot_segformer}
\subsection{InternImage}
\input{tables/sparsity_zero_shot_internimage}

\section{Ablation Studies}
\label{sec:ablation_appendix}
\subsection{Learned Sparsity}
For DeepLabv3+, Mask2Former, and UPerNet, we additionally report results for models explicitly trained with sparsity. Sparsity levels include 0\% and 40\% to 90\% in increments of 10\%. Results are provided for the baseline and all preprocessing variants on Cityscapes, Dark Zurich, all ACDC subsets, and the ACDC Mean. Specifically, DeepLabv3+ results are presented in \cref{tab:sparse_results_deeplabv3+_cityscapes_learned} to \cref{tab:sparse_results_deeplabv3+_acdc_full_learned}, Mask2Former in \cref{tab:sparse_results_mask2former_cityscapes_learned} to \cref{tab:sparse_results_mask2former_acdc_full_learned}, and UPerNet in \cref{tab:sparse_results_upernet_cityscapes_learned} to \cref{tab:sparse_results_upernet_acdc_full_learned}. We observed a peak of the Mask2Former baseline at 70\% sparsity in nighttime scenarios and conducted additional analyses using a frequency-weighted mean IoU (per-class IoU weighted by pixel frequency in the validation set). The sparsified RGB baseline shows a bias towards frequent classes (such as sky), suggesting reliance on class priors rather than robust sparse features, whereas our preprocessing does not exhibit this behavior, indicating greater robustness under sparsification. Results are visualized in \cref{fig:maks2former_peak_analysis}.

\subsubsection{DeepLabv3+}
\input{tables/sparsity_trained_deeplab}

\subsubsection{Mask2Former}
\input{tables/sparsity_trained_mask2former}
\input{tex_for_figures/mask2former_peak}
\clearpage
\subsubsection{UPerNet}
\input{tables/sparsity_trained_upernet}

\subsection{Threshold Sparsity}
In addition to the sparsification method described in \cref{sec:sparsity}, where a fixed percentage of the smallest absolute values is set to zero, we examine a threshold-based sparsification strategy for DeepLabv3+ and Mask2Former. In this variant, all values within the interval $[-\textit{threshold}, +\textit{threshold}]$ are set to zero. Results are reported for no sparsity as well as thresholds of 0.005, 0.0075, 0.01, and 0.02 on Cityscapes, Dark Zurich, all ACDC subsets, and the ACDC Mean. DeepLabv3+ results are shown in \cref{tab:sparse_results_deeplabv3+_cityscapes_threshold} to \cref{tab:sparse_results_deeplabv3+_acdc_full_threshold}, and Mask2Former results in \cref{tab:sparse_results_mask2former_cityscapes_threshold} to \cref{tab:sparse_results_mask2former_acdc_full_threshold}. As before, one table is provided per architecture and dataset.

We further report the average percentage of values set to zero for each sparsity threshold and dataset in \cref{tab:zero_percentages_threshold}. 

\subsubsection{DeepLabv3+}

\input{tables/sparsity_threshold_deeplab}

\subsubsection{Mask2Former}
\input{tables/sparsity_threshold_mask2former}
\subsubsection{Average Zero Percentages}
\input{tables/sparsity_threshold_zero_percentage}

\newpage
\subsection{Training from Scratch}
In addition to fine-tuning with our proposed preprocessing techniques, we trained the SegFormer architecture from scratch without relying on pretrained weights. For this purpose, we adapted the original training schedule (/\_base\_/schedules/\linebreak schedule\_160k.py) by increasing the number of iterations from 16k to 48k. The corresponding results are summarized in \cref{tab:training_from_scratch} to \cref{tab:training_from_scratch_2}.

\input{tables/training_from_scratch}

\subsection{Grayscale in Nighttime Situations}
Motivated by the strong nighttime performance of luminance-based preprocessing (\cref{sec:results}), we further analyze this approach independently of the \emph{spatial contrast extraction} (i.e., with $\mathrm{d}=0$).

We evaluate DeepLabv3+, SegFormer and UPerNet on two luminance variants: (i) a uniform version with all coefficients of $\mathrm{M}=\frac{1}{3}$, and (ii) a \emph{luminance green bias} formulation as defined in \cref{eq:matrix_luminance_green_bias_1x3}. Each variant is tested both as a single-channel and as a three-channel input. The results are visualized in (\cref{fig:Validation_results_grayscale_in_nighttime_situations_appendix}). Consistent with the main experiments, all luminance-based variants outperform the baseline under nighttime conditions. Detailed results for each dataset are provided in \cref{tab:going_beyond_nighttime_cityscapes} to \cref{tab:going_beyond_nighttime_acdc_snow}, covering Cityscapes, Dark Zurich, and all ACDC subsets. One table is presented for each dataset.

\input{tex_for_figures/appendix_ablation_grayscale_in_nighttime}
\input{tables/grayscale_evaluation}

\subsection{Integration Into AllWeatherNet}
In the main paper, we showed results for the integration of our preprocessing into AllWeatherNet. In this section, we report additional validation results for the Mask2Former architecture trained on Cityscapes and validated on Dark Zurich, each ACDC split and the ACDC Mean dataset comparing the results with and without the integration of AllWeatherNet. Resuls are shown in \cref{fig:allweathernet_results}. As discussed in the main paper, our preprocessing increases performance compared to the combination of the baseline and AllWeatherNet. In some cases applying our approach jointly with AllWeatherNet results in an even better performance. 
\input{tex_for_figures/allweathernet_results}

\subsection{Depth of Contrast Extraction}
\label{sec:appendix:depth_of_contrast_extraction}
We additionally ablate the contrast depth $d$, which controls the extent of center-surround aggregation in our DoG-style operator from \cref{sec:contrast_extraction}.
To enable a broader sweep under resource constraints, we use DeepLabV3+ with a smaller backbone (ResNetV1c-18) and evaluate $d \in \{0,\dots,10\}$. \cref{fig:Validation_results_depth_parameter_search_appendix} shows that depth has a substantial effect on both i.i.d. accuracy and robustness, with the optimal value depending mildly on the remapping variant and test domain.
Very small depths can under-emphasize contrast, while very large depths can over-smooth and remove details.
Balancing robustness under shift and performance on Cityscapes, $d{=}5$ offers a strong trade-off across variants, and we therefore use it as the default throughout the main experiments. For depth $d = 0$, we include results for the luminance and \emph{luminance green bias} variants with both single-channel and three-channel inputs. For depths $d = 0$ to $10$, results are presented for all three preprocessing variants. Detailed numerical results are provided in \cref{tab:depth_search_validation_on_cityscapes} to \cref{tab:depth_search_validation_on_acdc_snow}, with a separate table for each dataset.

\iffalse
In the main paper, we performed an ablation study on the depth parameter $d$. In this section, we report additional validation results for DeepLabv3+ (ResNetV1c-18). 
For depth $d = 0$, we include results for the luminance and \emph{luminance green bias} variants with both single-channel and three-channel inputs. For depths $d = 0$ to $10$, results are presented for all three preprocessing variants. The extended results are visualized in \cref{fig:Validation_results_depth_parameter_search_appendix}. 
Detailed numerical results are provided in \cref{tab:depth_search_validation_on_cityscapes} to \cref{tab:depth_search_validation_on_acdc_snow}, with a separate table for each dataset.
\fi

\input{tex_for_figures/appendix_ablation_depth_parameter_search}
\input{tables/depth_all_results}

\section{Classification}
\label{sec:appendix:further_experiments}
To extend the evaluation beyond street-scene segmentation, we also assessed the proposed preprocessing method on ConvNeXt-Tiny~\cite{liu2022convnet} for the task of image classification using the CIFAR-100 dataset~\cite{cifar100}, including robustness validation on CIFAR-100-C~\cite{cifar100c}. We adopted the implementation provided by OpenMixup \cite{li2022openmixup} and integrated our preprocessing pipeline into the framework. We used the configs configs/classification/cifar100/mixups/vits/convnext\_tiny/convnext\_t\_vanilla\_bs100\_ep200.py and configs/classification/cifar100/mixups/vits/convnext\_tiny/convnext\_t\_cutmix\_a2\_bs100\_ep200.py. The results are presented in \cref{fig:Validation results-cifar_vanilla} and \cref{fig:Validation results-cifar_mixup}, which show the Top-1 validation accuracy without and with CutMix~\cite{yun2019cutmix} augmentation, respectively. Overall, our preprocessing improves robustness to the corruptions in CIFAR-100-C, both with and without CutMix. We observe that robustness increases for all corruption types except impulse noise, certain blur variants, elastic transformations, pixelation, and JPEG compression when CutMix is not applied. In contrast, when combined with CutMix, our preprocessing even improves performance on impulse noise and pixelation.
\input{tex_for_figures/cifar_results_vanilla}
\input{tex_for_figures/cifar_results_mixup}

\clearpage
\input{tex_for_figures/extended_teaser_new}

\section{Additional Visualizations}
\label{sec:additional_visualizations_appendix}
In \cref{fig:extended_teaser}, we extend the visualizations presented in \cref{fig:teaser_new}. In \cref{fig:color-opponency_visualization_appendix}, we compare the original input images with those reparameterized using the color-opponency preprocessing at depth $d = 0$.

\input{tex_for_figures/color_opponency_depth0_new}

\section{Limitations}
\label{sec:appendix:limitations}
In our experiments, we aimed to identify the best-performing depth (\cref{sec:results}). While it is possible to determine a single overall best-performing depth, the optimal value can vary depending on the color-reparameterization used in the preprocessing. Moreover, the results from the depth search in the range of 0 to 10 may not be directly transferable to other architectures. This is illustrated by the depth search conducted on Deeplabv3+, as shown in \cref{fig:Validation_results_depth_parameter_search_appendix}: in this case, the luminance preprocessing with depth \(d=5\) yields better performance than luminance with depth \(d=0\) (i.e., luminance without contrast). By contrast, for all other architectures, luminance preprocessing with depth \(d=0\) surpasses depth \(d=5\) in nighttime scenarios.

\noindent Consequently, future research should focus on assessing how different architectures influence the performance across varying depths, or on developing methods that avoid using a fixed depth value during training.

%% file: tex_for_figures/appendix_main_results_mIoU_depth_5.tex
\begin{figure}[tb]
  \centering
  \includegraphics[width=\linewidth]{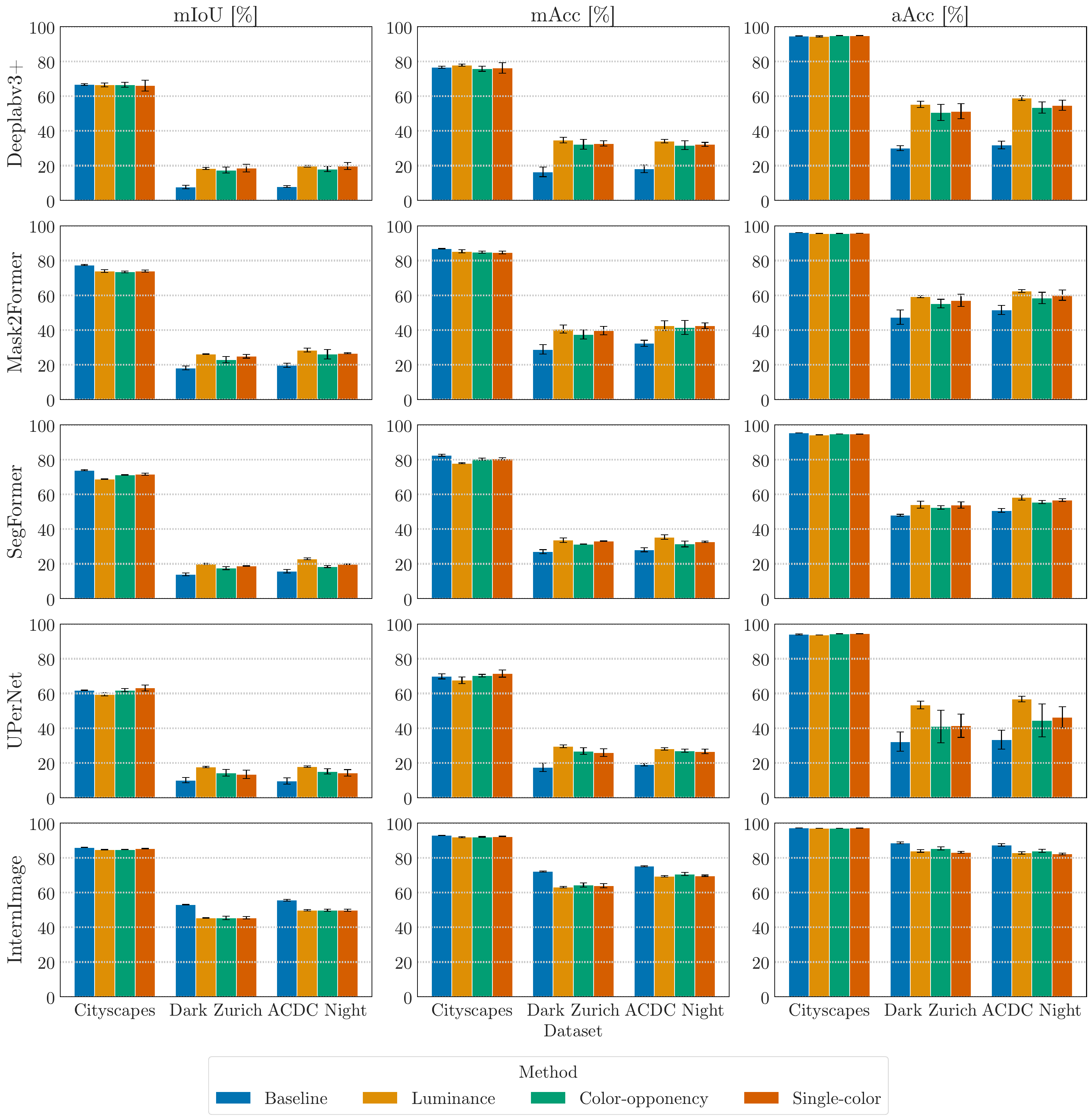}
  \caption{mIoU, mAcc and aAcc results of DeepLabv3+, Mask2Former, SegFormer, UPerNet, and InternImage architectures on Cityscapes, Dark Zurich and ACDC Night datasets.}
  \label{fig:validation_results_robustness_under_nighttime_conditions_appendix}
\end{figure}
\clearpage

%% file: tables/sparsity_zero_shot_deeplab.tex
% zero-shot tables
\begin{table}[h]
  \centering
  \caption{\textbf{Zero-shot} evaluation of the \textbf{DeepLabv3+} architecture trained on Cityscapes without sparsity and \textbf{validated on Cityscapes} at different sparsities. Results include the three preprocessing types (luminance, color-opponency, and single-color). Metrics reported are mean Intersection over Union (mIoU), mean Accuracy (mAcc), and average Accuracy (aAcc), averaged over three seeds. \textbf{Percent-based sparsity:} For a given \textit{percentage}, the \textit{percentage} of lowest absolute values are set to zero.}
  \scalebox{0.85}{
  \begin{tabular}{@{}lccrrr@{}}
    \toprule 
    Preprocessing &  Sparsity &\multicolumn{1}{c}{mIoU}&\multicolumn{1}{c}{mAcc}&\multicolumn{1}{c}{aAcc}  \\
    \midrule
    - & \multirow{4}{*}{-} &  66.68~$\pm$~0.44 & 76.58~$\pm$~0.60 & 94.68~$\pm$~0.05 \\
    Luminance   &  & 66.53~$\pm$~1.11 & 77.86~$\pm$~0.63 & 94.35~$\pm$~0.35 \\
     Color-opponency&   &  66.64~$\pm$~1.39 & 75.80~$\pm$~1.46 & 94.88~$\pm$~0.15 \\
     Single-color&   &  66.08~$\pm$~3.05 & 76.33~$\pm$~3.03 & 94.90~$\pm$~0.16 \\
    \midrule
     - & \multirow{4}{*}{10\%} & 65.36~$\pm$~0.71 & 75.88~$\pm$~0.62 & 94.16~$\pm$~0.07 \\
    Luminance & & 66.53~$\pm$~1.11 & 77.86~$\pm$~0.62 & 94.35~$\pm$~0.34 \\
    Color-opponency & & 66.64~$\pm$~1.38 & 75.80~$\pm$~1.45 & 94.88~$\pm$~0.15 \\
    Single-color & & 66.09~$\pm$~3.04 & 76.34~$\pm$~3.02 & 94.90~$\pm$~0.16 \\
    \midrule
    - & \multirow{4}{*}{20\%} & 61.43~$\pm$~0.85 & 74.39~$\pm$~0.60 & 91.92~$\pm$~0.26 \\
    Luminance & & 66.52~$\pm$~1.08 & 77.83~$\pm$~0.60 & 94.34~$\pm$~0.34 \\
    Color-opponency & & 66.61~$\pm$~1.38 & 75.77~$\pm$~1.45 & 94.87~$\pm$~0.16 \\
    Single-color & & 66.04~$\pm$~3.05 & 76.27~$\pm$~3.03 & 94.85~$\pm$~0.15 \\
    \midrule
    - & \multirow{4}{*}{30\%} & 54.59~$\pm$~1.95 & 70.93~$\pm$~0.62 & 86.70~$\pm$~1.25 \\
    Luminance & & 66.40~$\pm$~1.08 & 77.65~$\pm$~0.68 & 94.28~$\pm$~0.34 \\
    Color-opponency & & 66.49~$\pm$~1.38 & 75.65~$\pm$~1.48 & 94.81~$\pm$~0.17 \\
    Single-color & & 65.86~$\pm$~3.14 & 76.10~$\pm$~3.08 & 94.71~$\pm$~0.16 \\
    \midrule
    - & \multirow{4}{*}{40\%} & 46.53~$\pm$~3.44 & 64.78~$\pm$~1.98 & 77.46~$\pm$~3.85 \\
    Luminance & & 65.77~$\pm$~1.16 & 76.97~$\pm$~0.87 & 93.97~$\pm$~0.38 \\
    Color-opponency & & 66.27~$\pm$~1.35 & 75.41~$\pm$~1.51 & 94.70~$\pm$~0.22 \\
    Single-color & & 65.29~$\pm$~3.15 & 75.67~$\pm$~3.20 & 94.33~$\pm$~0.20 \\
    \midrule
    - & \multirow{4}{*}{50\%} & 36.99~$\pm$~4.02 & 54.64~$\pm$~4.10 & 65.06~$\pm$~8.11 \\
    Luminance & & 64.06~$\pm$~1.78 & 75.39~$\pm$~1.38 & 92.88~$\pm$~0.70 \\
    Color-opponency & & 65.73~$\pm$~1.21 & 74.94~$\pm$~1.47 & 94.44~$\pm$~0.31 \\
    Single-color & & 63.68~$\pm$~3.24 & 74.50~$\pm$~3.45 & 93.20~$\pm$~0.56 \\
    \midrule
    - & \multirow{4}{*}{60\%} & 26.89~$\pm$~4.51 & 41.59~$\pm$~5.82 & 52.68~$\pm$~12.23 \\
    Luminance & & 60.60~$\pm$~2.50 & 72.31~$\pm$~2.32 & 89.86~$\pm$~1.63 \\
    Color-opponency & & 64.56~$\pm$~0.81 & 73.95~$\pm$~1.36 & 93.78~$\pm$~0.63 \\
    Single-color & & 60.86~$\pm$~3.63 & 72.09~$\pm$~3.89 & 91.00~$\pm$~1.27 \\
    \midrule
    - & \multirow{4}{*}{70\%} & 16.07~$\pm$~4.45 & 26.48~$\pm$~6.17 & 38.71~$\pm$~14.95 \\
    Luminance & & 54.84~$\pm$~2.95 & 66.66~$\pm$~3.48 & 84.40~$\pm$~2.70 \\
    Color-opponency & & 62.89~$\pm$~0.63 & 72.47~$\pm$~1.46 & 92.93~$\pm$~0.99 \\
    Single-color & & 56.42~$\pm$~4.50 & 67.90~$\pm$~4.62 & 87.55~$\pm$~2.06 \\
    \midrule
    - & \multirow{4}{*}{80\%} & 7.51~$\pm$~3.57 & 13.96~$\pm$~4.50 & 24.99~$\pm$~15.90 \\
    Luminance & & 45.21~$\pm$~5.07 & 56.75~$\pm$~6.00 & 76.02~$\pm$~4.51 \\
    Color-opponency & & 59.55~$\pm$~0.84 & 69.20~$\pm$~1.63 & 91.82~$\pm$~1.37 \\
    Single-color & & 47.24~$\pm$~6.01 & 59.40~$\pm$~6.10 & 80.40~$\pm$~3.69 \\
    \midrule
    - & \multirow{4}{*}{90\%} & 3.84~$\pm$~2.48 & 8.44~$\pm$~2.78 & 17.29~$\pm$~13.35 \\
    Luminance & & 26.01~$\pm$~7.17 & 36.31~$\pm$~8.01 & 59.13~$\pm$~8.21 \\
    Color-opponency & & 48.04~$\pm$~2.05 & 59.18~$\pm$~2.91 & 83.64~$\pm$~4.83 \\
    Single-color & & 27.99~$\pm$~6.20 & 40.36~$\pm$~7.18 & 61.46~$\pm$~7.79 \\
     
    \bottomrule
  \end{tabular}
  }
  \label{tab:sparse_results_deeplabv3+_cityscapes_zero_shot}
\end{table}
\clearpage

\begin{table}[tb]
  \centering
  \caption{\textbf{Zero-shot} evaluation of the \textbf{DeepLabv3+} architecture trained on Cityscapes without sparsity and \textbf{validated on Dark Zurich} at different sparsities. Results include the three preprocessing types (luminance, color-opponency, and single-color). Metrics reported are mean Intersection over Union (mIoU), mean Accuracy (mAcc), and average Accuracy (aAcc), averaged over three seeds. \textbf{Percent-based sparsity:} For a given \textit{percentage}, the \textit{percentage} of lowest absolute values are set to zero.} 
  \scalebox{0.85}{
  \begin{tabular}{@{}lc rrr@{}}
    \toprule
    Preprocessing &  Sparsity &\multicolumn{1}{c}{mIoU}&\multicolumn{1}{c}{mAcc}&\multicolumn{1}{c}{aAcc}  \\
    \midrule
    - & \multirow{4}{*}{-}  &  7.64~$\pm$~1.00 & 16.40~$\pm$~2.74 & 30.13~$\pm$~1.41 \\
     Luminance  & & 18.39~$\pm$~0.54 & 34.73~$\pm$~1.68 & 55.35~$\pm$~1.81 \\
    Color-opponency&   &  17.46~$\pm$~1.70 & 32.32~$\pm$~2.85 & 50.68~$\pm$~4.61 \\
    Single-color &  & 18.54~$\pm$~2.18 & 32.79~$\pm$~1.49 & 51.35~$\pm$~4.26 \\
    \midrule
    - & \multirow{4}{*}{10\%} & 6.72~$\pm$~0.67 & 14.72~$\pm$~2.10 & 26.16~$\pm$~2.63 \\
    Luminance & & 18.40~$\pm$~0.56 & 34.73~$\pm$~1.70 & 55.35~$\pm$~1.83 \\
    Color-opponency & & 17.46~$\pm$~1.71 & 32.32~$\pm$~2.85 & 50.68~$\pm$~4.61 \\
    Single-color & & 18.57~$\pm$~2.18 & 32.83~$\pm$~1.50 & 51.36~$\pm$~4.25 \\
    \midrule
    - & \multirow{4}{*}{20\%} & 4.94~$\pm$~0.34 & 11.00~$\pm$~1.30 & 20.83~$\pm$~2.89 \\
    Luminance & & 18.59~$\pm$~0.64 & 34.78~$\pm$~1.95 & 55.40~$\pm$~1.66 \\
    Color-opponency & & 17.44~$\pm$~1.71 & 32.30~$\pm$~2.86 & 50.66~$\pm$~4.61 \\
    Single-color & & 18.62~$\pm$~2.16 & 32.84~$\pm$~1.46 & 51.36~$\pm$~4.11 \\
    \midrule
    - & \multirow{4}{*}{30\%} & 4.00~$\pm$~0.33 & 9.02~$\pm$~0.83 & 18.18~$\pm$~2.69 \\
    Luminance & & 18.98~$\pm$~1.00 & 34.69~$\pm$~2.48 & 54.95~$\pm$~1.06 \\
    Color-opponency & & 17.39~$\pm$~1.71 & 32.23~$\pm$~2.87 & 50.60~$\pm$~4.62 \\
    Single-color & & 18.80~$\pm$~2.09 & 32.88~$\pm$~1.31 & 51.38~$\pm$~4.12 \\
    \midrule
    - & \multirow{4}{*}{40\%} & 3.43~$\pm$~0.34 & 8.14~$\pm$~0.64 & 17.10~$\pm$~2.27 \\
    Luminance & & 19.08~$\pm$~1.35 & 34.03~$\pm$~2.79 & 53.86~$\pm$~0.98 \\
    Color-opponency & & 17.33~$\pm$~1.69 & 32.18~$\pm$~2.88 & 50.45~$\pm$~4.51 \\
    Single-color & & 19.06~$\pm$~1.98 & 32.98~$\pm$~1.21 & 51.00~$\pm$~4.09 \\
    \midrule
    - & \multirow{4}{*}{50\%} & 2.93~$\pm$~0.57 & 7.57~$\pm$~0.98 & 15.59~$\pm$~1.33 \\
    Luminance & & 18.91~$\pm$~1.83 & 32.93~$\pm$~3.28 & 52.36~$\pm$~1.44 \\
    Color-opponency & & 17.31~$\pm$~1.62 & 32.16~$\pm$~2.87 & 50.32~$\pm$~4.33 \\
    Single-color & & 18.90~$\pm$~1.82 & 32.45~$\pm$~0.81 & 49.73~$\pm$~3.77 \\
    \midrule
    - & \multirow{4}{*}{60\%} & 2.29~$\pm$~0.25 & 6.43~$\pm$~0.71 & 12.57~$\pm$~2.75 \\
    Luminance & & 18.35~$\pm$~2.61 & 31.62~$\pm$~4.06 & 49.88~$\pm$~2.55 \\
    Color-opponency & & 17.45~$\pm$~1.50 & 32.22~$\pm$~2.83 & 50.05~$\pm$~4.14 \\
    Single-color & & 18.37~$\pm$~1.77 & 31.29~$\pm$~0.97 & 48.09~$\pm$~3.72 \\
    \midrule
    - & \multirow{4}{*}{70\%} & 1.64~$\pm$~0.31 & 5.27~$\pm$~0.39 & 9.64~$\pm$~4.50 \\
    Luminance & & 17.04~$\pm$~3.30 & 29.42~$\pm$~4.69 & 46.76~$\pm$~3.99 \\
    Color-opponency & & 18.06~$\pm$~1.19 & 33.05~$\pm$~2.42 & 50.52~$\pm$~3.42 \\
    Single-color & & 17.32~$\pm$~2.09 & 29.47~$\pm$~1.76 & 46.18~$\pm$~4.06 \\
    \midrule
    - & \multirow{4}{*}{80\%} & 1.24~$\pm$~0.29 & 4.15~$\pm$~1.15 & 8.40~$\pm$~4.82 \\
    Luminance & & 14.59~$\pm$~3.63 & 26.00~$\pm$~5.14 & 41.51~$\pm$~5.29 \\
    Color-opponency & & 18.26~$\pm$~1.41 & 32.54~$\pm$~2.44 & 51.78~$\pm$~4.08 \\
    Single-color & & 15.76~$\pm$~2.32 & 27.18~$\pm$~1.68 & 43.49~$\pm$~6.08 \\
    \midrule
    - & \multirow{4}{*}{90\%} & 1.40~$\pm$~0.43 & 4.02~$\pm$~1.59 & 11.81~$\pm$~7.06 \\
    Luminance & & 11.03~$\pm$~3.30 & 20.16~$\pm$~4.75 & 33.85~$\pm$~4.96 \\
    Color-opponency & & 17.04~$\pm$~1.23 & 29.55~$\pm$~1.89 & 49.72~$\pm$~2.05 \\
    Single-color & & 12.30~$\pm$~2.34 & 22.36~$\pm$~1.99 & 36.99~$\pm$~7.95 \\
    \bottomrule
  \end{tabular}
  }
  \label{tab:sparse_results_deeplabv3+_dark_zurich_zero_shot}
\end{table}
\clearpage

\begin{table}[tb]
  \centering
  \caption{\textbf{Zero-shot} evaluation of the \textbf{DeepLabv3+} architecture trained on Cityscapes without sparsity and \textbf{validated on ACDC Night} at different sparsities. Results include the three preprocessing types (luminance, color-opponency, and single-color). Metrics reported are mean Intersection over Union (mIoU), mean Accuracy (mAcc), and average Accuracy (aAcc), averaged over three seeds. \textbf{Percent-based sparsity:} For a given \textit{percentage}, the \textit{percentage} of lowest absolute values are set to zero.}
  \scalebox{0.85}{
  \begin{tabular}{@{}lc rrr@{}}
    \toprule
    Preprocessing &  Sparsity &\multicolumn{1}{c}{mIoU}&\multicolumn{1}{c}{mAcc}&\multicolumn{1}{c}{aAcc}  \\
    \midrule
    - & \multirow{4}{*}{-} &  8.01~$\pm$~0.45 & 18.14~$\pm$~2.24 & 31.94~$\pm$~2.20 \\
     Luminance &  & 19.76~$\pm$~0.53 & 34.13~$\pm$~1.05 & 58.90~$\pm$~1.33 \\
    Color-opponency&   &  18.05~$\pm$~1.55 & 31.73~$\pm$~2.54 & 53.51~$\pm$~3.20 \\
    Single-color &  & 19.75~$\pm$~1.99 & 32.24~$\pm$~1.17 & 54.79~$\pm$~2.95 \\
    \midrule
    - & \multirow{4}{*}{10\%} & 6.80~$\pm$~0.06 & 16.05~$\pm$~1.12 & 27.84~$\pm$~3.49 \\
    Luminance & & 19.77~$\pm$~0.55 & 34.14~$\pm$~1.05 & 58.90~$\pm$~1.34 \\
    Color-opponency & & 18.05~$\pm$~1.55 & 31.73~$\pm$~2.54 & 53.52~$\pm$~3.20 \\
    Single-color & & 19.75~$\pm$~1.99 & 32.23~$\pm$~1.15 & 54.79~$\pm$~2.94 \\
    \midrule
    - & \multirow{4}{*}{20\%} & 4.99~$\pm$~0.36 & 12.38~$\pm$~0.97 & 22.63~$\pm$~3.25 \\
    Luminance & & 19.90~$\pm$~0.59 & 34.05~$\pm$~1.22 & 58.98~$\pm$~1.22 \\
    Color-opponency & & 18.03~$\pm$~1.56 & 31.72~$\pm$~2.56 & 53.50~$\pm$~3.23 \\
    Single-color & & 19.75~$\pm$~1.98 & 32.15~$\pm$~1.14 & 54.82~$\pm$~2.89 \\
    \midrule
    - & \multirow{4}{*}{30\%} & 3.94~$\pm$~0.41 & 10.35~$\pm$~0.85 & 19.50~$\pm$~2.79 \\
    Luminance & & 19.98~$\pm$~0.80 & 33.64~$\pm$~1.59 & 58.41~$\pm$~0.81 \\
    Color-opponency & & 18.01~$\pm$~1.56 & 31.71~$\pm$~2.56 & 53.45~$\pm$~3.21 \\
    Single-color & & 19.79~$\pm$~2.00 & 32.01~$\pm$~1.25 & 54.92~$\pm$~2.97 \\
    \midrule
    - & \multirow{4}{*}{40\%} & 3.51~$\pm$~0.44 & 8.85~$\pm$~0.72 & 18.52~$\pm$~3.00 \\
    Luminance & & 19.91~$\pm$~1.05 & 32.86~$\pm$~1.86 & 57.41~$\pm$~0.84 \\
    Color-opponency & & 17.98~$\pm$~1.45 & 31.66~$\pm$~2.50 & 53.31~$\pm$~2.92 \\
    Single-color & & 19.81~$\pm$~1.88 & 31.85~$\pm$~1.25 & 54.65~$\pm$~3.02 \\
    \midrule
    - & \multirow{4}{*}{50\%} & 3.20~$\pm$~0.44 & 8.52~$\pm$~0.68 & 18.09~$\pm$~2.09 \\
    Luminance & & 19.58~$\pm$~1.43 & 31.76~$\pm$~2.21 & 56.03~$\pm$~1.40 \\
    Color-opponency & & 17.95~$\pm$~1.30 & 31.51~$\pm$~2.43 & 53.16~$\pm$~2.40 \\
    Single-color & & 19.46~$\pm$~1.90 & 31.15~$\pm$~1.30 & 53.41~$\pm$~2.82 \\
    \midrule
    - & \multirow{4}{*}{60\%} & 2.56~$\pm$~0.20 & 8.12~$\pm$~1.39 & 15.12~$\pm$~4.02 \\
    Luminance & & 18.80~$\pm$~2.12 & 30.36~$\pm$~2.95 & 53.52~$\pm$~2.66 \\
    Color-opponency & & 18.01~$\pm$~1.01 & 31.35~$\pm$~2.18 & 52.88~$\pm$~1.78 \\
    Single-color & & 18.87~$\pm$~2.04 & 30.13~$\pm$~1.65 & 51.66~$\pm$~3.29 \\
    \midrule
    - & \multirow{4}{*}{70\%} & 1.75~$\pm$~0.44 & 6.65~$\pm$~0.99 & 10.59~$\pm$~5.43 \\
    Luminance & & 17.30~$\pm$~2.70 & 28.26~$\pm$~3.41 & 49.73~$\pm$~4.57 \\
    Color-opponency & & 18.35~$\pm$~0.85 & 31.55~$\pm$~2.11 & 53.31~$\pm$~0.98 \\
    Single-color & & 17.57~$\pm$~2.50 & 28.25~$\pm$~2.23 & 49.17~$\pm$~4.42 \\
    \midrule
    - & \multirow{4}{*}{80\%} & 1.24~$\pm$~0.50 & 6.09~$\pm$~0.60 & 8.38~$\pm$~5.66 \\
    Luminance & & 14.68~$\pm$~3.09 & 24.78~$\pm$~3.84 & 43.28~$\pm$~6.25 \\
    Color-opponency & & 18.64~$\pm$~1.21 & 31.06~$\pm$~2.32 & 54.36~$\pm$~3.06 \\
    Single-color & & 15.64~$\pm$~2.90 & 25.85~$\pm$~2.36 & 44.35~$\pm$~6.42 \\
    \midrule
    - & \multirow{4}{*}{90\%} & 1.32~$\pm$~0.57 & 5.44~$\pm$~0.54 & 11.04~$\pm$~7.32 \\
    Luminance & & 10.57~$\pm$~2.73 & 18.73~$\pm$~3.55 & 34.60~$\pm$~5.59 \\
    Color-opponency & & 17.18~$\pm$~1.09 & 28.61~$\pm$~2.48 & 51.77~$\pm$~1.03 \\
    Single-color & & 11.67~$\pm$~2.72 & 20.83~$\pm$~2.23 & 35.60~$\pm$~7.31 \\
    \bottomrule
  \end{tabular}
  }
  \label{tab:sparse_results_deeplabv3+_acdc_night_zero_shot}
\end{table}
\clearpage

\begin{table}[tb]
  \centering
  \caption{\textbf{Zero-shot} evaluation of the \textbf{DeepLabv3+} architecture trained on Cityscapes without sparsity and \textbf{validated on ACDC Fog} at different sparsities. Results include the three preprocessing types (luminance, color-opponency, and single-color). Metrics reported are mean Intersection over Union (mIoU), mean Accuracy (mAcc), and average Accuracy (aAcc), averaged over three seeds. \textbf{Percent-based sparsity:} For a given \textit{percentage}, the \textit{percentage} of lowest absolute values are set to zero.} 
  \scalebox{0.85}{
  \begin{tabular}{@{}lc rrr@{}}
    \toprule
    Preprocessing &  Sparsity &\multicolumn{1}{c}{mIoU}&\multicolumn{1}{c}{mAcc}&\multicolumn{1}{c}{aAcc}  \\
    \midrule
    - & \multirow{4}{*}{-} &  39.20~$\pm$~4.35 & 54.93~$\pm$~6.11 & 76.52~$\pm$~6.46 \\
    Luminance & & 48.90~$\pm$~3.67 & 62.73~$\pm$~3.23 & 84.55~$\pm$~3.74 \\
    Color-opponency&   &  49.46~$\pm$~3.07 & 60.40~$\pm$~4.01 & 87.64~$\pm$~2.93 \\
    Single-color &  & 50.66~$\pm$~3.38 & 62.22~$\pm$~2.75 & 83.45~$\pm$~1.46 \\
    \midrule
    - & \multirow{4}{*}{10\%} & 37.58~$\pm$~4.44 & 53.00~$\pm$~6.06 & 74.55~$\pm$~6.46 \\
    Luminance & & 48.98~$\pm$~3.66 & 62.77~$\pm$~3.19 & 84.90~$\pm$~3.64 \\
    Color-opponency & & 49.47~$\pm$~3.07 & 60.40~$\pm$~4.01 & 87.65~$\pm$~2.92 \\
    Single-color & & 50.73~$\pm$~3.33 & 62.25~$\pm$~2.68 & 83.73~$\pm$~1.48 \\
    \midrule
    - & \multirow{4}{*}{20\%} & 29.22~$\pm$~4.22 & 43.33~$\pm$~5.87 & 65.96~$\pm$~5.08 \\
    Luminance & & 49.27~$\pm$~3.60 & 62.88~$\pm$~3.04 & 86.13~$\pm$~3.00 \\
    Color-opponency & & 49.46~$\pm$~3.08 & 60.38~$\pm$~4.01 & 87.68~$\pm$~2.90 \\
    Single-color & & 51.13~$\pm$~3.42 & 62.49~$\pm$~2.69 & 84.91~$\pm$~1.67 \\
    \midrule
    - & \multirow{4}{*}{30\%} & 20.02~$\pm$~4.32 & 31.54~$\pm$~5.90 & 54.54~$\pm$~3.55 \\
    Luminance & & 49.26~$\pm$~3.73 & 62.83~$\pm$~3.01 & 86.38~$\pm$~2.41 \\
    Color-opponency & & 49.34~$\pm$~3.06 & 60.21~$\pm$~4.00 & 87.78~$\pm$~2.83 \\
    Single-color & & 50.86~$\pm$~3.84 & 62.34~$\pm$~2.96 & 84.05~$\pm$~1.41 \\
    \midrule
    - & \multirow{4}{*}{40\%} & 13.11~$\pm$~2.23 & 22.87~$\pm$~3.16 & 47.84~$\pm$~4.09 \\
    Luminance & & 48.50~$\pm$~3.81 & 62.26~$\pm$~2.88 & 84.91~$\pm$~2.52 \\
    Color-opponency & & 49.15~$\pm$~2.95 & 59.88~$\pm$~3.86 & 88.10~$\pm$~2.57 \\
    Single-color & & 49.66~$\pm$~4.12 & 61.39~$\pm$~3.17 & 81.90~$\pm$~0.37 \\
    \midrule
    - & \multirow{4}{*}{50\%} & 9.93~$\pm$~2.74 & 18.20~$\pm$~3.17 & 43.93~$\pm$~5.42 \\
    Luminance & & 47.52~$\pm$~3.33 & 61.28~$\pm$~2.90 & 84.14~$\pm$~1.87 \\
    Color-opponency & & 48.87~$\pm$~2.75 & 59.40~$\pm$~3.59 & 88.33~$\pm$~2.35 \\
    Single-color & & 48.52~$\pm$~3.80 & 60.13~$\pm$~2.77 & 81.59~$\pm$~1.02 \\
    \midrule
    - & \multirow{4}{*}{60\%} & 7.42~$\pm$~1.50 & 14.61~$\pm$~1.78 & 40.01~$\pm$~6.63 \\
    Luminance & & 45.42~$\pm$~2.45 & 58.63~$\pm$~3.21 & 83.51~$\pm$~0.93 \\
    Color-opponency & & 48.69~$\pm$~2.68 & 58.92~$\pm$~3.50 & 88.64~$\pm$~2.11 \\
    Single-color & & 46.71~$\pm$~4.02 & 58.04~$\pm$~3.02 & 82.30~$\pm$~1.86 \\
    \midrule
    - & \multirow{4}{*}{70\%} & 5.80~$\pm$~1.58 & 11.48~$\pm$~1.36 & 32.55~$\pm$~8.77 \\
    Luminance & & 41.44~$\pm$~1.50 & 53.23~$\pm$~4.30 & 81.66~$\pm$~1.21 \\
    Color-opponency & & 48.05~$\pm$~2.89 & 58.07~$\pm$~3.48 & 87.80~$\pm$~2.91 \\
    Single-color & & 42.67~$\pm$~4.60 & 52.76~$\pm$~3.58 & 81.54~$\pm$~1.90 \\
    \midrule
    - & \multirow{4}{*}{80\%} & 3.12~$\pm$~1.47 & 7.63~$\pm$~1.70 & 21.51~$\pm$~11.87 \\
    Luminance & & 33.17~$\pm$~1.85 & 42.57~$\pm$~5.63 & 76.55~$\pm$~1.55 \\
    Color-opponency & & 44.65~$\pm$~3.21 & 54.69~$\pm$~3.48 & 84.13~$\pm$~4.51 \\
    Single-color & & 36.12~$\pm$~5.17 & 44.43~$\pm$~4.84 & 78.28~$\pm$~2.59 \\
    \midrule
    - & \multirow{4}{*}{90\%} & 2.17~$\pm$~1.16 & 6.52~$\pm$~1.05 & 17.17~$\pm$~10.79 \\
    Luminance & & 20.35~$\pm$~2.49 & 26.25~$\pm$~4.01 & 67.86~$\pm$~2.29 \\
    Color-opponency & & 37.71~$\pm$~4.23 & 46.65~$\pm$~5.32 & 78.77~$\pm$~4.40 \\
    Single-color & & 24.51~$\pm$~5.43 & 30.55~$\pm$~5.56 & 71.68~$\pm$~3.99 \\
    \bottomrule
  \end{tabular}
  }
  \label{tab:sparse_results_deeplabv3+_acdc_fog_zero_shot}
\end{table}
\clearpage

\begin{table}[tb]
  \centering
  \caption{\textbf{Zero-shot} evaluation of the \textbf{DeepLabv3+} architecture trained on Cityscapes without sparsity and \textbf{validated on ACDC Rain} at different sparsities. Results include the three preprocessing types (luminance, color-opponency, and single-color). Metrics reported are mean Intersection over Union (mIoU), mean Accuracy (mAcc), and average Accuracy (aAcc), averaged over three seeds. \textbf{Percent-based sparsity:} For a given \textit{percentage}, the \textit{percentage} of lowest absolute values are set to zero.} 
  \scalebox{0.85}{
  \begin{tabular}{@{}lc rrr@{}}
    \toprule
    Preprocessing &  Sparsity &\multicolumn{1}{c}{mIoU}&\multicolumn{1}{c}{mAcc}&\multicolumn{1}{c}{aAcc}  \\
    \midrule
    - & \multirow{4}{*}{-} &  33.31~$\pm$~2.15 & 48.13~$\pm$~4.57 & 77.43~$\pm$~3.46 \\
     Luminance &  & 34.20~$\pm$~2.50 & 47.68~$\pm$~4.67 & 76.72~$\pm$~3.07 \\
    Color-opponency&   &  38.87~$\pm$~2.24 & 51.26~$\pm$~3.57 & 83.44~$\pm$~3.63 \\
    Single-color &  & 38.60~$\pm$~3.04 & 51.81~$\pm$~2.65 & 77.82~$\pm$~1.63 \\
    \midrule
    - & \multirow{4}{*}{10\%} & 30.82~$\pm$~1.95 & 45.70~$\pm$~4.11 & 74.52~$\pm$~3.71 \\
    Luminance & & 34.29~$\pm$~2.49 & 47.73~$\pm$~4.66 & 77.10~$\pm$~3.05 \\
    Color-opponency & & 38.87~$\pm$~2.24 & 51.26~$\pm$~3.57 & 83.45~$\pm$~3.63 \\
    Single-color & & 38.67~$\pm$~3.03 & 51.86~$\pm$~2.63 & 78.09~$\pm$~1.70 \\
    \midrule
    - & \multirow{4}{*}{20\%} & 23.48~$\pm$~1.11 & 36.90~$\pm$~3.33 & 65.63~$\pm$~1.92 \\
    Luminance & & 34.83~$\pm$~2.44 & 48.03~$\pm$~4.66 & 79.24~$\pm$~2.73 \\
    Color-opponency & & 38.88~$\pm$~2.23 & 51.26~$\pm$~3.55 & 83.50~$\pm$~3.61 \\
    Single-color & & 39.12~$\pm$~3.04 & 52.16~$\pm$~2.62 & 79.74~$\pm$~2.09 \\
    \midrule
    - & \multirow{4}{*}{30\%} & 15.75~$\pm$~0.44 & 25.42~$\pm$~2.15 & 57.84~$\pm$~0.62 \\
    Luminance & & 35.55~$\pm$~2.26 & 48.49~$\pm$~4.68 & 81.97~$\pm$~1.91 \\
    Color-opponency & & 38.93~$\pm$~2.22 & 51.28~$\pm$~3.57 & 83.69~$\pm$~3.52 \\
    Single-color & & 39.65~$\pm$~3.20 & 52.50~$\pm$~2.73 & 81.75~$\pm$~2.14 \\
    \midrule
    - & \multirow{4}{*}{40\%} & 10.61~$\pm$~0.90 & 18.84~$\pm$~0.73 & 51.69~$\pm$~2.31 \\
    Luminance & & 35.32~$\pm$~2.23 & 48.08~$\pm$~4.87 & 81.84~$\pm$~1.70 \\
    Color-opponency & & 39.03~$\pm$~2.19 & 51.27~$\pm$~3.60 & 84.23~$\pm$~3.20 \\
    Single-color & & 39.52~$\pm$~3.30 & 52.35~$\pm$~2.86 & 81.73~$\pm$~1.29 \\
    \midrule
    - & \multirow{4}{*}{50\%} & 8.21~$\pm$~0.59 & 15.41~$\pm$~0.67 & 45.80~$\pm$~4.15 \\
    Luminance & & 35.10~$\pm$~2.37 & 47.18~$\pm$~5.26 & 82.45~$\pm$~1.41 \\
    Color-opponency & & 39.07~$\pm$~2.14 & 51.17~$\pm$~3.64 & 84.78~$\pm$~2.76 \\
    Single-color & & 39.27~$\pm$~3.29 & 51.63~$\pm$~2.58 & 82.88~$\pm$~1.30 \\
    \midrule
    - & \multirow{4}{*}{60\%} & 6.45~$\pm$~0.62 & 12.33~$\pm$~0.49 & 38.30~$\pm$~6.78 \\
    Luminance & & 33.91~$\pm$~2.40 & 44.71~$\pm$~5.00 & 82.10~$\pm$~0.80 \\
    Color-opponency & & 39.07~$\pm$~1.83 & 50.82~$\pm$~3.33 & 85.81~$\pm$~1.86 \\
    Single-color & & 37.83~$\pm$~2.93 & 49.34~$\pm$~2.06 & 83.53~$\pm$~1.05 \\
    \midrule
    - & \multirow{4}{*}{70\%} & 4.58~$\pm$~1.06 & 9.46~$\pm$~1.27 & 27.54~$\pm$~10.54 \\
    Luminance & & 31.08~$\pm$~2.77 & 40.10~$\pm$~5.44 & 80.12~$\pm$~0.53 \\
    Color-opponency & & 37.97~$\pm$~1.99 & 49.60~$\pm$~3.45 & 85.11~$\pm$~2.21 \\
    Single-color & & 34.99~$\pm$~2.64 & 45.63~$\pm$~2.78 & 82.74~$\pm$~1.29 \\
    \midrule
    - & \multirow{4}{*}{80\%} & 2.96~$\pm$~1.34 & 7.67~$\pm$~2.15 & 18.19~$\pm$~12.30 \\
    Luminance & & 25.69~$\pm$~3.37 & 32.55~$\pm$~5.41 & 75.68~$\pm$~0.69 \\
    Color-opponency & & 35.28~$\pm$~1.86 & 46.76~$\pm$~3.42 & 82.90~$\pm$~2.83 \\
    Single-color & & 29.46~$\pm$~3.19 & 39.24~$\pm$~4.56 & 78.47~$\pm$~2.95 \\
    \midrule
    - & \multirow{4}{*}{90\%} & 2.15~$\pm$~1.37 & 5.53~$\pm$~2.03 & 14.84~$\pm$~12.56 \\
    Luminance & & 18.91~$\pm$~3.39 & 24.72~$\pm$~4.91 & 69.12~$\pm$~1.88 \\
    Color-opponency & & 29.41~$\pm$~2.47 & 39.73~$\pm$~5.02 & 78.09~$\pm$~3.27 \\
    Single-color & & 22.37~$\pm$~3.18 & 30.72~$\pm$~4.66 & 72.50~$\pm$~3.61 \\
    \bottomrule
  \end{tabular}
  }
  \label{tab:sparse_results_deeplabv3+_acdc_rain_zero_shot}
\end{table}
\clearpage

\begin{table}[tb]
  \centering
  \caption{\textbf{Zero-shot} evaluation of the \textbf{DeepLabv3+} architecture trained on Cityscapes without sparsity and \textbf{validated on ACDC Snow} at different sparsities. Results include the three preprocessing types (luminance, color-opponency, and single-color). Metrics reported are mean Intersection over Union (mIoU), mean Accuracy (mAcc), and average Accuracy (aAcc), averaged over three seeds. \textbf{Percent-based sparsity:} For a given \textit{percentage}, the \textit{percentage} of lowest absolute values are set to zero.} 
  \scalebox{0.85}{
  \begin{tabular}{@{}lc rrr@{}}
    \toprule
    Preprocessing &  Sparsity &\multicolumn{1}{c}{mIoU}&\multicolumn{1}{c}{mAcc}&\multicolumn{1}{c}{aAcc}  \\
    \midrule
    - & \multirow{4}{*}{-} &  24.18~$\pm$~3.60 & 36.05~$\pm$~4.01 & 64.03~$\pm$~10.17 \\
     Luminance &  & 31.58~$\pm$~1.88 & 45.09~$\pm$~3.17 & 70.05~$\pm$~3.19 \\
    Color-opponency&   &  32.97~$\pm$~2.97 & 44.40~$\pm$~3.82 & 76.43~$\pm$~5.42 \\
    Single-color &  & 32.60~$\pm$~3.50 & 44.66~$\pm$~2.88 & 69.89~$\pm$~1.47 \\
    \midrule
    - & \multirow{4}{*}{10\%} & 23.45~$\pm$~3.57 & 35.15~$\pm$~4.06 & 63.14~$\pm$~10.31 \\
    Luminance & & 31.75~$\pm$~1.87 & 45.21~$\pm$~3.14 & 70.72~$\pm$~3.13 \\
    Color-opponency & & 32.97~$\pm$~2.96 & 44.41~$\pm$~3.83 & 76.43~$\pm$~5.42 \\
    Single-color & & 32.73~$\pm$~3.49 & 44.76~$\pm$~2.86 & 70.42~$\pm$~1.43 \\
    \midrule
    - & \multirow{4}{*}{20\%} & 20.38~$\pm$~3.46 & 31.58~$\pm$~4.12 & 59.20~$\pm$~10.11 \\
    Luminance & & 32.55~$\pm$~1.80 & 45.74~$\pm$~3.03 & 73.90~$\pm$~2.62 \\
    Color-opponency & & 33.01~$\pm$~2.97 & 44.43~$\pm$~3.83 & 76.51~$\pm$~5.40 \\
    Single-color & & 33.43~$\pm$~3.59 & 45.29~$\pm$~2.78 & 73.33~$\pm$~1.99 \\
    \midrule
    - & \multirow{4}{*}{30\%} & 16.12~$\pm$~2.78 & 26.44~$\pm$~3.67 & 52.63~$\pm$~8.04 \\
    Luminance & & 32.71~$\pm$~1.64 & 45.78~$\pm$~2.89 & 74.86~$\pm$~1.85 \\
    Color-opponency & & 33.07~$\pm$~2.93 & 44.47~$\pm$~3.80 & 76.71~$\pm$~5.32 \\
    Single-color & & 33.59~$\pm$~3.72 & 45.43~$\pm$~2.87 & 74.17~$\pm$~2.04 \\
    \midrule
    - & \multirow{4}{*}{40\%} & 11.99~$\pm$~2.01 & 21.02~$\pm$~2.63 & 47.71~$\pm$~5.26 \\
    Luminance & & 32.58~$\pm$~1.28 & 45.45~$\pm$~2.81 & 75.03~$\pm$~1.12 \\
    Color-opponency & & 33.22~$\pm$~2.82 & 44.52~$\pm$~3.71 & 77.39~$\pm$~4.97 \\
    Single-color & & 33.81~$\pm$~3.51 & 45.47~$\pm$~2.82 & 75.32~$\pm$~1.67 \\
    \midrule
    - & \multirow{4}{*}{50\%} & 9.05~$\pm$~1.12 & 16.38~$\pm$~1.32 & 44.19~$\pm$~4.07 \\
    Luminance & & 32.20~$\pm$~1.09 & 44.48~$\pm$~2.94 & 75.45~$\pm$~0.45 \\
    Color-opponency & & 33.36~$\pm$~2.77 & 44.49~$\pm$~3.66 & 78.07~$\pm$~4.56 \\
    Single-color & & 33.75~$\pm$~3.11 & 45.01~$\pm$~2.98 & 76.42~$\pm$~1.34 \\
    \midrule
    - & \multirow{4}{*}{60\%} & 7.11~$\pm$~0.59 & 13.06~$\pm$~0.84 & 39.88~$\pm$~5.15 \\
    Luminance & & 30.83~$\pm$~1.43 & 42.55~$\pm$~3.47 & 74.53~$\pm$~0.66 \\
    Color-opponency & & 33.53~$\pm$~2.53 & 44.34~$\pm$~3.42 & 79.46~$\pm$~3.60 \\
    Single-color & & 32.76~$\pm$~3.31 & 43.66~$\pm$~3.86 & 76.52~$\pm$~0.71 \\
    \midrule
    - & \multirow{4}{*}{70\%} & 5.36~$\pm$~0.69 & 10.72~$\pm$~0.50 & 33.24~$\pm$~8.26 \\
    Luminance & & 28.59~$\pm$~1.97 & 39.40~$\pm$~4.45 & 72.60~$\pm$~0.75 \\
    Color-opponency & & 33.20~$\pm$~2.77 & 44.00~$\pm$~3.64 & 79.04~$\pm$~3.82 \\
    Single-color & & 30.59~$\pm$~2.90 & 40.88~$\pm$~4.09 & 75.54~$\pm$~1.39 \\
    \midrule
    - & \multirow{4}{*}{80\%} & 4.00~$\pm$~1.00 & 9.17~$\pm$~0.76 & 25.48~$\pm$~11.52 \\
    Luminance & & 24.55~$\pm$~2.76 & 33.74~$\pm$~5.81 & 68.51~$\pm$~1.15 \\
    Color-opponency & & 32.10~$\pm$~2.61 & 43.09~$\pm$~3.73 & 77.59~$\pm$~3.37 \\
    Single-color & & 26.94~$\pm$~3.17 & 36.47~$\pm$~4.96 & 72.39~$\pm$~3.01 \\
    \midrule
    - & \multirow{4}{*}{90\%} & 2.66~$\pm$~1.19 & 7.53~$\pm$~0.77 & 18.49~$\pm$~12.45 \\
    Luminance & & 17.99~$\pm$~3.26 & 24.99~$\pm$~5.50 & 62.95~$\pm$~2.96 \\
    Color-opponency & & 27.54~$\pm$~3.64 & 37.46~$\pm$~5.85 & 72.38~$\pm$~4.51 \\
    Single-color & & 20.44~$\pm$~2.92 & 28.91~$\pm$~5.88 & 67.22~$\pm$~2.92 \\
    \bottomrule
  \end{tabular}
  }
  \label{tab:sparse_results_deeplabv3+_acdc_snow_zero_shot}
\end{table}
\clearpage

\begin{table}[tb]
  \centering
  \caption{\textbf{Zero-shot} evaluation of the \textbf{DeepLabv3+} architecture trained on Cityscapes without sparsity and \textbf{validated on ACDC Mean} at different sparsities. Results include the three preprocessing types (luminance, color-opponency, and single-color). Metrics reported are mean Intersection over Union (mIoU), mean Accuracy (mAcc), and average Accuracy (aAcc), averaged over three seeds. \textbf{Percent-based sparsity:} For a given \textit{percentage}, the \textit{percentage} of lowest absolute values are set to zero.}
  \scalebox{0.85}{
  \begin{tabular}{@{}lc rrr@{}}
    \toprule
    Preprocessing & Sparsity &\multicolumn{1}{c}{mIoU}&\multicolumn{1}{c}{mAcc}&\multicolumn{1}{c}{aAcc}  \\
    \midrule
     - &\multirow{4}{*}{-}& 23.59~$\pm$~1.66 & 39.06~$\pm$~4.04 & 62.24~$\pm$~4.48\\
    Luminance & & 32.80~$\pm$~1.78 & 45.42~$\pm$~2.99 & 72.45~$\pm$~2.77 \\
    Color-opponency &  & 33.74~$\pm$~1.99 & 45.16~$\pm$~3.12 & 75.08~$\pm$~2.96 \\
    Single-color &  & 34.63~$\pm$~3.00 & 45.87~$\pm$~2.19 & 71.36~$\pm$~0.79 \\
    \midrule
    - & \multirow{4}{*}{10\%} & 21.90~$\pm$~1.39 & 37.29~$\pm$~3.81 & 59.76~$\pm$~4.21 \\
    Luminance & & 32.90~$\pm$~1.77 & 45.48~$\pm$~2.98 & 72.80~$\pm$~2.73 \\
    Color-opponency & & 33.75~$\pm$~1.99 & 45.16~$\pm$~3.12 & 75.09~$\pm$~2.96 \\
    Single-color & & 34.70~$\pm$~3.00 & 45.92~$\pm$~2.18 & 71.63~$\pm$~0.89 \\
    \midrule
    - & \multirow{4}{*}{20\%} & 17.49~$\pm$~1.08 & 31.27~$\pm$~3.51 & 53.11~$\pm$~3.22 \\
    Luminance & & 33.41~$\pm$~1.74 & 45.75~$\pm$~2.95 & 74.43~$\pm$~2.35 \\
    Color-opponency & & 33.75~$\pm$~1.98 & 45.16~$\pm$~3.12 & 75.13~$\pm$~2.94 \\
    Single-color & & 35.10~$\pm$~3.04 & 46.19~$\pm$~2.19 & 73.06~$\pm$~1.50 \\
    \midrule
    - & \multirow{4}{*}{30\%} & 12.75~$\pm$~1.01 & 24.10~$\pm$~2.88 & 45.92~$\pm$~2.26 \\
    Luminance & & 33.71~$\pm$~1.67 & 45.78~$\pm$~2.95 & 75.27~$\pm$~1.70 \\
    Color-opponency & & 33.75~$\pm$~1.97 & 45.12~$\pm$~3.10 & 75.24~$\pm$~2.87 \\
    Single-color & & 35.22~$\pm$~3.22 & 46.22~$\pm$~2.37 & 73.58~$\pm$~1.58 \\
    \midrule
    - & \multirow{4}{*}{40\%} & 9.33~$\pm$~0.87 & 18.93~$\pm$~1.59 & 41.26~$\pm$~2.32 \\
    Luminance & & 33.49~$\pm$~1.56 & 45.26~$\pm$~2.94 & 74.66~$\pm$~1.43 \\
    Color-opponency & & 33.78~$\pm$~1.91 & 45.03~$\pm$~3.04 & 75.58~$\pm$~2.63 \\
    Single-color & & 35.00~$\pm$~3.19 & 45.91~$\pm$~2.42 & 73.25~$\pm$~1.17 \\
    \midrule
    - & \multirow{4}{*}{50\%} & 7.38~$\pm$~0.55 & 15.62~$\pm$~0.89 & 37.84~$\pm$~3.57 \\
    Luminance & & 33.11~$\pm$~1.42 & 44.22~$\pm$~3.18 & 74.37~$\pm$~0.83 \\
    Color-opponency & & 33.79~$\pm$~1.87 & 44.88~$\pm$~2.95 & 75.91~$\pm$~2.38 \\
    Single-color & & 34.53~$\pm$~2.99 & 45.17~$\pm$~2.40 & 73.42~$\pm$~1.16 \\
    \midrule
    - & \multirow{4}{*}{60\%} & 5.94~$\pm$~0.42 & 12.64~$\pm$~0.75 & 33.18~$\pm$~5.60 \\
    Luminance & & 31.89~$\pm$~1.58 & 42.11~$\pm$~3.68 & 73.25~$\pm$~0.25 \\
    Color-opponency & & 33.82~$\pm$~1.77 & 44.59~$\pm$~2.76 & 76.51~$\pm$~1.87 \\
    Single-color & & 33.40~$\pm$~3.00 & 43.46~$\pm$~2.70 & 73.33~$\pm$~1.15 \\
    \midrule
    - & \multirow{4}{*}{70\%} & 4.47~$\pm$~0.75 & 9.77~$\pm$~0.85 & 25.86~$\pm$~8.19 \\
    Luminance & & 29.30~$\pm$~2.11 & 38.44~$\pm$~4.41 & 70.86~$\pm$~0.75 \\
    Color-opponency & & 33.50~$\pm$~1.99 & 43.97~$\pm$~2.91 & 76.14~$\pm$~2.18 \\
    Single-color & & 30.93~$\pm$~3.07 & 40.14~$\pm$~3.06 & 72.07~$\pm$~1.98 \\
    \midrule
    - & \multirow{4}{*}{80\%} & 2.94~$\pm$~1.07 & 7.47~$\pm$~0.88 & 18.31~$\pm$~10.28 \\
    Luminance & & 24.32~$\pm$~2.84 & 32.13~$\pm$~5.12 & 65.83~$\pm$~1.13 \\
    Color-opponency & & 32.18~$\pm$~1.95 & 42.12~$\pm$~2.82 & 74.58~$\pm$~2.61 \\
    Single-color & & 26.66~$\pm$~3.47 & 35.09~$\pm$~3.79 & 68.18~$\pm$~3.73 \\
    \midrule
    - & \multirow{4}{*}{90\%} & 2.11~$\pm$~1.16 & 6.35~$\pm$~0.90 & 15.35~$\pm$~10.68 \\
    Luminance & & 17.03~$\pm$~3.03 & 23.30~$\pm$~4.47 & 58.45~$\pm$~1.35 \\
    Color-opponency & & 27.72~$\pm$~2.54 & 36.47~$\pm$~4.09 & 70.11~$\pm$~2.80 \\
    Single-color & & 19.69~$\pm$~3.30 & 27.27~$\pm$~4.36 & 61.54~$\pm$~4.36 \\
    \bottomrule
  \end{tabular}
  }
  \label{tab:sparse_results_deeplabv3plus_acdc_full_zero_shot}
\end{table}
\clearpage

%% file: tables/sparsity_zero_shot_mask2former.tex
\begin{table}[h]
  \centering
  \caption{\textbf{Zero-shot} evaluation of the \textbf{Mask2Former} architecture trained on Cityscapes without sparsity and \textbf{validated on Cityscapes} at different sparsities. Results include the three preprocessing types (luminance, color-opponency, and single-color). Metrics reported are mean Intersection over Union (mIoU), mean Accuracy (mAcc), and average Accuracy (aAcc), averaged over three seeds. \textbf{Percent-based sparsity:} For a given \textit{percentage}, the \textit{percentage} of lowest absolute values are set to zero.}   
  \scalebox{0.85}{
  \begin{tabular}{@{}lc rrr@{}}
    \toprule
    Preprocessing &  Sparsity &\multicolumn{1}{c}{mIoU}&\multicolumn{1}{c}{mAcc}&\multicolumn{1}{c}{aAcc}  \\
    \midrule
    - & \multirow{4}{*}{-} &  77.43~$\pm$~0.35 & 86.83~$\pm$~0.19 & 96.10~$\pm$~0.05 \\
     Luminance  &  & 73.93~$\pm$~0.80 & 85.33~$\pm$~0.92 & 95.59~$\pm$~0.13 \\
    Color-opponency & &73.47~$\pm$~0.47 & 84.86~$\pm$~0.57 & 95.54~$\pm$~0.11 \\
    Single-color & &74.02~$\pm$~0.62 & 84.64~$\pm$~0.85 & 95.75~$\pm$~0.02 \\
    \midrule
    - & \multirow{4}{*}{10\%} & 74.94~$\pm$~1.70 & 85.20~$\pm$~1.29 & 95.67~$\pm$~0.09 \\
    Luminance & & 73.92~$\pm$~0.83 & 85.33~$\pm$~0.94 & 95.59~$\pm$~0.13 \\
    Color-opponency & & 73.47~$\pm$~0.46 & 84.88~$\pm$~0.56 & 95.54~$\pm$~0.11 \\
    Single-color & & 74.16~$\pm$~0.31 & 84.78~$\pm$~0.59 & 95.75~$\pm$~0.03 \\
    \midrule
    - & \multirow{4}{*}{20\%} & 72.52~$\pm$~1.74 & 84.52~$\pm$~0.66 & 94.64~$\pm$~0.29 \\
    Luminance & & 74.00~$\pm$~0.82 & 85.39~$\pm$~0.88 & 95.57~$\pm$~0.13 \\
    Color-opponency & & 73.46~$\pm$~0.46 & 84.86~$\pm$~0.56 & 95.54~$\pm$~0.11 \\
    Single-color & & 74.18~$\pm$~0.40 & 84.76~$\pm$~0.59 & 95.75~$\pm$~0.02 \\
    \midrule
    - & \multirow{4}{*}{30\%} & 65.14~$\pm$~3.37 & 81.57~$\pm$~0.78 & 92.17~$\pm$~1.18 \\
    Luminance & & 73.78~$\pm$~0.63 & 85.25~$\pm$~0.74 & 95.51~$\pm$~0.11 \\
    Color-opponency & & 73.03~$\pm$~0.33 & 84.52~$\pm$~0.08 & 95.52~$\pm$~0.14 \\
    Single-color & & 73.96~$\pm$~0.47 & 84.62~$\pm$~0.73 & 95.71~$\pm$~0.01 \\
    \midrule
    - & \multirow{4}{*}{40\%} & 58.81~$\pm$~2.95 & 77.28~$\pm$~1.12 & 88.27~$\pm$~1.51 \\
    Luminance & & 73.56~$\pm$~0.68 & 84.98~$\pm$~0.66 & 95.43~$\pm$~0.11 \\
    Color-opponency & & 72.96~$\pm$~0.31 & 84.50~$\pm$~0.12 & 95.51~$\pm$~0.14 \\
    Single-color & & 73.75~$\pm$~0.19 & 84.53~$\pm$~0.57 & 95.65~$\pm$~0.02 \\
    \midrule
    - & \multirow{4}{*}{50\%} & 50.36~$\pm$~2.42 & 68.62~$\pm$~2.73 & 81.81~$\pm$~1.50 \\
    Luminance & & 72.61~$\pm$~0.30 & 84.09~$\pm$~0.13 & 95.24~$\pm$~0.07 \\
    Color-opponency & & 72.71~$\pm$~0.65 & 84.36~$\pm$~0.27 & 95.48~$\pm$~0.16 \\
    Single-color & & 73.21~$\pm$~0.33 & 84.15~$\pm$~0.69 & 95.49~$\pm$~0.04 \\
    \midrule
    - & \multirow{4}{*}{60\%} & 39.41~$\pm$~1.06 & 54.93~$\pm$~0.47 & 71.38~$\pm$~2.53 \\
    Luminance & & 70.75~$\pm$~0.33 & 82.61~$\pm$~0.30 & 94.80~$\pm$~0.04 \\
    Color-opponency & & 72.97~$\pm$~0.12 & 84.65~$\pm$~0.32 & 95.47~$\pm$~0.12 \\
    Single-color & & 71.62~$\pm$~0.63 & 82.84~$\pm$~0.83 & 95.16~$\pm$~0.03 \\
    \midrule
    - & \multirow{4}{*}{70\%} & 25.74~$\pm$~1.31 & 37.17~$\pm$~0.48 & 56.72~$\pm$~3.10 \\
    Luminance & & 67.67~$\pm$~0.38 & 79.70~$\pm$~0.25 & 93.95~$\pm$~0.08 \\
    Color-opponency & & 72.20~$\pm$~0.93 & 84.11~$\pm$~0.42 & 95.38~$\pm$~0.12 \\
    Single-color & & 68.81~$\pm$~0.91 & 80.45~$\pm$~1.35 & 94.51~$\pm$~0.03 \\
    \midrule
    - & \multirow{4}{*}{80\%} & 13.83~$\pm$~0.41 & 21.74~$\pm$~1.26 & 43.10~$\pm$~2.35 \\
    Luminance & & 59.32~$\pm$~0.42 & 71.35~$\pm$~0.16 & 91.33~$\pm$~0.25 \\
    Color-opponency & & 70.92~$\pm$~0.28 & 83.22~$\pm$~0.42 & 95.04~$\pm$~0.03 \\
    Single-color & & 62.87~$\pm$~0.59 & 74.96~$\pm$~1.54 & 92.82~$\pm$~0.12 \\
    \midrule
    - & \multirow{4}{*}{90\%} & 9.12~$\pm$~0.12 & 15.85~$\pm$~1.22 & 39.65~$\pm$~2.68 \\
    Luminance & & 43.26~$\pm$~2.66 & 54.75~$\pm$~2.19 & 82.14~$\pm$~1.86 \\
    Color-opponency & & 64.29~$\pm$~0.94 & 76.95~$\pm$~0.94 & 93.18~$\pm$~0.06 \\
    Single-color & & 48.23~$\pm$~1.28 & 60.23~$\pm$~0.99 & 86.53~$\pm$~1.20 \\
    \bottomrule
  \end{tabular}
  }
  \label{tab:sparse_results_mask2Former_cityscapes_zero_shot}
\end{table}
\clearpage

\begin{table}[tb]
  \centering
  \caption{\textbf{Zero-shot} evaluation of the \textbf{Mask2Former} architecture trained on Cityscapes without sparsity and \textbf{validated on Dark Zurich} at different sparsities. Results include the three preprocessing types (luminance, color-opponency, and single-color). Metrics reported are mean Intersection over Union (mIoU), mean Accuracy (mAcc), and average Accuracy (aAcc), averaged over three seeds. \textbf{Percent-based sparsity:} For a given \textit{percentage}, the \textit{percentage} of lowest absolute values are set to zero.}   
  \scalebox{0.85}{
  \begin{tabular}{@{}lc rrr@{}}
    \toprule
    Preprocessing &  Sparsity &\multicolumn{1}{c}{mIoU}&\multicolumn{1}{c}{mAcc}&\multicolumn{1}{c}{aAcc}  \\
    \midrule
    - & \multirow{4}{*}{-} &  18.16~$\pm$~1.14 & 28.88~$\pm$~2.72 & 47.34~$\pm$~4.13 \\
     Luminance  &  &  26.21~$\pm$~0.18 & 40.56~$\pm$~2.23 & 59.16~$\pm$~0.58 \\
    Color-opponency & & 22.84~$\pm$~1.82 & 37.37~$\pm$~2.58 & 55.26~$\pm$~2.45 \\
    Single-color & & 24.92~$\pm$~1.09 & 39.64~$\pm$~2.37 & 57.09~$\pm$~3.52 \\
    \midrule
    - & \multirow{4}{*}{10\%} & 13.83~$\pm$~1.61 & 23.75~$\pm$~2.93 & 40.33~$\pm$~4.26 \\
    Luminance & & 26.19~$\pm$~0.17 & 40.53~$\pm$~2.14 & 59.06~$\pm$~0.53 \\
    Color-opponency & & 22.87~$\pm$~1.81 & 37.40~$\pm$~2.60 & 55.26~$\pm$~2.44 \\
    Single-color & & 24.85~$\pm$~1.24 & 39.67~$\pm$~2.37 & 57.14~$\pm$~3.58 \\
    \midrule
    - & \multirow{4}{*}{20\%} & 7.79~$\pm$~0.81 & 15.41~$\pm$~2.12 & 26.95~$\pm$~3.86 \\
    Luminance & & 26.05~$\pm$~0.14 & 40.74~$\pm$~1.77 & 59.09~$\pm$~0.26 \\
    Color-opponency & & 22.88~$\pm$~1.80 & 37.39~$\pm$~2.58 & 55.23~$\pm$~2.45 \\
    Single-color & & 25.05~$\pm$~1.21 & 39.88~$\pm$~2.61 & 57.52~$\pm$~3.48 \\
    \midrule
    - & \multirow{4}{*}{30\%} & 5.04~$\pm$~1.51 & 10.25~$\pm$~2.91 & 19.50~$\pm$~2.81 \\
    Luminance & & 26.63~$\pm$~0.35 & 41.29~$\pm$~1.67 & 59.19~$\pm$~0.98 \\
    Color-opponency & & 22.88~$\pm$~1.72 & 37.29~$\pm$~2.53 & 55.20~$\pm$~2.47 \\
    Single-color & & 25.18~$\pm$~0.97 & 39.97~$\pm$~2.54 & 57.51~$\pm$~3.45 \\
    \midrule
    - & \multirow{4}{*}{40\%} & 3.48~$\pm$~0.33 & 8.40~$\pm$~0.88 & 21.36~$\pm$~1.67 \\
    Luminance & & 26.85~$\pm$~0.13 & 42.01~$\pm$~0.83 & 59.33~$\pm$~1.37 \\
    Color-opponency & & 22.98~$\pm$~1.82 & 37.22~$\pm$~2.57 & 55.26~$\pm$~2.51 \\
    Single-color & & 25.10~$\pm$~1.01 & 39.91~$\pm$~2.45 & 57.91~$\pm$~3.48 \\
    \midrule
    - & \multirow{4}{*}{50\%} & 3.96~$\pm$~0.57 & 9.44~$\pm$~1.15 & 24.85~$\pm$~4.07 \\
    Luminance & & 26.76~$\pm$~0.79 & 41.85~$\pm$~1.37 & 59.25~$\pm$~2.00 \\
    Color-opponency & & 22.97~$\pm$~1.88 & 37.37~$\pm$~2.51 & 55.32~$\pm$~2.63 \\
    Single-color & & 26.07~$\pm$~1.04 & 40.80~$\pm$~2.53 & 59.32~$\pm$~3.72 \\
    \midrule
    - & \multirow{4}{*}{60\%} & 4.55~$\pm$~0.83 & 10.56~$\pm$~2.14 & 28.07~$\pm$~4.81 \\
    Luminance & & 27.18~$\pm$~1.09 & 42.29~$\pm$~1.06 & 60.40~$\pm$~4.53 \\
    Color-opponency & & 23.04~$\pm$~1.94 & 37.56~$\pm$~2.37 & 55.50~$\pm$~2.64 \\
    Single-color & & 25.60~$\pm$~1.50 & 40.68~$\pm$~3.58 & 60.02~$\pm$~3.87 \\
    \midrule
    - & \multirow{4}{*}{70\%} & 5.41~$\pm$~0.74 & 10.90~$\pm$~1.61 & 29.87~$\pm$~2.86 \\
    Luminance & & 26.99~$\pm$~2.30 & 42.28~$\pm$~2.89 & 62.33~$\pm$~7.31 \\
    Color-opponency & & 23.33~$\pm$~1.51 & 38.38~$\pm$~2.00 & 56.18~$\pm$~1.80 \\
    Single-color & & 25.54~$\pm$~1.22 & 40.75~$\pm$~2.62 & 60.40~$\pm$~4.41 \\
    \midrule
    - & \multirow{4}{*}{80\%} & 5.89~$\pm$~0.17 & 11.45~$\pm$~0.72 & 33.24~$\pm$~1.74 \\
    Luminance & & 24.92~$\pm$~0.70 & 40.67~$\pm$~2.24 & 60.82~$\pm$~4.94 \\
    Color-opponency & & 23.38~$\pm$~1.06 & 39.13~$\pm$~0.66 & 56.21~$\pm$~1.64 \\
    Single-color & & 24.73~$\pm$~1.05 & 39.21~$\pm$~2.25 & 59.46~$\pm$~2.54 \\
    \midrule
    - & \multirow{4}{*}{90\%} & 6.08~$\pm$~0.94 & 12.01~$\pm$~0.63 & 35.23~$\pm$~4.81 \\
    Luminance & & 17.67~$\pm$~1.79 & 32.38~$\pm$~1.46 & 47.49~$\pm$~3.98 \\
    Color-opponency & & 22.87~$\pm$~1.29 & 38.10~$\pm$~1.18 & 54.80~$\pm$~1.30 \\
    Single-color & & 19.90~$\pm$~1.80 & 33.11~$\pm$~1.05 & 50.59~$\pm$~7.18 \\
    \bottomrule
  \end{tabular}
  }
  \label{tab:sparse_results_mask2Former_dark_zurich_zero_shot}
\end{table}
\clearpage

\begin{table}[tb]
  \centering
  \caption{\textbf{Zero-shot} evaluation of the \textbf{Mask2Former} architecture trained on Cityscapes without sparsity and \textbf{validated on ACDC Night} at different sparsities. Results include the three preprocessing types (luminance, color-opponency, and single-color). Metrics reported are mean Intersection over Union (mIoU), mean Accuracy (mAcc), and average Accuracy (aAcc), averaged over three seeds. \textbf{Percent-based sparsity:} For a given \textit{percentage}, the \textit{percentage} of lowest absolute values are set to zero.}   
  \scalebox{0.85}{
  \begin{tabular}{@{}lc rrr@{}}
    \toprule
    Preprocessing &  Sparsity &\multicolumn{1}{c}{mIoU}&\multicolumn{1}{c}{mAcc}&\multicolumn{1}{c}{aAcc}  \\
    \midrule
    - & \multirow{4}{*}{-} &  19.65~$\pm$~1.22 & 32.32~$\pm$~1.80 & 51.57~$\pm$~2.56 \\
     Luminance  &  &  28.45~$\pm$~1.18 & 42.50~$\pm$~2.86 & 62.43~$\pm$~0.83 \\
    Color-opponency & & 26.08~$\pm$~2.72 & 41.45~$\pm$~3.98 & 58.44~$\pm$~3.31 \\
    Single-color & &26.64~$\pm$~0.26 & 42.38~$\pm$~1.67 & 60.03~$\pm$~2.94 \\
    \midrule
    - & \multirow{4}{*}{10\%} & 14.35~$\pm$~1.63 & 26.18~$\pm$~1.94 & 43.53~$\pm$~3.29 \\
    Luminance & & 28.56~$\pm$~1.32 & 42.62~$\pm$~3.00 & 62.48~$\pm$~0.79 \\
    Color-opponency & & 26.11~$\pm$~2.69 & 41.46~$\pm$~3.95 & 58.43~$\pm$~3.31 \\
    Single-color & & 26.63~$\pm$~0.19 & 42.32~$\pm$~1.74 & 60.04~$\pm$~2.93 \\
    \midrule
    - & \multirow{4}{*}{20\%} & 8.44~$\pm$~1.16 & 16.93~$\pm$~1.50 & 30.71~$\pm$~1.95 \\
    Luminance & & 28.51~$\pm$~1.31 & 42.41~$\pm$~2.59 & 62.63~$\pm$~0.39 \\
    Color-opponency & & 26.10~$\pm$~2.64 & 41.43~$\pm$~3.91 & 58.41~$\pm$~3.28 \\
    Single-color & & 26.95~$\pm$~0.52 & 42.48~$\pm$~2.24 & 60.40~$\pm$~3.10 \\
    \midrule
    - & \multirow{4}{*}{30\%} & 5.43~$\pm$~1.25 & 10.80~$\pm$~1.95 & 20.98~$\pm$~2.22 \\
    Luminance & & 29.05~$\pm$~1.26 & 42.50~$\pm$~2.28 & 62.85~$\pm$~0.57 \\
    Color-opponency & & 26.11~$\pm$~2.58 & 41.36~$\pm$~3.85 & 58.39~$\pm$~3.27 \\
    Single-color & & 27.17~$\pm$~0.82 & 42.25~$\pm$~2.30 & 60.67~$\pm$~3.11 \\
    \midrule
    - & \multirow{4}{*}{40\%} & 3.36~$\pm$~0.51 & 8.20~$\pm$~0.16 & 20.51~$\pm$~1.29 \\
    Luminance & & 29.29~$\pm$~1.74 & 42.91~$\pm$~1.94 & 62.93~$\pm$~1.24 \\
    Color-opponency & & 26.20~$\pm$~2.73 & 41.18~$\pm$~4.10 & 58.38~$\pm$~3.36 \\
    Single-color & & 27.01~$\pm$~0.88 & 42.02~$\pm$~2.12 & 61.08~$\pm$~2.87 \\
    \midrule
    - & \multirow{4}{*}{50\%} & 3.56~$\pm$~0.78 & 8.73~$\pm$~1.29 & 23.36~$\pm$~0.55 \\
    Luminance & & 29.17~$\pm$~1.78 & 42.57~$\pm$~1.65 & 63.12~$\pm$~1.84 \\
    Color-opponency & & 26.20~$\pm$~2.97 & 41.33~$\pm$~4.26 & 58.53~$\pm$~3.39 \\
    Single-color & & 27.70~$\pm$~0.83 & 42.92~$\pm$~2.48 & 62.36~$\pm$~2.97 \\
    \midrule
    - & \multirow{4}{*}{60\%} & 4.18~$\pm$~0.95 & 9.43~$\pm$~1.78 & 27.19~$\pm$~2.31 \\
    Luminance & & 29.64~$\pm$~2.59 & 43.26~$\pm$~2.37 & 63.54~$\pm$~3.17 \\
    Color-opponency & & 26.57~$\pm$~3.16 & 41.87~$\pm$~4.63 & 58.93~$\pm$~3.17 \\
    Single-color & & 27.72~$\pm$~0.57 & 42.84~$\pm$~2.57 & 62.87~$\pm$~2.98 \\
    \midrule
    - & \multirow{4}{*}{70\%} & 5.03~$\pm$~0.60 & 10.01~$\pm$~1.31 & 30.55~$\pm$~1.88 \\
    Luminance & & 28.98~$\pm$~3.17 & 42.52~$\pm$~3.16 & 64.59~$\pm$~4.83 \\
    Color-opponency & & 26.76~$\pm$~3.65 & 42.30~$\pm$~4.80 & 59.52~$\pm$~2.86 \\
    Single-color & & 26.97~$\pm$~0.40 & 42.11~$\pm$~2.19 & 62.96~$\pm$~3.07 \\
    \midrule
    - & \multirow{4}{*}{80\%} & 5.67~$\pm$~0.41 & 10.66~$\pm$~0.66 & 34.73~$\pm$~1.86 \\
    Luminance & & 25.30~$\pm$~1.70 & 39.40~$\pm$~2.86 & 62.55~$\pm$~3.36 \\
    Color-opponency & & 26.56~$\pm$~2.53 & 41.87~$\pm$~3.42 & 60.23~$\pm$~2.27 \\
    Single-color & & 24.52~$\pm$~1.06 & 38.99~$\pm$~1.16 & 60.96~$\pm$~3.10 \\
    \midrule
    - & \multirow{4}{*}{90\%} & 5.84~$\pm$~0.52 & 11.28~$\pm$~0.34 & 36.90~$\pm$~3.57 \\
    Luminance & & 18.07~$\pm$~1.32 & 30.86~$\pm$~0.98 & 51.70~$\pm$~2.14 \\
    Color-opponency & & 25.17~$\pm$~2.31 & 39.66~$\pm$~2.70 & 59.14~$\pm$~1.82 \\
    Single-color & & 19.28~$\pm$~2.23 & 32.35~$\pm$~2.02 & 53.07~$\pm$~6.76 \\
    \bottomrule
  \end{tabular}
  }
  \label{tab:sparse_results_mask2Former_acdc_night_zero_shot}
\end{table}
\clearpage

\begin{table}[tb]
  \centering
  \caption{\textbf{Zero-shot} evaluation of the \textbf{Mask2Former} architecture trained on Cityscapes without sparsity and \textbf{validated on ACDC Fog} at different sparsities. Results include the three preprocessing types (luminance, color-opponency, and single-color). Metrics reported are mean Intersection over Union (mIoU), mean Accuracy (mAcc), and average Accuracy (aAcc), averaged over three seeds. \textbf{Percent-based sparsity:} For a given \textit{percentage}, the \textit{percentage} of lowest absolute values are set to zero.}   
  \scalebox{0.85}{
  \begin{tabular}{@{}lc rrr@{}}
    \toprule
    Preprocessing &  Sparsity &\multicolumn{1}{c}{mIoU}&\multicolumn{1}{c}{mAcc}&\multicolumn{1}{c}{aAcc}  \\
    \midrule
    - & \multirow{4}{*}{-} &  64.70~$\pm$~3.04 & 79.77~$\pm$~3.06 & 91.08~$\pm$~1.46 \\
     Luminance  &  &  62.16~$\pm$~1.18 & 75.22~$\pm$~0.88 & 91.30~$\pm$~1.05 \\
    Color-opponency & & 61.58~$\pm$~0.95 & 77.08~$\pm$~1.58 & 89.45~$\pm$~1.28 \\
    Single-color & & 63.77~$\pm$~3.25 & 77.78~$\pm$~2.56 & 90.43~$\pm$~1.33 \\
    \midrule
    - & \multirow{4}{*}{10\%} & 56.66~$\pm$~5.07 & 75.13~$\pm$~2.24 & 87.53~$\pm$~5.10 \\
    Luminance & & 62.37~$\pm$~0.92 & 74.89~$\pm$~0.23 & 91.96~$\pm$~0.70 \\
    Color-opponency & & 56.47~$\pm$~1.79 & 72.78~$\pm$~1.83 & 88.36~$\pm$~1.78 \\
    Single-color & & 62.04~$\pm$~2.65 & 75.43~$\pm$~2.82 & 91.04~$\pm$~1.80 \\
    \midrule
    - & \multirow{4}{*}{20\%} & 50.54~$\pm$~4.09 & 71.45~$\pm$~1.50 & 83.39~$\pm$~5.77 \\
    Luminance & & 62.67~$\pm$~0.92 & 75.13~$\pm$~0.22 & 92.06~$\pm$~0.66 \\
    Color-opponency & & 56.48~$\pm$~1.81 & 72.80~$\pm$~1.85 & 88.37~$\pm$~1.77 \\
    Single-color & & 62.31~$\pm$~2.28 & 75.56~$\pm$~2.57 & 91.18~$\pm$~1.57 \\
    \midrule
    - & \multirow{4}{*}{30\%} & 39.96~$\pm$~4.66 & 61.79~$\pm$~3.93 & 75.51~$\pm$~5.36 \\
    Luminance & & 62.71~$\pm$~0.46 & 75.11~$\pm$~0.44 & 92.06~$\pm$~0.53 \\
    Color-opponency & & 56.50~$\pm$~1.84 & 72.81~$\pm$~1.90 & 88.39~$\pm$~1.77 \\
    Single-color & & 63.58~$\pm$~2.97 & 76.21~$\pm$~3.15 & 91.21~$\pm$~1.51 \\
    \midrule
    - & \multirow{4}{*}{40\%} & 32.37~$\pm$~4.34 & 51.83~$\pm$~4.55 & 66.09~$\pm$~2.89 \\
    Luminance & & 62.00~$\pm$~0.31 & 74.48~$\pm$~0.97 & 91.55~$\pm$~0.19 \\
    Color-opponency & & 56.44~$\pm$~1.77 & 72.70~$\pm$~1.79 & 88.45~$\pm$~1.71 \\
    Single-color & & 62.94~$\pm$~3.40 & 75.62~$\pm$~3.60 & 90.36~$\pm$~2.18 \\
    \midrule
    - & \multirow{4}{*}{50\%} & 24.66~$\pm$~2.63 & 38.18~$\pm$~3.16 & 59.58~$\pm$~1.39 \\
    Luminance & & 59.10~$\pm$~1.14 & 71.37~$\pm$~2.13 & 90.88~$\pm$~0.34 \\
    Color-opponency & & 56.22~$\pm$~1.64 & 72.48~$\pm$~1.69 & 88.46~$\pm$~1.73 \\
    Single-color & & 61.86~$\pm$~3.67 & 74.68~$\pm$~3.86 & 89.37~$\pm$~2.73 \\
    \midrule
    - & \multirow{4}{*}{60\%} & 16.78~$\pm$~1.87 & 25.75~$\pm$~1.39 & 50.65~$\pm$~2.49 \\
    Luminance & & 56.55~$\pm$~3.29 & 68.34~$\pm$~3.95 & 90.64~$\pm$~0.31 \\
    Color-opponency & & 56.25~$\pm$~1.56 & 72.43~$\pm$~1.59 & 88.31~$\pm$~1.86 \\
    Single-color & & 59.41~$\pm$~3.69 & 72.51~$\pm$~3.99 & 88.81~$\pm$~2.37 \\
    \midrule
    - & \multirow{4}{*}{70\%} & 9.40~$\pm$~1.47 & 15.58~$\pm$~0.64 & 40.72~$\pm$~2.05 \\
    Luminance & & 52.68~$\pm$~2.15 & 64.46~$\pm$~3.32 & 89.20~$\pm$~0.86 \\
    Color-opponency & & 55.79~$\pm$~1.17 & 71.62~$\pm$~1.46 & 87.90~$\pm$~1.88 \\
    Single-color & & 54.23~$\pm$~5.05 & 67.55~$\pm$~4.46 & 86.62~$\pm$~3.08 \\
    \midrule
    - & \multirow{4}{*}{80\%} & 6.67~$\pm$~1.14 & 12.26~$\pm$~1.41 & 30.76~$\pm$~3.65 \\
    Luminance & & 44.75~$\pm$~0.54 & 56.85~$\pm$~1.70 & 84.14~$\pm$~1.98 \\
    Color-opponency & & 55.65~$\pm$~0.78 & 71.76~$\pm$~1.68 & 86.79~$\pm$~1.69 \\
    Single-color & & 44.03~$\pm$~3.92 & 57.00~$\pm$~4.52 & 82.28~$\pm$~3.76 \\
    \midrule
    - & \multirow{4}{*}{90\%} & 4.39~$\pm$~1.64 & 8.72~$\pm$~1.59 & 22.10~$\pm$~4.98 \\
    Luminance & & 30.94~$\pm$~1.67 & 40.30~$\pm$~1.49 & 75.87~$\pm$~4.52 \\
    Color-opponency & & 49.09~$\pm$~4.13 & 64.21~$\pm$~1.85 & 85.25~$\pm$~2.29 \\
    Single-color & & 32.33~$\pm$~3.30 & 41.04~$\pm$~3.32 & 76.43~$\pm$~3.11 \\
    \bottomrule
  \end{tabular}
  }
  \label{tab:sparse_results_mask2Former_acdc_fog_zero_shot}
\end{table}
\clearpage

\begin{table}[tb]
  \centering
  \caption{\textbf{Zero-shot} evaluation of the \textbf{Mask2Former} architecture trained on Cityscapes without sparsity and \textbf{validated on ACDC Rain} at different sparsities. Results include the three preprocessing types (luminance, color-opponency, and single-color). Metrics reported are mean Intersection over Union (mIoU), mean Accuracy (mAcc), and average Accuracy (aAcc), averaged over three seeds. \textbf{Percent-based sparsity:} For a given \textit{percentage}, the \textit{percentage} of lowest absolute values are set to zero.}   
  \scalebox{0.85}{
  \begin{tabular}{@{}lc rrr@{}}
    \toprule
    Preprocessing &  Sparsity &\multicolumn{1}{c}{mIoU}&\multicolumn{1}{c}{mAcc}&\multicolumn{1}{c}{aAcc}  \\
    \midrule
    - & \multirow{4}{*}{-} &  49.91~$\pm$~1.32 & 71.13~$\pm$~1.71 & 87.97~$\pm$~0.38 \\
     Luminance  &  &  47.20~$\pm$~0.13 & 64.00~$\pm$~2.66 & 87.95~$\pm$~1.50 \\
    Color-opponency & & 44.45~$\pm$~0.48 & 60.91~$\pm$~0.56 & 86.78~$\pm$~1.08 \\
    Single-color & & 44.86~$\pm$~3.38 & 62.10~$\pm$~5.34 & 86.84~$\pm$~3.21 \\
    \midrule
    - & \multirow{4}{*}{10\%} & 44.78~$\pm$~3.62 & 67.73~$\pm$~3.49 & 84.67~$\pm$~3.57 \\
    Luminance & & 49.81~$\pm$~1.39 & 64.00~$\pm$~2.97 & 88.94~$\pm$~1.05 \\
    Color-opponency & & 43.89~$\pm$~0.43 & 59.30~$\pm$~0.90 & 86.08~$\pm$~1.64 \\
    Single-color & & 45.48~$\pm$~2.44 & 62.80~$\pm$~1.43 & 87.18~$\pm$~3.88 \\
    \midrule
    - & \multirow{4}{*}{20\%} & 36.89~$\pm$~2.03 & 59.22~$\pm$~2.36 & 77.55~$\pm$~1.50 \\
    Luminance & & 49.95~$\pm$~1.56 & 64.04~$\pm$~3.07 & 89.13~$\pm$~0.92 \\
    Color-opponency & & 43.89~$\pm$~0.42 & 59.31~$\pm$~0.90 & 86.09~$\pm$~1.64 \\
    Single-color & & 45.23~$\pm$~2.41 & 62.54~$\pm$~1.16 & 87.43~$\pm$~3.70 \\
    \midrule
    - & \multirow{4}{*}{30\%} & 26.12~$\pm$~0.40 & 40.11~$\pm$~2.10 & 67.73~$\pm$~1.22 \\
    Luminance & & 49.81~$\pm$~1.94 & 63.82~$\pm$~3.33 & 89.25~$\pm$~0.85 \\
    Color-opponency & & 43.93~$\pm$~0.45 & 59.36~$\pm$~0.82 & 86.10~$\pm$~1.62 \\
    Single-color & & 45.74~$\pm$~2.11 & 62.62~$\pm$~1.40 & 88.05~$\pm$~2.93 \\
    \midrule
    - & \multirow{4}{*}{40\%} & 16.11~$\pm$~0.91 & 24.24~$\pm$~2.61 & 57.85~$\pm$~3.54 \\
    Luminance & & 49.77~$\pm$~2.00 & 63.66~$\pm$~3.11 & 89.25~$\pm$~0.69 \\
    Color-opponency & & 43.99~$\pm$~0.44 & 59.36~$\pm$~0.84 & 86.14~$\pm$~1.52 \\
    Single-color & & 45.31~$\pm$~1.64 & 61.18~$\pm$~1.26 & 87.84~$\pm$~3.09 \\
    \midrule
    - & \multirow{4}{*}{50\%} & 11.39~$\pm$~1.14 & 18.62~$\pm$~2.59 & 48.64~$\pm$~4.69 \\
    Luminance & & 48.87~$\pm$~1.68 & 62.34~$\pm$~2.41 & 89.07~$\pm$~0.78 \\
    Color-opponency & & 43.83~$\pm$~0.53 & 59.14~$\pm$~1.09 & 86.13~$\pm$~1.45 \\
    Single-color & & 44.06~$\pm$~1.94 & 58.25~$\pm$~1.55 & 87.57~$\pm$~2.96 \\
    \midrule
    - & \multirow{4}{*}{60\%} & 8.68~$\pm$~1.26 & 14.75~$\pm$~1.60 & 41.97~$\pm$~4.15 \\
    Luminance & & 46.47~$\pm$~0.90 & 59.15~$\pm$~0.17 & 88.61~$\pm$~0.94 \\
    Color-opponency & & 43.84~$\pm$~0.54 & 58.98~$\pm$~1.20 & 86.22~$\pm$~1.50 \\
    Single-color & & 42.13~$\pm$~2.48 & 55.62~$\pm$~2.16 & 87.05~$\pm$~2.46 \\
    \midrule
    - & \multirow{4}{*}{70\%} & 7.41~$\pm$~1.11 & 13.41~$\pm$~1.24 & 38.51~$\pm$~2.60 \\
    Luminance & & 42.11~$\pm$~1.44 & 53.50~$\pm$~2.19 & 86.82~$\pm$~1.00 \\
    Color-opponency & & 43.47~$\pm$~0.35 & 58.50~$\pm$~1.76 & 86.08~$\pm$~1.60 \\
    Single-color & & 39.49~$\pm$~2.00 & 53.85~$\pm$~1.60 & 85.38~$\pm$~2.44 \\
    \midrule
    - & \multirow{4}{*}{80\%} & 6.73~$\pm$~0.43 & 12.91~$\pm$~0.77 & 36.25~$\pm$~1.46 \\
    Luminance & & 35.36~$\pm$~2.45 & 44.10~$\pm$~3.01 & 83.03~$\pm$~1.88 \\
    Color-opponency & & 42.57~$\pm$~0.66 & 56.83~$\pm$~1.37 & 86.03~$\pm$~1.40 \\
    Single-color & & 34.64~$\pm$~1.54 & 45.22~$\pm$~2.68 & 82.84~$\pm$~2.12 \\
    \midrule
    - & \multirow{4}{*}{90\%} & 5.21~$\pm$~1.01 & 10.54~$\pm$~0.93 & 29.13~$\pm$~2.66 \\
    Luminance & & 26.15~$\pm$~1.65 & 32.71~$\pm$~1.28 & 75.16~$\pm$~3.31 \\
    Color-opponency & & 39.48~$\pm$~2.87 & 53.26~$\pm$~1.83 & 84.81~$\pm$~2.55 \\
    Single-color & & 26.96~$\pm$~0.90 & 33.30~$\pm$~1.01 & 76.86~$\pm$~0.38 \\
    \bottomrule
  \end{tabular}
  }
  \label{tab:sparse_results_mask2Former_acdc_rain_zero_shot}
\end{table}
\clearpage

\begin{table}[tb]
  \centering
  \caption{\textbf{Zero-shot} evaluation of the \textbf{Mask2Former} architecture trained on Cityscapes without sparsity and \textbf{validated on ACDC Snow} at different sparsities. Results include the three preprocessing types (luminance, color-opponency, and single-color). Metrics reported are mean Intersection over Union (mIoU), mean Accuracy (mAcc), and average Accuracy (aAcc), averaged over three seeds. \textbf{Percent-based sparsity:} For a given \textit{percentage}, the \textit{percentage} of lowest absolute values are set to zero.}  
  \scalebox{0.85}{
  \begin{tabular}{@{}lc rrr@{}}
    \toprule
    Preprocessing &  Sparsity &\multicolumn{1}{c}{mIoU}&\multicolumn{1}{c}{mAcc}&\multicolumn{1}{c}{aAcc}  \\
    \midrule
    - & \multirow{4}{*}{-} &  48.20~$\pm$~2.09 & 62.94~$\pm$~2.16 & 81.98~$\pm$~1.71 \\
     Luminance  &  &  49.87~$\pm$~2.06 & 60.95~$\pm$~0.88 & 85.47~$\pm$~2.31 \\
    Color-opponency & & 47.48~$\pm$~0.67 & 60.15~$\pm$~0.81 & 82.73~$\pm$~1.13 \\
    Single-color & & 46.97~$\pm$~3.98 & 61.26~$\pm$~5.43 & 81.86~$\pm$~3.87 \\
    \midrule
    - & \multirow{4}{*}{10\%} & 46.47~$\pm$~2.96 & 61.69~$\pm$~3.31 & 80.94~$\pm$~2.56 \\
    Luminance & & 50.85~$\pm$~1.51 & 61.92~$\pm$~0.23 & 86.83~$\pm$~2.38 \\
    Color-opponency & & 45.88~$\pm$~1.31 & 59.25~$\pm$~0.83 & 80.72~$\pm$~2.05 \\
    Single-color & & 44.14~$\pm$~2.98 & 57.26~$\pm$~4.73 & 82.19~$\pm$~4.19 \\
    \midrule
    - & \multirow{4}{*}{20\%} & 41.44~$\pm$~3.69 & 56.73~$\pm$~2.54 & 77.15~$\pm$~2.98 \\
    Luminance & & 50.86~$\pm$~1.76 & 61.86~$\pm$~0.55 & 87.08~$\pm$~2.01 \\
    Color-opponency & & 45.91~$\pm$~1.27 & 59.27~$\pm$~0.83 & 80.73~$\pm$~2.04 \\
    Single-color & & 45.06~$\pm$~4.11 & 58.15~$\pm$~6.02 & 82.58~$\pm$~4.12 \\
    \midrule
    - & \multirow{4}{*}{30\%} & 33.75~$\pm$~4.44 & 48.99~$\pm$~3.15 & 69.63~$\pm$~4.86 \\
    Luminance & & 50.89~$\pm$~2.29 & 61.87~$\pm$~1.42 & 87.14~$\pm$~1.67 \\
    Color-opponency & & 45.95~$\pm$~1.28 & 59.29~$\pm$~0.85 & 80.79~$\pm$~2.02 \\
    Single-color & & 45.12~$\pm$~4.08 & 58.04~$\pm$~5.94 & 82.58~$\pm$~3.92 \\
    \midrule
    - & \multirow{4}{*}{40\%} & 24.86~$\pm$~3.29 & 37.03~$\pm$~1.15 & 61.50~$\pm$~4.96 \\
    Luminance & & 50.88~$\pm$~2.10 & 61.72~$\pm$~1.21 & 86.99~$\pm$~1.57 \\
    Color-opponency & & 45.63~$\pm$~1.72 & 58.82~$\pm$~1.22 & 80.94~$\pm$~1.98 \\
    Single-color & & 44.84~$\pm$~4.10 & 57.97~$\pm$~5.97 & 82.03~$\pm$~3.93 \\
    \midrule
    - & \multirow{4}{*}{50\%} & 19.64~$\pm$~3.13 & 29.13~$\pm$~1.13 & 55.93~$\pm$~5.46 \\
    Luminance & & 50.62~$\pm$~1.46 & 60.87~$\pm$~0.73 & 86.44~$\pm$~1.62 \\
    Color-opponency & & 45.89~$\pm$~1.51 & 58.97~$\pm$~1.22 & 81.06~$\pm$~1.95 \\
    Single-color & & 43.98~$\pm$~4.28 & 56.77~$\pm$~6.61 & 81.51~$\pm$~3.76 \\
    \midrule
    - & \multirow{4}{*}{60\%} & 12.97~$\pm$~1.88 & 19.80~$\pm$~2.15 & 50.95~$\pm$~3.79 \\
    Luminance & & 47.94~$\pm$~1.83 & 58.23~$\pm$~1.03 & 85.39~$\pm$~1.51 \\
    Color-opponency & & 46.38~$\pm$~0.90 & 59.51~$\pm$~0.98 & 81.28~$\pm$~2.15 \\
    Single-color & & 43.72~$\pm$~5.89 & 55.31~$\pm$~7.49 & 80.88~$\pm$~3.60 \\
    \midrule
    - & \multirow{4}{*}{70\%} & 9.64~$\pm$~0.89 & 16.16~$\pm$~0.65 & 45.32~$\pm$~2.09 \\
    Luminance & & 44.81~$\pm$~2.51 & 54.71~$\pm$~1.94 & 83.92~$\pm$~1.51 \\
    Color-opponency & & 46.79~$\pm$~1.58 & 59.86~$\pm$~1.54 & 81.15~$\pm$~2.29 \\
    Single-color & & 41.19~$\pm$~4.56 & 51.97~$\pm$~5.65 & 80.15~$\pm$~3.04 \\
    \midrule
    - & \multirow{4}{*}{80\%} & 7.22~$\pm$~0.84 & 13.74~$\pm$~0.52 & 38.81~$\pm$~1.96 \\
    Luminance & & 38.52~$\pm$~2.29 & 47.11~$\pm$~2.31 & 80.46~$\pm$~1.98 \\
    Color-opponency & & 46.14~$\pm$~2.61 & 59.40~$\pm$~1.15 & 80.92~$\pm$~1.91 \\
    Single-color & & 37.61~$\pm$~3.41 & 46.94~$\pm$~4.40 & 79.32~$\pm$~2.02 \\
    \midrule
    - & \multirow{4}{*}{90\%} & 5.28~$\pm$~0.71 & 10.84~$\pm$~1.01 & 29.77~$\pm$~1.22 \\
    Luminance & & 25.27~$\pm$~1.65 & 31.37~$\pm$~1.51 & 70.22~$\pm$~3.60 \\
    Color-opponency & & 41.72~$\pm$~4.16 & 53.81~$\pm$~3.24 & 78.75~$\pm$~1.97 \\
    Single-color & & 28.82~$\pm$~1.92 & 35.58~$\pm$~2.48 & 74.03~$\pm$~0.85 \\
    \bottomrule
  \end{tabular}
  }
  \label{tab:sparse_results_mask2Former_acdc_snow_zero_shot}
\end{table}
\clearpage

\begin{table}[tb]
  \centering
  \caption{\textbf{Zero-shot} evaluation of the \textbf{Mask2Former} architecture trained on Cityscapes without sparsity and \textbf{validated on ACDC Mean} at different sparsities. Results include the three preprocessing types (luminance, color-opponency, and single-color). Metrics reported are mean Intersection over Union (mIoU), mean Accuracy (mAcc), and average Accuracy (aAcc), averaged over three seeds. \textbf{Percent-based sparsity:} For a given \textit{percentage}, the \textit{percentage} of lowest absolute values are set to zero.}
  \scalebox{0.85}{
  \begin{tabular}{@{}lc rrr@{}}
    \toprule
    Preprocessing & Sparsity &\multicolumn{1}{c}{mIoU}&\multicolumn{1}{c}{mAcc}&\multicolumn{1}{c}{aAcc}  \\
    \midrule
     - & \multirow{4}{*}{-}& 42.05~$\pm$~1.41 & 60.13~$\pm$~2.21 & 77.94~$\pm$~1.35\\
     Luminance &  & 46.83~$\pm$~1.10 & 59.72~$\pm$~2.32 & 81.61~$\pm$~1.38 \\
    Color-opponency &  & 44.93~$\pm$~0.66 & 58.84~$\pm$~0.86 & 79.18~$\pm$~0.62 \\
    Single-color &  & 45.23~$\pm$~1.74 & 59.85~$\pm$~2.53 & 79.63~$\pm$~2.76 \\
    \midrule
    - & \multirow{4}{*}{10\%} & 38.27~$\pm$~1.45 & 56.73~$\pm$~0.79 & 75.13~$\pm$~1.95 \\
    Luminance & & 47.01~$\pm$~1.33 & 59.83~$\pm$~2.49 & 81.65~$\pm$~1.37 \\
    Color-opponency & & 44.94~$\pm$~0.65 & 58.84~$\pm$~0.85 & 79.18~$\pm$~0.62 \\
    Single-color & & 45.27~$\pm$~1.76 & 59.83~$\pm$~2.49 & 79.65~$\pm$~2.75 \\
    \midrule
    - & \multirow{4}{*}{20\%} & 32.15~$\pm$~2.13 & 49.95~$\pm$~0.74 & 68.24~$\pm$~2.58 \\
    Luminance & & 46.97~$\pm$~1.34 & 59.79~$\pm$~2.56 & 81.82~$\pm$~1.12 \\
    Color-opponency & & 44.87~$\pm$~0.55 & 58.74~$\pm$~0.82 & 79.17~$\pm$~0.63 \\
    Single-color & & 45.51~$\pm$~1.92 & 59.84~$\pm$~2.70 & 79.92~$\pm$~2.70 \\
    \midrule
    - & \multirow{4}{*}{30\%} & 25.57~$\pm$~2.81 & 40.17~$\pm$~2.19 & 59.54~$\pm$~3.74 \\
    Luminance & & 47.32~$\pm$~0.78 & 59.99~$\pm$~2.01 & 82.08~$\pm$~0.64 \\
    Color-opponency & & 44.88~$\pm$~0.55 & 58.74~$\pm$~0.83 & 79.19~$\pm$~0.63 \\
    Single-color & & 45.53~$\pm$~1.55 & 59.52~$\pm$~2.22 & 80.05~$\pm$~2.55 \\
    \midrule
    - & \multirow{4}{*}{40\%} & 19.58~$\pm$~2.98 & 31.71~$\pm$~1.94 & 52.56~$\pm$~4.96 \\
    Luminance & & 47.00~$\pm$~0.33 & 59.69~$\pm$~0.78 & 81.92~$\pm$~0.58 \\
    Color-opponency & & 44.86~$\pm$~0.56 & 58.66~$\pm$~0.86 & 79.27~$\pm$~0.65 \\
    Single-color & & 45.27~$\pm$~1.31 & 59.31~$\pm$~2.14 & 79.82~$\pm$~2.49 \\
    \midrule
    - & \multirow{4}{*}{50\%} & 15.05~$\pm$~3.03 & 24.86~$\pm$~2.15 & 47.90~$\pm$~5.13 \\
    Luminance & & 46.65~$\pm$~0.78 & 59.34~$\pm$~0.61 & 81.56~$\pm$~1.05 \\
    Color-opponency & & 44.75~$\pm$~0.70 & 58.56~$\pm$~0.92 & 79.37~$\pm$~0.66 \\
    Single-color & & 44.80~$\pm$~1.79 & 58.61~$\pm$~2.76 & 79.74~$\pm$~2.56 \\
    \midrule
    - & \multirow{4}{*}{60\%} & 10.84~$\pm$~1.58 & 17.41~$\pm$~1.68 & 42.70~$\pm$~4.40 \\
    Luminance & & 44.31~$\pm$~0.93 & 56.54~$\pm$~0.41 & 81.07~$\pm$~1.48 \\
    Color-opponency & & 45.09~$\pm$~0.88 & 58.76~$\pm$~1.03 & 79.57~$\pm$~0.67 \\
    Single-color & & 43.36~$\pm$~1.77 & 56.76~$\pm$~2.70 & 79.41~$\pm$~2.45 \\
    \midrule
    - & \multirow{4}{*}{70\%} & 7.99~$\pm$~0.67 & 13.63~$\pm$~1.25 & 38.36~$\pm$~2.15 \\
    Luminance & & 41.41~$\pm$~2.13 & 53.31~$\pm$~1.57 & 80.12~$\pm$~2.09 \\
    Color-opponency & & 44.92~$\pm$~1.50 & 58.20~$\pm$~1.69 & 79.51~$\pm$~0.95 \\
    Single-color & & 40.72~$\pm$~1.21 & 53.75~$\pm$~2.02 & 78.37~$\pm$~2.37 \\
    \midrule
    - & \multirow{4}{*}{80\%} & 6.79~$\pm$~0.34 & 11.97~$\pm$~0.31 & 35.69~$\pm$~0.64 \\
    Luminance & & 35.17~$\pm$~1.49 & 46.84~$\pm$~1.79 & 76.66~$\pm$~2.56 \\
    Color-opponency & & 43.83~$\pm$~1.86 & 56.82~$\pm$~1.68 & 79.23~$\pm$~1.31 \\
    Single-color & & 36.33~$\pm$~1.11 & 48.21~$\pm$~1.08 & 76.29~$\pm$~1.98 \\
    \midrule
    - & \multirow{4}{*}{90\%} & 5.75~$\pm$~0.53 & 10.47~$\pm$~0.14 & 32.19~$\pm$~2.22 \\
    Luminance & & 24.39~$\pm$~2.15 & 34.15~$\pm$~0.24 & 68.27~$\pm$~3.53 \\
    Color-opponency & & 40.37~$\pm$~2.31 & 52.47~$\pm$~1.90 & 77.87~$\pm$~1.96 \\
    Single-color & & 27.47~$\pm$~1.26 & 37.33~$\pm$~1.64 & 71.02~$\pm$~1.75 \\
    \bottomrule
  \end{tabular}
  }
  \label{tab:sparse_results_mask2former_acdc_full}
\end{table}
\clearpage

%% file: tables/sparsity_zero_shot_upernet.tex
\begin{table}[h]
  \centering
  \caption{\textbf{Zero-shot} evaluation of the \textbf{UPerNet} architecture trained on Cityscapes without sparsity and \textbf{validated on Cityscapes} at different sparsities. Results include the three preprocessing types (luminance, color-opponency, and single-color). Metrics reported are mean Intersection over Union (mIoU), mean Accuracy (mAcc), and average Accuracy (aAcc), averaged over three seeds. \textbf{Percent-based sparsity:} For a given \textit{percentage}, the \textit{percentage} of lowest absolute values are set to zero.}   
  \scalebox{0.85}{
  \begin{tabular}{@{}lc rrr@{}}
    \toprule
    Preprocessing &  Sparsity &\multicolumn{1}{c}{mIoU}&\multicolumn{1}{c}{mAcc}&\multicolumn{1}{c}{aAcc}  \\
    \midrule
    - & \multirow{4}{*}{-} &  64.64~$\pm$~1.93 & 72.63~$\pm$~1.25 & 94.42~$\pm$~0.17 \\
     Luminance &  & 62.63~$\pm$~1.79 & 71.37~$\pm$~1.79 & 93.85~$\pm$~0.11 \\
    Color-opponency&   & 64.33~$\pm$~1.32 & 72.12~$\pm$~1.90 & 94.63~$\pm$~0.13 \\
    Single-color &  & 63.37~$\pm$~2.80 & 72.26~$\pm$~2.80 & 94.47~$\pm$~0.20 \\
    \midrule
    - & \multirow{4}{*}{10\%} & 60.53~$\pm$~0.18 & 69.01~$\pm$~1.52 & 93.30~$\pm$~0.52 \\
    Luminance & & 59.50~$\pm$~0.93 & 67.65~$\pm$~1.90 & 93.70~$\pm$~0.04 \\
    Color-opponency & & 61.93~$\pm$~0.87 & 70.26~$\pm$~0.72 & 94.35~$\pm$~0.12 \\
    Single-color & & 63.17~$\pm$~1.80 & 71.44~$\pm$~2.07 & 94.45~$\pm$~0.07 \\
    \midrule
    - & \multirow{4}{*}{20\%} & 55.74~$\pm$~0.61 & 66.58~$\pm$~1.59 & 89.85~$\pm$~1.20 \\
    Luminance & & 59.49~$\pm$~0.94 & 67.64~$\pm$~1.92 & 93.67~$\pm$~0.06 \\
    Color-opponency & & 61.94~$\pm$~0.80 & 70.25~$\pm$~0.75 & 94.34~$\pm$~0.11 \\
    Single-color & & 63.14~$\pm$~1.85 & 71.44~$\pm$~2.13 & 94.39~$\pm$~0.09 \\
    \midrule
    - & \multirow{4}{*}{30\%} & 48.11~$\pm$~0.36 & 61.91~$\pm$~1.76 & 81.03~$\pm$~1.86 \\
    Luminance & & 59.31~$\pm$~1.03 & 67.44~$\pm$~1.94 & 93.52~$\pm$~0.13 \\
    Color-opponency & & 61.94~$\pm$~0.65 & 70.23~$\pm$~0.83 & 94.29~$\pm$~0.12 \\
    Single-color & & 62.91~$\pm$~1.76 & 71.35~$\pm$~2.13 & 94.14~$\pm$~0.30 \\
    \midrule
    - & \multirow{4}{*}{40\%} & 39.86~$\pm$~0.58 & 55.17~$\pm$~1.76 & 68.90~$\pm$~1.47 \\
    Luminance & & 58.57~$\pm$~1.21 & 66.89~$\pm$~1.88 & 92.90~$\pm$~0.63 \\
    Color-opponency & & 61.75~$\pm$~0.56 & 70.09~$\pm$~0.96 & 94.16~$\pm$~0.16 \\
    Single-color & & 62.25~$\pm$~1.48 & 70.99~$\pm$~2.04 & 93.59~$\pm$~0.83 \\
    \midrule
    - & \multirow{4}{*}{50\%} & 32.16~$\pm$~0.44 & 46.58~$\pm$~2.53 & 59.16~$\pm$~2.18 \\
    Luminance & & 56.44~$\pm$~1.83 & 65.02~$\pm$~1.76 & 90.75~$\pm$~2.56 \\
    Color-opponency & & 60.59~$\pm$~1.10 & 69.50~$\pm$~1.03 & 93.49~$\pm$~0.56 \\
    Single-color & & 60.32~$\pm$~0.95 & 69.87~$\pm$~1.84 & 91.88~$\pm$~2.25 \\
    \midrule
    - & \multirow{4}{*}{60\%} & 23.46~$\pm$~2.00 & 35.19~$\pm$~3.52 & 49.91~$\pm$~3.78 \\
    Luminance & & 52.71~$\pm$~2.30 & 61.72~$\pm$~1.93 & 85.49~$\pm$~5.59 \\
    Color-opponency & & 58.32~$\pm$~1.82 & 68.01~$\pm$~0.78 & 90.56~$\pm$~3.36 \\
    Single-color & & 57.14~$\pm$~0.66 & 67.34~$\pm$~1.43 & 89.02~$\pm$~3.81 \\
    \midrule
    - & \multirow{4}{*}{70\%} & 13.31~$\pm$~2.76 & 21.77~$\pm$~3.09 & 38.38~$\pm$~6.43 \\
    Luminance & & 46.46~$\pm$~2.51 & 55.98~$\pm$~2.30 & 76.71~$\pm$~5.48 \\
    Color-opponency & & 55.92~$\pm$~2.05 & 65.75~$\pm$~0.74 & 87.34~$\pm$~6.33 \\
    Single-color & & 52.75~$\pm$~1.20 & 63.31~$\pm$~1.05 & 85.33~$\pm$~4.88 \\
    \midrule
    - & \multirow{4}{*}{80\%} & 6.07~$\pm$~2.21 & 12.03~$\pm$~2.55 & 27.70~$\pm$~8.67 \\
    Luminance & & 35.23~$\pm$~3.05 & 45.19~$\pm$~2.80 & 62.44~$\pm$~3.21 \\
    Color-opponency & & 53.16~$\pm$~1.38 & 62.63~$\pm$~1.45 & 87.60~$\pm$~4.38 \\
    Single-color & & 44.34~$\pm$~1.67 & 55.23~$\pm$~0.76 & 78.06~$\pm$~6.37 \\
    \midrule
    - & \multirow{4}{*}{90\%} & 3.76~$\pm$~1.55 & 8.76~$\pm$~2.34 & 22.85~$\pm$~6.46 \\
    Luminance & & 18.82~$\pm$~3.08 & 26.69~$\pm$~2.70 & 45.14~$\pm$~1.91 \\
    Color-opponency & & 42.80~$\pm$~2.63 & 52.11~$\pm$~3.28 & 80.36~$\pm$~6.53 \\
    Single-color & & 26.04~$\pm$~2.50 & 37.00~$\pm$~1.32 & 58.31~$\pm$~9.31 \\
    \bottomrule
  \end{tabular}
  }
  \label{tab:sparse_results_upernet_cityscapes_zero_shot}
\end{table}
\clearpage

\begin{table}[tb]
  \centering
  \caption{\textbf{Zero-shot} evaluation of the \textbf{UPerNet} architecture trained on Cityscapes without sparsity and \textbf{validated on Dark Zurich} at different sparsities. Results include the three preprocessing types (luminance, color-opponency, and single-color). Metrics reported are mean Intersection over Union (mIoU), mean Accuracy (mAcc), and average Accuracy (aAcc), averaged over three seeds. \textbf{Percent-based sparsity:} For a given \textit{percentage}, the \textit{percentage} of lowest absolute values are set to zero.}   
  \scalebox{0.85}{
  \begin{tabular}{@{}lc rrr@{}}
    \toprule
    Preprocessing &  Sparsity &\multicolumn{1}{c}{mIoU}&\multicolumn{1}{c}{mAcc}&\multicolumn{1}{c}{aAcc}  \\
    \midrule
    - & \multirow{4}{*}{-} &  11.42~$\pm$~0.58 & 21.92~$\pm$~2.60 & 42.86~$\pm$~6.74 \\
     Luminance &   & 18.03~$\pm$~1.82 & 32.15~$\pm$~1.79 & 50.00~$\pm$~1.10 \\
    Color-opponency& & 16.10~$\pm$~0.95 & 29.13~$\pm$~1.21 & 46.56~$\pm$~4.08 \\
    Single-color& & 13.82~$\pm$~1.52 & 27.68~$\pm$~1.50 & 44.06~$\pm$~3.62 \\
    \midrule
    - & \multirow{4}{*}{10\%} & 8.28~$\pm$~0.97 & 15.14~$\pm$~2.07 & 29.34~$\pm$~3.08 \\
    Luminance & & 17.66~$\pm$~0.47 & 29.58~$\pm$~0.74 & 53.38~$\pm$~2.28 \\
    Color-opponency & & 14.31~$\pm$~1.89 & 26.85~$\pm$~1.98 & 40.99~$\pm$~9.40 \\
    Single-color & & 13.47~$\pm$~2.46 & 25.99~$\pm$~2.26 & 41.37~$\pm$~6.76 \\
    \midrule
    - & \multirow{4}{*}{20\%} & 6.19~$\pm$~0.36 & 12.43~$\pm$~1.23 & 26.52~$\pm$~2.59 \\
    Luminance & & 17.69~$\pm$~0.42 & 29.55~$\pm$~0.79 & 53.18~$\pm$~2.15 \\
    Color-opponency & & 14.29~$\pm$~1.90 & 26.81~$\pm$~2.03 & 40.88~$\pm$~9.58 \\
    Single-color & & 13.38~$\pm$~2.53 & 25.78~$\pm$~2.51 & 40.96~$\pm$~7.07 \\
    \midrule
    - & \multirow{4}{*}{30\%} & 4.34~$\pm$~0.07 & 9.72~$\pm$~1.23 & 23.57~$\pm$~2.05 \\
    Luminance & & 17.63~$\pm$~0.32 & 29.32~$\pm$~1.16 & 52.49~$\pm$~2.07 \\
    Color-opponency & & 14.17~$\pm$~2.09 & 26.63~$\pm$~2.37 & 40.21~$\pm$~10.48 \\
    Single-color & & 13.30~$\pm$~2.76 & 25.35~$\pm$~3.12 & 40.11~$\pm$~7.92 \\
    \midrule
    - & \multirow{4}{*}{40\%} & 3.36~$\pm$~0.32 & 8.42~$\pm$~0.98 & 21.53~$\pm$~2.25 \\
    Luminance & & 17.46~$\pm$~0.35 & 28.89~$\pm$~1.41 & 51.11~$\pm$~1.90 \\
    Color-opponency & & 13.78~$\pm$~2.57 & 26.05~$\pm$~3.19 & 38.32~$\pm$~12.57 \\
    Single-color & & 13.09~$\pm$~3.21 & 24.58~$\pm$~3.80 & 38.28~$\pm$~9.28 \\
    \midrule
    - & \multirow{4}{*}{50\%} & 2.57~$\pm$~0.57 & 7.06~$\pm$~0.29 & 19.36~$\pm$~2.48 \\
    Luminance & & 16.95~$\pm$~0.58 & 27.98~$\pm$~1.85 & 49.21~$\pm$~1.79 \\
    Color-opponency & & 13.47~$\pm$~3.16 & 25.65~$\pm$~4.16 & 36.43~$\pm$~14.57 \\
    Single-color & & 12.62~$\pm$~3.76 & 23.45~$\pm$~4.39 & 36.12~$\pm$~11.23 \\
    \midrule
    - & \multirow{4}{*}{60\%} & 2.29~$\pm$~0.35 & 6.53~$\pm$~0.14 & 18.52~$\pm$~2.15 \\
    Luminance & & 16.29~$\pm$~0.70 & 26.77~$\pm$~2.46 & 46.40~$\pm$~2.42 \\
    Color-opponency & & 13.28~$\pm$~3.54 & 25.34~$\pm$~4.59 & 35.20~$\pm$~15.38 \\
    Single-color & & 12.10~$\pm$~4.32 & 22.31~$\pm$~4.87 & 33.99~$\pm$~12.96 \\
    \midrule
    - & \multirow{4}{*}{70\%} & 2.21~$\pm$~0.40 & 6.12~$\pm$~0.30 & 18.37~$\pm$~2.80 \\
    Luminance & & 14.60~$\pm$~1.46 & 25.06~$\pm$~3.03 & 41.88~$\pm$~4.26 \\
    Color-opponency & & 13.26~$\pm$~3.83 & 25.27~$\pm$~4.74 & 35.31~$\pm$~15.25 \\
    Single-color & & 11.44~$\pm$~4.46 & 21.12~$\pm$~4.82 & 32.46~$\pm$~14.56 \\
    \midrule
    - & \multirow{4}{*}{80\%} & 2.39~$\pm$~0.42 & 6.36~$\pm$~0.63 & 18.95~$\pm$~4.74 \\
    Luminance & & 12.06~$\pm$~1.96 & 22.23~$\pm$~3.37 & 35.57~$\pm$~6.17 \\
    Color-opponency & & 14.49~$\pm$~1.88 & 26.86~$\pm$~1.70 & 43.31~$\pm$~8.20 \\
    Single-color & & 10.68~$\pm$~4.36 & 20.19~$\pm$~4.89 & 32.08~$\pm$~15.28 \\
    \midrule
    - & \multirow{4}{*}{90\%} & 2.97~$\pm$~0.84 & 6.18~$\pm$~1.83 & 20.57~$\pm$~8.10 \\
    Luminance & & 8.30~$\pm$~1.44 & 16.77~$\pm$~2.36 & 26.22~$\pm$~4.00 \\
    Color-opponency & & 13.96~$\pm$~1.78 & 25.09~$\pm$~3.02 & 45.90~$\pm$~5.05 \\
    Single-color & & 9.54~$\pm$~3.24 & 18.41~$\pm$~3.65 & 29.94~$\pm$~11.68 \\
    \bottomrule
  \end{tabular}
  }
  \label{tab:sparse_results_upernet_dark_zurich_zero_shot}
\end{table}
\clearpage

\begin{table}[tb]
  \centering
  \caption{\textbf{Zero-shot} evaluation of the \textbf{UPerNet} architecture trained on Cityscapes without sparsity and \textbf{validated on ACDC Night} at different sparsities. Results include the three preprocessing types (luminance, color-opponency, and single-color). Metrics reported are mean Intersection over Union (mIoU), mean Accuracy (mAcc), and average Accuracy (aAcc), averaged over three seeds. \textbf{Percent-based sparsity:} For a given \textit{percentage}, the \textit{percentage} of lowest absolute values are set to zero.} 
  \scalebox{0.85}{
  \begin{tabular}{@{}lc rrr@{}}
    \toprule
    Preprocessing &  Sparsity &\multicolumn{1}{c}{mIoU}&\multicolumn{1}{c}{mAcc}&\multicolumn{1}{c}{aAcc}  \\
    \midrule
    - & \multirow{4}{*}{-} &  11.53~$\pm$~1.24 & 21.96~$\pm$~2.20 & 45.24~$\pm$~7.17 \\
     Luminance &   & 18.31~$\pm$~1.15 & 31.05~$\pm$~0.94 & 54.27~$\pm$~1.21 \\
    Color-opponency& & 16.37~$\pm$~0.81 & 28.01~$\pm$~1.57 & 50.83~$\pm$~3.65 \\
    Single-color& & 14.64~$\pm$~1.36 & 28.31~$\pm$~1.02 & 49.00~$\pm$~4.06 \\
    \midrule
    - & \multirow{4}{*}{10\%} & 7.59~$\pm$~0.90 & 16.83~$\pm$~0.35 & 29.52~$\pm$~2.26 \\
    Luminance & & 17.89~$\pm$~0.43 & 28.09~$\pm$~0.75 & 56.80~$\pm$~1.62 \\
    Color-opponency & & 15.09~$\pm$~1.49 & 27.02~$\pm$~0.90 & 44.50~$\pm$~9.53 \\
    Single-color & & 14.29~$\pm$~1.88 & 26.62~$\pm$~1.36 & 46.23~$\pm$~6.08 \\
    \midrule
    - & \multirow{4}{*}{20\%} & 5.77~$\pm$~0.80 & 13.41~$\pm$~0.84 & 26.39~$\pm$~1.97 \\
    Luminance & & 17.89~$\pm$~0.35 & 28.00~$\pm$~0.83 & 56.55~$\pm$~1.56 \\
    Color-opponency & & 15.06~$\pm$~1.51 & 27.00~$\pm$~0.92 & 44.37~$\pm$~9.73 \\
    Single-color & & 14.21~$\pm$~2.00 & 26.42~$\pm$~1.59 & 45.83~$\pm$~6.50 \\
    \midrule
    - & \multirow{4}{*}{30\%} & 4.52~$\pm$~0.32 & 11.02~$\pm$~0.43 & 24.28~$\pm$~2.44 \\
    Luminance & & 17.76~$\pm$~0.20 & 27.65~$\pm$~0.98 & 55.85~$\pm$~1.54 \\
    Color-opponency & & 14.98~$\pm$~1.67 & 26.87~$\pm$~1.15 & 43.70~$\pm$~10.76 \\
    Single-color & & 13.98~$\pm$~2.25 & 25.79~$\pm$~2.22 & 44.80~$\pm$~7.57 \\
    \midrule
    - & \multirow{4}{*}{40\%} & 3.76~$\pm$~0.30 & 10.11~$\pm$~0.93 & 22.60~$\pm$~2.55 \\
    Luminance & & 17.52~$\pm$~0.11 & 27.18~$\pm$~1.26 & 54.71~$\pm$~1.51 \\
    Color-opponency & & 14.71~$\pm$~2.15 & 26.36~$\pm$~1.95 & 41.76~$\pm$~13.38 \\
    Single-color & & 13.69~$\pm$~2.59 & 24.94~$\pm$~2.87 & 42.86~$\pm$~8.98 \\
    \midrule
    - & \multirow{4}{*}{50\%} & 2.93~$\pm$~0.33 & 8.52~$\pm$~0.88 & 20.85~$\pm$~0.47 \\
    Luminance & & 16.92~$\pm$~0.40 & 26.33~$\pm$~1.63 & 52.86~$\pm$~1.75 \\
    Color-opponency & & 14.44~$\pm$~2.65 & 25.86~$\pm$~2.86 & 39.73~$\pm$~15.73 \\
    Single-color & & 13.06~$\pm$~2.98 & 23.70~$\pm$~3.32 & 40.26~$\pm$~10.95 \\
    \midrule
    - & \multirow{4}{*}{60\%} & 2.55~$\pm$~0.27 & 7.43~$\pm$~0.71 & 19.86~$\pm$~0.29 \\
    Luminance & & 15.79~$\pm$~0.98 & 25.05~$\pm$~2.20 & 49.20~$\pm$~3.24 \\
    Color-opponency & & 14.28~$\pm$~2.87 & 25.31~$\pm$~3.28 & 38.34~$\pm$~16.76 \\
    Single-color & & 12.47~$\pm$~3.39 & 22.69~$\pm$~3.63 & 37.57~$\pm$~13.14 \\
    \midrule
    - & \multirow{4}{*}{70\%} & 2.30~$\pm$~0.15 & 6.82~$\pm$~0.54 & 18.59~$\pm$~1.39 \\
    Luminance & & 14.05~$\pm$~1.61 & 23.30~$\pm$~2.82 & 43.52~$\pm$~5.57 \\
    Color-opponency & & 14.03~$\pm$~2.97 & 24.58~$\pm$~3.44 & 37.88~$\pm$~16.49 \\
    Single-color & & 11.70~$\pm$~3.51 & 21.53~$\pm$~3.66 & 35.40~$\pm$~14.84 \\
    \midrule
    - & \multirow{4}{*}{80\%} & 2.36~$\pm$~0.17 & 7.45~$\pm$~0.62 & 18.74~$\pm$~3.92 \\
    Luminance & & 11.60~$\pm$~1.86 & 20.66~$\pm$~3.15 & 36.25~$\pm$~7.09 \\
    Color-opponency & & 15.45~$\pm$~1.44 & 26.01~$\pm$~1.26 & 46.11~$\pm$~8.56 \\
    Single-color & & 11.04~$\pm$~3.58 & 20.89~$\pm$~3.59 & 34.56~$\pm$~14.96 \\
    \midrule
    - & \multirow{4}{*}{90\%} & 2.82~$\pm$~0.51 & 7.65~$\pm$~0.70 & 19.74~$\pm$~6.34 \\
    Luminance & & 7.97~$\pm$~1.31 & 15.38~$\pm$~2.04 & 27.44~$\pm$~4.11 \\
    Color-opponency & & 14.63~$\pm$~1.28 & 24.26~$\pm$~2.36 & 46.98~$\pm$~6.14 \\
    Single-color & & 9.83~$\pm$~2.75 & 19.31~$\pm$~1.94 & 32.89~$\pm$~11.10 \\
    \bottomrule
  \end{tabular}
  }
  \label{tab:sparse_results_upernet_acdc_night_zero_shot}
\end{table}
\clearpage

\begin{table}[tb]
  \centering
  \caption{\textbf{Zero-shot} evaluation of the \textbf{UPerNet} architecture trained on Cityscapes without sparsity and \textbf{validated on ACDC Fog} at different sparsities. Results include the three preprocessing types (luminance, color-opponency, and single-color). Metrics reported are mean Intersection over Union (mIoU), mean Accuracy (mAcc), and average Accuracy (aAcc), averaged over three seeds. \textbf{Percent-based sparsity:} For a given \textit{percentage}, the \textit{percentage} of lowest absolute values are set to zero.} 
  \scalebox{0.85}{
  \begin{tabular}{@{}lc rrr@{}}
    \toprule
    Preprocessing &  Sparsity &\multicolumn{1}{c}{mIoU}&\multicolumn{1}{c}{mAcc}&\multicolumn{1}{c}{aAcc}  \\
    \midrule
    - & \multirow{4}{*}{-} &  45.44~$\pm$~1.69 & 55.56~$\pm$~3.78 & 86.25~$\pm$~2.41 \\
     Luminance &   & 46.52~$\pm$~0.66 & 55.61~$\pm$~1.40 & 82.09~$\pm$~3.30 \\
    Color-opponency& & 46.10~$\pm$~0.60 & 55.97~$\pm$~0.97 & 84.47~$\pm$~2.68 \\
    Single-color& & 45.70~$\pm$~2.04 & 55.52~$\pm$~1.92 & 82.15~$\pm$~8.56 \\
    \midrule
    - & \multirow{4}{*}{10\%} & 38.60~$\pm$~1.36 & 49.61~$\pm$~2.42 & 80.09~$\pm$~2.50 \\
    Luminance & & 42.72~$\pm$~1.22 & 52.58~$\pm$~4.33 & 81.78~$\pm$~5.31 \\
    Color-opponency & & 44.94~$\pm$~0.41 & 54.20~$\pm$~1.19 & 84.52~$\pm$~1.64 \\
    Single-color & & 45.34~$\pm$~2.88 & 55.99~$\pm$~4.01 & 82.87~$\pm$~6.44 \\
    \midrule
    - & \multirow{4}{*}{20\%} & 30.80~$\pm$~1.86 & 41.92~$\pm$~2.47 & 69.88~$\pm$~4.14 \\
    Luminance & & 42.91~$\pm$~1.56 & 52.67~$\pm$~4.34 & 82.47~$\pm$~6.22 \\
    Color-opponency & & 44.95~$\pm$~0.40 & 54.20~$\pm$~1.19 & 84.57~$\pm$~1.63 \\
    Single-color & & 45.55~$\pm$~3.10 & 56.08~$\pm$~4.14 & 83.59~$\pm$~7.18 \\
    \midrule
    - & \multirow{4}{*}{30\%} & 23.64~$\pm$~1.70 & 33.76~$\pm$~2.66 & 61.10~$\pm$~4.89 \\
    Luminance & & 42.24~$\pm$~2.03 & 52.15~$\pm$~4.61 & 80.69~$\pm$~8.33 \\
    Color-opponency & & 44.92~$\pm$~0.39 & 54.14~$\pm$~1.17 & 84.71~$\pm$~1.59 \\
    Single-color & & 45.16~$\pm$~3.46 & 55.83~$\pm$~4.43 & 82.38~$\pm$~9.17 \\
    \midrule
    - & \multirow{4}{*}{40\%} & 18.46~$\pm$~1.04 & 26.97~$\pm$~1.26 & 56.65~$\pm$~5.04 \\
    Luminance & & 40.46~$\pm$~1.90 & 50.81~$\pm$~4.77 & 76.61~$\pm$~8.31 \\
    Color-opponency & & 44.76~$\pm$~0.41 & 53.95~$\pm$~1.12 & 85.13~$\pm$~1.43 \\
    Single-color & & 44.13~$\pm$~3.16 & 55.26~$\pm$~4.48 & 80.02~$\pm$~9.28 \\
    \midrule
    - & \multirow{4}{*}{50\%} & 13.26~$\pm$~1.20 & 20.24~$\pm$~1.35 & 53.02~$\pm$~5.13 \\
    Luminance & & 39.12~$\pm$~2.20 & 49.36~$\pm$~5.04 & 75.34~$\pm$~9.25 \\
    Color-opponency & & 44.26~$\pm$~0.47 & 53.44~$\pm$~0.99 & 85.43~$\pm$~1.17 \\
    Single-color & & 43.00~$\pm$~3.08 & 54.18~$\pm$~4.72 & 78.81~$\pm$~9.13 \\
    \midrule
    - & \multirow{4}{*}{60\%} & 10.32~$\pm$~0.55 & 16.59~$\pm$~0.19 & 47.78~$\pm$~5.47 \\
    Luminance & & 36.97~$\pm$~2.91 & 46.67~$\pm$~5.60 & 74.10~$\pm$~10.14 \\
    Color-opponency & & 43.80~$\pm$~0.65 & 52.97~$\pm$~1.16 & 85.71~$\pm$~0.92 \\
    Single-color & & 41.58~$\pm$~2.83 & 52.45~$\pm$~4.73 & 79.07~$\pm$~7.77 \\
    \midrule
    - & \multirow{4}{*}{70\%} & 6.75~$\pm$~1.83 & 11.74~$\pm$~2.37 & 36.85~$\pm$~10.31 \\
    Luminance & & 33.31~$\pm$~2.78 & 41.93~$\pm$~5.37 & 73.24~$\pm$~8.66 \\
    Color-opponency & & 42.50~$\pm$~0.80 & 51.82~$\pm$~1.16 & 83.58~$\pm$~1.29 \\
    Single-color & & 39.39~$\pm$~2.74 & 49.13~$\pm$~4.79 & 79.61~$\pm$~5.41 \\
    \midrule
    - & \multirow{4}{*}{80\%} & 5.00~$\pm$~1.95 & 9.62~$\pm$~1.85 & 25.75~$\pm$~12.60 \\
    Luminance & & 25.56~$\pm$~2.64 & 32.66~$\pm$~4.07 & 68.51~$\pm$~6.36 \\
    Color-opponency & & 40.71~$\pm$~1.20 & 49.43~$\pm$~0.40 & 79.25~$\pm$~4.57 \\
    Single-color & & 34.79~$\pm$~2.35 & 42.58~$\pm$~3.63 & 77.79~$\pm$~2.52 \\
    \midrule
    - & \multirow{4}{*}{90\%} & 4.63~$\pm$~1.67 & 8.67~$\pm$~1.71 & 23.59~$\pm$~12.14 \\
    Luminance & & 14.73~$\pm$~2.59 & 20.09~$\pm$~2.14 & 57.34~$\pm$~6.68 \\
    Color-opponency & & 32.68~$\pm$~3.06 & 39.33~$\pm$~2.48 & 73.92~$\pm$~8.43 \\
    Single-color & & 22.91~$\pm$~1.19 & 28.54~$\pm$~0.98 & 70.96~$\pm$~0.80 \\
    \bottomrule
  \end{tabular}
  }
  \label{tab:sparse_results_upernet_acdc_fog_zero_shot}
\end{table}
\clearpage

\begin{table}[tb]
  \centering
  \caption{\textbf{Zero-shot} evaluation of the \textbf{UPerNet} architecture trained on Cityscapes without sparsity and \textbf{validated on ACDC Rain} at different sparsities. Results include the three preprocessing types (luminance, color-opponency, and single-color). Metrics reported are mean Intersection over Union (mIoU), mean Accuracy (mAcc), and average Accuracy (aAcc), averaged over three seeds. \textbf{Percent-based sparsity:} For a given \textit{percentage}, the \textit{percentage} of lowest absolute values are set to zero.}   
  \scalebox{0.85}{
  \begin{tabular}{@{}lc rrr@{}}
    \toprule
    Preprocessing &  Sparsity &\multicolumn{1}{c}{mIoU}&\multicolumn{1}{c}{mAcc}&\multicolumn{1}{c}{aAcc}  \\
    \midrule
    - & \multirow{4}{*}{-} &  35.35~$\pm$~2.12 & 48.51~$\pm$~2.46 & 81.44~$\pm$~4.98 \\
     Luminance &   & 29.31~$\pm$~0.53 & 40.70~$\pm$~2.36 & 69.42~$\pm$~5.65 \\
    Color-opponency& & 34.28~$\pm$~1.23 & 45.48~$\pm$~1.77 & 76.43~$\pm$~5.73 \\
    Single-color& &  33.35~$\pm$~1.69 & 43.90~$\pm$~2.94 & 73.61~$\pm$~6.08 \\
    \midrule
    - & \multirow{4}{*}{10\%} & 28.38~$\pm$~2.27 & 42.24~$\pm$~3.28 & 75.46~$\pm$~3.14 \\
    Luminance & & 28.22~$\pm$~0.52 & 38.29~$\pm$~2.55 & 67.69~$\pm$~5.99 \\
    Color-opponency & & 33.96~$\pm$~2.67 & 44.00~$\pm$~2.42 & 75.14~$\pm$~5.91 \\
    Single-color & & 33.43~$\pm$~1.70 & 43.96~$\pm$~2.94 & 73.97~$\pm$~6.09 \\
    \midrule
    - & \multirow{4}{*}{20\%} & 22.42~$\pm$~1.75 & 35.15~$\pm$~1.95 & 68.30~$\pm$~3.71 \\
    Luminance & & 28.61~$\pm$~0.77 & 38.58~$\pm$~2.59 & 69.31~$\pm$~7.03 \\
    Color-opponency & & 33.99~$\pm$~2.67 & 44.01~$\pm$~2.42 & 75.24~$\pm$~5.92 \\
    Single-color & & 34.00~$\pm$~1.90 & 44.34~$\pm$~2.99 & 76.21~$\pm$~6.55 \\
    \midrule
    - & \multirow{4}{*}{30\%} & 16.06~$\pm$~0.96 & 24.63~$\pm$~0.66 & 62.46~$\pm$~4.30 \\
    Luminance & & 29.14~$\pm$~1.32 & 38.94~$\pm$~2.67 & 71.59~$\pm$~8.93 \\
    Color-opponency & & 34.06~$\pm$~2.67 & 44.04~$\pm$~2.42 & 75.54~$\pm$~5.92 \\
    Single-color & & 34.79~$\pm$~2.25 & 44.86~$\pm$~3.10 & 79.09~$\pm$~7.23 \\
    \midrule
    - & \multirow{4}{*}{40\%} & 11.78~$\pm$~0.78 & 19.16~$\pm$~0.79 & 56.41~$\pm$~5.53 \\
    Luminance & & 28.90~$\pm$~1.06 & 38.69~$\pm$~2.40 & 71.35~$\pm$~7.35 \\
    Color-opponency & & 34.17~$\pm$~2.72 & 44.06~$\pm$~2.44 & 76.40~$\pm$~5.91 \\
    Single-color & & 34.42~$\pm$~1.82 & 44.60~$\pm$~2.86 & 78.58~$\pm$~5.72 \\
    \midrule
    - & \multirow{4}{*}{50\%} & 8.70~$\pm$~1.18 & 15.31~$\pm$~0.64 & 49.12~$\pm$~7.52 \\
    Luminance & & 29.03~$\pm$~0.99 & 38.26~$\pm$~1.89 & 73.57~$\pm$~6.22 \\
    Color-opponency & & 34.22~$\pm$~2.67 & 43.93~$\pm$~2.38 & 77.59~$\pm$~5.78 \\
    Single-color & & 34.08~$\pm$~1.56 & 43.85~$\pm$~2.69 & 79.83~$\pm$~4.47 \\
    \midrule
    - & \multirow{4}{*}{60\%} & 6.48~$\pm$~1.62 & 12.22~$\pm$~1.62 & 40.16~$\pm$~10.24 \\
    Luminance & & 28.04~$\pm$~1.11 & 36.29~$\pm$~2.15 & 74.43~$\pm$~5.54 \\
    Color-opponency & & 34.62~$\pm$~2.36 & 43.90~$\pm$~2.16 & 80.08~$\pm$~5.18 \\
    Single-color & & 33.38~$\pm$~1.42 & 42.53~$\pm$~2.56 & 81.37~$\pm$~2.89 \\
    \midrule
    - & \multirow{4}{*}{70\%} & 4.74~$\pm$~1.98 & 9.93~$\pm$~2.67 & 30.42~$\pm$~12.85 \\
    Luminance & & 25.82~$\pm$~1.36 & 32.61~$\pm$~1.97 & 73.82~$\pm$~3.90 \\
    Color-opponency & & 33.64~$\pm$~2.34 & 42.46~$\pm$~1.98 & 79.27~$\pm$~5.97 \\
    Single-color & & 30.99~$\pm$~1.33 & 38.73~$\pm$~2.69 & 81.36~$\pm$~0.62 \\
    \midrule
    - & \multirow{4}{*}{80\%} & 4.10~$\pm$~2.21 & 9.03~$\pm$~2.87 & 25.34~$\pm$~14.06 \\
    Luminance & & 21.04~$\pm$~1.15 & 26.51~$\pm$~1.22 & 69.41~$\pm$~2.17 \\
    Color-opponency & & 31.70~$\pm$~2.05 & 40.01~$\pm$~1.43 & 77.81~$\pm$~7.54 \\
    Single-color & & 26.47~$\pm$~1.50 & 32.99~$\pm$~2.43 & 77.78~$\pm$~0.56 \\
    \midrule
    - & \multirow{4}{*}{90\%} & 4.06~$\pm$~1.96 & 9.13~$\pm$~2.52 & 23.08~$\pm$~13.04 \\
    Luminance & & 14.32~$\pm$~0.74 & 18.75~$\pm$~0.44 & 61.42~$\pm$~3.02 \\
    Color-opponency & & 25.40~$\pm$~2.84 & 31.96~$\pm$~2.78 & 73.98~$\pm$~8.45 \\
    Single-color & & 20.45~$\pm$~0.84 & 26.35~$\pm$~1.43 & 71.69~$\pm$~0.64 \\
    \bottomrule
  \end{tabular}
  }
  \label{tab:sparse_results_upernet_acdc_rain_zero_shot}
\end{table}
\clearpage

\begin{table}[tb]
  \centering
  \caption{\textbf{Zero-shot} evaluation of the \textbf{UPerNet} architecture trained on Cityscapes without sparsity and \textbf{validated on ACDC Snow} at different sparsities. Results include the three preprocessing types (luminance, color-opponency, and single-color). Metrics reported are mean Intersection over Union (mIoU), mean Accuracy (mAcc), and average Accuracy (aAcc), averaged over three seeds. \textbf{Percent-based sparsity:} For a given \textit{percentage}, the \textit{percentage} of lowest absolute values are set to zero.}   
  \scalebox{0.85}{
  \begin{tabular}{@{}lc rrr@{}}
    \toprule
    Preprocessing &  Sparsity &\multicolumn{1}{c}{mIoU}&\multicolumn{1}{c}{mAcc}&\multicolumn{1}{c}{aAcc}  \\
    \midrule
    - & \multirow{3}{*}{-} &  26.90~$\pm$~1.92 & 40.71~$\pm$~2.42 & 68.91~$\pm$~3.87 \\
     Luminance &   & 26.63~$\pm$~0.26 & 38.34~$\pm$~1.56 & 62.29~$\pm$~2.11 \\
    Color-opponency& & 26.57~$\pm$~1.75 & 37.94~$\pm$~1.75 & 68.68~$\pm$~1.71 \\
    Single-color& & 27.84~$\pm$~2.10 & 37.96~$\pm$~2.27 & 67.74~$\pm$~7.07 \\
    \midrule
    - & \multirow{4}{*}{10\%} & 25.68~$\pm$~2.06 & 38.60~$\pm$~2.56 & 66.72~$\pm$~3.71 \\
    Luminance & & 25.97~$\pm$~2.30 & 36.71~$\pm$~4.16 & 66.22~$\pm$~7.58 \\
    Color-opponency & & 26.74~$\pm$~1.62 & 37.03~$\pm$~1.48 & 69.18~$\pm$~1.68 \\
    Single-color & & 25.41~$\pm$~2.30 & 35.81~$\pm$~2.98 & 67.21~$\pm$~8.15 \\
    \midrule
    - & \multirow{4}{*}{20\%} & 21.85~$\pm$~2.65 & 33.37~$\pm$~2.84 & 60.27~$\pm$~4.57 \\
    Luminance & & 26.64~$\pm$~2.78 & 37.17~$\pm$~4.34 & 68.91~$\pm$~9.06 \\
    Color-opponency & & 26.78~$\pm$~1.62 & 37.06~$\pm$~1.46 & 69.29~$\pm$~1.66 \\
    Single-color & & 26.13~$\pm$~2.56 & 36.23~$\pm$~3.00 & 69.79~$\pm$~8.63 \\
    \midrule
    - & \multirow{4}{*}{30\%} & 17.63~$\pm$~2.84 & 27.60~$\pm$~3.19 & 54.39~$\pm$~5.41 \\
    Luminance & & 26.58~$\pm$~2.70 & 37.06~$\pm$~4.30 & 69.09~$\pm$~8.57 \\
    Color-opponency & & 26.83~$\pm$~1.60 & 37.09~$\pm$~1.45 & 69.51~$\pm$~1.62 \\
    Single-color & & 26.12~$\pm$~2.46 & 36.12~$\pm$~2.93 & 69.80~$\pm$~7.99 \\
    \midrule
    - & \multirow{4}{*}{40\%} & 13.35~$\pm$~2.47 & 21.50~$\pm$~2.67 & 50.42~$\pm$~6.14 \\
    Luminance & & 26.17~$\pm$~2.44 & 36.40~$\pm$~4.23 & 68.74~$\pm$~7.50 \\
    Color-opponency & & 27.00~$\pm$~1.58 & 37.17~$\pm$~1.44 & 70.25~$\pm$~1.47 \\
    Single-color & & 25.92~$\pm$~2.23 & 35.75~$\pm$~2.78 & 69.97~$\pm$~6.83 \\
    \midrule
    - & \multirow{4}{*}{50\%} & 10.11~$\pm$~2.03 & 17.19~$\pm$~2.00 & 47.18~$\pm$~6.46 \\
    Luminance & & 25.38~$\pm$~2.32 & 35.10~$\pm$~4.08 & 67.93~$\pm$~6.45 \\
    Color-opponency & & 27.17~$\pm$~1.46 & 37.21~$\pm$~1.39 & 71.16~$\pm$~1.12 \\
    Single-color & & 25.78~$\pm$~2.13 & 35.17~$\pm$~2.56 & 70.35~$\pm$~6.04 \\
    \midrule
    - & \multirow{4}{*}{60\%} & 7.43~$\pm$~1.47 & 13.82~$\pm$~1.13 & 40.95~$\pm$~6.93 \\
    Luminance & & 23.95~$\pm$~2.28 & 32.77~$\pm$~4.03 & 66.19~$\pm$~5.26 \\
    Color-opponency & & 27.52~$\pm$~1.14 & 37.27~$\pm$~1.19 & 72.87~$\pm$~0.23 \\
    Single-color & & 25.51~$\pm$~2.25 & 34.44~$\pm$~3.07 & 70.55~$\pm$~4.01 \\
    \midrule
    - & \multirow{4}{*}{70\%} & 5.51~$\pm$~1.43 & 11.10~$\pm$~1.88 & 33.07~$\pm$~7.09 \\
    Luminance & & 21.38~$\pm$~2.41 & 28.89~$\pm$~3.78 & 62.97~$\pm$~4.59 \\
    Color-opponency & & 26.97~$\pm$~1.19 & 36.72~$\pm$~1.15 & 72.10~$\pm$~0.44 \\
    Single-color & & 24.79~$\pm$~1.70 & 32.66~$\pm$~2.30 & 70.59~$\pm$~2.86 \\
    \midrule
    - & \multirow{4}{*}{80\%} & 4.12~$\pm$~1.48 & 8.49~$\pm$~2.30 & 24.92~$\pm$~10.41 \\
    Luminance & & 18.30~$\pm$~2.13 & 24.27~$\pm$~2.78 & 58.86~$\pm$~3.40 \\
    Color-opponency & & 26.23~$\pm$~0.88 & 36.01~$\pm$~0.63 & 71.49~$\pm$~0.84 \\
    Single-color & & 22.59~$\pm$~1.05 & 29.37~$\pm$~1.77 & 68.79~$\pm$~1.58 \\
    \midrule
    - & \multirow{4}{*}{90\%} & 3.86~$\pm$~1.19 & 7.58~$\pm$~1.86 & 21.17~$\pm$~8.69 \\
    Luminance & & 12.50~$\pm$~2.00 & 17.33~$\pm$~1.84 & 52.00~$\pm$~3.25 \\
    Color-opponency & & 22.48~$\pm$~2.37 & 30.43~$\pm$~2.34 & 66.63~$\pm$~4.87 \\
    Single-color & & 18.26~$\pm$~1.13 & 24.02~$\pm$~1.32 & 64.19~$\pm$~1.70 \\
    \bottomrule
  \end{tabular}
  }
  \label{tab:sparse_results_upernet_acdc_snow_zero_shot}
\end{table}
\clearpage

\begin{table}[tb]
  \centering
  \caption{\textbf{Zero-shot} evaluation of the \textbf{UPerNet} architecture trained on Cityscapes without sparsity and \textbf{validated on ACDC Mean} at different sparsities. Results include the three preprocessing types (luminance, color-opponency, and single-color). Metrics reported are mean Intersection over Union (mIoU), mean Accuracy (mAcc), and average Accuracy (aAcc), averaged over three seeds. \textbf{Percent-based sparsity:} For a given \textit{percentage}, the \textit{percentage} of lowest absolute values are set to zero.}
  \scalebox{0.85}{
  \begin{tabular}{@{}lc rrr@{}}
    \toprule
    Preprocessing & Sparsity &\multicolumn{1}{c}{mIoU}&\multicolumn{1}{c}{mAcc}&\multicolumn{1}{c}{aAcc}  \\
    \midrule
    - & \multirow{4}{*}{-} & 25.65~$\pm$~0.91 & 37.63~$\pm$~2.33 & 66.15~$\pm$~3.03\\
     Luminance &  & 28.34~$\pm$~0.81 & 37.54~$\pm$~2.80 & 67.73~$\pm$~4.73 \\
    Color-opponency &  & 29.15~$\pm$~1.62 & 38.43~$\pm$~1.61 & 68.14~$\pm$~2.79 \\
    Single-color &  & 28.22~$\pm$~0.89 & 38.73~$\pm$~2.45 & 67.17~$\pm$~3.82 \\
    \midrule
    - & \multirow{4}{*}{10\%} & 23.50~$\pm$~0.64 & 35.32~$\pm$~2.01 & 62.69~$\pm$~2.77 \\
    Luminance & & 28.42~$\pm$~0.85 & 37.61~$\pm$~2.80 & 68.04~$\pm$~4.83 \\
    Color-opponency & & 29.16~$\pm$~1.62 & 38.43~$\pm$~1.61 & 68.15~$\pm$~2.79 \\
    Single-color & & 28.29~$\pm$~0.90 & 38.77~$\pm$~2.44 & 67.40~$\pm$~3.82 \\
    \midrule
    - & \multirow{4}{*}{20\%} & 19.20~$\pm$~0.78 & 30.07~$\pm$~1.53 & 55.97~$\pm$~3.13 \\
    Luminance & & 28.74~$\pm$~1.16 & 37.81~$\pm$~2.87 & 69.20~$\pm$~5.75 \\
    Color-opponency & & 29.17~$\pm$~1.62 & 38.43~$\pm$~1.61 & 68.18~$\pm$~2.81 \\
    Single-color & & 28.66~$\pm$~1.02 & 38.95~$\pm$~2.45 & 68.68~$\pm$~4.03 \\
    \midrule
    - & \multirow{4}{*}{30\%} & 15.00~$\pm$~0.98 & 24.24~$\pm$~1.37 & 50.36~$\pm$~3.70 \\
    Luminance & & 28.68~$\pm$~1.50 & 37.66~$\pm$~3.03 & 69.19~$\pm$~6.72 \\
    Color-opponency & & 29.17~$\pm$~1.64 & 38.40~$\pm$~1.66 & 68.17~$\pm$~2.97 \\
    Single-color & & 28.78~$\pm$~1.14 & 38.86~$\pm$~2.41 & 68.83~$\pm$~4.14 \\
    \midrule
    - & \multirow{4}{*}{40\%} & 11.71~$\pm$~1.18 & 19.77~$\pm$~1.06 & 46.33~$\pm$~4.48 \\
    Luminance & & 28.05~$\pm$~1.38 & 36.96~$\pm$~3.08 & 67.74~$\pm$~6.10 \\
    Color-opponency & & 29.11~$\pm$~1.73 & 38.26~$\pm$~1.84 & 68.17~$\pm$~3.42 \\
    Single-color & & 28.39~$\pm$~0.90 & 38.34~$\pm$~2.21 & 67.66~$\pm$~3.17 \\
    \midrule
    - & \multirow{4}{*}{50\%} & 8.79~$\pm$~1.00 & 15.77~$\pm$~1.04 & 42.37~$\pm$~4.73 \\
    Luminance & & 27.48~$\pm$~1.45 & 36.01~$\pm$~3.11 & 67.30~$\pm$~5.88 \\
    Color-opponency & & 29.00~$\pm$~1.80 & 38.01~$\pm$~2.04 & 68.25~$\pm$~3.82 \\
    Single-color & & 27.93~$\pm$~0.79 & 37.54~$\pm$~2.17 & 67.10~$\pm$~2.10 \\
    \midrule
    - & \multirow{4}{*}{60\%} & 6.86~$\pm$~0.78 & 13.14~$\pm$~0.39 & 37.05~$\pm$~5.63 \\
    Luminance & & 26.16~$\pm$~1.75 & 34.06~$\pm$~3.36 & 65.84~$\pm$~5.94 \\
    Color-opponency & & 29.06~$\pm$~1.65 & 37.79~$\pm$~2.05 & 69.00~$\pm$~3.67 \\
    Single-color & & 27.32~$\pm$~0.52 & 36.27~$\pm$~2.04 & 66.91~$\pm$~0.41 \\
    \midrule
    - & \multirow{4}{*}{70\%} & 5.07~$\pm$~1.20 & 10.37~$\pm$~1.22 & 29.65~$\pm$~7.81 \\
    Luminance & & 23.64~$\pm$~2.13 & 30.75~$\pm$~3.42 & 63.23~$\pm$~5.37 \\
    Color-opponency & & 28.35~$\pm$~1.66 & 36.87~$\pm$~1.99 & 67.97~$\pm$~3.50 \\
    Single-color & & 25.89~$\pm$~0.24 & 33.76~$\pm$~1.63 & 66.50~$\pm$~1.65 \\
    \midrule
    - & \multirow{4}{*}{80\%} & 4.06~$\pm$~1.45 & 9.02~$\pm$~1.20 & 23.65~$\pm$~10.12 \\
    Luminance & & 19.17~$\pm$~1.99 & 25.50~$\pm$~2.64 & 58.08~$\pm$~4.51 \\
    Color-opponency & & 27.89~$\pm$~0.94 & 35.90~$\pm$~0.82 & 68.49~$\pm$~2.21 \\
    Single-color & & 23.27~$\pm$~0.36 & 30.03~$\pm$~0.76 & 64.49~$\pm$~2.76 \\
    \midrule
    - & \multirow{4}{*}{90\%} & 3.93~$\pm$~1.40 & 8.21~$\pm$~1.36 & 21.88~$\pm$~9.75 \\
    Luminance & & 12.65~$\pm$~1.66 & 17.91~$\pm$~1.57 & 49.37~$\pm$~4.13 \\
    Color-opponency & & 23.56~$\pm$~2.23 & 29.96~$\pm$~2.26 & 65.24~$\pm$~6.54 \\
    Single-color & & 17.98~$\pm$~0.42 & 23.85~$\pm$~0.10 & 59.72~$\pm$~2.51 \\
    \bottomrule
  \end{tabular}
  }
  \label{tab:sparse_results_upernet_acdc_full}
\end{table}
\clearpage

%% file: tables/sparsity_zero_shot_segformer.tex
\begin{table}[h]
  \centering
  \caption{\textbf{Zero-shot} evaluation of the \textbf{SegFormer} architecture trained on Cityscapes without sparsity and \textbf{validated on Cityscapes} at different sparsities. Results include the three preprocessing types (luminance, color-opponency, and single-color). Metrics reported are mean Intersection over Union (mIoU), mean Accuracy (mAcc), and average Accuracy (aAcc), averaged over three seeds. \textbf{Percent-based sparsity:} For a given \textit{percentage}, the \textit{percentage} of lowest absolute values are set to zero.}
  \scalebox{0.85}{
  \begin{tabular}{@{}lc rrr@{}}
    \toprule
    Preprocessing & Sparsity &\multicolumn{1}{c}{mIoU}&\multicolumn{1}{c}{mAcc}&\multicolumn{1}{c}{aAcc}  \\
    \midrule
    - & \multirow{4}{*}{-} &  73.86~$\pm$~0.29 & 82.58~$\pm$~0.59 & 95.36~$\pm$~0.03 \\
     Luminance &  & 68.73~$\pm$~0.20 & 77.97~$\pm$~0.38 & 94.30~$\pm$~0.10 \\
    Color-opponency &  & 71.26~$\pm$~0.16 & 80.22~$\pm$~0.58 & 94.80~$\pm$~0.05 \\
    Single-color &  & 71.59~$\pm$~0.67 & 80.56~$\pm$~0.59 & 94.79~$\pm$~0.09 \\
    \midrule
    - & \multirow{4}{*}{10\%} & 72.72~$\pm$~0.25 & 81.93~$\pm$~0.54 & 95.06~$\pm$~0.01 \\
    Luminance & & 68.73~$\pm$~0.20 & 77.97~$\pm$~0.38 & 94.30~$\pm$~0.10 \\
    Color-opponency & & 71.25~$\pm$~0.15 & 80.21~$\pm$~0.58 & 94.80~$\pm$~0.05 \\
    Single-color & & 71.58~$\pm$~0.67 & 80.55~$\pm$~0.60 & 94.79~$\pm$~0.09 \\
    \midrule
    - & \multirow{4}{*}{20\%} & 69.69~$\pm$~0.73 & 80.40~$\pm$~0.63 & 94.10~$\pm$~0.10 \\
    Luminance & & 68.74~$\pm$~0.19 & 77.96~$\pm$~0.36 & 94.29~$\pm$~0.10 \\
    Color-opponency & & 71.19~$\pm$~0.14 & 80.15~$\pm$~0.58 & 94.78~$\pm$~0.04 \\
    Single-color & & 71.47~$\pm$~0.63 & 80.44~$\pm$~0.59 & 94.75~$\pm$~0.09 \\
    \midrule
    - & \multirow{4}{*}{30\%} & 64.52~$\pm$~0.57 & 78.08~$\pm$~0.55 & 91.30~$\pm$~0.38 \\
    Luminance & & 68.58~$\pm$~0.19 & 77.79~$\pm$~0.35 & 94.22~$\pm$~0.10 \\
    Color-opponency & & 71.04~$\pm$~0.11 & 80.00~$\pm$~0.58 & 94.74~$\pm$~0.03 \\
    Single-color & & 71.10~$\pm$~0.53 & 80.22~$\pm$~0.51 & 94.63~$\pm$~0.08 \\
    \midrule
    - & \multirow{4}{*}{40\%} & 56.54~$\pm$~0.47 & 74.10~$\pm$~0.67 & 85.09~$\pm$~0.75 \\
    Luminance & & 67.89~$\pm$~0.10 & 77.18~$\pm$~0.28 & 93.96~$\pm$~0.17 \\
    Color-opponency & & 70.83~$\pm$~0.14 & 79.83~$\pm$~0.63 & 94.69~$\pm$~0.02 \\
    Single-color & & 70.33~$\pm$~0.45 & 79.65~$\pm$~0.42 & 94.49~$\pm$~0.08 \\
    \midrule
    - & \multirow{4}{*}{50\%} & 47.61~$\pm$~0.78 & 68.35~$\pm$~0.75 & 76.24~$\pm$~0.83 \\
    Luminance & & 66.19~$\pm$~0.26 & 75.85~$\pm$~0.17 & 93.24~$\pm$~0.37 \\
    Color-opponency & & 70.37~$\pm$~0.17 & 79.47~$\pm$~0.70 & 94.59~$\pm$~0.03 \\
    Single-color & & 69.10~$\pm$~0.40 & 78.77~$\pm$~0.44 & 94.17~$\pm$~0.07 \\
    \midrule
    - & \multirow{4}{*}{60\%} & 39.62~$\pm$~1.34 & 60.16~$\pm$~1.22 & 67.23~$\pm$~1.11 \\
    Luminance & & 62.63~$\pm$~0.84 & 72.97~$\pm$~0.23 & 91.33~$\pm$~1.01 \\
    Color-opponency & & 69.53~$\pm$~0.14 & 78.79~$\pm$~0.79 & 94.38~$\pm$~0.04 \\
    Single-color & & 66.73~$\pm$~0.24 & 77.05~$\pm$~0.64 & 93.31~$\pm$~0.06 \\
    \midrule
    - & \multirow{4}{*}{70\%} & 31.09~$\pm$~1.63 & 48.37~$\pm$~1.61 & 58.02~$\pm$~1.93 \\
    Luminance & & 56.01~$\pm$~1.46 & 67.72~$\pm$~0.35 & 86.59~$\pm$~2.43 \\
    Color-opponency & & 68.10~$\pm$~0.12 & 77.48~$\pm$~0.88 & 93.98~$\pm$~0.03 \\
    Single-color & & 61.45~$\pm$~0.04 & 73.46~$\pm$~0.53 & 90.69~$\pm$~0.28 \\
    \midrule
    - & \multirow{4}{*}{80\%} & 20.19~$\pm$~1.72 & 32.48~$\pm$~1.80 & 47.23~$\pm$~2.80 \\
    Luminance & & 45.20~$\pm$~2.14 & 58.31~$\pm$~0.87 & 77.21~$\pm$~3.77 \\
    Color-opponency & & 64.40~$\pm$~0.37 & 73.99~$\pm$~0.98 & 92.72~$\pm$~0.10 \\
    Single-color & & 50.33~$\pm$~0.46 & 65.03~$\pm$~0.24 & 83.26~$\pm$~0.24 \\
    \midrule
    - & \multirow{4}{*}{90\%} & 9.70~$\pm$~0.98 & 17.40~$\pm$~0.76 & 36.14~$\pm$~3.95 \\
    Luminance & & 27.86~$\pm$~1.47 & 38.58~$\pm$~0.05 & 61.60~$\pm$~3.88 \\
    Color-opponency & & 50.51~$\pm$~0.79 & 62.30~$\pm$~1.30 & 84.03~$\pm$~1.15 \\
    Single-color & & 30.35~$\pm$~0.63 & 44.47~$\pm$~1.11 & 65.21~$\pm$~0.97 \\
    \bottomrule
  \end{tabular}
  }
  \label{tab:sparse_results_segformer_cityscapes_zero_shot}
\end{table}
\clearpage

\begin{table}[tb]
  \centering
  \caption{\textbf{Zero-shot} evaluation of the \textbf{SegFormer} architecture trained on Cityscapes without sparsity and \textbf{validated on Dark Zurich} at different sparsities. Results include the three preprocessing types (luminance, color-opponency, and single-color). Metrics reported are mean Intersection over Union (mIoU), mean Accuracy (mAcc), and average Accuracy (aAcc), averaged over three seeds. \textbf{Percent-based sparsity:} For a given \textit{percentage}, the \textit{percentage} of lowest absolute values are set to zero.}
  \scalebox{0.85}{
  \begin{tabular}{@{}lc rrr@{}}
    \toprule
    Preprocessing & Sparsity &\multicolumn{1}{c}{mIoU}&\multicolumn{1}{c}{mAcc}&\multicolumn{1}{c}{aAcc}  \\
    \midrule
    - & \multirow{4}{*}{-} &  13.95~$\pm$~0.79 & 26.98~$\pm$~1.19 & 47.96~$\pm$~0.57 \\
     Luminance &  & 19.99~$\pm$~0.31 & 33.67~$\pm$~1.31 & 54.02~$\pm$~1.98 \\
    Color-opponency &  & 17.50~$\pm$~0.82 & 31.33~$\pm$~0.12 & 52.46~$\pm$~1.01 \\
    Single-color &  & 18.76~$\pm$~0.13 & 33.11~$\pm$~0.19 & 53.88~$\pm$~1.79 \\
    \midrule
    - & \multirow{4}{*}{10\%} & 14.46~$\pm$~0.94 & 28.12~$\pm$~1.86 & 44.74~$\pm$~1.68 \\
    Luminance & & 19.98~$\pm$~0.32 & 33.69~$\pm$~1.31 & 53.99~$\pm$~1.98 \\
    Color-opponency & & 17.50~$\pm$~0.82 & 31.34~$\pm$~0.12 & 52.46~$\pm$~1.01 \\
    Single-color & & 18.75~$\pm$~0.13 & 33.09~$\pm$~0.21 & 53.85~$\pm$~1.79 \\
    \midrule
    - & \multirow{4}{*}{20\%} & 11.45~$\pm$~0.50 & 23.43~$\pm$~1.60 & 36.48~$\pm$~1.12 \\
    Luminance & & 19.97~$\pm$~0.32 & 33.64~$\pm$~1.39 & 53.84~$\pm$~1.99 \\
    Color-opponency & & 17.50~$\pm$~0.83 & 31.33~$\pm$~0.11 & 52.48~$\pm$~1.01 \\
    Single-color & & 18.72~$\pm$~0.13 & 32.99~$\pm$~0.23 & 53.77~$\pm$~1.72 \\
    \midrule
    - & \multirow{4}{*}{30\%} & 9.98~$\pm$~0.52 & 20.33~$\pm$~1.41 & 32.23~$\pm$~1.22 \\
    Luminance & & 20.09~$\pm$~0.31 & 33.54~$\pm$~1.42 & 53.53~$\pm$~2.10 \\
    Color-opponency & & 17.48~$\pm$~0.85 & 31.28~$\pm$~0.10 & 52.50~$\pm$~1.00 \\
    Single-color & & 18.81~$\pm$~0.12 & 32.87~$\pm$~0.34 & 53.65~$\pm$~1.60 \\
    \midrule
    - & \multirow{4}{*}{40\%} & 8.76~$\pm$~0.74 & 17.77~$\pm$~1.58 & 31.44~$\pm$~2.01 \\
    Luminance & & 20.17~$\pm$~0.59 & 33.26~$\pm$~1.72 & 53.31~$\pm$~2.22 \\
    Color-opponency & & 17.43~$\pm$~0.88 & 31.09~$\pm$~0.10 & 52.42~$\pm$~0.92 \\
    Single-color & & 18.82~$\pm$~0.09 & 32.63~$\pm$~0.43 & 53.39~$\pm$~1.55 \\
    \midrule
    - & \multirow{4}{*}{50\%} & 7.33~$\pm$~0.77 & 15.30~$\pm$~1.34 & 28.52~$\pm$~2.68 \\
    Luminance & & 19.99~$\pm$~0.76 & 32.80~$\pm$~1.72 & 52.76~$\pm$~2.46 \\
    Color-opponency & & 17.37~$\pm$~0.83 & 30.80~$\pm$~0.24 & 52.24~$\pm$~0.72 \\
    Single-color & & 18.73~$\pm$~0.09 & 32.36~$\pm$~0.33 & 52.95~$\pm$~1.38 \\
    \midrule
    - & \multirow{4}{*}{60\%} & 6.48~$\pm$~0.86 & 13.27~$\pm$~1.00 & 27.60~$\pm$~2.32 \\
    Luminance & & 19.56~$\pm$~0.44 & 32.06~$\pm$~1.52 & 52.01~$\pm$~2.50 \\
    Color-opponency & & 17.38~$\pm$~0.76 & 30.51~$\pm$~0.38 & 51.99~$\pm$~0.57 \\
    Single-color & & 18.53~$\pm$~0.09 & 31.86~$\pm$~0.47 & 52.57~$\pm$~1.20 \\
    \midrule
    - & \multirow{4}{*}{70\%} & 5.92~$\pm$~0.20 & 12.19~$\pm$~0.53 & 28.07~$\pm$~0.66 \\
    Luminance & & 18.59~$\pm$~0.88 & 30.38~$\pm$~1.48 & 50.94~$\pm$~2.21 \\
    Color-opponency & & 17.54~$\pm$~0.73 & 30.47~$\pm$~0.75 & 51.97~$\pm$~0.28 \\
    Single-color & & 18.10~$\pm$~0.23 & 30.94~$\pm$~0.54 & 51.94~$\pm$~0.77 \\
    \midrule
    - & \multirow{4}{*}{80\%} & 5.08~$\pm$~0.18 & 10.85~$\pm$~0.73 & 26.89~$\pm$~1.43 \\
    Luminance & & 16.85~$\pm$~0.56 & 27.71~$\pm$~0.81 & 48.73~$\pm$~1.97 \\
    Color-opponency & & 17.62~$\pm$~0.37 & 30.05~$\pm$~1.00 & 52.51~$\pm$~0.32 \\
    Single-color & & 16.90~$\pm$~0.32 & 28.74~$\pm$~1.14 & 50.12~$\pm$~1.12 \\
    \midrule
    - & \multirow{4}{*}{90\%} & 3.06~$\pm$~0.49 & 8.21~$\pm$~0.77 & 21.88~$\pm$~2.34 \\
    Luminance & & 12.68~$\pm$~0.23 & 21.58~$\pm$~0.56 & 42.18~$\pm$~2.12 \\
    Color-opponency & & 15.91~$\pm$~0.27 & 26.91~$\pm$~0.99 & 50.81~$\pm$~0.83 \\
    Single-color & & 12.97~$\pm$~0.85 & 22.60~$\pm$~1.58 & 43.02~$\pm$~2.42 \\
    \bottomrule
  \end{tabular}
  }
  \label{tab:sparse_results_segformer_dark_zurich}
\end{table}
\clearpage

\begin{table}[tb]
  \centering
  \caption{\textbf{Zero-shot} evaluation of the \textbf{SegFormer} architecture trained on Cityscapes without sparsity and \textbf{validated on ACDC Night} at different sparsities. Results include the three preprocessing types (luminance, color-opponency, and single-color). Metrics reported are mean Intersection over Union (mIoU), mean Accuracy (mAcc), and average Accuracy (aAcc), averaged over three seeds. \textbf{Percent-based sparsity:} For a given \textit{percentage}, the \textit{percentage} of lowest absolute values are set to zero.}
  \scalebox{0.85}{
  \begin{tabular}{@{}lc rrr@{}}
    \toprule
    Preprocessing & Sparsity &\multicolumn{1}{c}{mIoU}&\multicolumn{1}{c}{mAcc}&\multicolumn{1}{c}{aAcc}  \\
    \midrule
     - & \multirow{3}{*}{-} &  15.76~$\pm$~1.09 & 28.11~$\pm$~1.16 & 50.68~$\pm$~1.11 \\
     Luminance &  & 22.84~$\pm$~0.50 & 35.40~$\pm$~1.27 & 58.25~$\pm$~1.50 \\
    Color-opponency &  & 18.40~$\pm$~0.51 & 31.41~$\pm$~1.63 & 55.52~$\pm$~0.89 \\
    Single-color &  & 19.78~$\pm$~0.13 & 32.70~$\pm$~0.48 & 56.70~$\pm$~0.86 \\
    \midrule
    - & \multirow{4}{*}{10\%} & 15.10~$\pm$~1.07 & 28.17~$\pm$~1.70 & 46.52~$\pm$~1.73 \\
    Luminance & & 22.84~$\pm$~0.51 & 35.40~$\pm$~1.28 & 58.24~$\pm$~1.51 \\
    Color-opponency & & 18.40~$\pm$~0.51 & 31.41~$\pm$~1.63 & 55.52~$\pm$~0.89 \\
    Single-color & & 19.77~$\pm$~0.12 & 32.70~$\pm$~0.45 & 56.68~$\pm$~0.86 \\
    \midrule
    - & \multirow{4}{*}{20\%} & 12.16~$\pm$~0.61 & 24.19~$\pm$~1.42 & 38.40~$\pm$~1.15 \\
    Luminance & & 22.84~$\pm$~0.56 & 35.33~$\pm$~1.31 & 58.11~$\pm$~1.52 \\
    Color-opponency & & 18.40~$\pm$~0.51 & 31.41~$\pm$~1.62 & 55.53~$\pm$~0.90 \\
    Single-color & & 19.73~$\pm$~0.15 & 32.58~$\pm$~0.44 & 56.59~$\pm$~0.83 \\
    \midrule
    - & \multirow{4}{*}{30\%} & 10.69~$\pm$~0.39 & 21.11~$\pm$~1.10 & 34.34~$\pm$~1.07 \\
    Luminance & & 22.79~$\pm$~0.67 & 35.08~$\pm$~1.27 & 57.82~$\pm$~1.59 \\
    Color-opponency & & 18.39~$\pm$~0.51 & 31.38~$\pm$~1.60 & 55.55~$\pm$~0.90 \\
    Single-color & & 19.77~$\pm$~0.20 & 32.46~$\pm$~0.43 & 56.48~$\pm$~0.78 \\
    \midrule
    - & \multirow{4}{*}{40\%} & 9.25~$\pm$~0.41 & 17.57~$\pm$~1.28 & 34.10~$\pm$~2.03 \\
    Luminance & & 22.69~$\pm$~0.80 & 34.66~$\pm$~1.32 & 57.53~$\pm$~1.71 \\
    Color-opponency & & 18.32~$\pm$~0.51 & 31.19~$\pm$~1.59 & 55.50~$\pm$~0.85 \\
    Single-color & & 19.69~$\pm$~0.21 & 32.30~$\pm$~0.43 & 56.20~$\pm$~0.70 \\
    \midrule
    - & \multirow{4}{*}{50\%} & 7.95~$\pm$~0.78 & 14.97~$\pm$~1.52 & 32.61~$\pm$~2.73 \\
    Luminance & & 22.24~$\pm$~0.73 & 33.83~$\pm$~1.21 & 56.91~$\pm$~1.87 \\
    Color-opponency & & 18.29~$\pm$~0.48 & 30.98~$\pm$~1.67 & 55.36~$\pm$~0.72 \\
    Single-color & & 19.48~$\pm$~0.23 & 31.90~$\pm$~0.51 & 55.83~$\pm$~0.54 \\
    \midrule
    - & \multirow{4}{*}{60\%} & 6.89~$\pm$~0.81 & 12.66~$\pm$~1.02 & 30.36~$\pm$~2.28 \\
    Luminance & & 21.56~$\pm$~0.59 & 32.60~$\pm$~0.89 & 56.05~$\pm$~1.97 \\
    Color-opponency & & 18.22~$\pm$~0.45 & 30.70~$\pm$~1.73 & 55.24~$\pm$~0.59 \\
    Single-color & & 19.30~$\pm$~0.19 & 31.42~$\pm$~0.56 & 55.53~$\pm$~0.49 \\
    \midrule
    - & \multirow{4}{*}{70\%} & 5.94~$\pm$~0.37 & 10.98~$\pm$~0.43 & 27.59~$\pm$~0.92 \\
    Luminance & & 20.28~$\pm$~0.50 & 30.73~$\pm$~0.53 & 54.81~$\pm$~1.75 \\
    Color-opponency & & 18.33~$\pm$~0.11 & 30.45~$\pm$~1.96 & 55.31~$\pm$~0.19 \\
    Single-color & & 18.56~$\pm$~0.35 & 30.18~$\pm$~0.21 & 54.82~$\pm$~0.78 \\
    \midrule
    - & \multirow{4}{*}{80\%} & 4.91~$\pm$~0.48 & 9.68~$\pm$~0.70 & 26.94~$\pm$~1.55 \\
    Luminance & & 18.00~$\pm$~0.30 & 27.59~$\pm$~0.78 & 52.48~$\pm$~1.68 \\
    Color-opponency & & 18.53~$\pm$~0.18 & 30.01~$\pm$~2.11 & 55.93~$\pm$~0.43 \\
    Single-color & & 16.80~$\pm$~1.06 & 27.91~$\pm$~0.54 & 52.41~$\pm$~2.08 \\
    \midrule
    - & \multirow{4}{*}{90\%} & 2.89~$\pm$~0.58 & 7.40~$\pm$~0.54 & 22.56~$\pm$~1.38 \\
    Luminance & & 13.17~$\pm$~0.18 & 21.03~$\pm$~0.55 & 45.50~$\pm$~2.19 \\
    Color-opponency & & 17.00~$\pm$~0.26 & 27.60~$\pm$~1.78 & 54.27~$\pm$~0.90 \\
    Single-color & & 12.49~$\pm$~1.97 & 21.81~$\pm$~1.41 & 44.98~$\pm$~4.37 \\
    \bottomrule
  \end{tabular}
  }
  \label{tab:sparse_results_segformer_acdc_night}
\end{table}
\clearpage

\begin{table}[tb]
  \centering
  \caption{\textbf{Zero-shot} evaluation of the \textbf{SegFormer} architecture trained on Cityscapes without sparsity and \textbf{validated on ADCD Fog} at different sparsities. Results include the three preprocessing types (luminance, color-opponency, and single-color). Metrics reported are mean Intersection over Union (mIoU), mean Accuracy (mAcc), and average Accuracy (aAcc), averaged over three seeds. \textbf{Percent-based sparsity:} For a given \textit{percentage}, the \textit{percentage} of lowest absolute values are set to zero.}
  \scalebox{0.85}{
  \begin{tabular}{@{}lc rrr@{}}
    \toprule
    Preprocessing & Sparsity &\multicolumn{1}{c}{mIoU}&\multicolumn{1}{c}{mAcc}&\multicolumn{1}{c}{aAcc}  \\
    \midrule
    - & \multirow{4}{*}{-} &  63.67~$\pm$~0.94 & 74.53~$\pm$~0.83 & 92.79~$\pm$~0.19 \\
     Luminance &  & 52.70~$\pm$~3.44 & 63.26~$\pm$~2.82 & 87.70~$\pm$~2.11 \\
    Color-opponency &  & 57.79~$\pm$~1.80 & 67.87~$\pm$~2.60 & 90.20~$\pm$~0.80 \\
    Single-color &  & 54.55~$\pm$~0.14 & 66.95~$\pm$~0.67 & 90.42~$\pm$~0.54 \\
    \midrule
    - & \multirow{4}{*}{10\%} & 61.82~$\pm$~0.75 & 74.08~$\pm$~0.65 & 92.03~$\pm$~0.30 \\
    Luminance & & 52.68~$\pm$~3.44 & 63.22~$\pm$~2.82 & 87.79~$\pm$~2.07 \\
    Color-opponency & & 57.79~$\pm$~1.80 & 67.87~$\pm$~2.60 & 90.20~$\pm$~0.80 \\
    Single-color & & 54.55~$\pm$~0.15 & 66.92~$\pm$~0.67 & 90.44~$\pm$~0.53 \\
    \midrule
    - & \multirow{4}{*}{20\%} & 51.93~$\pm$~1.00 & 67.53~$\pm$~1.96 & 86.52~$\pm$~0.76 \\
    Luminance & & 52.80~$\pm$~3.42 & 63.24~$\pm$~2.82 & 88.23~$\pm$~1.95 \\
    Color-opponency & & 57.78~$\pm$~1.81 & 67.86~$\pm$~2.60 & 90.22~$\pm$~0.80 \\
    Single-color & & 54.58~$\pm$~0.23 & 66.91~$\pm$~0.71 & 90.52~$\pm$~0.47 \\
    \midrule
    - & \multirow{4}{*}{30\%} & 42.20~$\pm$~1.72 & 58.86~$\pm$~1.91 & 77.74~$\pm$~0.89 \\
    Luminance & & 52.63~$\pm$~3.39 & 62.94~$\pm$~2.81 & 88.76~$\pm$~1.78 \\
    Color-opponency & & 57.75~$\pm$~1.84 & 67.79~$\pm$~2.62 & 90.25~$\pm$~0.76 \\
    Single-color & & 54.25~$\pm$~0.23 & 66.55~$\pm$~0.82 & 90.50~$\pm$~0.39 \\
    \midrule
    - & \multirow{4}{*}{40\%} & 33.45~$\pm$~1.05 & 49.84~$\pm$~0.97 & 70.40~$\pm$~0.77 \\
    Luminance & & 52.37~$\pm$~3.32 & 62.58~$\pm$~2.85 & 89.09~$\pm$~1.57 \\
    Color-opponency & & 57.65~$\pm$~1.85 & 67.60~$\pm$~2.64 & 90.33~$\pm$~0.67 \\
    Single-color & & 53.86~$\pm$~0.17 & 66.20~$\pm$~0.57 & 90.31~$\pm$~0.32 \\
    \midrule
    - & \multirow{4}{*}{50\%} & 26.89~$\pm$~0.56 & 38.36~$\pm$~0.59 & 64.33~$\pm$~0.24 \\
    Luminance & & 51.21~$\pm$~3.40 & 61.55~$\pm$~2.96 & 88.65~$\pm$~1.43 \\
    Color-opponency & & 57.39~$\pm$~1.98 & 67.19~$\pm$~2.75 & 90.41~$\pm$~0.54 \\
    Single-color & & 52.77~$\pm$~0.57 & 65.16~$\pm$~0.81 & 89.81~$\pm$~0.29 \\
    \midrule
    - & \multirow{4}{*}{60\%} & 16.79~$\pm$~1.36 & 23.68~$\pm$~2.18 & 60.21~$\pm$~0.90 \\
    Luminance & & 48.32~$\pm$~2.72 & 59.15~$\pm$~2.48 & 86.71~$\pm$~1.80 \\
    Color-opponency & & 56.92~$\pm$~2.06 & 66.60~$\pm$~2.86 & 90.35~$\pm$~0.47 \\
    Single-color & & 49.59~$\pm$~0.72 & 61.96~$\pm$~0.71 & 88.72~$\pm$~0.33 \\
    \midrule
    - & \multirow{4}{*}{70\%} & 12.65~$\pm$~0.79 & 17.70~$\pm$~0.80 & 57.43~$\pm$~1.66 \\
    Luminance & & 42.22~$\pm$~2.12 & 53.46~$\pm$~2.11 & 82.23~$\pm$~2.45 \\
    Color-opponency & & 55.80~$\pm$~2.09 & 65.34~$\pm$~2.87 & 90.11~$\pm$~0.54 \\
    Single-color & & 44.50~$\pm$~1.04 & 56.71~$\pm$~0.51 & 86.04~$\pm$~0.89 \\
    \midrule
    - & \multirow{4}{*}{80\%} & 9.19~$\pm$~0.40 & 13.50~$\pm$~0.34 & 53.86~$\pm$~1.73 \\
    Luminance & & 33.89~$\pm$~1.35 & 44.29~$\pm$~1.36 & 76.04~$\pm$~3.24 \\
    Color-opponency & & 52.83~$\pm$~2.16 & 62.11~$\pm$~2.98 & 89.50~$\pm$~0.89 \\
    Single-color & & 36.72~$\pm$~1.69 & 47.91~$\pm$~0.34 & 80.80~$\pm$~2.84 \\
    \midrule
    - & \multirow{4}{*}{90\%} & 6.22~$\pm$~0.76 & 10.05~$\pm$~0.96 & 45.19~$\pm$~1.55 \\
    Luminance & & 21.11~$\pm$~1.09 & 28.01~$\pm$~0.45 & 66.01~$\pm$~3.15 \\
    Color-opponency & & 43.07~$\pm$~3.07 & 51.39~$\pm$~3.57 & 82.99~$\pm$~2.50 \\
    Single-color & & 23.39~$\pm$~3.28 & 30.96~$\pm$~2.63 & 70.53~$\pm$~7.11 \\
    \bottomrule
  \end{tabular}
  }
  \label{tab:sparse_results_segformer_acdc_fog}
\end{table}
\clearpage

\begin{table}[tb]
  \centering
  \caption{\textbf{Zero-shot} evaluation of the \textbf{SegFormer} architecture trained on Cityscapes without sparsity and \textbf{validated on ACDC Rain} at different sparsities. Results include the three preprocessing types (luminance, color-opponency, and single-color). Metrics reported are mean Intersection over Union (mIoU), mean Accuracy (mAcc), and average Accuracy (aAcc), averaged over three seeds. \textbf{Percent-based sparsity:} For a given \textit{percentage}, the \textit{percentage} of lowest absolute values are set to zero.}
  \scalebox{0.85}{
  \begin{tabular}{@{}lc rrr@{}}
    \toprule
    Preprocessing & Sparsity &\multicolumn{1}{c}{mIoU}&\multicolumn{1}{c}{mAcc}&\multicolumn{1}{c}{aAcc}  \\
    \midrule
    - & \multirow{4}{*}{-} &  45.50~$\pm$~0.87 & 64.84~$\pm$~1.90 & 87.16~$\pm$~0.69 \\
     Luminance &  & 38.12~$\pm$~0.91 & 50.77~$\pm$~1.20 & 82.60~$\pm$~2.43 \\
    Color-opponency &  & 43.38~$\pm$~2.29 & 57.70~$\pm$~5.15 & 86.78~$\pm$~1.07 \\
    Single-color &  & 42.22~$\pm$~1.25 & 58.07~$\pm$~1.27 & 85.91~$\pm$~1.78 \\
    \midrule
    - & \multirow{4}{*}{10\%} & 44.59~$\pm$~0.74 & 64.66~$\pm$~1.57 & 85.74~$\pm$~0.45 \\
    Luminance & & 38.16~$\pm$~0.89 & 50.79~$\pm$~1.21 & 82.73~$\pm$~2.38 \\
    Color-opponency & & 43.38~$\pm$~2.28 & 57.71~$\pm$~5.15 & 86.78~$\pm$~1.07 \\
    Single-color & & 42.24~$\pm$~1.25 & 58.07~$\pm$~1.28 & 85.93~$\pm$~1.76 \\
    \midrule
    - & \multirow{4}{*}{20\%} & 38.81~$\pm$~0.33 & 56.55~$\pm$~0.32 & 81.20~$\pm$~0.53 \\
    Luminance & & 38.43~$\pm$~0.78 & 50.91~$\pm$~1.22 & 83.43~$\pm$~2.15 \\
    Color-opponency & & 43.39~$\pm$~2.29 & 57.71~$\pm$~5.15 & 86.78~$\pm$~1.07 \\
    Single-color & & 42.26~$\pm$~1.22 & 58.01~$\pm$~1.32 & 86.02~$\pm$~1.70 \\
    \midrule
    - & \multirow{4}{*}{30\%} & 30.91~$\pm$~1.19 & 44.39~$\pm$~1.51 & 74.89~$\pm$~0.41 \\
    Luminance & & 38.90~$\pm$~0.65 & 51.16~$\pm$~1.26 & 84.37~$\pm$~1.89 \\
    Color-opponency & & 43.38~$\pm$~2.29 & 57.65~$\pm$~5.15 & 86.81~$\pm$~1.04 \\
    Single-color & & 42.27~$\pm$~1.23 & 57.89~$\pm$~1.48 & 86.12~$\pm$~1.63 \\
    \midrule
    - & \multirow{4}{*}{40\%} & 22.80~$\pm$~1.39 & 31.61~$\pm$~1.87 & 69.32~$\pm$~0.48 \\
    Luminance & & 39.20~$\pm$~0.61 & 51.13~$\pm$~1.37 & 84.95~$\pm$~1.66 \\
    Color-opponency & & 43.29~$\pm$~2.32 & 57.43~$\pm$~5.19 & 86.84~$\pm$~0.98 \\
    Single-color & & 42.14~$\pm$~1.16 & 57.38~$\pm$~1.54 & 86.20~$\pm$~1.51 \\
    \midrule
    - & \multirow{4}{*}{50\%} & 17.65~$\pm$~0.92 & 24.25~$\pm$~0.69 & 65.73~$\pm$~0.42 \\
    Luminance & & 38.66~$\pm$~0.40 & 49.82~$\pm$~1.18 & 84.80~$\pm$~1.57 \\
    Color-opponency & & 43.14~$\pm$~2.39 & 57.04~$\pm$~5.28 & 86.85~$\pm$~0.91 \\
    Single-color & & 41.56~$\pm$~0.96 & 56.25~$\pm$~1.56 & 86.01~$\pm$~1.33 \\
    \midrule
    - & \multirow{4}{*}{60\%} & 13.71~$\pm$~0.29 & 19.21~$\pm$~0.05 & 62.82~$\pm$~0.63 \\
    Luminance & & 36.08~$\pm$~0.13 & 45.87~$\pm$~0.63 & 83.18~$\pm$~1.91 \\
    Color-opponency & & 42.76~$\pm$~2.46 & 56.30~$\pm$~5.22 & 86.75~$\pm$~0.85 \\
    Single-color & & 39.67~$\pm$~0.96 & 53.31~$\pm$~1.14 & 84.89~$\pm$~1.31 \\
    \midrule
    - & \multirow{4}{*}{70\%} & 10.97~$\pm$~0.18 & 15.27~$\pm$~0.24 & 60.23~$\pm$~1.52 \\
    Luminance & & 32.04~$\pm$~0.48 & 41.01~$\pm$~0.41 & 79.59~$\pm$~2.09 \\
    Color-opponency & & 41.60~$\pm$~2.57 & 54.59~$\pm$~5.31 & 86.35~$\pm$~0.91 \\
    Single-color & & 35.60~$\pm$~0.81 & 47.19~$\pm$~1.20 & 82.09~$\pm$~1.49 \\
    \midrule
    - & \multirow{4}{*}{80\%} & 9.00~$\pm$~0.84 & 12.99~$\pm$~1.15 & 56.88~$\pm$~2.13 \\
    Luminance & & 25.09~$\pm$~0.59 & 32.32~$\pm$~0.49 & 72.97~$\pm$~1.49 \\
    Color-opponency & & 39.19~$\pm$~2.90 & 50.58~$\pm$~5.34 & 85.62~$\pm$~0.99 \\
    Single-color & & 27.94~$\pm$~1.08 & 36.63~$\pm$~0.30 & 76.73~$\pm$~2.34 \\
    \midrule
    - & \multirow{4}{*}{90\%} & 6.58~$\pm$~0.76 & 10.58~$\pm$~1.08 & 47.02~$\pm$~1.51 \\
    Luminance & & 16.71~$\pm$~0.47 & 22.05~$\pm$~0.73 & 66.02~$\pm$~1.59 \\
    Color-opponency & & 30.25~$\pm$~3.98 & 38.24~$\pm$~5.34 & 79.24~$\pm$~2.08 \\
    Single-color & & 18.96~$\pm$~0.84 & 25.02~$\pm$~0.84 & 68.38~$\pm$~2.93 \\
    \bottomrule
  \end{tabular}
  }
  \label{tab:sparse_results_segformer_acdc_rain}
\end{table}
\clearpage

\begin{table}[tb]
  \centering
  \caption{\textbf{Zero-shot} evaluation of the \textbf{SegFormer} architecture trained on Cityscapes without sparsity and \textbf{validated on ACDC Snow} at different sparsities. Results include the three preprocessing types (luminance, color-opponency, and single-color). Metrics reported are mean Intersection over Union (mIoU), mean Accuracy (mAcc), and average Accuracy (aAcc), averaged over three seeds. \textbf{Percent-based sparsity:} For a given \textit{percentage}, the \textit{percentage} of lowest absolute values are set to zero.}
  \scalebox{0.85}{
  \begin{tabular}{@{}lc rrr@{}}
    \toprule
    Preprocessing & Sparsity &\multicolumn{1}{c}{mIoU}&\multicolumn{1}{c}{mAcc}&\multicolumn{1}{c}{aAcc}  \\
    \midrule
    - & \multirow{4}{*}{-} &  46.92~$\pm$~0.51 & 57.57~$\pm$~0.80 & 86.05~$\pm$~0.19 \\
     Luminance &  & 38.94~$\pm$~1.05 & 47.85~$\pm$~1.28 & 78.59~$\pm$~2.57 \\
    Color-opponency &  & 42.60~$\pm$~0.63 & 50.74~$\pm$~1.28 & 83.58~$\pm$~1.17 \\
    Single-color &  & 41.89~$\pm$~2.11 & 50.58~$\pm$~1.07 & 83.20~$\pm$~2.76 \\
    \midrule
    - & \multirow{4}{*}{10\%} & 46.19~$\pm$~0.54 & 57.79~$\pm$~0.45 & 85.48~$\pm$~0.23 \\
    Luminance & & 38.99~$\pm$~1.04 & 47.87~$\pm$~1.26 & 78.75~$\pm$~2.55 \\
    Color-opponency & & 42.60~$\pm$~0.63 & 50.74~$\pm$~1.28 & 83.59~$\pm$~1.16 \\
    Single-color & & 41.89~$\pm$~2.11 & 50.58~$\pm$~1.07 & 83.23~$\pm$~2.75 \\
    \midrule
    - & \multirow{4}{*}{20\%} & 41.51~$\pm$~0.68 & 53.83~$\pm$~1.13 & 81.98~$\pm$~0.80 \\
    Luminance & & 39.24~$\pm$~0.95 & 48.03~$\pm$~1.22 & 79.53~$\pm$~2.48 \\
    Color-opponency & & 42.61~$\pm$~0.63 & 50.75~$\pm$~1.28 & 83.60~$\pm$~1.16 \\
    Single-color & & 41.93~$\pm$~2.08 & 50.60~$\pm$~1.05 & 83.37~$\pm$~2.65 \\
    \midrule
    - & \multirow{4}{*}{30\%} & 35.28~$\pm$~1.40 & 47.06~$\pm$~1.76 & 76.54~$\pm$~1.15 \\
    Luminance & & 39.55~$\pm$~0.85 & 48.14~$\pm$~1.18 & 80.57~$\pm$~2.30 \\
    Color-opponency & & 42.61~$\pm$~0.64 & 50.75~$\pm$~1.28 & 83.64~$\pm$~1.13 \\
    Single-color & & 42.02~$\pm$~2.05 & 50.57~$\pm$~0.99 & 83.54~$\pm$~2.47 \\
    \midrule
    - & \multirow{4}{*}{40\%} & 27.93~$\pm$~1.63 & 38.20~$\pm$~2.20 & 70.80~$\pm$~1.00 \\
    Luminance & & 39.19~$\pm$~0.74 & 47.56~$\pm$~1.04 & 80.82~$\pm$~2.19 \\
    Color-opponency & & 42.62~$\pm$~0.67 & 50.75~$\pm$~1.28 & 83.73~$\pm$~1.06 \\
    Single-color & & 41.38~$\pm$~1.83 & 49.71~$\pm$~0.79 & 83.44~$\pm$~2.29 \\
    \midrule
    - & \multirow{4}{*}{50\%} & 20.79~$\pm$~1.14 & 28.53~$\pm$~1.44 & 65.53~$\pm$~0.28 \\
    Luminance & & 37.56~$\pm$~0.95 & 45.74~$\pm$~0.79 & 79.43~$\pm$~2.65 \\
    Color-opponency & & 42.54~$\pm$~0.74 & 50.66~$\pm$~1.27 & 83.83~$\pm$~0.94 \\
    Single-color & & 40.09~$\pm$~1.74 & 48.07~$\pm$~0.99 & 82.58~$\pm$~2.30 \\
    \midrule
    - & \multirow{4}{*}{60\%} & 15.05~$\pm$~0.33 & 21.05~$\pm$~0.77 & 60.77~$\pm$~1.28 \\
    Luminance & & 34.64~$\pm$~1.22 & 42.46~$\pm$~0.64 & 76.69~$\pm$~3.38 \\
    Color-opponency & & 42.32~$\pm$~0.80 & 50.38~$\pm$~1.34 & 83.85~$\pm$~0.86 \\
    Single-color & & 37.61~$\pm$~1.82 & 45.31~$\pm$~1.13 & 80.81~$\pm$~2.69 \\
    \midrule
    - & \multirow{4}{*}{70\%} & 12.51~$\pm$~0.64 & 18.10~$\pm$~1.28 & 57.34~$\pm$~2.13 \\
    Luminance & & 30.17~$\pm$~1.58 & 37.22~$\pm$~1.02 & 71.88~$\pm$~4.35 \\
    Color-opponency & & 41.93~$\pm$~0.79 & 49.85~$\pm$~1.43 & 83.86~$\pm$~0.86 \\
    Single-color & & 33.99~$\pm$~2.69 & 41.11~$\pm$~2.05 & 77.66~$\pm$~3.86 \\
    \midrule
    - & \multirow{4}{*}{80\%} & 10.45~$\pm$~1.02 & 15.13~$\pm$~1.70 & 55.23~$\pm$~2.48 \\
    Luminance & & 24.58~$\pm$~1.67 & 30.94~$\pm$~1.16 & 65.31~$\pm$~3.53 \\
    Color-opponency & & 40.13~$\pm$~1.33 & 47.57~$\pm$~2.07 & 83.16~$\pm$~1.01 \\
    Single-color & & 28.21~$\pm$~3.53 & 34.52~$\pm$~3.05 & 71.81~$\pm$~5.30 \\
    \midrule
    - & \multirow{4}{*}{90\%} & 7.03~$\pm$~1.13 & 11.04~$\pm$~1.47 & 47.37~$\pm$~2.26 \\
    Luminance & & 16.85~$\pm$~1.05 & 22.29~$\pm$~1.17 & 58.75~$\pm$~3.41 \\
    Color-opponency & & 31.19~$\pm$~2.59 & 37.47~$\pm$~3.19 & 75.32~$\pm$~1.87 \\
    Single-color & & 19.08~$\pm$~3.26 & 24.46~$\pm$~3.01 & 62.83~$\pm$~6.45 \\
    \bottomrule
  \end{tabular}
  }
  \label{tab:sparse_results_segformer_acdc_snow}
\end{table}
\clearpage

\begin{table}[tb]
  \centering
  \caption{\textbf{Zero-shot} evaluation of the \textbf{SegFormer} architecture trained on Cityscapes without sparsity and \textbf{validated on ACDC Mean} at different sparsities. Results include the three preprocessing types (luminance, color-opponency, and single-color). Metrics reported are mean Intersection over Union (mIoU), mean Accuracy (mAcc), and average Accuracy (aAcc), averaged over three seeds. \textbf{Percent-based sparsity:} For a given \textit{percentage}, the \textit{percentage} of lowest absolute values are set to zero.}
  \scalebox{0.85}{
  \begin{tabular}{@{}lc rrr@{}}
    \toprule
    Preprocessing & Sparsity &\multicolumn{1}{c}{mIoU}&\multicolumn{1}{c}{mAcc}&\multicolumn{1}{c}{aAcc}  \\
    \midrule
    -& \multirow{4}{*}{-}& 41.11~$\pm$~0.89 & 55.22~$\pm$~1.13 & 78.94~$\pm$~0.39\\
     Luminance &  & 38.00~$\pm$~0.58 & 48.82~$\pm$~1.28 & 76.64~$\pm$~1.65 \\
    Color-opponency &  & 38.87~$\pm$~0.36 & 50.58~$\pm$~2.39 & 78.83~$\pm$~0.96 \\
    Single-color &  & 38.74~$\pm$~0.98 & 51.29~$\pm$~0.14 & 78.88~$\pm$~1.46 \\
    \midrule
    - & \multirow{4}{*}{10\%} & 39.34~$\pm$~0.71 & 55.27~$\pm$~0.71 & 77.19~$\pm$~0.63 \\
    Luminance & & 38.03~$\pm$~0.56 & 48.83~$\pm$~1.27 & 76.73~$\pm$~1.64 \\
    Color-opponency & & 38.87~$\pm$~0.35 & 50.59~$\pm$~2.38 & 78.83~$\pm$~0.96 \\
    Single-color & & 38.74~$\pm$~0.97 & 51.28~$\pm$~0.13 & 78.89~$\pm$~1.44 \\
    \midrule
    - & \multirow{4}{*}{20\%} & 33.85~$\pm$~0.19 & 49.60~$\pm$~0.80 & 71.76~$\pm$~0.73 \\
    Luminance & & 38.23~$\pm$~0.51 & 48.90~$\pm$~1.23 & 77.17~$\pm$~1.57 \\
    Color-opponency & & 38.87~$\pm$~0.34 & 50.58~$\pm$~2.38 & 78.85~$\pm$~0.96 \\
    Single-color & & 38.76~$\pm$~0.95 & 51.25~$\pm$~0.13 & 78.95~$\pm$~1.38 \\
    \midrule
    - & \multirow{4}{*}{30\%} & 28.18~$\pm$~0.28 & 42.52~$\pm$~1.04 & 65.62~$\pm$~0.77 \\
    Luminance & & 38.46~$\pm$~0.44 & 48.88~$\pm$~1.21 & 77.72~$\pm$~1.47 \\
    Color-opponency & & 38.87~$\pm$~0.34 & 50.55~$\pm$~2.38 & 78.87~$\pm$~0.94 \\
    Single-color & & 38.81~$\pm$~0.94 & 51.14~$\pm$~0.13 & 78.98~$\pm$~1.28 \\
    \midrule
    - & \multirow{4}{*}{40\%} & 22.76~$\pm$~0.50 & 34.38~$\pm$~1.17 & 60.94~$\pm$~0.88 \\
    Luminance & & 38.45~$\pm$~0.34 & 48.59~$\pm$~1.12 & 77.94~$\pm$~1.41 \\
    Color-opponency & & 38.83~$\pm$~0.31 & 50.41~$\pm$~2.37 & 78.91~$\pm$~0.87 \\
    Single-color & & 38.60~$\pm$~0.85 & 50.70~$\pm$~0.26 & 78.86~$\pm$~1.18 \\
    \midrule
    - & \multirow{4}{*}{50\%} & 17.80~$\pm$~0.56 & 26.68~$\pm$~0.73 & 56.85~$\pm$~0.61 \\
    Luminance & & 37.63~$\pm$~0.31 & 47.57~$\pm$~1.01 & 77.29~$\pm$~1.58 \\
    Color-opponency & & 38.75~$\pm$~0.29 & 50.20~$\pm$~2.40 & 78.92~$\pm$~0.77 \\
    Single-color & & 37.93~$\pm$~0.78 & 49.71~$\pm$~0.35 & 78.38~$\pm$~1.08 \\
    \midrule
    - & \multirow{4}{*}{60\%} & 13.06~$\pm$~0.38 & 19.63~$\pm$~0.75 & 53.36~$\pm$~0.72 \\
    Luminance & & 35.65~$\pm$~0.29 & 45.33~$\pm$~0.73 & 75.51~$\pm$~2.04 \\
    Color-opponency & & 38.58~$\pm$~0.36 & 49.75~$\pm$~2.45 & 78.85~$\pm$~0.68 \\
    Single-color & & 36.40~$\pm$~0.84 & 47.63~$\pm$~0.33 & 77.31~$\pm$~1.18 \\
    \midrule
    - & \multirow{4}{*}{70\%} & 10.57~$\pm$~0.44 & 15.96~$\pm$~0.68 & 50.47~$\pm$~1.46 \\
    Luminance & & 32.03~$\pm$~0.45 & 41.21~$\pm$~0.45 & 71.99~$\pm$~2.50 \\
    Color-opponency & & 38.25~$\pm$~0.57 & 48.86~$\pm$~2.51 & 78.72~$\pm$~0.61 \\
    Single-color & & 33.42~$\pm$~1.23 & 43.92~$\pm$~0.21 & 75.00~$\pm$~1.71 \\
    \midrule
    - & \multirow{4}{*}{80\%} & 8.65~$\pm$~0.69 & 13.23~$\pm$~0.99 & 48.06~$\pm$~1.91 \\
    Luminance & & 26.35~$\pm$~0.34 & 34.32~$\pm$~0.52 & 66.59~$\pm$~2.29 \\
    Color-opponency & & 37.04~$\pm$~1.01 & 46.68~$\pm$~2.78 & 78.37~$\pm$~0.78 \\
    Single-color & & 27.64~$\pm$~1.68 & 36.76~$\pm$~0.89 & 70.30~$\pm$~3.02 \\
    \midrule
    - & \multirow{4}{*}{90\%} & 5.99~$\pm$~0.80 & 9.90~$\pm$~0.91 & 40.39~$\pm$~1.27 \\
    Luminance & & 17.76~$\pm$~0.41 & 23.56~$\pm$~0.69 & 58.96~$\pm$~1.74 \\
    Color-opponency & & 30.37~$\pm$~1.63 & 38.71~$\pm$~3.38 & 72.81~$\pm$~1.49 \\
    Single-color & & 18.66~$\pm$~2.22 & 25.39~$\pm$~2.10 & 61.55~$\pm$~4.96 \\
    \bottomrule
  \end{tabular}
  }
  \label{tab:sparse_results_segformer_acdc_full}
\end{table}
\clearpage

%% file: tables/sparsity_zero_shot_internimage.tex
\begin{table}[h]
  \centering
  \caption{\textbf{Zero-shot} evaluation of the \textbf{InternImage} architecture trained on Cityscapes without sparsity and \textbf{validated on Cityscapes} at different sparsities. Results include the three preprocessing types (luminance, color-opponency, and single-color). Metrics reported are mean Intersection over Union (mIoU), mean Accuracy (mAcc), and average Accuracy (aAcc), averaged over three seeds. \textbf{Percent-based sparsity:} For a given \textit{percentage}, the \textit{percentage} of lowest absolute values are set to zero.}
  \scalebox{0.85}{
  \begin{tabular}{@{}lc rrr@{}}
    \toprule
    Preprocessing & Sparsity &\multicolumn{1}{c}{mIoU}&\multicolumn{1}{c}{mAcc}&\multicolumn{1}{c}{aAcc}  \\
    \midrule
    - & \multirow{4}{*}{0\%} & 86.01~$\pm$~0.17 & 92.90~$\pm$~0.02 & 97.29~$\pm$~0.01 \\
    Luminance &  & 84.69~$\pm$~0.13 & 91.91~$\pm$~0.27 & 97.05~$\pm$~0.03 \\
    Color-opponency &  & 84.79~$\pm$~0.10 & 92.06~$\pm$~0.21 & 97.05~$\pm$~0.00 \\
    Single-color &  & 85.25~$\pm$~0.17 & 92.28~$\pm$~0.19 & 97.14~$\pm$~0.04 \\
    \midrule
    - & \multirow{4}{*}{10\%} & 85.68~$\pm$~0.20 & 92.65~$\pm$~0.02 & 97.21~$\pm$~0.02 \\
    Luminance &  & 84.70~$\pm$~0.13 & 91.90~$\pm$~0.27 & 97.05~$\pm$~0.03 \\
    Color-opponency &  & 84.79~$\pm$~0.10 & 92.06~$\pm$~0.21 & 97.05~$\pm$~0.00 \\
    Single-color &  & 85.25~$\pm$~0.17 & 92.28~$\pm$~0.19 & 97.14~$\pm$~0.04 \\
    \midrule
    - & \multirow{4}{*}{20\%} & 84.79~$\pm$~0.28 & 92.09~$\pm$~0.03 & 96.98~$\pm$~0.03 \\
    Luminance &  & 84.68~$\pm$~0.13 & 91.89~$\pm$~0.28 & 97.04~$\pm$~0.03 \\
    Color-opponency &  & 84.79~$\pm$~0.10 & 92.06~$\pm$~0.21 & 97.05~$\pm$~0.00 \\
    Single-color &  & 85.24~$\pm$~0.16 & 92.28~$\pm$~0.19 & 97.14~$\pm$~0.04 \\
    \midrule
    - & \multirow{4}{*}{30\%} & 83.83~$\pm$~0.33 & 91.39~$\pm$~0.03 & 96.69~$\pm$~0.04 \\
    Luminance &  & 84.65~$\pm$~0.14 & 91.87~$\pm$~0.29 & 97.04~$\pm$~0.03 \\
    Color-opponency &  & 84.78~$\pm$~0.10 & 92.06~$\pm$~0.21 & 97.05~$\pm$~0.01 \\
    Single-color &  & 85.26~$\pm$~0.16 & 92.30~$\pm$~0.19 & 97.14~$\pm$~0.05 \\
    \midrule
    - & \multirow{4}{*}{40\%} & 82.49~$\pm$~0.33 & 90.29~$\pm$~0.06 & 96.09~$\pm$~0.16 \\
    Luminance &  & 84.52~$\pm$~0.16 & 91.78~$\pm$~0.33 & 97.01~$\pm$~0.02 \\
    Color-opponency &  & 84.76~$\pm$~0.10 & 92.05~$\pm$~0.21 & 97.04~$\pm$~0.00 \\
    Single-color &  & 85.25~$\pm$~0.16 & 92.29~$\pm$~0.19 & 97.13~$\pm$~0.04 \\
    \midrule
    - & \multirow{4}{*}{50\%} & 78.89~$\pm$~1.21 & 88.02~$\pm$~0.08 & 94.17~$\pm$~0.61 \\
    Luminance &  & 84.27~$\pm$~0.13 & 91.61~$\pm$~0.28 & 96.95~$\pm$~0.03 \\
    Color-opponency &  & 84.73~$\pm$~0.08 & 92.03~$\pm$~0.20 & 97.03~$\pm$~0.01 \\
    Single-color &  & 85.13~$\pm$~0.17 & 92.20~$\pm$~0.20 & 97.09~$\pm$~0.05 \\
    \midrule
    - & \multirow{4}{*}{60\%} & 73.43~$\pm$~2.24 & 84.71~$\pm$~0.06 & 89.57~$\pm$~1.36 \\
    Luminance &  & 83.74~$\pm$~0.18 & 91.22~$\pm$~0.35 & 96.82~$\pm$~0.02 \\
    Color-opponency &  & 84.66~$\pm$~0.10 & 92.01~$\pm$~0.21 & 97.02~$\pm$~0.01 \\
    Single-color &  & 84.55~$\pm$~0.33 & 91.78~$\pm$~0.35 & 96.97~$\pm$~0.03 \\
    \midrule
    - & \multirow{4}{*}{70\%} & 66.03~$\pm$~2.20 & 78.77~$\pm$~0.22 & 81.03~$\pm$~1.85 \\
    Luminance &  & 82.33~$\pm$~0.15 & 90.17~$\pm$~0.14 & 96.49~$\pm$~0.02 \\
    Color-opponency &  & 84.60~$\pm$~0.08 & 91.95~$\pm$~0.17 & 96.98~$\pm$~0.01 \\
    Single-color &  & 83.72~$\pm$~0.20 & 91.06~$\pm$~0.25 & 96.76~$\pm$~0.03 \\
    \midrule
    - & \multirow{4}{*}{80\%} & 55.54~$\pm$~2.61 & 70.12~$\pm$~0.52 & 69.40~$\pm$~1.64 \\
    Luminance &  & 78.63~$\pm$~0.14 & 87.30~$\pm$~0.21 & 95.53~$\pm$~0.01 \\
    Color-opponency &  & 84.04~$\pm$~0.04 & 91.58~$\pm$~0.14 & 96.86~$\pm$~0.02 \\
    Single-color &  & 80.69~$\pm$~0.23 & 88.84~$\pm$~0.01 & 96.07~$\pm$~0.02 \\
    \midrule
    - & \multirow{4}{*}{90\%} & 35.19~$\pm$~1.36 & 48.82~$\pm$~0.98 & 52.64~$\pm$~0.64 \\
    Luminance &  & 68.85~$\pm$~0.17 & 78.84~$\pm$~0.43 & 92.59~$\pm$~0.05 \\
    Color-opponency &  & 81.00~$\pm$~0.09 & 89.29~$\pm$~0.14 & 96.19~$\pm$~0.01 \\
    Single-color &  & 71.77~$\pm$~0.11 & 81.80~$\pm$~0.06 & 93.47~$\pm$~0.11 \\
    \bottomrule
  \end{tabular}
  }
  \label{tab:sparse_results_internimage_cityscapes_val}
\end{table}
\clearpage
\begin{table}[tb]
  \centering
  \caption{\textbf{Zero-shot} evaluation of the \textbf{InternImage} architecture trained on Cityscapes without sparsity and \textbf{validated on Dark Zurich} at different sparsities. Results include the three preprocessing types (luminance, color-opponency, and single-color). Metrics reported are mean Intersection over Union (mIoU), mean Accuracy (mAcc), and average Accuracy (aAcc), averaged over three seeds. \textbf{Percent-based sparsity:} For a given \textit{percentage}, the \textit{percentage} of lowest absolute values are set to zero.}
  \scalebox{0.85}{
  \begin{tabular}{@{}lc rrr@{}}
    \toprule
    Preprocessing & Sparsity &\multicolumn{1}{c}{mIoU}&\multicolumn{1}{c}{mAcc}&\multicolumn{1}{c}{aAcc}  \\
    \midrule
    - & \multirow{4}{*}{0\%} & 53.05~$\pm$~0.13 & 72.13~$\pm$~0.33 & 88.65~$\pm$~0.52 \\
    Luminance &  & 45.38~$\pm$~0.19 & 63.15~$\pm$~0.33 & 84.00~$\pm$~0.79 \\
    Color-opponency &  & 45.34~$\pm$~0.99 & 64.37~$\pm$~1.25 & 85.40~$\pm$~0.90 \\
    Single-color &  & 45.46~$\pm$~0.69 & 63.96~$\pm$~1.17 & 83.17~$\pm$~0.47 \\
    \midrule
    - & \multirow{4}{*}{10\%} & 48.28~$\pm$~0.82 & 69.85~$\pm$~0.69 & 87.09~$\pm$~0.95 \\
    Luminance &  & 45.39~$\pm$~0.19 & 63.16~$\pm$~0.35 & 83.99~$\pm$~0.79 \\
    Color-opponency &  & 45.34~$\pm$~0.99 & 64.36~$\pm$~1.25 & 85.40~$\pm$~0.90 \\
    Single-color &  & 45.44~$\pm$~0.69 & 63.94~$\pm$~1.18 & 83.14~$\pm$~0.47 \\
    \midrule
    - & \multirow{4}{*}{20\%} & 43.09~$\pm$~1.08 & 62.55~$\pm$~0.99 & 85.32~$\pm$~1.20 \\
    Luminance &  & 45.37~$\pm$~0.19 & 63.16~$\pm$~0.32 & 83.94~$\pm$~0.82 \\
    Color-opponency &  & 45.34~$\pm$~0.99 & 64.36~$\pm$~1.25 & 85.40~$\pm$~0.90 \\
    Single-color &  & 45.46~$\pm$~0.67 & 63.93~$\pm$~1.15 & 83.09~$\pm$~0.48 \\
    \midrule
    - & \multirow{4}{*}{30\%} & 39.40~$\pm$~0.75 & 56.69~$\pm$~0.55 & 82.32~$\pm$~1.27 \\
    Luminance &  & 45.34~$\pm$~0.25 & 63.11~$\pm$~0.32 & 83.87~$\pm$~0.86 \\
    Color-opponency &  & 45.32~$\pm$~1.00 & 64.36~$\pm$~1.26 & 85.39~$\pm$~0.91 \\
    Single-color &  & 45.44~$\pm$~0.60 & 63.89~$\pm$~1.07 & 82.99~$\pm$~0.51 \\
    \midrule
    - & \multirow{4}{*}{40\%} & 33.21~$\pm$~0.72 & 49.38~$\pm$~0.70 & 76.17~$\pm$~1.45 \\
    Luminance &  & 45.31~$\pm$~0.42 & 63.08~$\pm$~0.20 & 83.81~$\pm$~1.03 \\
    Color-opponency &  & 45.30~$\pm$~0.99 & 64.34~$\pm$~1.24 & 85.39~$\pm$~0.92 \\
    Single-color &  & 45.36~$\pm$~0.55 & 63.69~$\pm$~0.97 & 82.89~$\pm$~0.58 \\
    \midrule
    - & \multirow{4}{*}{50\%} & 28.10~$\pm$~0.85 & 42.78~$\pm$~0.71 & 69.60~$\pm$~1.51 \\
    Luminance &  & 45.19~$\pm$~0.48 & 63.28~$\pm$~0.42 & 83.69~$\pm$~1.22 \\
    Color-opponency &  & 45.29~$\pm$~1.01 & 64.36~$\pm$~1.28 & 85.36~$\pm$~0.93 \\
    Single-color &  & 45.27~$\pm$~0.61 & 63.67~$\pm$~0.97 & 82.70~$\pm$~0.54 \\
    \midrule
    - & \multirow{4}{*}{60\%} & 23.02~$\pm$~0.36 & 35.98~$\pm$~1.16 & 61.62~$\pm$~2.07 \\
    Luminance &  & 45.00~$\pm$~0.54 & 63.26~$\pm$~0.66 & 83.12~$\pm$~1.40 \\
    Color-opponency &  & 45.19~$\pm$~1.02 & 64.29~$\pm$~1.27 & 85.31~$\pm$~0.98 \\
    Single-color &  & 45.31~$\pm$~0.84 & 63.47~$\pm$~1.21 & 82.75~$\pm$~0.69 \\
    \midrule
    - & \multirow{4}{*}{70\%} & 18.38~$\pm$~1.00 & 30.23~$\pm$~1.40 & 55.45~$\pm$~2.20 \\
    Luminance &  & 44.63~$\pm$~0.73 & 63.11~$\pm$~0.57 & 82.25~$\pm$~1.71 \\
    Color-opponency &  & 45.07~$\pm$~1.11 & 64.14~$\pm$~1.40 & 85.14~$\pm$~0.97 \\
    Single-color &  & 44.59~$\pm$~0.97 & 62.72~$\pm$~1.28 & 82.13~$\pm$~0.79 \\
    \midrule
    - & \multirow{4}{*}{80\%} & 13.82~$\pm$~0.55 & 23.06~$\pm$~0.75 & 47.04~$\pm$~1.64 \\
    Luminance &  & 43.00~$\pm$~0.67 & 62.03~$\pm$~0.48 & 79.80~$\pm$~1.95 \\
    Color-opponency &  & 44.75~$\pm$~0.90 & 63.76~$\pm$~1.27 & 84.69~$\pm$~0.99 \\
    Single-color &  & 43.52~$\pm$~0.78 & 61.67~$\pm$~1.21 & 80.44~$\pm$~0.75 \\
    \midrule
    - & \multirow{4}{*}{90\%} & 8.10~$\pm$~0.32 & 15.48~$\pm$~0.62 & 35.52~$\pm$~1.07 \\
    Luminance &  & 37.98~$\pm$~0.56 & 56.36~$\pm$~0.66 & 76.15~$\pm$~1.15 \\
    Color-opponency &  & 43.77~$\pm$~0.74 & 62.71~$\pm$~1.27 & 82.89~$\pm$~0.95 \\
    Single-color &  & 39.19~$\pm$~0.84 & 57.40~$\pm$~1.26 & 75.84~$\pm$~1.41 \\
    \bottomrule
  \end{tabular}
  }
  \label{tab:sparse_results_internimage_darkzurich_val}
\end{table}
\clearpage
\begin{table}[tb]
  \centering
  \caption{\textbf{Zero-shot} evaluation of the \textbf{InternImage} architecture trained on Cityscapes without sparsity and \textbf{validated on ACDC Night} at different sparsities. Results include the three preprocessing types (luminance, color-opponency, and single-color). Metrics reported are mean Intersection over Union (mIoU), mean Accuracy (mAcc), and average Accuracy (aAcc), averaged over three seeds. \textbf{Percent-based sparsity:} For a given \textit{percentage}, the \textit{percentage} of lowest absolute values are set to zero.}
  \scalebox{0.85}{
  \begin{tabular}{@{}lc rrr@{}}
    \toprule
    Preprocessing & Sparsity &\multicolumn{1}{c}{mIoU}&\multicolumn{1}{c}{mAcc}&\multicolumn{1}{c}{aAcc}  \\
    \midrule
    - & \multirow{4}{*}{0\%} & 55.63~$\pm$~0.49 & 75.15~$\pm$~0.35 & 87.43~$\pm$~0.76 \\
    Luminance &  & 49.83~$\pm$~0.34 & 69.39~$\pm$~0.37 & 82.82~$\pm$~0.69 \\
    Color-opponency &  & 49.91~$\pm$~0.61 & 70.66~$\pm$~0.87 & 83.97~$\pm$~0.89 \\
    Single-color &  & 49.78~$\pm$~0.59 & 69.69~$\pm$~0.40 & 82.21~$\pm$~0.45 \\
    \midrule
    - & \multirow{4}{*}{10\%} & 51.04~$\pm$~0.94 & 72.46~$\pm$~1.05 & 85.30~$\pm$~0.97 \\
    Luminance &  & 49.83~$\pm$~0.33 & 69.40~$\pm$~0.38 & 82.81~$\pm$~0.70 \\
    Color-opponency &  & 49.91~$\pm$~0.60 & 70.66~$\pm$~0.87 & 83.97~$\pm$~0.89 \\
    Single-color &  & 49.78~$\pm$~0.59 & 69.68~$\pm$~0.40 & 82.21~$\pm$~0.45 \\
    \midrule
    - & \multirow{4}{*}{20\%} & 46.33~$\pm$~1.14 & 66.96~$\pm$~1.16 & 83.27~$\pm$~1.12 \\
    Luminance &  & 49.85~$\pm$~0.37 & 69.39~$\pm$~0.39 & 82.79~$\pm$~0.72 \\
    Color-opponency &  & 49.91~$\pm$~0.60 & 70.66~$\pm$~0.87 & 83.97~$\pm$~0.89 \\
    Single-color &  & 49.79~$\pm$~0.55 & 69.65~$\pm$~0.37 & 82.19~$\pm$~0.44 \\
    \midrule
    - & \multirow{4}{*}{30\%} & 42.21~$\pm$~0.79 & 61.40~$\pm$~0.63 & 80.60~$\pm$~1.13 \\
    Luminance &  & 49.86~$\pm$~0.43 & 69.32~$\pm$~0.47 & 82.78~$\pm$~0.78 \\
    Color-opponency &  & 49.89~$\pm$~0.61 & 70.66~$\pm$~0.87 & 83.97~$\pm$~0.90 \\
    Single-color &  & 49.78~$\pm$~0.44 & 69.60~$\pm$~0.33 & 82.13~$\pm$~0.46 \\
    \midrule
    - & \multirow{4}{*}{40\%} & 36.96~$\pm$~1.21 & 54.32~$\pm$~2.09 & 76.15~$\pm$~1.56 \\
    Luminance &  & 49.72~$\pm$~0.61 & 69.13~$\pm$~0.39 & 82.62~$\pm$~1.06 \\
    Color-opponency &  & 49.88~$\pm$~0.62 & 70.65~$\pm$~0.87 & 83.96~$\pm$~0.90 \\
    Single-color &  & 49.61~$\pm$~0.44 & 69.34~$\pm$~0.28 & 81.94~$\pm$~0.54 \\
    \midrule
    - & \multirow{4}{*}{50\%} & 31.61~$\pm$~0.80 & 46.69~$\pm$~0.57 & 70.07~$\pm$~1.35 \\
    Luminance &  & 49.63~$\pm$~0.73 & 68.92~$\pm$~0.57 & 82.40~$\pm$~1.28 \\
    Color-opponency &  & 49.88~$\pm$~0.63 & 70.66~$\pm$~0.86 & 83.94~$\pm$~0.91 \\
    Single-color &  & 49.57~$\pm$~0.50 & 69.23~$\pm$~0.38 & 81.72~$\pm$~0.64 \\
    \midrule
    - & \multirow{4}{*}{60\%} & 25.92~$\pm$~1.17 & 38.03~$\pm$~0.99 & 62.63~$\pm$~1.97 \\
    Luminance &  & 49.33~$\pm$~0.95 & 68.59~$\pm$~0.66 & 81.80~$\pm$~1.46 \\
    Color-opponency &  & 49.90~$\pm$~0.53 & 70.62~$\pm$~0.86 & 83.92~$\pm$~0.90 \\
    Single-color &  & 49.19~$\pm$~0.67 & 68.69~$\pm$~0.53 & 81.46~$\pm$~0.85 \\
    \midrule
    - & \multirow{4}{*}{70\%} & 20.26~$\pm$~0.93 & 30.85~$\pm$~0.67 & 53.23~$\pm$~2.15 \\
    Luminance &  & 48.75~$\pm$~1.09 & 67.98~$\pm$~0.46 & 81.05~$\pm$~1.77 \\
    Color-opponency &  & 49.92~$\pm$~0.46 & 70.48~$\pm$~0.83 & 83.76~$\pm$~0.94 \\
    Single-color &  & 48.55~$\pm$~0.80 & 67.81~$\pm$~0.69 & 80.73~$\pm$~1.07 \\
    \midrule
    - & \multirow{4}{*}{80\%} & 14.37~$\pm$~0.52 & 22.74~$\pm$~0.38 & 43.71~$\pm$~1.51 \\
    Luminance &  & 46.56~$\pm$~0.99 & 65.93~$\pm$~0.43 & 78.89~$\pm$~1.80 \\
    Color-opponency &  & 49.72~$\pm$~0.39 & 70.06~$\pm$~0.85 & 83.32~$\pm$~0.96 \\
    Single-color &  & 47.02~$\pm$~1.17 & 66.31~$\pm$~1.15 & 79.09~$\pm$~1.16 \\
    \midrule
    - & \multirow{4}{*}{90\%} & 7.97~$\pm$~0.40 & 13.66~$\pm$~0.54 & 33.08~$\pm$~0.90 \\
    Luminance &  & 40.62~$\pm$~0.94 & 58.98~$\pm$~0.92 & 74.68~$\pm$~1.09 \\
    Color-opponency &  & 48.36~$\pm$~0.21 & 68.04~$\pm$~0.78 & 81.69~$\pm$~0.79 \\
    Single-color &  & 42.07~$\pm$~1.15 & 60.56~$\pm$~1.07 & 74.09~$\pm$~1.55 \\
    \bottomrule
  \end{tabular}
  }
  \label{tab:sparse_results_internimage_acdc_night_val}
\end{table}
\clearpage
\begin{table}[tb]
  \centering
  \caption{\textbf{Zero-shot} evaluation of the \textbf{InternImage} architecture trained on Cityscapes without sparsity and \textbf{validated on ACDC Fog} at different sparsities. Results include the three preprocessing types (luminance, color-opponency, and single-color). Metrics reported are mean Intersection over Union (mIoU), mean Accuracy (mAcc), and average Accuracy (aAcc), averaged over three seeds. \textbf{Percent-based sparsity:} For a given \textit{percentage}, the \textit{percentage} of lowest absolute values are set to zero.}
  \scalebox{0.85}{
  \begin{tabular}{@{}lc rrr@{}}
    \toprule
    Preprocessing & Sparsity &\multicolumn{1}{c}{mIoU}&\multicolumn{1}{c}{mAcc}&\multicolumn{1}{c}{aAcc}  \\
    \midrule
    - & \multirow{4}{*}{0\%} & 84.12~$\pm$~0.25 & 92.57~$\pm$~0.23 & 96.84~$\pm$~0.05 \\
    Luminance &  & 80.44~$\pm$~0.45 & 90.11~$\pm$~0.23 & 96.31~$\pm$~0.03 \\
    Color-opponency &  & 80.33~$\pm$~0.72 & 90.26~$\pm$~0.06 & 96.43~$\pm$~0.02 \\
    Single-color &  & 81.38~$\pm$~0.92 & 90.57~$\pm$~0.20 & 96.38~$\pm$~0.03 \\
    \midrule
    - & \multirow{4}{*}{10\%} & 84.05~$\pm$~0.23 & 92.38~$\pm$~0.20 & 96.79~$\pm$~0.02 \\
    Luminance &  & 80.44~$\pm$~0.43 & 90.11~$\pm$~0.22 & 96.31~$\pm$~0.03 \\
    Color-opponency &  & 80.33~$\pm$~0.73 & 90.26~$\pm$~0.06 & 96.43~$\pm$~0.02 \\
    Single-color &  & 81.40~$\pm$~0.93 & 90.57~$\pm$~0.20 & 96.38~$\pm$~0.03 \\
    \midrule
    - & \multirow{4}{*}{20\%} & 83.16~$\pm$~0.20 & 91.51~$\pm$~0.29 & 96.47~$\pm$~0.03 \\
    Luminance &  & 80.39~$\pm$~0.34 & 90.11~$\pm$~0.23 & 96.31~$\pm$~0.03 \\
    Color-opponency &  & 80.33~$\pm$~0.73 & 90.26~$\pm$~0.06 & 96.43~$\pm$~0.02 \\
    Single-color &  & 81.15~$\pm$~0.84 & 90.55~$\pm$~0.18 & 96.38~$\pm$~0.03 \\
    \midrule
    - & \multirow{4}{*}{30\%} & 81.22~$\pm$~0.09 & 89.80~$\pm$~0.28 & 95.88~$\pm$~0.02 \\
    Luminance &  & 80.40~$\pm$~0.38 & 90.14~$\pm$~0.24 & 96.30~$\pm$~0.03 \\
    Color-opponency &  & 80.35~$\pm$~0.70 & 90.26~$\pm$~0.06 & 96.43~$\pm$~0.02 \\
    Single-color &  & 80.59~$\pm$~0.87 & 90.50~$\pm$~0.14 & 96.37~$\pm$~0.02 \\
    \midrule
    - & \multirow{4}{*}{40\%} & 78.90~$\pm$~0.15 & 87.63~$\pm$~0.35 & 94.91~$\pm$~0.08 \\
    Luminance &  & 80.18~$\pm$~0.10 & 90.08~$\pm$~0.23 & 96.27~$\pm$~0.03 \\
    Color-opponency &  & 80.40~$\pm$~0.63 & 90.29~$\pm$~0.08 & 96.43~$\pm$~0.02 \\
    Single-color &  & 80.06~$\pm$~0.46 & 90.40~$\pm$~0.12 & 96.33~$\pm$~0.02 \\
    \midrule
    - & \multirow{4}{*}{50\%} & 70.38~$\pm$~0.84 & 81.52~$\pm$~0.37 & 90.16~$\pm$~0.55 \\
    Luminance &  & 80.19~$\pm$~0.16 & 90.11~$\pm$~0.31 & 96.19~$\pm$~0.07 \\
    Color-opponency &  & 80.46~$\pm$~0.49 & 90.26~$\pm$~0.07 & 96.42~$\pm$~0.02 \\
    Single-color &  & 79.84~$\pm$~0.51 & 90.28~$\pm$~0.11 & 96.27~$\pm$~0.02 \\
    \midrule
    - & \multirow{4}{*}{60\%} & 56.35~$\pm$~0.98 & 70.11~$\pm$~0.77 & 79.65~$\pm$~0.65 \\
    Luminance &  & 79.07~$\pm$~0.51 & 89.70~$\pm$~0.35 & 95.93~$\pm$~0.14 \\
    Color-opponency &  & 80.58~$\pm$~0.28 & 90.25~$\pm$~0.07 & 96.43~$\pm$~0.02 \\
    Single-color &  & 79.25~$\pm$~0.49 & 89.91~$\pm$~0.17 & 96.08~$\pm$~0.06 \\
    \midrule
    - & \multirow{4}{*}{70\%} & 39.42~$\pm$~1.47 & 50.95~$\pm$~2.53 & 67.26~$\pm$~0.36 \\
    Luminance &  & 78.45~$\pm$~0.32 & 89.00~$\pm$~0.39 & 95.36~$\pm$~0.10 \\
    Color-opponency &  & 80.58~$\pm$~0.17 & 90.21~$\pm$~0.11 & 96.41~$\pm$~0.01 \\
    Single-color &  & 78.95~$\pm$~0.10 & 89.16~$\pm$~0.20 & 95.53~$\pm$~0.04 \\
    \midrule
    - & \multirow{4}{*}{80\%} & 16.53~$\pm$~1.28 & 22.71~$\pm$~1.70 & 55.93~$\pm$~1.76 \\
    Luminance &  & 73.01~$\pm$~0.29 & 83.16~$\pm$~0.17 & 93.84~$\pm$~0.09 \\
    Color-opponency &  & 80.11~$\pm$~0.14 & 89.75~$\pm$~0.31 & 96.34~$\pm$~0.03 \\
    Single-color &  & 74.86~$\pm$~0.65 & 84.75~$\pm$~0.17 & 94.43~$\pm$~0.09 \\
    \midrule
    - & \multirow{4}{*}{90\%} & 7.66~$\pm$~0.16 & 11.97~$\pm$~0.29 & 47.01~$\pm$~1.37 \\
    Luminance &  & 57.24~$\pm$~0.96 & 68.47~$\pm$~0.65 & 90.02~$\pm$~0.19 \\
    Color-opponency &  & 77.61~$\pm$~0.14 & 88.09~$\pm$~0.12 & 95.37~$\pm$~0.06 \\
    Single-color &  & 59.60~$\pm$~1.15 & 70.86~$\pm$~1.13 & 90.46~$\pm$~0.25 \\
    \bottomrule
  \end{tabular}
  }
  \label{tab:sparse_results_internimage_acdc_fog_val}
\end{table}
\clearpage
\begin{table}[tb]
  \centering
  \caption{\textbf{Zero-shot} evaluation of the \textbf{InternImage} architecture trained on Cityscapes without sparsity and \textbf{validated on ACDC Rain} at different sparsities. Results include the three preprocessing types (luminance, color-opponency, and single-color). Metrics reported are mean Intersection over Union (mIoU), mean Accuracy (mAcc), and average Accuracy (aAcc), averaged over three seeds. \textbf{Percent-based sparsity:} For a given \textit{percentage}, the \textit{percentage} of lowest absolute values are set to zero.}
  \scalebox{0.85}{
  \begin{tabular}{@{}lc rrr@{}}
    \toprule
    Preprocessing & Sparsity &\multicolumn{1}{c}{mIoU}&\multicolumn{1}{c}{mAcc}&\multicolumn{1}{c}{aAcc}  \\
    \midrule
    - & \multirow{4}{*}{0\%} & 74.16~$\pm$~1.03 & 89.09~$\pm$~0.75 & 95.67~$\pm$~0.13 \\
    Luminance &  & 70.28~$\pm$~0.54 & 86.72~$\pm$~0.39 & 95.07~$\pm$~0.21 \\
    Color-opponency &  & 70.11~$\pm$~2.56 & 86.28~$\pm$~1.53 & 94.79~$\pm$~0.48 \\
    Single-color &  & 71.90~$\pm$~1.04 & 87.47~$\pm$~0.28 & 95.32~$\pm$~0.24 \\
    \midrule
    - & \multirow{4}{*}{10\%} & 74.70~$\pm$~1.14 & 89.10~$\pm$~0.63 & 95.54~$\pm$~0.10 \\
    Luminance &  & 70.28~$\pm$~0.54 & 86.72~$\pm$~0.39 & 95.07~$\pm$~0.22 \\
    Color-opponency &  & 70.11~$\pm$~2.56 & 86.28~$\pm$~1.53 & 94.79~$\pm$~0.48 \\
    Single-color &  & 71.90~$\pm$~1.03 & 87.47~$\pm$~0.26 & 95.32~$\pm$~0.23 \\
    \midrule
    - & \multirow{4}{*}{20\%} & 73.75~$\pm$~1.03 & 88.08~$\pm$~0.83 & 94.99~$\pm$~0.10 \\
    Luminance &  & 70.24~$\pm$~0.53 & 86.72~$\pm$~0.38 & 95.06~$\pm$~0.22 \\
    Color-opponency &  & 70.13~$\pm$~2.56 & 86.30~$\pm$~1.54 & 94.79~$\pm$~0.48 \\
    Single-color &  & 71.87~$\pm$~0.98 & 87.47~$\pm$~0.25 & 95.30~$\pm$~0.23 \\
    \midrule
    - & \multirow{4}{*}{30\%} & 70.30~$\pm$~0.77 & 85.86~$\pm$~0.81 & 94.05~$\pm$~0.03 \\
    Luminance &  & 70.15~$\pm$~0.50 & 86.72~$\pm$~0.39 & 95.05~$\pm$~0.22 \\
    Color-opponency &  & 70.09~$\pm$~2.53 & 86.29~$\pm$~1.53 & 94.79~$\pm$~0.48 \\
    Single-color &  & 71.82~$\pm$~0.96 & 87.45~$\pm$~0.22 & 95.24~$\pm$~0.26 \\
    \midrule
    - & \multirow{4}{*}{40\%} & 63.45~$\pm$~0.43 & 80.72~$\pm$~0.76 & 91.19~$\pm$~0.27 \\
    Luminance &  & 70.07~$\pm$~0.42 & 86.69~$\pm$~0.35 & 95.01~$\pm$~0.22 \\
    Color-opponency &  & 70.06~$\pm$~2.54 & 86.28~$\pm$~1.52 & 94.79~$\pm$~0.48 \\
    Single-color &  & 71.78~$\pm$~0.95 & 87.44~$\pm$~0.23 & 95.18~$\pm$~0.24 \\
    \midrule
    - & \multirow{4}{*}{50\%} & 54.40~$\pm$~0.71 & 69.59~$\pm$~1.38 & 87.05~$\pm$~0.64 \\
    Luminance &  & 69.84~$\pm$~0.55 & 86.42~$\pm$~0.19 & 94.89~$\pm$~0.15 \\
    Color-opponency &  & 70.00~$\pm$~2.55 & 86.28~$\pm$~1.53 & 94.78~$\pm$~0.48 \\
    Single-color &  & 71.70~$\pm$~0.90 & 87.32~$\pm$~0.19 & 95.15~$\pm$~0.27 \\
    \midrule
    - & \multirow{4}{*}{60\%} & 46.74~$\pm$~0.37 & 60.67~$\pm$~0.81 & 81.52~$\pm$~0.84 \\
    Luminance &  & 69.30~$\pm$~0.29 & 86.02~$\pm$~0.26 & 94.74~$\pm$~0.14 \\
    Color-opponency &  & 69.69~$\pm$~2.48 & 86.27~$\pm$~1.52 & 94.74~$\pm$~0.48 \\
    Single-color &  & 71.18~$\pm$~0.38 & 86.86~$\pm$~0.13 & 95.03~$\pm$~0.15 \\
    \midrule
    - & \multirow{4}{*}{70\%} & 37.29~$\pm$~0.51 & 47.64~$\pm$~1.07 & 74.50~$\pm$~0.65 \\
    Luminance &  & 68.09~$\pm$~0.70 & 84.88~$\pm$~0.09 & 94.20~$\pm$~0.05 \\
    Color-opponency &  & 69.69~$\pm$~2.47 & 86.22~$\pm$~1.52 & 94.66~$\pm$~0.52 \\
    Single-color &  & 70.04~$\pm$~0.66 & 86.16~$\pm$~0.41 & 94.68~$\pm$~0.21 \\
    \midrule
    - & \multirow{4}{*}{80\%} & 22.44~$\pm$~0.21 & 28.87~$\pm$~0.44 & 66.58~$\pm$~0.36 \\
    Luminance &  & 63.15~$\pm$~0.44 & 79.96~$\pm$~0.31 & 92.02~$\pm$~0.16 \\
    Color-opponency &  & 69.30~$\pm$~2.48 & 86.00~$\pm$~1.57 & 94.60~$\pm$~0.53 \\
    Single-color &  & 64.45~$\pm$~1.10 & 82.12~$\pm$~0.02 & 93.09~$\pm$~0.27 \\
    \midrule
    - & \multirow{4}{*}{90\%} & 13.05~$\pm$~0.12 & 17.85~$\pm$~0.24 & 54.69~$\pm$~0.59 \\
    Luminance &  & 53.27~$\pm$~0.68 & 67.56~$\pm$~0.17 & 89.31~$\pm$~0.27 \\
    Color-opponency &  & 66.71~$\pm$~2.24 & 84.89~$\pm$~1.39 & 93.54~$\pm$~0.50 \\
    Single-color &  & 55.29~$\pm$~0.55 & 69.69~$\pm$~0.18 & 89.87~$\pm$~0.25 \\
    \bottomrule
  \end{tabular}
  }
  \label{tab:sparse_results_internimage_acdc_rain_val}
\end{table}
\clearpage
\begin{table}[tb]
  \centering
  \caption{\textbf{Zero-shot} evaluation of the \textbf{InternImage} architecture trained on Cityscapes without sparsity and \textbf{validated on ACDC Snow} at different sparsities. Results include the three preprocessing types (luminance, color-opponency, and single-color). Metrics reported are mean Intersection over Union (mIoU), mean Accuracy (mAcc), and average Accuracy (aAcc), averaged over three seeds. \textbf{Percent-based sparsity:} For a given \textit{percentage}, the \textit{percentage} of lowest absolute values are set to zero.}
  \scalebox{0.85}{
  \begin{tabular}{@{}lc rrr@{}}
    \toprule
    Preprocessing & Sparsity &\multicolumn{1}{c}{mIoU}&\multicolumn{1}{c}{mAcc}&\multicolumn{1}{c}{aAcc}  \\
    \midrule
    - & \multirow{4}{*}{0\%} & 74.18~$\pm$~0.57 & 81.96~$\pm$~0.69 & 94.91~$\pm$~0.09 \\
    Luminance &  & 71.16~$\pm$~1.02 & 79.84~$\pm$~0.73 & 94.42~$\pm$~0.14 \\
    Color-opponency &  & 70.28~$\pm$~0.98 & 79.25~$\pm$~0.63 & 94.35~$\pm$~0.18 \\
    Single-color &  & 70.71~$\pm$~0.81 & 79.44~$\pm$~0.76 & 94.24~$\pm$~0.17 \\
    \midrule
    - & \multirow{4}{*}{10\%} & 73.86~$\pm$~0.91 & 81.64~$\pm$~0.80 & 94.85~$\pm$~0.12 \\
    Luminance &  & 71.18~$\pm$~1.05 & 79.87~$\pm$~0.76 & 94.42~$\pm$~0.13 \\
    Color-opponency &  & 70.28~$\pm$~0.98 & 79.25~$\pm$~0.63 & 94.35~$\pm$~0.18 \\
    Single-color &  & 70.71~$\pm$~0.81 & 79.44~$\pm$~0.76 & 94.24~$\pm$~0.18 \\
    \midrule
    - & \multirow{4}{*}{20\%} & 73.63~$\pm$~0.68 & 81.67~$\pm$~0.70 & 94.60~$\pm$~0.12 \\
    Luminance &  & 71.13~$\pm$~1.04 & 79.84~$\pm$~0.74 & 94.42~$\pm$~0.13 \\
    Color-opponency &  & 70.28~$\pm$~0.98 & 79.25~$\pm$~0.63 & 94.35~$\pm$~0.18 \\
    Single-color &  & 70.71~$\pm$~0.82 & 79.45~$\pm$~0.75 & 94.24~$\pm$~0.18 \\
    \midrule
    - & \multirow{4}{*}{30\%} & 71.42~$\pm$~0.64 & 80.28~$\pm$~0.35 & 93.95~$\pm$~0.14 \\
    Luminance &  & 71.12~$\pm$~1.07 & 79.86~$\pm$~0.75 & 94.41~$\pm$~0.13 \\
    Color-opponency &  & 70.27~$\pm$~0.98 & 79.25~$\pm$~0.63 & 94.34~$\pm$~0.18 \\
    Single-color &  & 70.73~$\pm$~0.79 & 79.48~$\pm$~0.72 & 94.24~$\pm$~0.19 \\
    \midrule
    - & \multirow{4}{*}{40\%} & 69.13~$\pm$~0.92 & 77.75~$\pm$~0.37 & 92.84~$\pm$~0.16 \\
    Luminance &  & 70.67~$\pm$~1.35 & 79.82~$\pm$~0.70 & 94.28~$\pm$~0.17 \\
    Color-opponency &  & 70.26~$\pm$~0.97 & 79.24~$\pm$~0.63 & 94.34~$\pm$~0.17 \\
    Single-color &  & 70.37~$\pm$~1.20 & 79.39~$\pm$~0.76 & 94.13~$\pm$~0.24 \\
    \midrule
    - & \multirow{4}{*}{50\%} & 64.70~$\pm$~0.35 & 73.61~$\pm$~0.43 & 90.63~$\pm$~0.32 \\
    Luminance &  & 70.83~$\pm$~0.93 & 79.84~$\pm$~0.66 & 94.16~$\pm$~0.11 \\
    Color-opponency &  & 70.22~$\pm$~0.96 & 79.21~$\pm$~0.60 & 94.34~$\pm$~0.17 \\
    Single-color &  & 70.19~$\pm$~1.12 & 79.30~$\pm$~0.73 & 94.02~$\pm$~0.23 \\
    \midrule
    - & \multirow{4}{*}{60\%} & 55.94~$\pm$~0.75 & 66.24~$\pm$~0.21 & 86.25~$\pm$~0.57 \\
    Luminance &  & 69.67~$\pm$~1.59 & 79.90~$\pm$~0.74 & 93.64~$\pm$~0.04 \\
    Color-opponency &  & 70.46~$\pm$~0.54 & 79.46~$\pm$~0.15 & 94.36~$\pm$~0.14 \\
    Single-color &  & 69.09~$\pm$~0.55 & 79.01~$\pm$~0.62 & 93.78~$\pm$~0.14 \\
    \midrule
    - & \multirow{4}{*}{70\%} & 43.84~$\pm$~0.96 & 53.67~$\pm$~0.19 & 78.10~$\pm$~0.60 \\
    Luminance &  & 68.71~$\pm$~0.42 & 79.32~$\pm$~0.12 & 92.67~$\pm$~0.09 \\
    Color-opponency &  & 70.46~$\pm$~0.54 & 79.49~$\pm$~0.15 & 94.36~$\pm$~0.13 \\
    Single-color &  & 68.39~$\pm$~1.23 & 78.93~$\pm$~1.14 & 93.14~$\pm$~0.25 \\
    \midrule
    - & \multirow{4}{*}{80\%} & 27.28~$\pm$~0.43 & 35.30~$\pm$~0.32 & 65.93~$\pm$~0.29 \\
    Luminance &  & 65.76~$\pm$~0.55 & 76.60~$\pm$~0.37 & 91.31~$\pm$~0.13 \\
    Color-opponency &  & 70.04~$\pm$~0.74 & 79.43~$\pm$~0.14 & 94.24~$\pm$~0.14 \\
    Single-color &  & 65.91~$\pm$~1.08 & 76.96~$\pm$~0.65 & 92.15~$\pm$~0.28 \\
    \midrule
    - & \multirow{4}{*}{90\%} & 13.35~$\pm$~0.36 & 18.75~$\pm$~0.40 & 53.56~$\pm$~0.64 \\
    Luminance &  & 60.84~$\pm$~0.98 & 70.90~$\pm$~1.00 & 89.01~$\pm$~0.20 \\
    Color-opponency &  & 66.65~$\pm$~0.41 & 78.12~$\pm$~0.44 & 92.33~$\pm$~0.11 \\
    Single-color &  & 60.74~$\pm$~0.24 & 71.77~$\pm$~1.02 & 89.74~$\pm$~0.28 \\
    \bottomrule
  \end{tabular}
  }
  \label{tab:sparse_results_internimage_acdc_snow_val}
\end{table}
\clearpage
\begin{table}[tb]
  \centering
  \caption{\textbf{Zero-shot} evaluation of the \textbf{InternImage} architecture trained on Cityscapes without sparsity and \textbf{validated on ACDC Mean} at different sparsities. Results include the three preprocessing types (luminance, color-opponency, and single-color). Metrics reported are mean Intersection over Union (mIoU), mean Accuracy (mAcc), and average Accuracy (aAcc), averaged over three seeds. \textbf{Percent-based sparsity:} For a given \textit{percentage}, the \textit{percentage} of lowest absolute values are set to zero.}
  \scalebox{0.85}{
  \begin{tabular}{@{}lc rrr@{}}
    \toprule
    Preprocessing & Sparsity &\multicolumn{1}{c}{mIoU}&\multicolumn{1}{c}{mAcc}&\multicolumn{1}{c}{aAcc}  \\
    \midrule
    - & \multirow{4}{*}{0\%} & 72.02~$\pm$~0.29 & 84.69~$\pm$~0.43 & 93.71~$\pm$~0.15 \\
    Luminance &  & 67.93~$\pm$~0.35 & 81.52~$\pm$~0.22 & 92.15~$\pm$~0.25 \\
    Color-opponency &  & 67.66~$\pm$~0.95 & 81.61~$\pm$~0.48 & 92.38~$\pm$~0.11 \\
    Single-color &  & 68.44~$\pm$~0.66 & 81.79~$\pm$~0.37 & 92.04~$\pm$~0.19 \\
    \midrule
    - & \multirow{4}{*}{10\%} & 70.91~$\pm$~0.38 & 83.90~$\pm$~0.54 & 93.12~$\pm$~0.23 \\
    Luminance &  & 67.93~$\pm$~0.36 & 81.53~$\pm$~0.21 & 92.15~$\pm$~0.25 \\
    Color-opponency &  & 67.66~$\pm$~0.95 & 81.61~$\pm$~0.48 & 92.38~$\pm$~0.11 \\
    Single-color &  & 68.45~$\pm$~0.67 & 81.79~$\pm$~0.37 & 92.04~$\pm$~0.18 \\
    \midrule
    - & \multirow{4}{*}{20\%} & 69.22~$\pm$~0.39 & 82.06~$\pm$~0.63 & 92.33~$\pm$~0.27 \\
    Luminance &  & 67.90~$\pm$~0.39 & 81.52~$\pm$~0.21 & 92.15~$\pm$~0.26 \\
    Color-opponency &  & 67.66~$\pm$~0.95 & 81.62~$\pm$~0.49 & 92.38~$\pm$~0.11 \\
    Single-color &  & 68.38~$\pm$~0.54 & 81.78~$\pm$~0.36 & 92.03~$\pm$~0.18 \\
    \midrule
    - & \multirow{4}{*}{30\%} & 66.29~$\pm$~0.14 & 79.34~$\pm$~0.40 & 91.12~$\pm$~0.29 \\
    Luminance &  & 67.88~$\pm$~0.39 & 81.51~$\pm$~0.21 & 92.14~$\pm$~0.27 \\
    Color-opponency &  & 67.65~$\pm$~0.93 & 81.61~$\pm$~0.48 & 92.38~$\pm$~0.11 \\
    Single-color &  & 68.23~$\pm$~0.34 & 81.76~$\pm$~0.32 & 92.00~$\pm$~0.18 \\
    \midrule
    - & \multirow{4}{*}{40\%} & 62.11~$\pm$~0.56 & 75.10~$\pm$~0.70 & 88.77~$\pm$~0.50 \\
    Luminance &  & 67.66~$\pm$~0.54 & 81.43~$\pm$~0.17 & 92.05~$\pm$~0.35 \\
    Color-opponency &  & 67.65~$\pm$~0.91 & 81.62~$\pm$~0.47 & 92.38~$\pm$~0.11 \\
    Single-color &  & 67.96~$\pm$~0.45 & 81.64~$\pm$~0.32 & 91.90~$\pm$~0.20 \\
    \midrule
    - & \multirow{4}{*}{50\%} & 55.27~$\pm$~0.31 & 67.85~$\pm$~0.38 & 84.48~$\pm$~0.70 \\
    Luminance &  & 67.62~$\pm$~0.50 & 81.32~$\pm$~0.23 & 91.91~$\pm$~0.40 \\
    Color-opponency &  & 67.64~$\pm$~0.87 & 81.60~$\pm$~0.47 & 92.37~$\pm$~0.11 \\
    Single-color &  & 67.82~$\pm$~0.41 & 81.53~$\pm$~0.33 & 91.79~$\pm$~0.25 \\
    \midrule
    - & \multirow{4}{*}{60\%} & 46.24~$\pm$~0.64 & 58.77~$\pm$~0.62 & 77.51~$\pm$~0.98 \\
    Luminance &  & 66.84~$\pm$~0.38 & 81.05~$\pm$~0.12 & 91.53~$\pm$~0.43 \\
    Color-opponency &  & 67.66~$\pm$~0.71 & 81.65~$\pm$~0.37 & 92.36~$\pm$~0.10 \\
    Single-color &  & 67.18~$\pm$~0.24 & 81.12~$\pm$~0.33 & 91.59~$\pm$~0.27 \\
    \midrule
    - & \multirow{4}{*}{70\%} & 35.20~$\pm$~0.23 & 45.78~$\pm$~0.93 & 68.27~$\pm$~0.71 \\
    Luminance &  & 66.00~$\pm$~0.09 & 80.30~$\pm$~0.15 & 90.82~$\pm$~0.46 \\
    Color-opponency &  & 67.66~$\pm$~0.71 & 81.60~$\pm$~0.37 & 92.30~$\pm$~0.09 \\
    Single-color &  & 66.48~$\pm$~0.66 & 80.51~$\pm$~0.50 & 91.02~$\pm$~0.36 \\
    \midrule
    - & \multirow{4}{*}{80\%} & 20.16~$\pm$~0.16 & 27.40~$\pm$~0.50 & 58.04~$\pm$~0.18 \\
    Luminance &  & 62.12~$\pm$~0.26 & 76.42~$\pm$~0.14 & 89.02~$\pm$~0.51 \\
    Color-opponency &  & 67.29~$\pm$~0.66 & 81.31~$\pm$~0.32 & 92.13~$\pm$~0.07 \\
    Single-color &  & 63.06~$\pm$~0.55 & 77.53~$\pm$~0.44 & 89.69~$\pm$~0.44 \\
    \midrule
    - & \multirow{4}{*}{90\%} & 10.51~$\pm$~0.07 & 15.56~$\pm$~0.24 & 47.08~$\pm$~0.46 \\
    Luminance &  & 52.99~$\pm$~0.83 & 66.48~$\pm$~0.58 & 85.76~$\pm$~0.41 \\
    Color-opponency &  & 64.83~$\pm$~0.65 & 79.79~$\pm$~0.43 & 90.73~$\pm$~0.16 \\
    Single-color &  & 54.42~$\pm$~0.54 & 68.22~$\pm$~0.59 & 86.04~$\pm$~0.52 \\
    \bottomrule
  \end{tabular}
  }
  \label{tab:sparse_results_internimage_acdc_4_mean}
\end{table}
\clearpage

%% file: tables/sparsity_trained_deeplab.tex
% percentage sparsity results
\begin{table}[h]
  \centering
  \caption{Evaluation of the \textbf{DeepLabv3+} architecture trained on Cityscapes and \textbf{validated on Cityscapes} at different sparsities. Results include the three preprocessing types (luminance, color-opponency, and single-color). Metrics reported are mean Intersection over Union (mIoU), mean Accuracy (mAcc), and average Accuracy (aAcc), averaged over three seeds. \textbf{Percent-based sparsity:} For a given \textit{percentage}, the \textit{percentage} of lowest absolute values are set to zero.}   
  \begin{tabular}{@{}lcrrr@{}}
    \toprule
    Preprocessing  & Sparsity &\multicolumn{1}{c}{mIoU}&\multicolumn{1}{c}{mAcc}&\multicolumn{1}{c}{aAcc}  \\
    \midrule
     - & \multirow{3}{*}{-} &  66.68~$\pm$~0.44 & 76.58~$\pm$~0.60 & 94.68~$\pm$~0.05 \\
    Luminance   &  & 66.53~$\pm$~1.11 & 77.86~$\pm$~0.63 & 94.35~$\pm$~0.35 \\
     Color-opponency&   &  66.64~$\pm$~1.39 & 75.80~$\pm$~1.46 & 94.88~$\pm$~0.15 \\
     Single-color&   &  66.08~$\pm$~3.05 & 76.33~$\pm$~3.03 & 94.90~$\pm$~0.16 \\
    \midrule
     - & \multirow{4}{*}{40\%} & 62.77~$\pm$~1.19 & 74.36~$\pm$~0.57 & 93.08~$\pm$~0.16 \\
    Luminance &  & 64.89~$\pm$~1.39 & 75.69~$\pm$~1.08 & 94.49~$\pm$~0.18 \\
    Color-opponency &  & 66.91~$\pm$~1.44 & 76.74~$\pm$~2.18 & 94.83~$\pm$~0.05 \\
    Single-color &  & 66.42~$\pm$~2.94 & 76.23~$\pm$~2.96 & 94.95~$\pm$~0.17 \\
    \midrule
     - & \multirow{4}{*}{50\%} & 58.56~$\pm$~2.58 & 70.48~$\pm$~0.77 & 92.03~$\pm$~1.09 \\
    Luminance &  & 64.75~$\pm$~2.07 & 75.20~$\pm$~2.18 & 94.27~$\pm$~0.22 \\
    Color-opponency &  & 67.71~$\pm$~0.79 & 77.90~$\pm$~1.05 & 94.85~$\pm$~0.21 \\
    Single-color &  & 63.40~$\pm$~3.44 & 73.92~$\pm$~3.55 & 94.69~$\pm$~0.24 \\
    \midrule
     - & \multirow{4}{*}{60\%} & 57.19~$\pm$~1.67 & 69.26~$\pm$~1.34 & 91.58~$\pm$~0.51 \\
    Luminance &  & 63.14~$\pm$~2.62 & 73.47~$\pm$~2.20 & 94.02~$\pm$~0.18 \\
    Color-opponency &  & 64.88~$\pm$~0.98 & 75.02~$\pm$~1.09 & 94.75~$\pm$~0.27 \\
    Single-color &  & 64.90~$\pm$~2.54 & 74.79~$\pm$~3.41 & 94.68~$\pm$~0.17 \\
    \midrule
     - & \multirow{4}{*}{70\%} & 52.91~$\pm$~2.60 & 63.29~$\pm$~1.94 & 90.83~$\pm$~0.53 \\
    Luminance &  & 61.52~$\pm$~0.93 & 71.76~$\pm$~0.71 & 93.25~$\pm$~0.40 \\
    Color-opponency &  & 65.90~$\pm$~0.79 & 76.43~$\pm$~0.68 & 94.53~$\pm$~0.18 \\
    Single-color &  & 63.09~$\pm$~1.82 & 73.29~$\pm$~1.79 & 94.06~$\pm$~0.49 \\
    \midrule
     - & \multirow{4}{*}{80\%} & 49.44~$\pm$~1.34 & 60.29~$\pm$~1.27 & 89.45~$\pm$~0.23 \\
    Luminance &  & 57.32~$\pm$~3.98 & 67.75~$\pm$~3.99 & 91.76~$\pm$~0.90 \\
    Color-opponency &  & 64.98~$\pm$~0.32 & 75.37~$\pm$~0.81 & 94.20~$\pm$~0.14 \\
    Single-color &  & 59.84~$\pm$~3.22 & 70.17~$\pm$~3.32 & 93.49~$\pm$~0.29 \\
    \midrule
     - & \multirow{4}{*}{90\%} & 38.60~$\pm$~1.10 & 48.03~$\pm$~1.54 & 84.61~$\pm$~0.36 \\
    Luminance &  & 51.13~$\pm$~0.20 & 62.14~$\pm$~0.55 & 89.06~$\pm$~0.23 \\
    Color-opponency &  & 60.89~$\pm$~0.80 & 71.16~$\pm$~1.17 & 93.17~$\pm$~0.42 \\
    Single-color &  & 53.42~$\pm$~2.35 & 63.05~$\pm$~3.27 & 90.99~$\pm$~0.68 \\
    \bottomrule
  \end{tabular}
  \label{tab:sparse_results_deeplabv3+_cityscapes_learned}
\end{table}
\clearpage

\begin{table}[tb]
  \centering
  \caption{Evaluation of the \textbf{DeepLabv3+} architecture trained on Cityscapes and \textbf{validated on Dark Zurich} at different sparsities. Results include the three preprocessing types (luminance, color-opponency, and single-color). Metrics reported are mean Intersection over Union (mIoU), mean Accuracy (mAcc), and average Accuracy (aAcc), averaged over three seeds. \textbf{Percent-based sparsity:} For a given \textit{percentage}, the \textit{percentage} of lowest absolute values are set to zero.}   
  \begin{tabular}{@{}lcrrr@{}}
    \toprule
    Preprocessing  & Sparsity &\multicolumn{1}{c}{mIoU}&\multicolumn{1}{c}{mAcc}&\multicolumn{1}{c}{aAcc}  \\
    \midrule
     - & \multirow{4}{*}{-} &  7.64~$\pm$~1.00 & 16.40~$\pm$~2.74 & 30.13~$\pm$~1.41 \\
     Luminance &   & 18.39~$\pm$~0.54 & 34.73~$\pm$~1.68 & 55.35~$\pm$~1.81 \\
    Color-opponency&   &  17.64~$\pm$~1.70 & 32.32~$\pm$~2.85 & 50.68~$\pm$~4.61 \\
    Single-color &   &  18.54~$\pm$~2.18 & 32.79~$\pm$~1.49 & 51.35~$\pm$~4.26 \\
    \midrule
     - & \multirow{4}{*}{40\%} & 6.37~$\pm$~0.89 & 14.46~$\pm$~0.97 & 27.49~$\pm$~3.89 \\
    Luminance &  & 20.66~$\pm$~0.62 & 36.33~$\pm$~1.60 & 56.14~$\pm$~1.29 \\
    Color-opponency &  & 18.81~$\pm$~0.59 & 32.57~$\pm$~1.32 & 52.77~$\pm$~1.51 \\
    Single-color &  & 17.46~$\pm$~2.43 & 32.20~$\pm$~3.21 & 48.79~$\pm$~5.74 \\
    \midrule
     - & \multirow{4}{*}{50\%} & 5.18~$\pm$~1.66 & 12.20~$\pm$~3.35 & 20.13~$\pm$~7.87 \\
    Luminance &  & 20.62~$\pm$~1.15 & 34.74~$\pm$~2.33 & 55.74~$\pm$~3.00 \\
    Color-opponency &  & 19.40~$\pm$~1.27 & 33.43~$\pm$~1.89 & 54.35~$\pm$~1.28 \\
    Single-color &  & 18.20~$\pm$~0.69 & 31.88~$\pm$~1.36 & 51.38~$\pm$~3.91 \\
    \midrule
     - & \multirow{4}{*}{60\%} & 5.03~$\pm$~0.83 & 11.15~$\pm$~0.22 & 19.55~$\pm$~4.84 \\
    Luminance &  & 20.01~$\pm$~0.60 & 33.19~$\pm$~1.60 & 53.62~$\pm$~1.70 \\
    Color-opponency &  & 19.58~$\pm$~1.60 & 33.11~$\pm$~3.94 & 56.08~$\pm$~1.01 \\
    Single-color &  & 18.69~$\pm$~0.52 & 33.40~$\pm$~1.29 & 54.20~$\pm$~1.16 \\
    \midrule
     - & \multirow{4}{*}{70\%} & 6.63~$\pm$~1.15 & 14.96~$\pm$~2.74 & 28.43~$\pm$~3.40 \\
    Luminance &  & 17.81~$\pm$~1.49 & 31.31~$\pm$~2.27 & 51.87~$\pm$~3.06 \\
    Color-opponency &  & 17.68~$\pm$~0.53 & 31.69~$\pm$~0.64 & 52.86~$\pm$~0.53 \\
    Single-color &  & 18.83~$\pm$~1.72 & 32.65~$\pm$~2.37 & 51.47~$\pm$~3.88 \\
    \midrule
     - & \multirow{4}{*}{80\%} & 7.62~$\pm$~0.84 & 18.35~$\pm$~2.50 & 35.79~$\pm$~5.24 \\
    Luminance &  & 16.16~$\pm$~0.81 & 26.41~$\pm$~0.92 & 47.81~$\pm$~2.17 \\
    Color-opponency &  & 19.95~$\pm$~0.57 & 33.97~$\pm$~0.42 & 54.24~$\pm$~0.89 \\
    Single-color &  & 18.53~$\pm$~0.74 & 31.80~$\pm$~1.79 & 51.14~$\pm$~1.43 \\
    \midrule
     - & \multirow{4}{*}{90\%} & 6.50~$\pm$~0.53 & 17.24~$\pm$~1.81 & 35.30~$\pm$~3.21 \\
    Luminance &  & 15.25~$\pm$~1.64 & 26.18~$\pm$~1.46 & 45.41~$\pm$~2.06 \\
    Color-opponency &  & 19.46~$\pm$~0.77 & 33.18~$\pm$~0.59 & 51.55~$\pm$~1.88 \\
    Single-color &  & 15.06~$\pm$~1.28 & 26.01~$\pm$~1.88 & 45.65~$\pm$~1.32 \\
    \bottomrule
  \end{tabular}
  \label{tab:sparse_results_deeplabv3+_dark_zurich_learned}
\end{table}
\clearpage

\begin{table}[tb]
  \centering
  \caption{Evaluation of the \textbf{DeepLabv3+} architecture trained on Cityscapes and \textbf{validated on ACDC Night} at different sparsities. Results include the three preprocessing types (luminance, color-opponency, and single-color). Metrics reported are mean Intersection over Union (mIoU), mean Accuracy (mAcc), and average Accuracy (aAcc), averaged over three seeds. \textbf{Percent-based sparsity:} For a given \textit{percentage}, the \textit{percentage} of lowest absolute values are set to zero.}   
  \begin{tabular}{@{}lc rrr@{}}
    \toprule
    Preprocessing  & Sparsity &\multicolumn{1}{c}{mIoU}&\multicolumn{1}{c}{mAcc}&\multicolumn{1}{c}{aAcc}  \\
    \midrule
     - & \multirow{4}{*}{-} &  8.01~$\pm$~0.45 & 18.14~$\pm$~2.24 & 31.94~$\pm$~2.20 \\
     Luminance  &  & 19.76~$\pm$~0.53 & 34.13~$\pm$~1.05 & 58.90~$\pm$~1.33 \\
    Color-opponency&   &  18.05~$\pm$~1.55 & 31.73~$\pm$~2.54 & 53.51~$\pm$~3.20 \\
    Single-color &   &  19.75~$\pm$~1.99 & 32.24~$\pm$~1.17 & 54.79~$\pm$~2.95 \\
    \midrule
     - & \multirow{4}{*}{40\%} & 12.99~$\pm$~1.08 & 20.03~$\pm$~2.42 & 47.38~$\pm$~3.81 \\
    Luminance &  & 21.38~$\pm$~0.54 & 36.08~$\pm$~1.08 & 59.28~$\pm$~0.92 \\
    Color-opponency &  & 19.98~$\pm$~0.28 & 32.59~$\pm$~0.85 & 56.45~$\pm$~1.19 \\
    Single-color &  & 18.15~$\pm$~1.92 & 31.34~$\pm$~2.39 & 53.05~$\pm$~4.76 \\
    \midrule
     - & \multirow{4}{*}{50\%} & 10.24~$\pm$~1.35 & 18.17~$\pm$~2.25 & 42.60~$\pm$~4.08 \\
    Luminance &  & 21.64~$\pm$~1.10 & 34.38~$\pm$~1.46 & 59.34~$\pm$~2.44 \\
    Color-opponency &  & 20.04~$\pm$~0.98 & 32.63~$\pm$~1.86 & 57.94~$\pm$~0.92 \\
    Single-color &  & 18.96~$\pm$~0.72 & 31.29~$\pm$~1.11 & 54.72~$\pm$~3.40 \\
    \midrule
     - & \multirow{4}{*}{60\%} & 10.37~$\pm$~1.91 & 17.92~$\pm$~2.65 & 42.16~$\pm$~10.41 \\
    Luminance &  & 20.98~$\pm$~0.40 & 32.72~$\pm$~1.59 & 57.80~$\pm$~0.97 \\
    Color-opponency &  & 20.15~$\pm$~1.24 & 33.06~$\pm$~3.39 & 58.75~$\pm$~0.49 \\
    Single-color &  & 19.72~$\pm$~1.15 & 33.21~$\pm$~0.67 & 57.51~$\pm$~0.64 \\
    \midrule
     - & \multirow{4}{*}{70\%} & 13.44~$\pm$~1.63 & 23.68~$\pm$~1.11 & 51.01~$\pm$~5.20 \\
    Luminance &  & 19.40~$\pm$~1.33 & 31.40~$\pm$~2.27 & 56.10~$\pm$~2.09 \\
    Color-opponency &  & 19.22~$\pm$~1.01 & 32.11~$\pm$~1.34 & 56.58~$\pm$~0.73 \\
    Single-color &  & 19.64~$\pm$~1.52 & 31.18~$\pm$~1.17 & 54.54~$\pm$~3.43 \\
    \midrule
     - & \multirow{4}{*}{80\%} & 17.69~$\pm$~2.58 & 29.01~$\pm$~2.92 & 55.15~$\pm$~4.11 \\
    Luminance &  & 17.52~$\pm$~1.48 & 26.65~$\pm$~1.79 & 50.97~$\pm$~2.71 \\
    Color-opponency &  & 21.61~$\pm$~0.48 & 33.73~$\pm$~0.44 & 57.92~$\pm$~0.79 \\
    Single-color &  & 19.30~$\pm$~0.77 & 30.72~$\pm$~0.90 & 54.56~$\pm$~1.18 \\
    \midrule
     - & \multirow{4}{*}{90\%} & 8.99~$\pm$~1.60& 17.82~$\pm$~1.81 & 38.86~$\pm$~4.17 \\
    Luminance &  & 16.35~$\pm$~1.89 & 26.35~$\pm$~1.95 & 47.75~$\pm$~2.43 \\
    Color-opponency &  & 20.81~$\pm$~0.32 & 32.21~$\pm$~0.61 & 55.78~$\pm$~1.76 \\
    Single-color &  & 15.82~$\pm$~0.84 & 25.46~$\pm$~1.24 & 49.29~$\pm$~1.00 \\

    \bottomrule
  \end{tabular}
  \label{tab:sparse_results_deeplabv3+_acdc_night_learned}
\end{table}
\clearpage

\begin{table}[tb]
  \centering
  \caption{Evaluation of the \textbf{DeepLabv3+} architecture trained on Cityscapes and \textbf{validated on ACDC Fog} at different sparsities. Results include the three preprocessing types (luminance, color-opponency, and single-color). Metrics reported are mean Intersection over Union (mIoU), mean Accuracy (mAcc), and average Accuracy (aAcc), averaged over three seeds. \textbf{Percent-based sparsity:} For a given \textit{percentage}, the \textit{percentage} of lowest absolute values are set to zero.}   
  \begin{tabular}{@{}lcrrr@{}}
    \toprule
    Preprocessing  & Sparsity &\multicolumn{1}{c}{mIoU}&\multicolumn{1}{c}{mAcc}&\multicolumn{1}{c}{aAcc}  \\
    \midrule
     - & \multirow{3}{*}{-} &  39.20~$\pm$~4.35 & 54.93~$\pm$~6.11 & 76.52~$\pm$~6.46 \\
     Luminance  &   & 48.90~$\pm$~3.67 & 62.73~$\pm$~3.23 & 84.55~$\pm$~3.74 \\
    Color-opponency&   &  49.46~$\pm$~3.07 & 60.40~$\pm$~4.01 & 87.64~$\pm$~2.93 \\
    Single-color &   &  50.66~$\pm$~3.38 & 62.22~$\pm$~2.75 & 83.45~$\pm$~1.46 \\    
    \midrule
     - & \multirow{4}{*}{40\%} & 30.38~$\pm$~0.84 & 42.41~$\pm$~2.09 & 62.36~$\pm$~5.16 \\
    Luminance &  & 50.42~$\pm$~1.66 & 61.57~$\pm$~0.55 & 88.39~$\pm$~2.30 \\
    Color-opponency &  & 47.48~$\pm$~2.49 & 58.95~$\pm$~0.51 & 84.76~$\pm$~3.27 \\
    Single-color &  & 49.08~$\pm$~1.73 & 61.04~$\pm$~1.97 & 85.04~$\pm$~1.59 \\
    \midrule
     - & \multirow{4}{*}{50\%} & 21.99~$\pm$~1.14 & 34.20~$\pm$~2.70 & 53.79~$\pm$~2.14 \\
    Luminance &  & 50.86~$\pm$~2.88 & 61.20~$\pm$~0.56 & 87.60~$\pm$~3.22 \\
    Color-opponency &  & 48.78~$\pm$~0.28 & 61.79~$\pm$~0.33 & 87.85~$\pm$~1.41 \\
    Single-color &  & 47.97~$\pm$~1.97 & 58.80~$\pm$~1.06 & 87.54~$\pm$~2.16 \\
    \midrule
     - & \multirow{4}{*}{60\%} & 20.36~$\pm$~6.98 & 31.28~$\pm$~5.96 & 59.57~$\pm$~11.10 \\
    Luminance &  & 47.34~$\pm$~2.41 & 57.15~$\pm$~2.38 & 83.85~$\pm$~4.06 \\
    Color-opponency &  & 49.49~$\pm$~4.20 & 59.95~$\pm$~4.83 & 87.81~$\pm$~3.85 \\
    Single-color &  & 47.77~$\pm$~4.80 & 58.99~$\pm$~4.04 & 84.19~$\pm$~5.52 \\
    \midrule
     - & \multirow{4}{*}{70\%} & 17.87~$\pm$~1.80 & 27.75~$\pm$~2.98 & 60.58~$\pm$~6.43 \\
    Luminance &  & 43.19~$\pm$~1.11 & 52.84~$\pm$~0.76 & 84.45~$\pm$~1.48 \\
    Color-opponency &  & 48.82~$\pm$~0.73 & 60.57~$\pm$~0.68 & 85.84~$\pm$~3.06 \\
    Single-color &  & 43.25~$\pm$~0.81 & 54.20~$\pm$~1.63 & 82.91~$\pm$~3.58 \\
    \midrule
     - & \multirow{4}{*}{80\%} & 24.06~$\pm$~3.01 & 34.60~$\pm$~3.33 & 63.52~$\pm$~7.05 \\
    Luminance &  & 40.49~$\pm$~5.60 & 50.60~$\pm$~5.61 & 78.02~$\pm$~8.16 \\
    Color-opponency &  & 49.72~$\pm$~1.32 & 59.13~$\pm$~1.64 & 87.19~$\pm$~2.43 \\
    Single-color &  & 41.44~$\pm$~2.28 & 50.95~$\pm$~0.97 & 85.34~$\pm$~1.15 \\
    \midrule
     - & \multirow{4}{*}{90\%} & 18.05~$\pm$~4.81 & 29.58~$\pm$~5.91 & 57.06~$\pm$~15.13 \\
    Luminance &  & 37.17~$\pm$~2.02 & 47.19~$\pm$~0.69 & 79.19~$\pm$~5.05 \\
    Color-opponency &  & 45.16~$\pm$~2.50 & 55.36~$\pm$~1.13 & 85.07~$\pm$~3.89 \\
    Single-color &  & 29.53~$\pm$~6.07 & 37.44~$\pm$~4.50 & 71.89~$\pm$~12.76 \\
    \bottomrule
  \end{tabular}
  \label{tab:sparse_results_deeplabv3+_acdc_fog_learned}
\end{table}
\clearpage

\begin{table}[tb]
  \centering
  \caption{Evaluation of the \textbf{DeepLabv3+} architecture trained on Cityscapes and \textbf{validated on ACDC Rain} at different sparsities. Results include the three preprocessing types (luminance, color-opponency, and single-color). Metrics reported are mean Intersection over Union (mIoU), mean Accuracy (mAcc), and average Accuracy (aAcc), averaged over three seeds. \textbf{Percent-based sparsity:} For a given \textit{percentage}, the \textit{percentage} of lowest absolute values are set to zero.}   
  \begin{tabular}{@{}lcrrr@{}}
    \toprule
    Preprocessing  & Sparsity &\multicolumn{1}{c}{mIoU}&\multicolumn{1}{c}{mAcc}&\multicolumn{1}{c}{aAcc}  \\
    \midrule
     - & \multirow{4}{*}{-} &  33.31~$\pm$~2.15 & 48.13~$\pm$~4.57 & 77.43~$\pm$~3.46 \\
     Luminance  &   & 34.20~$\pm$~2.50 & 47.68~$\pm$~4.67 & 76.72~$\pm$~3.07 \\
    Color-opponency&   &  38.87~$\pm$~2.24 & 51.26~$\pm$~3.57 & 83.44~$\pm$~3.64 \\
    Single-color &   &  38.60~$\pm$~3.04 & 51.81~$\pm$~2.65 & 77.82~$\pm$~1.63 \\
    \midrule
     - & \multirow{4}{*}{40\%} & 23.15~$\pm$~2.04 & 32.85~$\pm$~3.37 & 66.35~$\pm$~4.42 \\
    Luminance &  & 36.25~$\pm$~0.65 & 49.47~$\pm$~0.94 & 84.51~$\pm$~1.37 \\
    Color-opponency &  & 36.42~$\pm$~1.80 & 49.09~$\pm$~1.45 & 80.44~$\pm$~3.71 \\
    Single-color &  & 38.71~$\pm$~1.35 & 51.81~$\pm$~1.73 & 84.47~$\pm$~1.45 \\
    \midrule
     - & \multirow{4}{*}{50\%} & 19.45~$\pm$~1.93 & 30.13~$\pm$~1.20 & 62.62~$\pm$~6.16 \\
    Luminance &  & 38.50~$\pm$~1.96 & 50.51~$\pm$~1.16 & 85.52~$\pm$~2.42 \\
    Color-opponency &  & 37.90~$\pm$~1.75 & 50.42~$\pm$~1.79 & 84.23~$\pm$~1.64 \\
    Single-color &  & 39.63~$\pm$~1.62 & 52.81~$\pm$~2.60 & 86.99~$\pm$~1.35 \\
    \midrule
     - & \multirow{4}{*}{60\%} & 20.90~$\pm$~3.05 & 30.88~$\pm$~3.77 & 68.83~$\pm$~4.90 \\
    Luminance &  & 36.37~$\pm$~0.91 & 47.49~$\pm$~1.11 & 83.76~$\pm$~2.75 \\
    Color-opponency &  & 38.64~$\pm$~3.13 & 51.87~$\pm$~5.13 & 84.06~$\pm$~3.80 \\
    Single-color &  & 38.46~$\pm$~3.77 & 50.28~$\pm$~3.44 & 84.83~$\pm$~3.75 \\
    \midrule
     - & \multirow{4}{*}{70\%} & 18.22~$\pm$~2.32 & 26.29~$\pm$~3.06 & 61.93~$\pm$~6.97 \\
    Luminance &  & 34.96~$\pm$~1.33 & 43.91~$\pm$~2.27 & 84.64~$\pm$~1.30 \\
    Color-opponency &  & 36.90~$\pm$~0.86 & 48.63~$\pm$~2.20 & 82.04~$\pm$~3.93 \\
    Single-color &  & 36.12~$\pm$~0.51 & 45.95~$\pm$~1.69 & 83.64~$\pm$~1.88 \\
    \midrule
     - & \multirow{4}{*}{80\%} & 21.78~$\pm$~3.78 & 30.07~$\pm$~4.90 & 65.71~$\pm$~8.19 \\
    Luminance &  & 33.60~$\pm$~5.53 & 43.00~$\pm$~4.99 & 79.51~$\pm$~8.46 \\
    Color-opponency &  & 37.05~$\pm$~1.20 & 47.28~$\pm$~2.09 & 84.30~$\pm$~2.37 \\
    Single-color &  & 35.33~$\pm$~1.38 & 43.74~$\pm$~0.93 & 85.11~$\pm$~1.58 \\
    \midrule
     - & \multirow{4}{*}{90\%} & 16.36~$\pm$~3.43 & 28.21~$\pm$~4.71 & 59.82~$\pm$~11.14 \\
    Luminance &  & 31.10~$\pm$~0.98 & 39.23~$\pm$~0.74 & 79.62~$\pm$~3.03 \\
    Color-opponency &  & 36.12~$\pm$~0.91 & 46.09~$\pm$~0.54 & 83.82~$\pm$~2.74 \\
    Single-color &  & 27.80~$\pm$~2.85 & 34.98~$\pm$~2.29 & 77.08~$\pm$~6.87 \\
    \bottomrule
  \end{tabular}
  \label{tab:sparse_results_deeplabv3+_acdc_rain_learned}
\end{table}
\clearpage

\begin{table}[tb]
  \centering
  \caption{Evaluation of the \textbf{DeepLabv3+} architecture trained on Cityscapes and \textbf{validated on ACDC Snow} at different sparsities. Results include the three preprocessing types (luminance, color-opponency, and single-color). Metrics reported are mean Intersection over Union (mIoU), mean Accuracy (mAcc), and average Accuracy (aAcc), averaged over three seeds. \textbf{Percent-based sparsity:} For a given \textit{percentage}, the \textit{percentage} of lowest absolute values are set to zero.}   
  \begin{tabular}{@{}lcrrr@{}}
    \toprule
    Preprocessing  & Sparsity &\multicolumn{1}{c}{mIoU}&\multicolumn{1}{c}{mAcc}&\multicolumn{1}{c}{aAcc}  \\
    \midrule
     - & \multirow{4}{*}{-} & 24.18~$\pm$~3.60 & 36.05~$\pm$~4.01 & 64.03~$\pm$~10.17 \\
    Luminance  &   &31.58~$\pm$~1.88 & 45.09~$\pm$~3.17 & 70.05~$\pm$~3.19 \\
    Color-opponency&   &  32.97~$\pm$~2.97 & 44.40~$\pm$~3.83 & 76.43~$\pm$~5.42 \\
    Single-color &   &  32.60~$\pm$~3.50 & 44.66~$\pm$~2.88 & 69.89~$\pm$~1.47 \\
    \midrule
     - & \multirow{4}{*}{40\%} & 22.14~$\pm$~0.77 & 33.26~$\pm$~1.67 & 54.66~$\pm$~4.28 \\
    Luminance &  & 34.44~$\pm$~2.07 & 47.02~$\pm$~1.72 & 80.22~$\pm$~2.08 \\
    Color-opponency &  & 31.25~$\pm$~1.78 & 43.37~$\pm$~1.46 & 73.62~$\pm$~3.85 \\
    Single-color &  & 33.07~$\pm$~1.26 & 45.41~$\pm$~0.82 & 77.46~$\pm$~1.59 \\
    \midrule
     - & \multirow{4}{*}{50\%} & 20.24~$\pm$~0.95 & 31.68~$\pm$~1.24 & 49.25~$\pm$~1.84 \\
    Luminance &  & 34.48~$\pm$~3.67 & 46.21~$\pm$~2.39 & 77.88~$\pm$~3.40 \\
    Color-opponency &  & 33.80~$\pm$~1.15 & 46.57~$\pm$~0.99 & 77.37~$\pm$~4.23 \\
    Single-color &  & 34.15~$\pm$~2.94 & 46.03~$\pm$~3.01 & 80.14~$\pm$~2.63 \\
    \midrule
     - & \multirow{4}{*}{60\%} & 22.09~$\pm$~3.51 & 33.47~$\pm$~2.01 & 60.41~$\pm$~3.37 \\
    Luminance &  & 33.76~$\pm$~1.12 & 45.29~$\pm$~0.71 & 76.64~$\pm$~3.44 \\
    Color-opponency &  & 34.99~$\pm$~2.65 & 47.24~$\pm$~3.50 & 77.02~$\pm$~5.02 \\
    Single-color &  & 33.18~$\pm$~4.05 & 44.35~$\pm$~3.66 & 78.12~$\pm$~6.05 \\
    \midrule
     - & \multirow{4}{*}{70\%} & 21.76~$\pm$~2.10 & 33.40~$\pm$~1.96 & 59.33~$\pm$~9.39 \\
    Luminance &  & 31.26~$\pm$~1.60 & 41.34~$\pm$~1.34 & 76.87~$\pm$~2.19 \\
    Color-opponency &  & 31.40~$\pm$~1.62 & 43.50~$\pm$~1.28 & 73.48~$\pm$~3.83 \\
    Single-color &  & 29.68~$\pm$~2.71 & 39.80~$\pm$~3.14 & 76.30~$\pm$~5.17 \\
    \midrule
     - & \multirow{4}{*}{80\%} & 22.84~$\pm$~3.02 & 36.32~$\pm$~3.74 & 61.18~$\pm$~6.48 \\
    Luminance &  & 30.30~$\pm$~5.64 & 40.69~$\pm$~4.49 & 71.96~$\pm$~7.89 \\
    Color-opponency &  & 34.22~$\pm$~0.64 & 45.42~$\pm$~0.33 & 79.59~$\pm$~0.79 \\
    Single-color &  & 30.32~$\pm$~1.46 & 39.56~$\pm$~1.21 & 77.99~$\pm$~1.56 \\
    \midrule
     - & \multirow{4}{*}{90\%} & 18.05~$\pm$~3.12 & 33.15~$\pm$~3.11 & 55.06~$\pm$~11.79 \\
    Luminance &  & 28.22~$\pm$~1.15 & 37.40~$\pm$~1.22 & 73.18~$\pm$~4.81 \\
    Color-opponency &  & 31.92~$\pm$~2.34 & 42.72~$\pm$~2.23 & 75.36~$\pm$~3.48 \\
    Single-color &  & 22.49~$\pm$~4.00 & 29.74~$\pm$~3.24 & 65.88~$\pm$~11.30 \\
    \bottomrule
  \end{tabular}
  \label{tab:sparse_results_deeplabv3+_acdc_snow_learned}
\end{table}
\clearpage

\begin{table}[tb]
  \centering
  \caption{Evaluation of the \textbf{DeepLabv3+} architecture trained on Cityscapes and \textbf{validated on ACDC Mean} at different sparsities. Results include the three preprocessing types (luminance, color-opponency, and single-color). Metrics reported are mean Intersection over Union (mIoU), mean Accuracy (mAcc), and average Accuracy (aAcc), averaged over three seeds. \textbf{Percent-based sparsity:} For a given \textit{percentage}, the \textit{percentage} of lowest absolute values are set to zero.}   
  \begin{tabular}{@{}lcrrr@{}}
    \toprule
    Preprocessing  & Sparsity &\multicolumn{1}{c}{mIoU}&\multicolumn{1}{c}{mAcc}&\multicolumn{1}{c}{aAcc}  \\
    \midrule
     - & \multirow{4}{*}{0\%} & 23.59~$\pm$~1.66 & 39.06~$\pm$~4.04 & 62.24~$\pm$~4.48\\
    Luminance &  & 32.80~$\pm$~1.78 & 45.42~$\pm$~2.99 & 72.45~$\pm$~2.77 \\
    Color-opponency &  & 33.74~$\pm$~1.99 & 45.16~$\pm$~3.12 & 75.08~$\pm$~2.96 \\
    Single-color &  & 34.63~$\pm$~3.00 & 45.87~$\pm$~2.19 & 71.36~$\pm$~0.79 \\
    \midrule
     - & \multirow{4}{*}{40\%} & 22.86~$\pm$~0.96 & 37.07~$\pm$~3.11 & 62.10~$\pm$~1.49 \\
    Luminance &  & 34.95~$\pm$~0.76 & 46.55~$\pm$~0.46 & 77.95~$\pm$~1.16 \\
    Color-opponency &  & 33.16~$\pm$~1.36 & 44.36~$\pm$~0.68 & 73.68~$\pm$~2.38 \\
    Single-color &  & 33.87~$\pm$~1.66 & 45.48~$\pm$~1.60 & 74.83~$\pm$~2.23 \\
    \midrule
     - & \multirow{4}{*}{50\%} & 18.87~$\pm$~2.19 & 33.56~$\pm$~3.25 & 56.54~$\pm$~3.41 \\
    Luminance &  & 35.86~$\pm$~2.28 & 46.36~$\pm$~1.12 & 77.44~$\pm$~2.59 \\
    Color-opponency &  & 34.58~$\pm$~1.26 & 46.03~$\pm$~0.52 & 76.70~$\pm$~1.86 \\
    Single-color &  & 34.36~$\pm$~1.37 & 45.14~$\pm$~1.79 & 77.17~$\pm$~0.70 \\
    \midrule
     - & \multirow{4}{*}{60\%} & 18.15~$\pm$~1.10 & 31.93~$\pm$~1.42 & 59.37~$\pm$~1.58 \\
    Luminance &  & 34.17~$\pm$~0.96 & 44.02~$\pm$~1.51 & 75.38~$\pm$~2.72 \\
    Color-opponency &  & 35.58~$\pm$~2.58 & 46.26~$\pm$~4.20 & 76.77~$\pm$~3.23 \\
    Single-color &  & 33.99~$\pm$~3.25 & 44.66~$\pm$~3.14 & 76.02~$\pm$~3.73 \\
    \midrule
     - & \multirow{4}{*}{70\%} & 17.11~$\pm$~2.12 & 29.87~$\pm$~3.01 & 59.10~$\pm$~3.31 \\
    Luminance &  & 31.93~$\pm$~1.61 & 40.82~$\pm$~1.84 & 75.36~$\pm$~1.42 \\
    Color-opponency &  & 33.55~$\pm$~0.17 & 44.37~$\pm$~0.47 & 74.35~$\pm$~2.64 \\
    Single-color &  & 32.04~$\pm$~0.54 & 41.14~$\pm$~1.18 & 74.19~$\pm$~2.08 \\
    \midrule
     - & \multirow{4}{*}{80\%} & 13.57~$\pm$~1.99 & 21.96~$\pm$~2.85 & 51.08~$\pm$~7.02 \\
    Luminance &  & 30.63~$\pm$~4.56 & 39.05~$\pm$~3.92 & 69.96~$\pm$~6.46 \\
    Color-opponency &  & 35.00~$\pm$~0.47 & 44.28~$\pm$~0.45 & 77.10~$\pm$~1.23 \\
    Single-color &  & 31.44~$\pm$~1.36 & 39.82~$\pm$~0.76 & 75.59~$\pm$~1.15 \\
    \midrule
     - & \multirow{4}{*}{90\%} & 10.48~$\pm$~0.88 & 16.83~$\pm$~1.26 & 48.62~$\pm$~3.93 \\
    Luminance &  & 28.38~$\pm$~1.64 & 36.92~$\pm$~0.95 & 69.76~$\pm$~3.54 \\
    Color-opponency &  & 33.39~$\pm$~1.33 & 42.63~$\pm$~1.10 & 74.86~$\pm$~2.69 \\
    Single-color &  & 24.19~$\pm$~3.24 & 31.66~$\pm$~2.62 & 65.91~$\pm$~7.92 \\
    \bottomrule
  \end{tabular}
  \label{tab:sparse_results_deeplabv3+_acdc_full_learned}
\end{table}
\clearpage

%% file: tables/sparsity_trained_mask2former.tex
\begin{table}[h]
  \centering
  \caption{Evaluation of the \textbf{Mask2Former} architecture trained on Cityscapes and \textbf{validated on Cityscapes} at different sparsities. Results include the three preprocessing types (luminance, color-opponency, and single-color). Metrics reported are mean Intersection over Union (mIoU), mean Accuracy (mAcc), and average Accuracy (aAcc), averaged over three seeds. \textbf{Percent-based sparsity:} For a given \textit{percentage}, the \textit{percentage} of lowest absolute values are set to zero.}   
  \begin{tabular}{@{}lcrrr@{}}
    \toprule
    Preprocessing  & Sparsity &\multicolumn{1}{c}{mIoU}&\multicolumn{1}{c}{mAcc}&\multicolumn{1}{c}{aAcc}  \\
    \midrule
     - & \multirow{4}{*}{-} & 77.43~$\pm$~0.35 & 86.83~$\pm$~0.19 & 96.10~$\pm$~0.05 \\
    Luminance  &  & 73.93~$\pm$~0.80 & 85.33~$\pm$~0.92 & 95.59~$\pm$~0.13 \\
    Color-opponency &   &  73.47~$\pm$~0.47 & 84.86~$\pm$~0.57 & 95.54~$\pm$~0.11 \\
    Single-color &   &  74.02~$\pm$~0.62 & 84.64~$\pm$~0.85 & 95.75~$\pm$~0.02 \\
    \midrule
     - & \multirow{4}{*}{40\%} & 69.23~$\pm$~1.59 & 79.78~$\pm$~1.50 & 93.96~$\pm$~0.49 \\
    Luminance &  & 74.06~$\pm$~0.09 & 85.29~$\pm$~0.82 & 95.50~$\pm$~0.03 \\
    Color-opponency &  & 73.42~$\pm$~0.91 & 84.50~$\pm$~0.81 & 95.55~$\pm$~0.08 \\
    Single-color &  & 74.75~$\pm$~1.02 & 85.46~$\pm$~0.47 & 95.73~$\pm$~0.05 \\
    \midrule
     - & \multirow{4}{*}{50\%} & 67.78~$\pm$~0.51 & 78.83~$\pm$~0.39 & 93.92~$\pm$~0.23 \\
    Luminance &  & 72.90~$\pm$~0.75 & 84.76~$\pm$~0.30 & 95.29~$\pm$~0.03 \\
    Color-opponency &  & 73.56~$\pm$~0.62 & 84.49~$\pm$~0.36 & 95.51~$\pm$~0.11 \\
    Single-color &  & 74.45~$\pm$~0.33 & 85.31~$\pm$~0.38 & 95.65~$\pm$~0.06 \\
    \midrule
     - & \multirow{4}{*}{60\%} & 68.43~$\pm$~0.20 & 78.80~$\pm$~0.25 & 93.92~$\pm$~0.13 \\
    Luminance &  & 71.84~$\pm$~1.53 & 83.29~$\pm$~1.00 & 95.19~$\pm$~0.07 \\
    Color-opponency &  & 74.07~$\pm$~1.16 & 85.30~$\pm$~0.99 & 95.52~$\pm$~0.07 \\
    Single-color &  & 72.34~$\pm$~0.04 & 83.42~$\pm$~0.51 & 95.45~$\pm$~0.07 \\
    \midrule
     - & \multirow{4}{*}{70\%} & 65.90~$\pm$~0.53 & 76.35~$\pm$~1.05 & 93.30~$\pm$~0.12 \\
    Luminance &  & 70.24~$\pm$~0.31 & 82.91~$\pm$~0.47 & 94.77~$\pm$~0.07 \\
    Color-opponency &  & 72.44~$\pm$~1.47 & 84.32~$\pm$~1.65 & 95.36~$\pm$~0.04 \\
    Single-color &  & 71.62~$\pm$~1.24 & 82.80~$\pm$~1.23 & 95.24~$\pm$~0.08 \\
    \midrule
     - & \multirow{4}{*}{80\%} & 63.27~$\pm$~0.12 & 74.13~$\pm$~0.31 & 92.71~$\pm$~0.30 \\
    Luminance &  & 66.79~$\pm$~1.28 & 79.36~$\pm$~1.93 & 94.00~$\pm$~0.05 \\
    Color-opponency &  & 71.08~$\pm$~0.47 & 82.86~$\pm$~0.68 & 95.14~$\pm$~0.10 \\
    Single-color &  & 66.68~$\pm$~1.23 & 78.50~$\pm$~0.85 & 94.46~$\pm$~0.12 \\
    \midrule
     - & \multirow{4}{*}{90\%} & 57.55~$\pm$~1.51 & 69.81~$\pm$~2.00 & 91.21~$\pm$~0.26 \\
    Luminance &  & 59.43~$\pm$~0.80 & 72.07~$\pm$~0.80 & 92.01~$\pm$~0.20 \\
    Color-opponency &  & 67.95~$\pm$~1.74 & 79.78~$\pm$~1.85 & 94.40~$\pm$~0.05 \\
    Single-color &  & 62.07~$\pm$~0.88 & 75.22~$\pm$~1.06 & 92.82~$\pm$~0.16 \\

    \bottomrule
  \end{tabular}
  \label{tab:sparse_results_mask2former_cityscapes_learned}
\end{table}
\clearpage

\begin{table}[tb]
  \centering
  \caption{Evaluation of the \textbf{Mask2Former} architecture trained on Cityscapes and \textbf{validated on Dark Zurich} at different sparsities. Results include the three preprocessing types (luminance, color-opponency, and single-color). Metrics reported are mean Intersection over Union (mIoU), mean Accuracy (mAcc), and average Accuracy (aAcc), averaged over three seeds. \textbf{Percent-based sparsity:} For a given \textit{percentage}, the \textit{percentage} of lowest absolute values are set to zero.}   
  \begin{tabular}{@{}lcrrr@{}}
    \toprule
    Preprocessing  & Sparsity &\multicolumn{1}{c}{mIoU}&\multicolumn{1}{c}{mAcc}&\multicolumn{1}{c}{aAcc}  \\
    \midrule
     - & \multirow{4}{*}{-} & 18.16~$\pm$~1.14 & 28.88~$\pm$~2.72 & 47.34~$\pm$~4.13 \\
     Luminance  &  & 26.21~$\pm$~0.18 & 40.56~$\pm$~2.23 & 59.16~$\pm$~0.58 \\
    Color-opponency &   &  22.84~$\pm$~1.82 & 37.37~$\pm$~2.58 & 55.26~$\pm$~2.45 \\
    Single-color &   &  24.92~$\pm$~1.09 & 39.64~$\pm$~2.37 & 57.09~$\pm$~3.52 \\
    \midrule
     - & \multirow{4}{*}{40\%} & 20.59~$\pm$~0.19 & 30.55~$\pm$~0.95 & 53.65~$\pm$~3.14 \\
    Luminance &  & 24.55~$\pm$~2.41 & 39.29~$\pm$~1.75 & 59.32~$\pm$~2.57 \\
    Color-opponency &  & 20.78~$\pm$~1.05 & 36.26~$\pm$~1.06 & 55.13~$\pm$~1.13 \\
    Single-color &  & 24.46~$\pm$~2.08 & 39.46~$\pm$~3.27 & 56.76~$\pm$~0.93 \\
    \midrule
     - & \multirow{4}{*}{50\%} & 21.81~$\pm$~0.94 & 34.36~$\pm$~1.78 & 59.98~$\pm$~3.27 \\
    Luminance &  & 22.81~$\pm$~2.14 & 36.33~$\pm$~2.36 & 55.45~$\pm$~2.46 \\
    Color-opponency &  & 24.61~$\pm$~0.28 & 38.77~$\pm$~1.86 & 57.76~$\pm$~1.76 \\
    Single-color &  & 24.19~$\pm$~1.63 & 38.39~$\pm$~1.16 & 56.10~$\pm$~0.89 \\
    \midrule
     - & \multirow{4}{*}{60\%} & 21.38~$\pm$~2.10 & 33.55~$\pm$~3.35 & 58.65~$\pm$~2.42 \\
    Luminance &  & 22.80~$\pm$~0.92 & 36.77~$\pm$~0.97 & 53.45~$\pm$~1.25 \\
    Color-opponency &  & 23.24~$\pm$~0.85 & 37.25~$\pm$~0.33 & 54.82~$\pm$~1.71 \\
    Single-color &  & 21.68~$\pm$~0.08 & 34.46~$\pm$~1.59 & 50.24~$\pm$~0.91 \\
    \midrule
     - & \multirow{4}{*}{70\%} & 24.87~$\pm$~1.69 & 38.37~$\pm$~1.01 & 62.85~$\pm$~4.05 \\
    Luminance &  & 22.84~$\pm$~1.29 & 35.82~$\pm$~1.15 & 50.30~$\pm$~0.85 \\
    Color-opponency &  & 24.14~$\pm$~1.11 & 38.30~$\pm$~1.14 & 54.56~$\pm$~1.13 \\
    Single-color &  & 20.05~$\pm$~2.31 & 31.85~$\pm$~1.33 & 50.29~$\pm$~1.61 \\
    \midrule
     - & \multirow{4}{*}{80\%} & 24.31~$\pm$~2.06 & 36.53~$\pm$~1.92 & 62.52~$\pm$~5.32 \\
    Luminance &  & 22.62~$\pm$~1.32 & 35.79~$\pm$~1.20 & 50.96~$\pm$~2.38 \\
    Color-opponency &  & 22.96~$\pm$~1.13 & 36.54~$\pm$~0.32 & 51.63~$\pm$~1.03 \\
    Single-color &  & 20.68~$\pm$~1.93 & 32.78~$\pm$~4.22 & 51.49~$\pm$~6.55 \\
    \midrule
     - & \multirow{4}{*}{90\%} & 20.56~$\pm$~1.47 & 30.86~$\pm$~1.53 & 56.34~$\pm$~0.70 \\
    Luminance &  & 23.36~$\pm$~1.45 & 37.42~$\pm$~2.61 & 54.55~$\pm$~0.47 \\
    Color-opponency &  & 20.89~$\pm$~1.85 & 34.90~$\pm$~2.10 & 49.28~$\pm$~1.34 \\
    Single-color &  & 20.50~$\pm$~1.31 & 32.93~$\pm$~2.42 & 50.61~$\pm$~3.04 \\

    \bottomrule
  \end{tabular}
  \label{tab:sparse_results_mask2former_dark_zurich_learned}
\end{table}
\clearpage

\begin{table}[tb]
  \centering
  \caption{Evaluation of the \textbf{Mask2Former} architecture trained on Cityscapes and \textbf{validated on ACDC Night} at different sparsities. Results include the three preprocessing types (luminance, color-opponency, and single-color). Metrics reported are mean Intersection over Union (mIoU), mean Accuracy (mAcc), and average Accuracy (aAcc), averaged over three seeds. \textbf{Percent-based sparsity:} For a given \textit{percentage}, the \textit{percentage} of lowest absolute values are set to zero.}   
  \begin{tabular}{@{}lcrrr@{}}
    \toprule
    Preprocessing  & Sparsity &\multicolumn{1}{c}{mIoU}&\multicolumn{1}{c}{mAcc}&\multicolumn{1}{c}{aAcc}  \\
    \midrule
     - & \multirow{4}{*}{-} & 19.65~$\pm$~1.22 & 32.32~$\pm$~1.80 & 51.57~$\pm$~2.56 \\
     Luminance  &  & 28.45~$\pm$~1.18 & 42.50~$\pm$~2.86 & 62.43~$\pm$~0.83 \\
    Color-opponency &   &  26.08~$\pm$~2.72 & 41.45~$\pm$~3.98 & 58.44~$\pm$~3.31 \\
    Single-color &   &  26.64~$\pm$~0.26 & 42.38~$\pm$~1.67 & 60.03~$\pm$~2.94 \\
    \midrule
     - & \multirow{4}{*}{40\%} & 22.55~$\pm$~0.88 & 31.59~$\pm$~1.75 & 56.46~$\pm$~3.59 \\
    Luminance &  & 27.07~$\pm$~1.74 & 42.51~$\pm$~1.40 & 62.35~$\pm$~2.17 \\
    Color-opponency &  & 23.22~$\pm$~0.91 & 39.71~$\pm$~1.47 & 58.40~$\pm$~1.19 \\
    Single-color &  & 25.93~$\pm$~2.51 & 41.67~$\pm$~3.50 & 60.80~$\pm$~1.72 \\
    \midrule
     - & \multirow{4}{*}{50\%} & 24.80~$\pm$~1.47 & 35.70~$\pm$~2.24 & 63.10~$\pm$~3.06 \\
    Luminance &  & 24.34~$\pm$~0.93 & 38.79~$\pm$~2.71 & 59.01~$\pm$~1.95 \\
    Color-opponency &  & 25.45~$\pm$~0.81 & 40.41~$\pm$~2.15 & 59.86~$\pm$~1.45 \\
    Single-color &  & 26.29~$\pm$~2.48 & 41.11~$\pm$~2.18 & 59.86~$\pm$~0.78 \\
    \midrule
     - & \multirow{4}{*}{60\%} & 24.95~$\pm$~0.29 & 35.83~$\pm$~0.76 & 62.47~$\pm$~1.67 \\
    Luminance &  & 24.27~$\pm$~1.09 & 38.86~$\pm$~0.65 & 56.42~$\pm$~0.58 \\
    Color-opponency &  & 24.81~$\pm$~0.52 & 39.19~$\pm$~1.34 & 57.58~$\pm$~1.46 \\
    Single-color &  & 22.86~$\pm$~0.23 & 35.90~$\pm$~2.18 & 52.94~$\pm$~1.00 \\
    \midrule
     - & \multirow{4}{*}{70\%} & 27.93~$\pm$~1.50 & 39.28~$\pm$~1.30 & 66.71~$\pm$~2.90 \\
    Luminance &  & 23.89~$\pm$~0.86 & 36.49~$\pm$~0.67 & 52.93~$\pm$~1.53 \\
    Color-opponency &  & 25.45~$\pm$~0.53 & 40.28~$\pm$~1.59 & 57.29~$\pm$~1.06 \\
    Single-color &  & 21.14~$\pm$~1.42 & 32.91~$\pm$~1.82 & 52.75~$\pm$~2.29 \\
    \midrule
     - & \multirow{4}{*}{80\%} & 26.99~$\pm$~1.07 & 37.94~$\pm$~1.75 & 66.37~$\pm$~3.92 \\
    Luminance &  & 22.97~$\pm$~1.33 & 36.05~$\pm$~1.63 & 53.66~$\pm$~1.90 \\
    Color-opponency &  & 24.94~$\pm$~0.03 & 37.37~$\pm$~1.12 & 53.87~$\pm$~2.04 \\
    Single-color &  & 21.63~$\pm$~1.45 & 33.30~$\pm$~1.76 & 51.15~$\pm$~4.91 \\
    \midrule
     - & \multirow{4}{*}{90\%} & 22.99~$\pm$~2.08 & 31.95~$\pm$~2.82 & 60.11~$\pm$~1.71 \\
    Luminance &  & 23.23~$\pm$~1.18 & 36.59~$\pm$~2.00 & 55.72~$\pm$~0.65 \\
    Color-opponency &  & 21.76~$\pm$~2.68 & 35.45~$\pm$~2.78 & 51.42~$\pm$~1.85 \\
    Single-color &  & 21.82~$\pm$~0.95 & 34.18~$\pm$~2.92 & 52.20~$\pm$~2.15 \\

    \bottomrule
  \end{tabular}
  \label{tab:sparse_results_mask2former_acdc_night_learned}
\end{table}
\clearpage

\begin{table}[tb]
  \centering
  \caption{Evaluation of the \textbf{Mask2Former} architecture trained on Cityscapes and \textbf{validated on ACDC Fog} at different sparsities. Results include the three preprocessing types (luminance, color-opponency, and single-color). Metrics reported are mean Intersection over Union (mIoU), mean Accuracy (mAcc), and average Accuracy (aAcc), averaged over three seeds. \textbf{Percent-based sparsity:} For a given \textit{percentage}, the \textit{percentage} of lowest absolute values are set to zero. }   
  \begin{tabular}{@{}lcrrr@{}}
    \toprule
    Preprocessing  & Sparsity &\multicolumn{1}{c}{mIoU}&\multicolumn{1}{c}{mAcc}&\multicolumn{1}{c}{aAcc}  \\
    \midrule
     - & \multirow{4}{*}{-} & 64.70~$\pm$~3.04 & 79.77~$\pm$~3.06 & 91.08~$\pm$~1.46 \\
     Luminance  & & 62.16~$\pm$~1.18 & 75.22~$\pm$~0.88 & 91.30~$\pm$~1.05 \\
    Color-opponency &   &  61.58~$\pm$~0.95 & 77.08~$\pm$~1.58 & 89.45~$\pm$~1.28 \\
    Single-color &   &  63.77~$\pm$~3.25 & 77.78~$\pm$~2.56 & 90.43~$\pm$~1.33 \\
    \midrule
     - & \multirow{4}{*}{40\%} & 55.36~$\pm$~1.97 & 67.82~$\pm$~1.12 & 84.92~$\pm$~1.04 \\
    Luminance &  & 61.67~$\pm$~1.74 & 74.15~$\pm$~2.53 & 91.50~$\pm$~1.83 \\
    Color-opponency &  & 61.39~$\pm$~2.98 & 75.37~$\pm$~2.02 & 90.01~$\pm$~2.33 \\
    Single-color &  & 60.47~$\pm$~1.85 & 75.37~$\pm$~1.48 & 90.37~$\pm$~1.76 \\
    \midrule
     - & \multirow{4}{*}{50\%} & 55.07~$\pm$~1.49 & 67.67~$\pm$~1.98 & 86.40~$\pm$~0.75 \\
    Luminance &  & 62.58~$\pm$~4.62 & 75.27~$\pm$~4.88 & 90.36~$\pm$~2.12 \\
    Color-opponency &  & 63.87~$\pm$~4.72 & 76.91~$\pm$~2.56 & 91.62~$\pm$~0.78 \\
    Single-color &  & 65.42~$\pm$~2.17 & 78.72~$\pm$~1.13 & 89.37~$\pm$~1.50 \\
    \midrule
     - & \multirow{4}{*}{60\%} & 54.83~$\pm$~0.96 & 68.30~$\pm$~1.12 & 84.97~$\pm$~1.73 \\
    Luminance &  & 60.28~$\pm$~1.50 & 73.88~$\pm$~0.69 & 89.93~$\pm$~0.13 \\
    Color-opponency &  & 62.79~$\pm$~4.87 & 75.55~$\pm$~5.82 & 91.77~$\pm$~1.39 \\
    Single-color &  & 60.59~$\pm$~3.55 & 74.24~$\pm$~4.35 & 89.71~$\pm$~1.75 \\
    \midrule
     - & \multirow{4}{*}{70\%} & 52.08~$\pm$~2.88 & 64.66~$\pm$~2.04 & 85.53~$\pm$~2.14 \\
    Luminance &  & 57.08~$\pm$~1.55 & 70.03~$\pm$~0.96 & 88.70~$\pm$~0.83 \\
    Color-opponency &  & 64.41~$\pm$~0.62 & 77.02~$\pm$~1.18 & 91.63~$\pm$~0.45 \\
    Single-color &  & 57.85~$\pm$~3.61 & 71.28~$\pm$~2.90 & 87.34~$\pm$~2.14 \\
    \midrule
     - & \multirow{4}{*}{80\%} & 46.10~$\pm$~3.99 & 62.30~$\pm$~2.73 & 81.43~$\pm$~2.50 \\
    Luminance &  & 54.75~$\pm$~0.80 & 67.52~$\pm$~1.15 & 89.02~$\pm$~1.21 \\
    Color-opponency &  & 58.61~$\pm$~1.96 & 71.97~$\pm$~0.90 & 88.20~$\pm$~1.30 \\
    Single-color &  & 53.41~$\pm$~6.33 & 66.78~$\pm$~5.48 & 86.12~$\pm$~1.56 \\
    \midrule
     - & \multirow{4}{*}{90\%} & 35.54~$\pm$~2.67 & 55.17~$\pm$~0.58 & 65.14~$\pm$~9.60 \\
    Luminance &  & 49.65~$\pm$~3.13 & 61.94~$\pm$~1.46 & 87.69~$\pm$~0.73 \\
    Color-opponency &  & 51.86~$\pm$~2.74 & 66.80~$\pm$~1.66 & 86.83~$\pm$~2.13 \\
    Single-color &  & 45.67~$\pm$~2.21 & 58.55~$\pm$~2.00 & 85.22~$\pm$~1.24 \\
    \bottomrule
  \end{tabular}
  \label{tab:sparse_results_mask2former_acdc_fog_learned}
\end{table}
\clearpage

\begin{table}[tb]
  \centering
  \caption{Evaluation of the \textbf{Mask2Former} architecture trained on Cityscapes and \textbf{validated on ACDC Rain} at different sparsities. Results include the three preprocessing types (luminance, color-opponency, and single-color). Metrics reported are mean Intersection over Union (mIoU), mean Accuracy (mAcc), and average Accuracy (aAcc), averaged over three seeds. \textbf{Percent-based sparsity:} For a given \textit{percentage}, the \textit{percentage} of lowest absolute values are set to zero. }   
  \begin{tabular}{@{}lcrrr@{}}
    \toprule
    Preprocessing  & Sparsity &\multicolumn{1}{c}{mIoU}&\multicolumn{1}{c}{mAcc}&\multicolumn{1}{c}{aAcc}  \\
    \midrule
     - & \multirow{4}{*}{-} & 49.91~$\pm$~1.32 & 71.13~$\pm$~1.71 & 87.97~$\pm$~0.38 \\
     Luminance  &  & 47.20~$\pm$~0.13 & 64.00~$\pm$~2.66 & 87.95~$\pm$~1.50 \\
    Color-opponency &   &  44.45~$\pm$~0.48 & 60.91~$\pm$~0.56 & 86.78~$\pm$~1.08 \\
    Single-color &   &  44.86~$\pm$~3.38 & 62.10~$\pm$~5.34 & 86.84~$\pm$~3.21 \\
    \midrule
     - & \multirow{4}{*}{40\%} & 45.01~$\pm$~3.27 & 58.35~$\pm$~3.42 & 83.09~$\pm$~1.74 \\
    Luminance &  & 46.69~$\pm$~2.35 & 62.64~$\pm$~4.76 & 89.06~$\pm$~1.22 \\
    Color-opponency &  & 46.39~$\pm$~1.65 & 63.18~$\pm$~3.58 & 86.97~$\pm$~1.23 \\
    Single-color &  & 46.45~$\pm$~2.40 & 64.05~$\pm$~3.47 & 87.37~$\pm$~2.14 \\
    \midrule
     - & \multirow{4}{*}{50\%} & 44.44~$\pm$~3.10 & 56.97~$\pm$~2.83 & 85.47~$\pm$~1.20 \\
    Luminance &  & 46.44~$\pm$~2.85 & 61.78~$\pm$~3.37 & 87.68~$\pm$~1.25 \\
    Color-opponency &  & 47.07~$\pm$~1.86 & 63.86~$\pm$~1.82 & 88.02~$\pm$~0.99 \\
    Single-color &  & 45.40~$\pm$~2.76 & 62.42~$\pm$~3.43 & 87.43~$\pm$~2.00 \\
    \midrule
     - & \multirow{4}{*}{60\%} & 44.59~$\pm$~1.43 & 57.51~$\pm$~2.47 & 84.78~$\pm$~0.93 \\
    Luminance &  & 45.36~$\pm$~1.90 & 60.63~$\pm$~1.27 & 88.65~$\pm$~0.13 \\
    Color-opponency &  & 48.65~$\pm$~3.58 & 64.84~$\pm$~1.25 & 88.52~$\pm$~1.65 \\
    Single-color &  & 46.56~$\pm$~2.41 & 64.06~$\pm$~1.58 & 87.17~$\pm$~2.16 \\
    \midrule
     - & \multirow{4}{*}{70\%} & 43.97~$\pm$~1.74 & 56.31~$\pm$~1.18 & 84.74~$\pm$~0.87 \\
    Luminance &  & 45.76~$\pm$~1.95 & 58.34~$\pm$~3.24 & 88.28~$\pm$~0.62 \\
    Color-opponency &  & 48.09~$\pm$~0.74 & 64.77~$\pm$~1.10 & 89.05~$\pm$~0.58 \\
    Single-color &  & 42.96~$\pm$~1.14 & 58.32~$\pm$~0.68 & 85.73~$\pm$~2.01 \\
    \midrule
     - & \multirow{4}{*}{80\%} & 42.27~$\pm$~0.62 & 55.94~$\pm$~1.28 & 83.73~$\pm$~1.68 \\
    Luminance &  & 42.66~$\pm$~0.97 & 53.26~$\pm$~0.97 & 87.76~$\pm$~0.79 \\
    Color-opponency &  & 47.35~$\pm$~2.45 & 63.45~$\pm$~2.54 & 86.93~$\pm$~1.64 \\
    Single-color &  & 41.54~$\pm$~1.62 & 55.37~$\pm$~1.65 & 86.10~$\pm$~0.67 \\
    \midrule
     - & \multirow{4}{*}{90\%} & 35.86~$\pm$~4.55 & 51.33~$\pm$~4.30 & 72.05~$\pm$~7.57 \\
    Luminance &  & 39.03~$\pm$~0.44 & 48.96~$\pm$~0.71 & 85.64~$\pm$~0.86 \\
    Color-opponency &  & 44.76~$\pm$~3.67 & 59.80~$\pm$~4.44 & 85.64~$\pm$~4.08 \\
    Single-color &  & 36.25~$\pm$~2.08 & 46.05~$\pm$~1.50 & 84.15~$\pm$~2.15 \\
    \bottomrule
  \end{tabular}
  \label{tab:sparse_results_mask2former_acdc_rain_learned}
\end{table}
\clearpage

\begin{table}[tb]
  \centering
  \caption{Evaluation of the \textbf{Mask2Former} architecture trained on Cityscapes and \textbf{validated on ACDC Snow} at different sparsities. Results include the three preprocessing types (luminance, color-opponency, and single-color). Metrics reported are mean Intersection over Union (mIoU), mean Accuracy (mAcc), and average Accuracy (aAcc), averaged over three seeds. \textbf{Percent-based sparsity:} For a given \textit{percentage}, the \textit{percentage} of lowest absolute values are set to zero.}   
  \begin{tabular}{@{}lcrrr@{}}
    \toprule
    Preprocessing  & Sparsity &\multicolumn{1}{c}{mIoU}&\multicolumn{1}{c}{mAcc}&\multicolumn{1}{c}{aAcc}  \\
    \midrule
     - & \multirow{4}{*}{-} & 48.20~$\pm$~2.09 & 62.94~$\pm$~2.16 & 81.98~$\pm$~1.71 \\
     Luminance  &  & 49.87~$\pm$~2.06 & 60.95~$\pm$~0.88 & 85.47~$\pm$~2.31 \\
    Color-opponency &   &  47.48~$\pm$~0.67 & 60.15~$\pm$~0.81 & 82.73~$\pm$~1.13 \\
    Single-color &   &  46.97~$\pm$~3.98 & 61.26~$\pm$~5.43 & 81.86~$\pm$~3.87 \\
    \midrule
     - & \multirow{4}{*}{40\%} & 48.03~$\pm$~2.26 & 58.69~$\pm$~2.26 & 82.53~$\pm$~1.41 \\
    Luminance &  & 47.90~$\pm$~1.03 & 59.27~$\pm$~0.82 & 86.74~$\pm$~0.59 \\
    Color-opponency &  & 49.63~$\pm$~2.27 & 61.88~$\pm$~1.75 & 84.48~$\pm$~1.46 \\
    Single-color &  & 44.80~$\pm$~2.42 & 58.75~$\pm$~3.27 & 82.85~$\pm$~2.57 \\
    \midrule
     - & \multirow{4}{*}{50\%} & 48.62~$\pm$~0.51 & 59.69~$\pm$~1.25 & 83.61~$\pm$~0.80 \\
    Luminance &  & 50.68~$\pm$~1.87 & 61.81~$\pm$~3.10 & 85.23~$\pm$~1.95 \\
    Color-opponency &  & 48.38~$\pm$~3.60 & 60.82~$\pm$~2.84 & 84.51~$\pm$~1.46 \\
    Single-color &  & 50.07~$\pm$~2.06 & 61.91~$\pm$~0.91 & 83.67~$\pm$~0.81 \\
    \midrule
     - & \multirow{4}{*}{60\%} & 48.69~$\pm$~1.44 & 61.44~$\pm$~1.32 & 83.31~$\pm$~1.23 \\
    Luminance &  & 49.40~$\pm$~0.67 & 60.49~$\pm$~0.90 & 86.69~$\pm$~0.87 \\
    Color-opponency &  & 50.51~$\pm$~3.84 & 63.37~$\pm$~2.88 & 84.58~$\pm$~1.84 \\
    Single-color &  & 49.37~$\pm$~0.99 & 61.09~$\pm$~0.48 & 83.02~$\pm$~1.61 \\
    \midrule
     - & \multirow{4}{*}{70\%} & 47.30~$\pm$~0.99 & 60.51~$\pm$~1.29 & 82.18~$\pm$~0.26 \\
    Luminance &  & 44.55~$\pm$~1.93 & 53.86~$\pm$~1.71 & 85.38~$\pm$~1.65 \\
    Color-opponency &  & 52.23~$\pm$~1.57 & 64.33~$\pm$~1.51 & 85.84~$\pm$~1.44 \\
    Single-color &  & 43.33~$\pm$~1.25 & 53.03~$\pm$~0.53 & 83.10~$\pm$~1.57 \\
    \midrule
     - & \multirow{4}{*}{80\%} & 43.34~$\pm$~2.05 & 56.14~$\pm$~2.53 & 78.92~$\pm$~3.46 \\
    Luminance &  & 48.46~$\pm$~0.53 & 58.07~$\pm$~1.04 & 85.64~$\pm$~1.71 \\
    Color-opponency &  & 49.06~$\pm$~0.70 & 61.06~$\pm$~0.18 & 82.80~$\pm$~2.37 \\
    Single-color &  & 43.82~$\pm$~1.19 & 53.70~$\pm$~1.15 & 83.92~$\pm$~0.96 \\
    \midrule
     - & \multirow{4}{*}{90\%} & 32.19~$\pm$~1.77 & 47.01~$\pm$~1.50 & 63.27~$\pm$~6.83 \\
    Luminance &  & 42.24~$\pm$~0.51 & 52.30~$\pm$~0.91 & 82.44~$\pm$~0.50 \\
    Color-opponency &  & 46.86~$\pm$~2.86 & 58.37~$\pm$~3.12 & 82.19~$\pm$~2.53 \\
    Single-color &  & 39.28~$\pm$~1.46 & 47.81~$\pm$~1.64 & 80.70~$\pm$~2.18 \\
    \bottomrule
  \end{tabular}
  \label{tab:sparse_results_mask2former_acdc_snow_learned}
\end{table}
\clearpage

\begin{table}[tb]
  \centering
  \caption{Evaluation of the \textbf{Mask2Former} architecture trained on Cityscapes and \textbf{validated on ACDC Mean} at different sparsities. Results include the three preprocessing types (luminance, color-opponency, and single-color). Metrics reported are mean Intersection over Union (mIoU), mean Accuracy (mAcc), and average Accuracy (aAcc), averaged over three seeds. \textbf{Percent-based sparsity:} For a given \textit{percentage}, the \textit{percentage} of lowest absolute values are set to zero.}   
  \begin{tabular}{@{}lcrrr@{}}
    \toprule
    Preprocessing  & Sparsity &\multicolumn{1}{c}{mIoU}&\multicolumn{1}{c}{mAcc}&\multicolumn{1}{c}{aAcc}  \\
    \midrule
     - & \multirow{4}{*}{-} & 42.05~$\pm$~1.41 & 60.13~$\pm$~2.21 & 77.94~$\pm$~1.35\\
    Luminance &  & 46.83~$\pm$~1.10 & 59.72~$\pm$~2.32 & 81.61~$\pm$~1.38 \\
    Color-opponency &  & 44.93~$\pm$~0.66 & 58.84~$\pm$~0.86 & 79.18~$\pm$~0.62 \\
    Single-color &  & 45.23~$\pm$~1.74 & 59.85~$\pm$~2.53 & 79.63~$\pm$~2.76 \\
    \midrule
     - & \multirow{4}{*}{40\%} & 40.06~$\pm$~2.03 & 56.98~$\pm$~1.54 & 77.01~$\pm$~1.26 \\
    Luminance &  & 45.30~$\pm$~1.80 & 58.31~$\pm$~1.43 & 82.25~$\pm$~1.36 \\
    Color-opponency &  & 44.03~$\pm$~1.65 & 58.48~$\pm$~1.05 & 79.80~$\pm$~1.01 \\
    Single-color &  & 44.22~$\pm$~2.67 & 58.76~$\pm$~3.18 & 80.19~$\pm$~2.04 \\
    \midrule
     - & \multirow{4}{*}{50\%} & 37.31~$\pm$~2.00 & 54.79~$\pm$~0.35 & 76.21~$\pm$~1.44 \\
    Luminance &  & 45.07~$\pm$~2.74 & 58.64~$\pm$~3.78 & 80.40~$\pm$~1.76 \\
    Color-opponency &  & 45.20~$\pm$~1.90 & 58.70~$\pm$~0.96 & 80.83~$\pm$~0.61 \\
    Single-color &  & 46.11~$\pm$~2.09 & 59.69~$\pm$~0.85 & 79.92~$\pm$~1.00 \\
    \midrule
     - & \multirow{4}{*}{60\%} & 33.24~$\pm$~1.17 & 51.66~$\pm$~0.67 & 75.09~$\pm$~1.64 \\
    Luminance &  & 44.65~$\pm$~0.77 & 57.25~$\pm$~0.81 & 80.23~$\pm$~0.12 \\
    Color-opponency &  & 46.88~$\pm$~2.88 & 60.02~$\pm$~1.43 & 80.43~$\pm$~0.91 \\
    Single-color &  & 45.06~$\pm$~1.40 & 58.73~$\pm$~0.69 & 78.01~$\pm$~1.43 \\
    \midrule
     - & \multirow{4}{*}{70\%} & 29.98~$\pm$~1.35 & 47.38~$\pm$~1.74 & 72.65~$\pm$~0.84 \\
    Luminance &  & 42.85~$\pm$~1.13 & 53.43~$\pm$~1.46 & 78.62~$\pm$~0.97 \\
    Color-opponency &  & 47.44~$\pm$~0.52 & 60.34~$\pm$~0.97 & 80.76~$\pm$~0.37 \\
    Single-color &  & 40.09~$\pm$~0.83 & 52.30~$\pm$~1.55 & 77.03~$\pm$~1.97 \\
    \midrule
     - & \multirow{4}{*}{80\%} & 23.81~$\pm$~1.42 & 38.26~$\pm$~0.92 & 68.70~$\pm$~2.25 \\
    Luminance &  & 42.39~$\pm$~0.64 & 53.14~$\pm$~0.54 & 78.82~$\pm$~1.12 \\
    Color-opponency &  & 45.95~$\pm$~0.96 & 57.95~$\pm$~1.04 & 77.76~$\pm$~1.67 \\
    Single-color &  & 39.90~$\pm$~1.85 & 50.28~$\pm$~2.43 & 76.62~$\pm$~1.37 \\
    \midrule
     - & \multirow{4}{*}{90\%} & 17.03~$\pm$~1.10 & 26.92~$\pm$~1.09 & 59.84~$\pm$~4.43 \\
    Luminance &  & 37.84~$\pm$~0.49 & 48.54~$\pm$~1.21 & 77.70~$\pm$~0.56 \\
    Color-opponency &  & 41.39~$\pm$~1.19 & 53.57~$\pm$~1.46 & 76.32~$\pm$~1.67 \\
    Single-color &  & 35.69~$\pm$~0.14 & 45.48~$\pm$~0.84 & 75.38~$\pm$~1.20 \\
    \bottomrule
  \end{tabular}
  \label{tab:sparse_results_mask2former_acdc_full_learned}
\end{table}
\clearpage

%% file: tex_for_figures/mask2former_peak.tex
\begin{figure*}[tb]
  \centering
  \includegraphics[width=\linewidth]{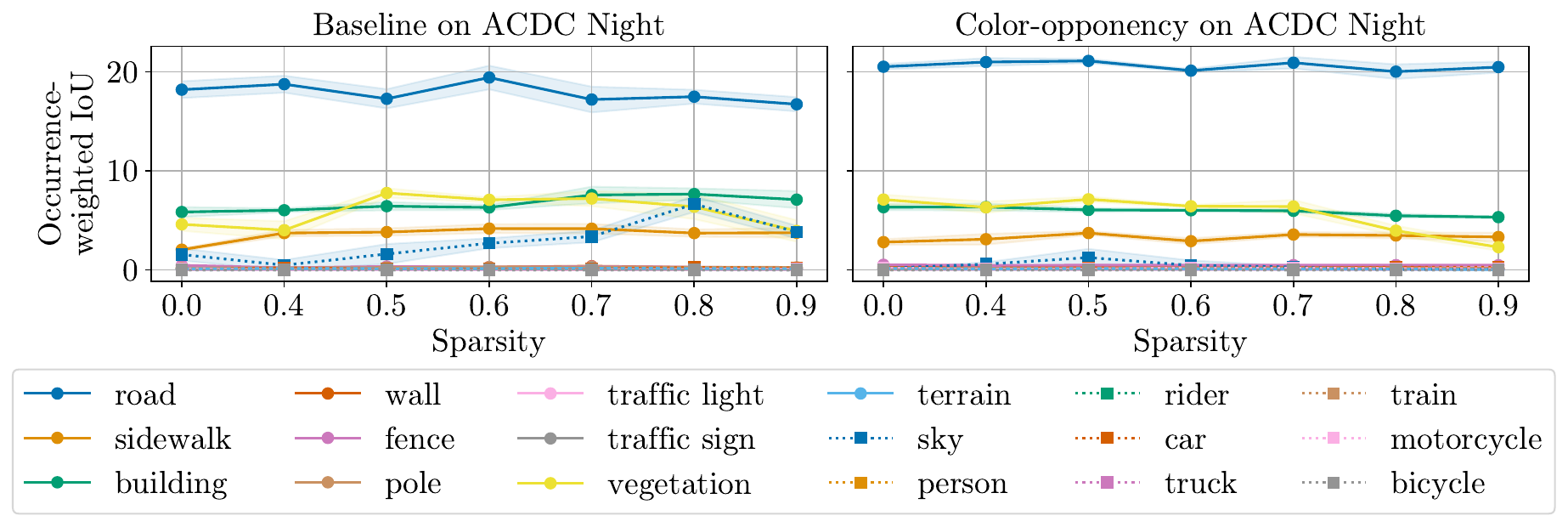}
  \caption{Frequency-weighted mean IoU (per-class IoU weighted by pixel frequency in the validation set) for the Mask2Former architecture. The results cover 0\%  sparsity as well as sparsities from 40\% to 90\% in increments of 10\%, where 0\% denotes testing without sparsity. All preprocessing variations, luminance, color-opponency, and single-color, were trained including contrast extraction.}
  \label{fig:maks2former_peak_analysis}
\end{figure*}

%The Mask2Former Baseline Peak at about 70\% Sparsity.} We conducted additional analyses using a frequency-weighted mean IoU (per-class IoU weighted by pixel frequency in the validation set). The sparsified RGB baseline shows a bias towards frequent classes (such as sky), suggesting reliance on class priors rather than robust sparse features, whereas our preprocessing does not exhibit this behavior, indicating greater robustness under sparsification.

%% file: tables/sparsity_trained_upernet.tex
\begin{table}[h]
  \centering
  \caption{Evaluation of the \textbf{UPerNet} architecture trained on Cityscapes and \textbf{validated on Cityscapes} at different sparsities. Results include the three preprocessing types (luminance, color-opponency, and single-color). Metrics reported are mean Intersection over Union (mIoU), mean Accuracy (mAcc), and average Accuracy (aAcc), averaged over three seeds. \textbf{Percent-based sparsity:} For a given \textit{percentage}, the \textit{percentage} of lowest absolute values are set to zero. }   
  \begin{tabular}{@{}lcrrr@{}}
    \toprule
    Preprocessing  & Sparsity &\multicolumn{1}{c}{mIoU}&\multicolumn{1}{c}{mAcc}&\multicolumn{1}{c}{aAcc}  \\
    \midrule
     - & \multirow{4}{*}{-} & 64.64~$\pm$~1.93 & 72.63~$\pm$~1.25 & 94.42~$\pm$~0.17 \\
    Luminance &  & 62.63~$\pm$~1.79 & 71.37~$\pm$~1.79 & 93.85~$\pm$~0.11 \\
    Color-opponency&   &  64.33~$\pm$~1.32 & 72.12~$\pm$~1.90 & 94.63~$\pm$~0.13 \\
    Single-color &   &  63.37~$\pm$~2.80 & 72.26~$\pm$~2.80 & 94.47~$\pm$~0.20 \\
    \midrule
    - & \multirow{4}{*}{40\%} & 42.44~$\pm$~2.37 & 51.36~$\pm$~2.96 & 86.13~$\pm$~2.55 \\
    Luminance &  & 60.42~$\pm$~0.47 & 68.63~$\pm$~0.50 & 93.68~$\pm$~0.25 \\
    Color-opponency &  & 63.46~$\pm$~2.25 & 71.83~$\pm$~2.66 & 94.54~$\pm$~0.13 \\
    Single-color &  & 62.69~$\pm$~0.82 & 70.38~$\pm$~0.37 & 94.32~$\pm$~0.19 \\
    \midrule
    - & \multirow{4}{*}{50\%} & 41.17~$\pm$~2.84 & 52.06~$\pm$~3.91 & 83.57~$\pm$~1.51 \\
    Luminance &  & 58.99~$\pm$~0.94 & 66.86~$\pm$~0.93 & 93.52~$\pm$~0.18 \\
    Color-opponency &  & 63.96~$\pm$~4.59 & 72.60~$\pm$~4.19 & 94.50~$\pm$~0.33 \\
    Single-color &  & 63.31~$\pm$~3.50 & 71.88~$\pm$~4.06 & 94.42~$\pm$~0.27 \\
    \midrule
    - & \multirow{4}{*}{60\%} & 41.25~$\pm$~7.88 & 51.40~$\pm$~7.23 & 81.63~$\pm$~7.82 \\
    Luminance &  & 57.66~$\pm$~0.84 & 66.39~$\pm$~1.55 & 92.79~$\pm$~0.19 \\
    Color-opponency &  & 61.61~$\pm$~0.97 & 69.28~$\pm$~1.74 & 94.21~$\pm$~0.06 \\
    Single-color &  & 61.19~$\pm$~2.55 & 69.67~$\pm$~3.14 & 93.83~$\pm$~0.27 \\
    \midrule
    - & \multirow{4}{*}{70\%} & 26.33~$\pm$~7.12 & 38.14~$\pm$~6.71 & 65.08~$\pm$~13.53 \\
    Luminance &  & 59.66~$\pm$~3.53 & 68.90~$\pm$~3.68 & 92.61~$\pm$~0.68 \\
    Color-opponency &  & 62.99~$\pm$~3.01 & 71.25~$\pm$~3.06 & 94.22~$\pm$~0.26 \\
    Single-color &  & 60.92~$\pm$~0.63 & 70.05~$\pm$~1.43 & 93.17~$\pm$~0.29 \\
    \midrule
    - & \multirow{4}{*}{80\%} & 26.54~$\pm$~10.02 & 38.83~$\pm$~7.95 & 64.37~$\pm$~10.63 \\
    Luminance &  & 56.31~$\pm$~2.63 & 66.07~$\pm$~2.84 & 91.51~$\pm$~0.45 \\
    Color-opponency &  & 59.32~$\pm$~0.58 & 67.48~$\pm$~0.46 & 93.47~$\pm$~0.14 \\
    Single-color &  & 60.12~$\pm$~1.14 & 69.00~$\pm$~1.35 & 92.57~$\pm$~0.29 \\
    \midrule
    - & \multirow{4}{*}{90\%} & 29.54~$\pm$~0.95 & 40.41~$\pm$~2.01 & 73.29~$\pm$~2.49 \\
    Luminance &  & 49.14~$\pm$~0.72 & 58.79~$\pm$~2.34 & 88.53~$\pm$~0.23 \\
    Color-opponency &  & 59.10~$\pm$~3.33 & 67.97~$\pm$~4.00 & 92.57~$\pm$~0.25 \\
    Single-color &  & 53.51~$\pm$~0.83 & 63.21~$\pm$~1.34 & 90.20~$\pm$~0.26 \\
    \bottomrule
  \end{tabular}
  \label{tab:sparse_results_upernet_cityscapes_learned}
\end{table}
\clearpage

\begin{table}[tb]
  \centering
  \caption{Evaluation of the \textbf{UPerNet} architecture trained on Cityscapes and \textbf{validated on Dark Zurich} at different sparsities. Results include the three preprocessing types (luminance, color-opponency, and single-color). Metrics reported are mean Intersection over Union (mIoU), mean Accuracy (mAcc), and average Accuracy (aAcc), averaged over three seeds. \textbf{Percent-based sparsity:} For a given \textit{percentage}, the \textit{percentage} of lowest absolute values are set to zero.}   
  \begin{tabular}{@{}lcrrr@{}}
    \toprule
    Preprocessing  & Sparsity &\multicolumn{1}{c}{mIoU}&\multicolumn{1}{c}{mAcc}&\multicolumn{1}{c}{aAcc}  \\
    \midrule
    - & \multirow{4}{*}{-} & 11.42~$\pm$~0.58 & 21.92~$\pm$~2.60 & 42.86~$\pm$~6.74 \\
     Luminance &   & 18.03~$\pm$~1.82 & 32.15~$\pm$~1.79 & 50.00~$\pm$~1.10 \\
    Color-opponency&   &  16.10~$\pm$~0.95 & 29.13~$\pm$~1.21 & 46.56~$\pm$~4.08 \\
    Single-color &   &  13.82~$\pm$~1.52 & 27.68~$\pm$~1.50 & 44.06~$\pm$~3.62 \\
    \midrule
    - & \multirow{4}{*}{40\%} & 11.68~$\pm$~1.77 & 17.55~$\pm$~3.61 & 43.33~$\pm$~8.15 \\
    Luminance &  & 17.00~$\pm$~0.84 & 28.83~$\pm$~2.63 & 50.89~$\pm$~0.98 \\
    Color-opponency &  & 15.66~$\pm$~1.77 & 28.69~$\pm$~1.37 & 48.64~$\pm$~1.52 \\
    Single-color &  & 16.01~$\pm$~2.47 & 28.70~$\pm$~2.72 & 49.09~$\pm$~8.97 \\
    \midrule
    - & \multirow{4}{*}{50\%} & 11.63~$\pm$~1.88 & 18.67~$\pm$~2.59 & 45.84~$\pm$~3.95 \\
    Luminance &  & 18.41~$\pm$~1.45 & 31.65~$\pm$~0.84 & 53.84~$\pm$~0.65 \\
    Color-opponency &  & 17.18~$\pm$~0.97 & 30.47~$\pm$~0.73 & 49.65~$\pm$~4.33 \\
    Single-color &  & 16.47~$\pm$~0.89 & 29.15~$\pm$~2.23 & 46.33~$\pm$~5.58 \\
    \midrule
    - & \multirow{4}{*}{60\%} & 12.87~$\pm$~2.79 & 20.23~$\pm$~4.37 & 48.49~$\pm$~4.16 \\
    Luminance &  & 17.66~$\pm$~0.19 & 28.73~$\pm$~0.66 & 54.70~$\pm$~0.77 \\
    Color-opponency &  & 17.42~$\pm$~1.41 & 29.71~$\pm$~1.94 & 49.31~$\pm$~4.95 \\
    Single-color &  & 16.20~$\pm$~0.76 & 28.99~$\pm$~1.20 & 49.84~$\pm$~2.38 \\
    \midrule
    - & \multirow{4}{*}{70\%} & 11.12~$\pm$~2.68 & 18.94~$\pm$~4.07 & 47.82~$\pm$~5.20 \\
    Luminance &  & 16.83~$\pm$~1.87 & 27.68~$\pm$~1.57 & 52.38~$\pm$~1.66 \\
    Color-opponency &  & 17.84~$\pm$~2.02 & 31.46~$\pm$~0.88 & 54.26~$\pm$~2.32 \\
    Single-color &  & 14.74~$\pm$~0.76 & 26.73~$\pm$~1.84 & 44.97~$\pm$~0.28 \\
    \midrule
    - & \multirow{4}{*}{80\%} & 9.65~$\pm$~1.97 & 18.60~$\pm$~1.54 & 42.93~$\pm$~3.83 \\
    Luminance &  & 15.18~$\pm$~1.25 & 26.69~$\pm$~2.62 & 47.55~$\pm$~3.37 \\
    Color-opponency &  & 16.71~$\pm$~0.44 & 28.19~$\pm$~2.11 & 50.47~$\pm$~2.57 \\
    Single-color &  & 12.24~$\pm$~1.40 & 22.90~$\pm$~3.00 & 33.62~$\pm$~10.31 \\
    \midrule
    - & \multirow{4}{*}{90\%} & 9.39~$\pm$~0.33 & 17.40~$\pm$~1.22 & 45.29~$\pm$~1.43 \\
    Luminance &  & 11.54~$\pm$~0.67 & 20.59~$\pm$~1.84 & 38.06~$\pm$~4.23 \\
    Color-opponency &  & 15.01~$\pm$~2.05 & 26.59~$\pm$~1.51 & 47.53~$\pm$~3.17 \\
    Single-color &  & 12.20~$\pm$~1.10 & 23.65~$\pm$~1.55 & 39.98~$\pm$~6.99 \\
    \bottomrule
  \end{tabular}
  \label{tab:sparse_results_upernet_dark_zurich_learned}
\end{table}
\clearpage

\begin{table}[tb]
  \centering
  \caption{Evaluation of the \textbf{UPerNet} architecture trained on Cityscapes and \textbf{validated on ACDC Night} at different sparsities. Results include the three preprocessing types (luminance, color-opponency, and single-color). Metrics reported are mean Intersection over Union (mIoU), mean Accuracy (mAcc), and average Accuracy (aAcc), averaged over three seeds. \textbf{Percent-based sparsity:} For a given \textit{percentage}, the \textit{percentage} of lowest absolute values are set to zero. }   
  \begin{tabular}{@{}lcrrr@{}}
    \toprule
    Preprocessing  & Sparsity &\multicolumn{1}{c}{mIoU}&\multicolumn{1}{c}{mAcc}&\multicolumn{1}{c}{aAcc}  \\
    \midrule
    - & \multirow{4}{*}{-} & 11.53~$\pm$~1.24 & 21.96~$\pm$~2.20 & 45.24~$\pm$~7.17 \\
     Luminance &   & 18.31~$\pm$~1.15 & 31.05~$\pm$~0.94 & 54.27~$\pm$~1.21 \\
    Color-opponency&   &  16.37~$\pm$~0.81 & 28.01~$\pm$~1.57 & 50.83~$\pm$~3.65 \\
    Single-color &   &  14.64~$\pm$~1.36 & 28.31~$\pm$~1.02 & 49.00~$\pm$~4.06 \\
    \midrule
    - & \multirow{4}{*}{40\%} & 11.05~$\pm$~1.22 & 16.39~$\pm$~2.70 & 45.81~$\pm$~6.39 \\
    Luminance &  & 17.82~$\pm$~0.62 & 28.65~$\pm$~2.76 & 55.15~$\pm$~0.84 \\
    Color-opponency &  & 16.16~$\pm$~1.52 & 28.69~$\pm$~1.63 & 52.35~$\pm$~1.64 \\
    Single-color &  & 16.51~$\pm$~1.71 & 28.97~$\pm$~1.69 & 51.97~$\pm$~8.24 \\
    \midrule
    - & \multirow{4}{*}{50\%} & 11.40~$\pm$~1.92 & 17.44~$\pm$~2.63 & 47.32~$\pm$~3.33 \\
    Luminance &  & 18.69~$\pm$~0.87 & 30.40~$\pm$~0.56 & 57.27~$\pm$~0.46 \\
    Color-opponency &  & 17.40~$\pm$~1.07 & 28.69~$\pm$~0.46 & 53.96~$\pm$~2.80 \\
    Single-color &  & 16.84~$\pm$~1.03 & 29.43~$\pm$~2.17 & 50.51~$\pm$~5.45 \\
    \midrule
    - & \multirow{4}{*}{60\%} & 12.00~$\pm$~3.01 & 18.80~$\pm$~5.03 & 48.04~$\pm$~7.58 \\
    Luminance &  & 18.28~$\pm$~0.42 & 29.25~$\pm$~0.67 & 58.24~$\pm$~0.78 \\
    Color-opponency &  & 17.74~$\pm$~0.77 & 30.61~$\pm$~0.79 & 53.67~$\pm$~3.00 \\
    Single-color &  & 16.55~$\pm$~0.84 & 28.55~$\pm$~1.16 & 52.81~$\pm$~2.36 \\
    \midrule
    - & \multirow{4}{*}{70\%} & 10.67~$\pm$~2.90 & 17.65~$\pm$~4.26 & 46.94~$\pm$~7.53 \\
    Luminance &  & 18.44~$\pm$~1.38 & 28.44~$\pm$~1.13 & 56.24~$\pm$~0.96 \\
    Color-opponency &  & 18.58~$\pm$~1.48 & 31.41~$\pm$~1.12 & 57.70~$\pm$~1.77 \\
    Single-color &  & 15.08~$\pm$~1.07 & 25.69~$\pm$~2.28 & 48.09~$\pm$~2.05 \\
    \midrule
    - & \multirow{4}{*}{80\%} & 9.45~$\pm$~1.97 & 17.29~$\pm$~1.73 & 41.78~$\pm$~2.29 \\
    Luminance &  & 15.89~$\pm$~1.61 & 25.73~$\pm$~2.78 & 50.38~$\pm$~4.55 \\
    Color-opponency &  & 17.84~$\pm$~0.76 & 28.57~$\pm$~1.56 & 54.88~$\pm$~2.32 \\
    Single-color &  & 12.51~$\pm$~1.74 & 21.63~$\pm$~2.53 & 33.80~$\pm$~8.30 \\
    \midrule
    - & \multirow{4}{*}{90\%} & 9.40~$\pm$~0.12 & 16.54~$\pm$~0.96 & 49.37~$\pm$~1.08 \\
    Luminance &  & 12.65~$\pm$~1.03 & 21.12~$\pm$~1.54 & 40.99~$\pm$~3.42 \\
    Color-opponency &  & 16.23~$\pm$~1.22 & 25.63~$\pm$~1.14 & 52.37~$\pm$~2.29 \\
    Single-color &  & 12.53~$\pm$~0.98 & 22.34~$\pm$~1.58 & 42.31~$\pm$~8.28 \\
    \bottomrule
  \end{tabular}
  \label{tab:sparse_results_upernet_acdc_night_learned}
\end{table}
\clearpage

\begin{table}[tb]
  \centering
  \caption{Evaluation of the \textbf{UPerNet} architecture trained on Cityscapes and \textbf{validated on ACDC Fog} at different sparsities. Results include the three preprocessing types (luminance, color-opponency, and single-color). Metrics reported are mean Intersection over Union (mIoU), mean Accuracy (mAcc), and average Accuracy (aAcc), averaged over three seeds. \textbf{Percent-based sparsity:} For a given \textit{percentage}, the \textit{percentage} of lowest absolute values are set to zero.}   
  \begin{tabular}{@{}lcrrr@{}}
    \toprule
    Preprocessing  & Sparsity &\multicolumn{1}{c}{mIoU}&\multicolumn{1}{c}{mAcc}&\multicolumn{1}{c}{aAcc}  \\
    \midrule
     - & \multirow{3}{*}{-} & 45.44~$\pm$~1.69 & 55.56~$\pm$~3.78 & 86.25~$\pm$~2.41 \\
    Luminance &   & 46.52~$\pm$~0.66 & 55.61~$\pm$~1.40 & 82.09~$\pm$~3.30 \\
    Color-opponency&   &  46.10~$\pm$~0.60 & 55.97~$\pm$~0.97 & 84.47~$\pm$~2.68 \\
    Single-color &   &  45.70~$\pm$~2.04 & 55.52~$\pm$~1.92 & 82.15~$\pm$~8.56 \\
    \midrule
    - & \multirow{4}{*}{40\%} & 27.04~$\pm$~1.09 & 35.50~$\pm$~3.06 & 57.00~$\pm$~4.29 \\
    Luminance &  & 45.90~$\pm$~1.61 & 53.27~$\pm$~3.75 & 87.47~$\pm$~1.27 \\
    Color-opponency &  & 48.29~$\pm$~1.41 & 57.73~$\pm$~1.37 & 87.37~$\pm$~2.36 \\
    Single-color &  & 45.29~$\pm$~1.18 & 55.16~$\pm$~0.81 & 84.22~$\pm$~2.55 \\
    \midrule
    - & \multirow{4}{*}{50\%} & 28.62~$\pm$~2.76 & 36.59~$\pm$~1.62 & 66.55~$\pm$~1.18 \\
    Luminance &  & 41.85~$\pm$~0.63 & 49.43~$\pm$~0.69 & 79.81~$\pm$~5.05 \\
    Color-opponency &  & 46.77~$\pm$~1.67 & 56.42~$\pm$~1.41 & 82.89~$\pm$~3.53 \\
    Single-color &  & 44.27~$\pm$~1.23 & 55.69~$\pm$~2.10 & 81.80~$\pm$~4.19 \\
    \midrule
    - & \multirow{4}{*}{60\%} & 26.85~$\pm$~5.17 & 33.48~$\pm$~5.88 & 70.27~$\pm$~6.04 \\
    Luminance &  & 43.36~$\pm$~0.91 & 51.45~$\pm$~0.53 & 81.56~$\pm$~4.56 \\
    Color-opponency &  & 46.22~$\pm$~2.05 & 55.32~$\pm$~1.75 & 83.04~$\pm$~3.84 \\
    Single-color &  & 44.92~$\pm$~1.56 & 54.73~$\pm$~2.24 & 82.89~$\pm$~1.22 \\
    \midrule
    - & \multirow{4}{*}{70\%} & 14.36~$\pm$~6.43 & 22.03~$\pm$~6.59 & 51.04~$\pm$~15.71 \\
    Luminance &  & 42.07~$\pm$~3.14 & 49.56~$\pm$~4.36 & 80.78~$\pm$~4.65 \\
    Color-opponency &  & 45.65~$\pm$~3.84 & 54.74~$\pm$~3.39 & 83.38~$\pm$~4.42 \\
    Single-color &  & 42.35~$\pm$~1.16 & 50.06~$\pm$~1.05 & 81.25~$\pm$~4.51 \\
    \midrule
    - & \multirow{4}{*}{80\%} & 13.88~$\pm$~7.01 & 21.21~$\pm$~7.72 & 49.82~$\pm$~15.01 \\
    Luminance &  & 40.94~$\pm$~1.29 & 47.57~$\pm$~2.28 & 84.82~$\pm$~1.28 \\
    Color-opponency &  & 43.47~$\pm$~2.35 & 52.05~$\pm$~2.06 & 79.62~$\pm$~8.05 \\
    Single-color &  & 36.92~$\pm$~2.08 & 45.42~$\pm$~0.53 & 74.63~$\pm$~6.83 \\
    \midrule
    - & \multirow{4}{*}{90\%} & 15.92~$\pm$~1.73 & 24.72~$\pm$~0.90 & 61.17~$\pm$~6.09 \\
    Luminance &  & 32.84~$\pm$~0.40 & 40.14~$\pm$~1.07 & 76.69~$\pm$~3.83 \\
    Color-opponency &  & 44.45~$\pm$~2.71 & 52.80~$\pm$~3.52 & 82.98~$\pm$~0.19 \\
    Single-color &  & 26.18~$\pm$~2.19 & 33.97~$\pm$~1.82 & 68.86~$\pm$~3.67 \\
    \bottomrule
  \end{tabular}
  \label{tab:sparse_results_upernet_acdc_fog_learned}
\end{table}
\clearpage

\begin{table}[tb]
  \centering
  \caption{Evaluation of the \textbf{UPerNet} architecture trained on Cityscapes and \textbf{validated on ACDC Rain} at different sparsities. Results include the three preprocessing types (luminance, color-opponency, and single-color). Metrics reported are mean Intersection over Union (mIoU), mean Accuracy (mAcc), and average Accuracy (aAcc), averaged over three seeds. \textbf{Percent-based sparsity:} For a given \textit{percentage}, the \textit{percentage} of lowest absolute values are set to zero.}   
  \begin{tabular}{@{}lcrrr@{}}
    \toprule
    Preprocessing  & Sparsity &\multicolumn{1}{c}{mIoU}&\multicolumn{1}{c}{mAcc}&\multicolumn{1}{c}{aAcc}  \\
    \midrule
    - & \multirow{4}{*}{-} & 35.35~$\pm$~2.12 & 48.51~$\pm$~2.46 & 81.44~$\pm$~4.98 \\
     Luminance &   & 29.31~$\pm$~0.53 & 40.70~$\pm$~2.36 & 69.42~$\pm$~5.65 \\
    Color-opponency&   &  34.28~$\pm$~1.23 & 45.48~$\pm$~1.77 & 76.43~$\pm$~5.73 \\
    Single-color &   &   33.35~$\pm$~1.69 & 43.90~$\pm$~2.94 & 73.61~$\pm$~6.08 \\
    \midrule
    - & \multirow{4}{*}{40\%} & 15.27~$\pm$~0.84 & 22.42~$\pm$~2.54 & 45.28~$\pm$~2.60 \\
    Luminance &  & 32.19~$\pm$~1.31 & 40.87~$\pm$~2.74 & 80.24~$\pm$~1.87 \\
    Color-opponency &  & 35.98~$\pm$~0.50 & 46.98~$\pm$~1.03 & 81.20~$\pm$~2.18 \\
    Single-color &  & 34.77~$\pm$~0.99 & 45.03~$\pm$~1.84 & 79.59~$\pm$~2.34 \\
    \midrule
    - & \multirow{4}{*}{50\%} & 18.72~$\pm$~1.03 & 25.69~$\pm$~1.30 & 60.83~$\pm$~3.05 \\
    Luminance &  & 30.74~$\pm$~0.55 & 39.64~$\pm$~0.65 & 75.50~$\pm$~3.60 \\
    Color-opponency &  & 33.28~$\pm$~0.57 & 45.36~$\pm$~1.80 & 72.85~$\pm$~2.16 \\
    Single-color &  & 35.58~$\pm$~1.04 & 47.21~$\pm$~1.46 & 81.34~$\pm$~0.56 \\
    \midrule
    - & \multirow{4}{*}{60\%} & 20.08~$\pm$~1.91 & 26.46~$\pm$~2.77 & 65.28~$\pm$~1.74 \\
    Luminance &  & 31.45~$\pm$~0.67 & 40.19~$\pm$~1.60 & 78.10~$\pm$~3.85 \\
    Color-opponency &  & 33.13~$\pm$~2.48 & 43.80~$\pm$~2.08 & 78.27~$\pm$~5.70 \\
    Single-color &  & 35.34~$\pm$~1.64 & 44.60~$\pm$~1.51 & 81.55~$\pm$~2.66 \\
    \midrule
    - & \multirow{4}{*}{70\%} & 13.91~$\pm$~5.23 & 19.89~$\pm$~6.45 & 51.91~$\pm$~17.81 \\
    Luminance &  & 30.99~$\pm$~2.91 & 38.28~$\pm$~4.15 & 78.02~$\pm$~2.54 \\
    Color-opponency &  & 32.82~$\pm$~1.47 & 42.89~$\pm$~1.39 & 76.19~$\pm$~3.50 \\
    Single-color &  & 33.69~$\pm$~1.22 & 41.47~$\pm$~1.78 & 82.07~$\pm$~2.90 \\
    \midrule
    - & \multirow{4}{*}{80\%} & 14.86~$\pm$~3.84 & 22.42~$\pm$~2.49 & 58.06~$\pm$~9.43 \\
    Luminance &  & 30.26~$\pm$~1.60 & 36.81~$\pm$~2.63 & 81.88~$\pm$~2.14 \\
    Color-opponency &  & 31.61~$\pm$~2.16 & 40.70~$\pm$~1.83 & 76.77~$\pm$~6.40 \\
    Single-color &  & 30.43~$\pm$~1.11 & 37.56~$\pm$~0.55 & 77.78~$\pm$~4.12 \\
    \midrule
    - & \multirow{4}{*}{90\%} & 16.36~$\pm$~0.89 & 23.53~$\pm$~1.83 & 66.59~$\pm$~3.07 \\
    Luminance &  & 26.99~$\pm$~0.79 & 33.21~$\pm$~0.84 & 78.19~$\pm$~2.00 \\
    Color-opponency &  & 32.63~$\pm$~1.36 & 40.50~$\pm$~2.01 & 80.90~$\pm$~1.04 \\
    Single-color &  & 23.46~$\pm$~1.39 & 29.43~$\pm$~1.93 & 74.28~$\pm$~2.73 \\
    \bottomrule
  \end{tabular}
  \label{tab:sparse_results_upernet_acdc_rain_learned}
\end{table}
\clearpage

\begin{table}[tb]
  \centering
  \caption{Evaluation of the \textbf{UPerNet} architecture trained on Cityscapes and \textbf{validated on ACDC Snow} at different sparsities. Results include the three preprocessing types (luminance, color-opponency, and single-color). Metrics reported are mean Intersection over Union (mIoU), mean Accuracy (mAcc), and average Accuracy (aAcc), averaged over three seeds. \textbf{Percent-based sparsity:} For a given \textit{percentage}, the \textit{percentage} of lowest absolute values are set to zero.}   
  \begin{tabular}{@{}lcrrr@{}}
    \toprule
    Preprocessing  & Sparsity &\multicolumn{1}{c}{mIoU}&\multicolumn{1}{c}{mAcc}&\multicolumn{1}{c}{aAcc}  \\
    \midrule
    - & \multirow{4}{*}{-}  & 26.90~$\pm$~1.92 & 40.71~$\pm$~2.42 & 68.91~$\pm$~3.87 \\
     Luminance &  & 26.63~$\pm$~0.26 & 38.34~$\pm$~1.56 & 62.29~$\pm$~2.11 \\
    Color-opponency&   &  26.57~$\pm$~1.75 & 37.94~$\pm$~1.75 & 68.68~$\pm$~1.71 \\
    Single-color &   &  27.84~$\pm$~2.10 & 37.96~$\pm$~2.27 & 67.74~$\pm$~7.07 \\
    \midrule
    - & \multirow{4}{*}{40\%} & 19.83~$\pm$~1.14 & 28.18~$\pm$~1.58 & 49.93~$\pm$~5.16 \\
    Luminance &  & 27.93~$\pm$~1.38 & 37.46~$\pm$~2.77 & 72.73~$\pm$~2.13 \\
    Color-opponency &  & 28.34~$\pm$~2.95 & 39.13~$\pm$~3.36 & 71.96~$\pm$~4.61 \\
    Single-color &  & 27.17~$\pm$~1.20 & 37.11~$\pm$~1.49 & 70.81~$\pm$~2.70 \\
    \midrule
    - & \multirow{4}{*}{50\%} & 21.07~$\pm$~1.52 & 29.51~$\pm$~0.29 & 56.48~$\pm$~2.68 \\
    Luminance &  & 26.69~$\pm$~1.79 & 35.25~$\pm$~1.78 & 65.63~$\pm$~5.43 \\
    Color-opponency &  & 28.85~$\pm$~2.98 & 40.47~$\pm$~1.83 & 66.80~$\pm$~4.11 \\
    Single-color &  & 28.08~$\pm$~0.63 & 39.01~$\pm$~1.38 & 71.33~$\pm$~3.56 \\
    \midrule
    - & \multirow{4}{*}{60\%} & 20.40~$\pm$~1.55 & 28.35~$\pm$~2.47 & 59.59~$\pm$~3.24 \\
    Luminance &  & 26.44~$\pm$~0.47 & 35.82~$\pm$~0.94 & 67.14~$\pm$~3.91 \\
    Color-opponency &  & 28.23~$\pm$~1.54 & 39.55~$\pm$~1.55 & 70.19~$\pm$~6.39 \\
    Single-color &  & 28.64~$\pm$~0.32 & 38.33~$\pm$~0.35 & 72.86~$\pm$~1.10 \\
    \midrule
    - & \multirow{4}{*}{70\%} & 17.58~$\pm$~4.84 & 25.98~$\pm$~6.92 & 51.58~$\pm$~13.35 \\
    Luminance &  & 21.90~$\pm$~2.24 & 29.42~$\pm$~3.98 & 61.26~$\pm$~4.73 \\
    Color-opponency &  & 27.99~$\pm$~2.83 & 38.42~$\pm$~2.35 & 68.47~$\pm$~6.07 \\
    Single-color &  & 27.21~$\pm$~1.24 & 35.07~$\pm$~1.23 & 72.49~$\pm$~6.13 \\
    \midrule
    - & \multirow{4}{*}{80\%} & 17.46~$\pm$~6.38 & 27.13~$\pm$~5.39 & 53.92~$\pm$~14.20 \\
    Luminance &  & 24.09~$\pm$~3.29 & 30.40~$\pm$~4.13 & 70.88~$\pm$~4.09 \\
    Color-opponency &  & 27.48~$\pm$~2.43 & 37.76~$\pm$~1.77 & 71.02~$\pm$~6.47 \\
    Single-color &  & 23.63~$\pm$~0.54 & 30.23~$\pm$~0.77 & 69.08~$\pm$~3.31 \\
    \midrule
    - & \multirow{4}{*}{90\%} & 17.49~$\pm$~2.12 & 27.53~$\pm$~1.70 & 59.57~$\pm$~6.83 \\
    Luminance &  & 23.93~$\pm$~1.08 & 30.71~$\pm$~1.52 & 69.38~$\pm$~0.99 \\
    Color-opponency &  & 26.63~$\pm$~2.06 & 35.94~$\pm$~2.88 & 70.43~$\pm$~1.62 \\
    Single-color &  & 18.21~$\pm$~2.27 & 24.19~$\pm$~2.60 & 63.61~$\pm$~4.87 \\
    \bottomrule
  \end{tabular}
  \label{tab:sparse_results_upernet_acdc_snow_learned}
\end{table}
\clearpage

\begin{table}[tb]
  \centering
  \caption{Evaluation of the \textbf{UPerNet} architecture trained on Cityscapes and \textbf{validated on ACDC Mean} at different sparsities. Results include the three preprocessing types (luminance, color-opponency, and single-color). Metrics reported are mean Intersection over Union (mIoU), mean Accuracy (mAcc), and average Accuracy (aAcc), averaged over three seeds. \textbf{Percent-based sparsity:} For a given \textit{percentage}, the \textit{percentage} of lowest absolute values are set to zero.}   
  \begin{tabular}{@{}lcrrr@{}}
    \toprule
    Preprocessing  & Sparsity &\multicolumn{1}{c}{mIoU}&\multicolumn{1}{c}{mAcc}&\multicolumn{1}{c}{aAcc}  \\
    \midrule
    - & \multirow{4}{*}{-} & 25.65~$\pm$~0.91 & 37.63~$\pm$~2.33 & 66.15~$\pm$~3.03\\
    Luminance &  & 28.34~$\pm$~0.81 & 37.54~$\pm$~2.80 & 67.73~$\pm$~4.73 \\
    Color-opponency &  & 29.15~$\pm$~1.62 & 38.43~$\pm$~1.61 & 68.14~$\pm$~2.79 \\
    Single-color &  & 28.22~$\pm$~0.89 & 38.73~$\pm$~2.45 & 67.17~$\pm$~3.82 \\
    \midrule
    - & \multirow{4}{*}{40\%} & 21.89~$\pm$~1.40 & 34.43~$\pm$~1.00 & 58.03~$\pm$~3.66 \\
    Luminance &  & 30.45~$\pm$~0.79 & 38.69~$\pm$~2.85 & 73.75~$\pm$~1.39 \\
    Color-opponency &  & 31.28~$\pm$~1.43 & 41.44~$\pm$~1.58 & 73.06~$\pm$~2.59 \\
    Single-color &  & 30.45~$\pm$~0.90 & 39.88~$\pm$~0.63 & 71.49~$\pm$~2.40 \\
    \midrule
    - & \multirow{4}{*}{50\%} & 19.02~$\pm$~0.77 & 34.07~$\pm$~1.96 & 59.18~$\pm$~2.80 \\
    Luminance &  & 29.16~$\pm$~0.90 & 37.27~$\pm$~0.79 & 69.46~$\pm$~3.38 \\
    Color-opponency &  & 30.80~$\pm$~2.11 & 40.82~$\pm$~0.80 & 69.01~$\pm$~2.79 \\
    Single-color &  & 30.87~$\pm$~0.52 & 40.67~$\pm$~1.52 & 71.09~$\pm$~2.88 \\
    \midrule
    - & \multirow{4}{*}{60\%} & 18.94~$\pm$~1.27 & 31.55~$\pm$~0.38 & 60.68~$\pm$~1.34 \\
    Luminance &  & 29.53~$\pm$~0.18 & 37.58~$\pm$~0.45 & 71.16~$\pm$~3.20 \\
    Color-opponency &  & 30.36~$\pm$~1.70 & 40.29~$\pm$~1.26 & 71.16~$\pm$~4.68 \\
    Single-color &  & 30.70~$\pm$~0.43 & 39.72~$\pm$~0.73 & 72.37~$\pm$~0.75 \\
    \midrule
    - & \multirow{4}{*}{70\%} & 15.76~$\pm$~0.85 & 26.97~$\pm$~1.27 & 57.04~$\pm$~3.51 \\
    Luminance &  & 28.39~$\pm$~1.89 & 35.37~$\pm$~2.50 & 68.98~$\pm$~2.67 \\
    Color-opponency &  & 30.52~$\pm$~2.11 & 39.92~$\pm$~1.32 & 71.33~$\pm$~3.27 \\
    Single-color &  & 29.23~$\pm$~0.73 & 36.53~$\pm$~1.61 & 70.80~$\pm$~3.08 \\
    \midrule
    - & \multirow{4}{*}{80\%} & 13.37~$\pm$~0.38 & 22.31~$\pm$~1.27 & 54.26~$\pm$~0.17 \\
    Luminance &  & 27.58~$\pm$~2.30 & 34.65~$\pm$~2.57 & 71.82~$\pm$~1.78 \\
    Color-opponency &  & 29.47~$\pm$~1.97 & 38.28~$\pm$~1.85 & 70.45~$\pm$~5.77 \\
    Single-color &  & 25.61~$\pm$~1.17 & 32.70~$\pm$~0.80 & 63.59~$\pm$~5.54 \\
    \midrule
    - & \multirow{4}{*}{90\%} & 9.88~$\pm$~0.32 & 15.83~$\pm$~0.65 & 46.79~$\pm$~3.25 \\
    Luminance &  & 24.17~$\pm$~0.08 & 31.06~$\pm$~0.13 & 66.11~$\pm$~0.66 \\
    Color-opponency &  & 29.84~$\pm$~1.75 & 37.66~$\pm$~2.17 & 71.52~$\pm$~0.60 \\
    Single-color &  & 20.24~$\pm$~1.55 & 27.26~$\pm$~1.80 & 62.11~$\pm$~4.65 \\
    \bottomrule
  \end{tabular}
  \label{tab:sparse_results_upernet_acdc_full_learned}
\end{table}
\clearpage

%% file: tables/sparsity_threshold_deeplab.tex
% percentage results table!!!
\begin{table}[h]
  \centering
  \caption{Evaluation of the \textbf{DeepLabv3+} architecture trained on Cityscapes and \textbf{validated on Cityscapes} at different sparsities. Results include the three preprocessing types (luminance, color-opponency, and single-color). Metrics reported are mean Intersection over Union (mIoU), mean Accuracy (mAcc), and average Accuracy (aAcc), averaged over three seeds. \textbf{Threshold-based sparsity:} For a given \textit{threshold}, all values within the range $[-\textit{threshold}, +\textit{threshold}]$ are set to zero.}   
  \begin{tabular}{@{}lc rrr@{}}
    \toprule
    Preprocessing &  Sparsity &\multicolumn{1}{c}{mIoU}&\multicolumn{1}{c}{mAcc}&\multicolumn{1}{c}{aAcc}  \\
    \midrule
    - & - & 66.68~$\pm$~0.44 & 76.58~$\pm$~0.60 & 94.68~$\pm$~0.05 \\
    \midrule
    \midrule
    Luminance   & \multirow{3}{*}{-} & 66.53~$\pm$~1.11 & 77.86~$\pm$~0.63 & 94.35~$\pm$~0.35 \\
     Color-opponency& & 66.64~$\pm$~1.39 & 75.80~$\pm$~1.46 & 94.88~$\pm$~0.15 \\
     Single-color& & 66.08~$\pm$~3.05 & 76.33~$\pm$~3.03 & 94.90~$\pm$~0.16 \\
    \midrule
    Luminance  & \multirow{3}{*}{0.005} & 66.27~$\pm$~1.06 & 76.50~$\pm$~1.54 & 94.52~$\pm$~0.26 \\
    Color-opponency& & 67.37~$\pm$~1.35 & 77.08~$\pm$~1.14 & 94.87~$\pm$~0.18 \\
    Single-color & & 68.15~$\pm$~2.54 & 78.26~$\pm$~1.97 & 94.99~$\pm$~0.32 \\
    \midrule
    Luminance  & \multirow{3}{*}{0.0075} & 65.29~$\pm$~2.59 & 76.77~$\pm$~3.70 & 94.31~$\pm$~0.48 \\
    Color-opponency& & 65.04~$\pm$~1.31 & 75.32~$\pm$~1.34 & 94.70~$\pm$~0.24 \\
    Single-color & & 67.93~$\pm$~1.65 & 78.33~$\pm$~1.60 & 95.00~$\pm$~0.18 \\
    \midrule
    Luminance  & \multirow{3}{*}{0.01} & 63.11~$\pm$~1.53 & 73.82~$\pm$~2.06 & 94.28~$\pm$~0.19 \\
    Color-opponency& & 63.69~$\pm$~2.66 & 73.98~$\pm$~3.63 & 94.28~$\pm$~0.17 \\
    Single-color & & 67.51~$\pm$~0.85 & 78.08~$\pm$~0.50 & 95.00~$\pm$~0.09 \\
    \midrule
    Luminance  & \multirow{3}{*}{0.02} & 63.68~$\pm$~1.13 & 74.67~$\pm$~0.93 & 93.60~$\pm$~0.32 \\
    Color-opponency& & 64.37~$\pm$~1.03 & 74.71~$\pm$~1.64 & 94.26~$\pm$~0.18 \\
    Single-color & & 64.20~$\pm$~0.96 & 73.87~$\pm$~1.86 & 94.69~$\pm$~0.12 \\
    \bottomrule
  \end{tabular}
  \label{tab:sparse_results_deeplabv3+_cityscapes_threshold}
\end{table}
\clearpage

\begin{table}[tb]
  \centering
  \caption{Evaluation of the \textbf{DeepLabv3+} architecture trained on Cityscapes and \textbf{validated on Dark Zurich} at different sparsities. Results include the three preprocessing types (luminance, color-opponency, and single-color). Metrics reported are mean Intersection over Union (mIoU), mean Accuracy (mAcc), and average Accuracy (aAcc), averaged over three seeds. \textbf{Threshold-based sparsity:} For a given \textit{threshold}, all values within the range $[-\textit{threshold}, +\textit{threshold}]$ are set to zero. }   
  \begin{tabular}{@{}lc rrr@{}}
    \toprule
    Preprocessing &  Sparsity &\multicolumn{1}{c}{mIoU}&\multicolumn{1}{c}{mAcc}&\multicolumn{1}{c}{aAcc}  \\
    \midrule
    - & - & 7.64~$\pm$~1.00 & 16.40~$\pm$~2.74 & 30.13~$\pm$~1.41 \\
    \midrule
    \midrule
    Luminance  &  \multirow{3}{*}{-}& 18.39~$\pm$~0.54 & 34.73~$\pm$~1.68 & 55.35~$\pm$~1.81 \\
    Color-opponency& & 17.64~$\pm$~1.70 & 32.32~$\pm$~2.85 & 50.68~$\pm$~4.61 \\
    Single-color & & 18.54~$\pm$~2.18 & 32.79~$\pm$~1.49 & 51.35~$\pm$~4.26 \\
    \midrule
    Luminance  & \multirow{3}{*}{0.005} & 19.88~$\pm$~0.61 & 33.61~$\pm$~0.46 & 54.70~$\pm$~2.52 \\
    Color-opponency& & 17.80~$\pm$~0.95 & 31.30~$\pm$~1.03 & 52.15~$\pm$~3.77 \\
    Single-color & & 18.66~$\pm$~0.58 & 32.37~$\pm$~0.75 & 52.17~$\pm$~2.58 \\
    \midrule
    Luminance  & \multirow{3}{*}{0.0075} & 21.17~$\pm$~0.29 & 36.03~$\pm$~2.24 & 56.19~$\pm$~1.62 \\
    Color-opponency& & 19.54~$\pm$~1.12 & 32.92~$\pm$~2.68 & 54.89~$\pm$~1.50 \\
    Single-color & & 18.24~$\pm$~1.15 & 32.60~$\pm$~2.86 & 52.80~$\pm$~1.76 \\
    \midrule
    Luminance  & \multirow{3}{*}{0.01} & 22.28~$\pm$~1.72 & 36.73~$\pm$~1.97 & 56.89~$\pm$~0.75 \\
    Color-opponency& & 19.75~$\pm$~0.77 & 33.28~$\pm$~1.00 & 55.23~$\pm$~2.86 \\
    Single-color & & 18.79~$\pm$~1.26 & 31.51~$\pm$~0.62 & 52.92~$\pm$~1.56 \\
    \midrule
    Luminance  & \multirow{3}{*}{0.02} & 20.08~$\pm$~1.60 & 35.35~$\pm$~1.51 & 53.19~$\pm$~5.72 \\
    Color-opponency& & 20.96~$\pm$~1.07 & 35.42~$\pm$~1.47 & 55.50~$\pm$~1.08 \\
    Single-color & & 19.28~$\pm$~1.58 & 32.46~$\pm$~2.11 & 53.14~$\pm$~1.84 \\

    \bottomrule
  \end{tabular}
  \label{tab:sparse_results_deeplabv3+_dark_zurich_threshold}
\end{table}

\begin{table}[tb]
  \centering
  \caption{Evaluation of the \textbf{DeepLabv3+} architecture trained on Cityscapes and \textbf{validated on ACDC Night} at different sparsities. Results include the three preprocessing types (luminance, color-opponency, and single-color). Metrics reported are mean Intersection over Union (mIoU), mean Accuracy (mAcc), and average Accuracy (aAcc), averaged over three seeds. \textbf{Threshold-based sparsity:} For a given \textit{threshold}, all values within the range $[-\textit{threshold}, +\textit{threshold}]$ are set to zero.  }   
  \begin{tabular}{@{}lc rrr@{}}
    \toprule
    Preprocessing &  Sparsity &\multicolumn{1}{c}{mIoU}&\multicolumn{1}{c}{mAcc}&\multicolumn{1}{c}{aAcc}  \\
    \midrule
    - & - & 8.01~$\pm$~0.45 & 18.14~$\pm$~2.24 & 31.94~$\pm$~2.20 \\
    \midrule
    \midrule
    Luminance & \multirow{3}{*}{-} & 19.76~$\pm$~0.53 & 34.13~$\pm$~1.05 & 58.90~$\pm$~1.33 \\
    Color-opponency& & 18.05~$\pm$~1.55 & 31.73~$\pm$~2.54 & 53.51~$\pm$~3.20 \\
    Single-color & & 19.75~$\pm$~1.99 & 32.24~$\pm$~1.17 & 54.79~$\pm$~2.95 \\
    \midrule
    Luminance  & \multirow{3}{*}{0.005} & 21.16~$\pm$~1.02 & 33.59~$\pm$~0.31 & 58.52~$\pm$~1.52 \\
    Color-opponency& & 18.92~$\pm$~0.97 & 30.95~$\pm$~1.61 & 55.82~$\pm$~3.22 \\
    Single-color & & 19.24~$\pm$~0.77 & 31.58~$\pm$~1.13 & 55.46~$\pm$~1.27 \\
    \midrule
    Luminance  & \multirow{3}{*}{0.0075} & 22.57~$\pm$~0.21 & 36.25~$\pm$~1.10 & 59.81~$\pm$~1.51 \\
    Color-opponency& & 20.63~$\pm$~1.03 & 32.62~$\pm$~1.82 & 58.47~$\pm$~1.80 \\
    Single-color & & 19.19~$\pm$~1.10 & 32.46~$\pm$~1.46 & 55.89~$\pm$~1.54 \\
    \midrule
    Luminance  & \multirow{3}{*}{0.01} & 22.39~$\pm$~1.41 & 35.16~$\pm$~1.73 & 60.13~$\pm$~0.93 \\
    Color-opponency& & 20.91~$\pm$~0.94 & 32.62~$\pm$~1.09 & 58.66~$\pm$~2.50 \\
    Single-color & & 19.05~$\pm$~0.71 & 30.87~$\pm$~0.55 & 56.07~$\pm$~0.95 \\
    \midrule
    Luminance  & \multirow{3}{*}{0.02} & 20.49~$\pm$~1.36 & 33.49~$\pm$~1.75 & 56.65~$\pm$~4.45 \\
    Color-opponency& & 22.36~$\pm$~1.19 & 33.97~$\pm$~1.34 & 58.44~$\pm$~1.03 \\
    Single-color & & 19.81~$\pm$~1.37 & 31.65~$\pm$~1.62 & 56.02~$\pm$~1.63 \\
    
    \bottomrule
  \end{tabular}
  \label{tab:sparse_results_deeplabv3+_acdc_night_threshold}
\end{table}

\begin{table}[tb]
  \centering
  \caption{Evaluation of the \textbf{DeepLabv3+} architecture trained on Cityscapes and \textbf{validated on ACDC Fog} at different sparsities. Results include the three preprocessing types (luminance, color-opponency, and single-color). Metrics reported are mean Intersection over Union (mIoU), mean Accuracy (mAcc), and average Accuracy (aAcc), averaged over three seeds. \textbf{Threshold-based sparsity:} For a given \textit{threshold}, all values within the range $[-\textit{threshold}, +\textit{threshold}]$ are set to zero.  }   
  \begin{tabular}{@{}lc rrr@{}}
    \toprule
    Preprocessing &  Sparsity &\multicolumn{1}{c}{mIoU}&\multicolumn{1}{c}{mAcc}&\multicolumn{1}{c}{aAcc}  \\
    \midrule
    - & - & 39.20~$\pm$~4.35 & 54.93~$\pm$~6.11 & 76.52~$\pm$~6.46 \\
    \midrule
    \midrule
    Luminance &  \multirow{3}{*}{-}& 48.90~$\pm$~3.67 & 62.73~$\pm$~3.23 & 84.55~$\pm$~3.74 \\
    Color-opponency& & 49.46~$\pm$~3.07 & 60.40~$\pm$~4.01 & 87.64~$\pm$~2.93 \\
    Single-color & & 50.66~$\pm$~3.38 & 62.22~$\pm$~2.75 & 83.45~$\pm$~1.46 \\ 
    \midrule
    Luminance  & \multirow{3}{*}{0.005} & 51.65~$\pm$~0.95 & 62.43~$\pm$~2.83 & 87.73~$\pm$~2.05 \\
    Color-opponency& & 48.79~$\pm$~1.32 & 60.34~$\pm$~0.34 & 83.25~$\pm$~5.84 \\
    Single-color & & 50.87~$\pm$~3.32 & 63.80~$\pm$~3.54 & 84.19~$\pm$~6.37 \\
    \midrule
    Luminance  & \multirow{3}{*}{0.0075} & 51.63~$\pm$~3.78 & 62.90~$\pm$~4.51 & 85.76~$\pm$~3.59 \\
    Color-opponency& & 45.73~$\pm$~2.79 & 57.39~$\pm$~1.07 & 81.38~$\pm$~7.73 \\
    Single-color & & 51.10~$\pm$~2.35 & 64.43~$\pm$~2.91 & 84.97~$\pm$~3.30 \\
    \midrule
    Luminance  & \multirow{3}{*}{0.01} & 50.59~$\pm$~1.11 & 61.00~$\pm$~1.13 & 88.94~$\pm$~2.13 \\
    Color-opponency& & 44.68~$\pm$~1.69 & 55.83~$\pm$~1.06 & 85.04~$\pm$~2.66 \\
    Single-color & & 47.67~$\pm$~3.06 & 60.09~$\pm$~0.42 & 76.90~$\pm$~3.08 \\
    \midrule
    Luminance  & \multirow{3}{*}{0.02} & 50.73~$\pm$~1.55 & 60.07~$\pm$~2.48 & 89.72~$\pm$~0.42 \\
    Color-opponency& & 46.02~$\pm$~2.37 & 55.61~$\pm$~2.90 & 86.76~$\pm$~1.83 \\
    Single-color & & 46.46~$\pm$~2.73 & 57.29~$\pm$~2.89 & 86.42~$\pm$~2.34 \\
  \bottomrule
  \end{tabular}
  \label{tab:sparse_results_deeplabv3+_acdc_fog_threshold}
\end{table}

\begin{table}[tb]
  \centering
  \caption{Evaluation of the \textbf{DeepLabv3+} architecture trained on Cityscapes and \textbf{validated on ACDC Rain} at different sparsities. Results include the three preprocessing types (luminance, color-opponency, and single-color). Metrics reported are mean Intersection over Union (mIoU), mean Accuracy (mAcc), and average Accuracy (aAcc), averaged over three seeds. \textbf{Threshold-based sparsity:} For a given \textit{threshold}, all values within the range $[-\textit{threshold}, +\textit{threshold}]$ are set to zero. }   
  \begin{tabular}{@{}lc rrr@{}}
    \toprule
    Preprocessing &  Sparsity &\multicolumn{1}{c}{mIoU}&\multicolumn{1}{c}{mAcc}&\multicolumn{1}{c}{aAcc}  \\
    \midrule
     - & - & 33.31~$\pm$~2.15 & 48.13~$\pm$~4.57 & 77.43~$\pm$~3.46 \\
    \midrule
    \midrule
    Luminance  &  \multirow{3}{*}{-} &34.20~$\pm$~2.50 & 47.68~$\pm$~4.67 & 76.72~$\pm$~3.07 \\
    Color-opponency& & 38.87~$\pm$~2.24 & 51.26~$\pm$~3.57 & 83.44~$\pm$~3.64 \\
    Single-color & & 38.60~$\pm$~3.04 & 51.81~$\pm$~2.65 & 77.82~$\pm$~1.63 \\
    \midrule
    Luminance  & \multirow{3}{*}{0.005} & 38.00~$\pm$~1.16 & 48.85~$\pm$~2.81 & 85.89~$\pm$~1.73 \\
    Color-opponency& & 37.53~$\pm$~1.10 & 49.74~$\pm$~0.29 & 82.10~$\pm$~4.63 \\
    Single-color & & 38.97~$\pm$~1.92 & 53.64~$\pm$~2.61 & 82.88~$\pm$~5.07 \\
    \midrule
    Luminance  & \multirow{3}{*}{0.0075} & 37.72~$\pm$~2.17 & 49.83~$\pm$~4.56 & 83.16~$\pm$~3.54 \\
    Color-opponency& & 35.23~$\pm$~2.50 & 47.00~$\pm$~1.18 & 80.37~$\pm$~6.10 \\
    Single-color & & 39.92~$\pm$~2.48 & 53.51~$\pm$~2.34 & 85.03~$\pm$~3.45 \\
    \midrule
    Luminance  & \multirow{3}{*}{0.01} & 37.99~$\pm$~1.54 & 50.94~$\pm$~4.15 & 86.40~$\pm$~1.24 \\
    Color-opponency& & 35.79~$\pm$~0.67 & 46.29~$\pm$~0.59 & 84.20~$\pm$~1.40 \\
    Single-color & & 39.69~$\pm$~1.76 & 54.51~$\pm$~0.87 & 79.75~$\pm$~3.18 \\
    \midrule
    Luminance  & \multirow{3}{*}{0.02} & 38.63~$\pm$~0.69 & 48.92~$\pm$~1.49 & 87.06~$\pm$~0.47 \\
    Color-opponency& & 38.58~$\pm$~1.53 & 48.35~$\pm$~2.01 & 86.35~$\pm$~1.16 \\
    Single-color & & 38.52~$\pm$~1.46 & 50.00~$\pm$~2.50 & 86.10~$\pm$~1.31 \\
  \bottomrule
  \end{tabular}
  \label{tab:sparse_results_deeplabv3+_acdc_rain_threshold}
\end{table}

\begin{table}[tb]
  \centering
  \caption{Evaluation of the \textbf{DeepLabv3+} architecture trained on Cityscapes and \textbf{validated on ACDC Snow} at different sparsities. Results include the three preprocessing types (luminance, color-opponency, and single-color). Metrics reported are mean Intersection over Union (mIoU), mean Accuracy (mAcc), and average Accuracy (aAcc), averaged over three seeds. \textbf{Threshold-based sparsity:} For a given \textit{threshold}, all values within the range $[-\textit{threshold}, +\textit{threshold}]$ are set to zero. }   
  \begin{tabular}{@{}lc rrr@{}}
    \toprule
    Preprocessing &  Sparsity &\multicolumn{1}{c}{mIoU}&\multicolumn{1}{c}{mAcc}&\multicolumn{1}{c}{aAcc}  \\
    \midrule
    - & - & 24.18~$\pm$~3.60 & 36.05~$\pm$~4.01 & 64.03~$\pm$~10.17 \\
    \midrule
    \midrule
    Luminance & \multirow{3}{*}{-}& 31.58~$\pm$~1.88 & 45.09~$\pm$~3.17 & 70.05~$\pm$~3.19 \\
    Color-opponency &  & 32.97~$\pm$~2.97 & 44.40~$\pm$~3.82 & 76.43~$\pm$~5.42 \\
    Single-color &  & 32.60~$\pm$~3.50 & 44.66~$\pm$~2.88 & 69.89~$\pm$~1.47 \\
    \midrule
    Luminance  & \multirow{3}{*}{0.005} & 34.30~$\pm$~0.82 & 45.97~$\pm$~1.85 & 77.55~$\pm$~4.78 \\
     Color-opponency& & 31.74~$\pm$~1.43 & 44.15~$\pm$~0.54 & 72.84~$\pm$~7.60 \\
     Single-color & & 32.22~$\pm$~3.20 & 46.32~$\pm$~2.50 & 74.30~$\pm$~8.40 \\
    \midrule
    Luminance  & \multirow{3}{*}{0.0075} & 34.54~$\pm$~3.28 & 47.48~$\pm$~3.81 & 76.77~$\pm$~6.98 \\
     Color-opponency& & 31.40~$\pm$~2.96 & 43.69~$\pm$~1.64 & 73.56~$\pm$~7.48 \\
     Single-color & & 33.57~$\pm$~2.42 & 46.43~$\pm$~3.89 & 77.83~$\pm$~3.10 \\
    \midrule
    Luminance  & \multirow{3}{*}{0.01} & 34.86~$\pm$~0.79 & 47.06~$\pm$~1.73 & 81.31~$\pm$~2.72 \\
     Color-opponency& & 31.14~$\pm$~1.71 & 42.93~$\pm$~0.80 & 76.98~$\pm$~3.86 \\
     Single-color & & 31.76~$\pm$~4.00 & 44.64~$\pm$~1.40 & 70.62~$\pm$~3.49 \\
    \midrule
    Luminance  & \multirow{3}{*}{0.02} & 33.04~$\pm$~2.08 & 44.92~$\pm$~2.47 & 78.93~$\pm$~2.40 \\
     Color-opponency& & 33.54~$\pm$~2.17 & 45.06~$\pm$~2.17 & 79.02~$\pm$~1.45 \\
     Single-color & & 33.04~$\pm$~2.08 & 44.92~$\pm$~2.47 & 78.93~$\pm$~2.40 \\

  \bottomrule
  \end{tabular}
  \label{tab:sparse_results_deeplabv3+_acdc_snow_threshold}
\end{table}

\begin{table}[tb]
  \centering
  \caption{Evaluation of the \textbf{DeepLabv3+} architecture trained on Cityscapes and \textbf{validated on ACDC Mean} at different sparsities. Results include the three preprocessing types (luminance, color-opponency, and single-color). Metrics reported are mean Intersection over Union (mIoU), mean Accuracy (mAcc), and average Accuracy (aAcc), averaged over three seeds. \textbf{Threshold-based sparsity:} For a given \textit{threshold}, all values within the range $[-\textit{threshold}, +\textit{threshold}]$ are set to zero. }   
  \begin{tabular}{@{}lc rrr@{}}
    \toprule
    Preprocessing &  Sparsity &\multicolumn{1}{c}{mIoU}&\multicolumn{1}{c}{mAcc}&\multicolumn{1}{c}{aAcc}  \\
    \midrule
    - & - & 23.59~$\pm$~1.66 & 39.06~$\pm$~4.04 & 62.24~$\pm$~4.48\\
    \midrule
    \midrule
    Luminance & \multirow{3}{*}{-} & 32.80~$\pm$~1.78 & 45.42~$\pm$~2.99 & 72.45~$\pm$~2.77 \\
    Color-opponency & & 33.74~$\pm$~1.99 & 45.16~$\pm$~3.12 & 75.08~$\pm$~2.96 \\
    Single-color & & 34.63~$\pm$~3.00 & 45.87~$\pm$~2.19 & 71.36~$\pm$~0.79 \\
    \midrule
    Luminance & \multirow{3}{*}{0.005} & 35.82~$\pm$~0.96 & 45.87~$\pm$~1.79 & 77.27~$\pm$~2.44 \\
    Color-opponency & & 33.60~$\pm$~1.38 & 44.12~$\pm$~0.84 & 73.37~$\pm$~4.99 \\
    Single-color & & 34.36~$\pm$~2.05 & 46.82~$\pm$~1.89 & 74.06~$\pm$~4.99 \\
    \midrule
    Luminance & \multirow{3}{*}{0.0075} & 36.12~$\pm$~2.37 & 47.35~$\pm$~3.54 & 76.25~$\pm$~3.87 \\
    Color-opponency & & 32.81~$\pm$~2.19 & 42.99~$\pm$~1.41 & 73.33~$\pm$~5.65 \\
    Single-color & & 35.10~$\pm$~2.17 & 47.32~$\pm$~2.56 & 75.77~$\pm$~2.67 \\
    \midrule
    Luminance & \multirow{3}{*}{0.01} & 35.87~$\pm$~1.04 & 46.92~$\pm$~2.14 & 79.04~$\pm$~1.28 \\
    Color-opponency & & 32.69~$\pm$~1.14 & 42.49~$\pm$~0.07 & 76.08~$\pm$~2.53 \\
    Single-color & & 33.82~$\pm$~2.44 & 45.52~$\pm$~0.94 & 70.72~$\pm$~2.56 \\
    \midrule
    Luminance & \multirow{3}{*}{0.02} & 35.52~$\pm$~0.16 & 45.38~$\pm$~0.56 & 78.61~$\pm$~1.50 \\
    Color-opponency & & 35.37~$\pm$~1.55 & 44.23~$\pm$~1.97 & 77.49~$\pm$~1.33 \\
    Single-color & & 33.92~$\pm$~1.60 & 44.08~$\pm$~2.60 & 76.71~$\pm$~1.72 \\

  \bottomrule
  \end{tabular}
  \label{tab:sparse_results_deeplabv3+_acdc_full_threshold}
\end{table}
\clearpage

%% file: tables/sparsity_threshold_mask2former.tex
\begin{table}[h]
  \centering
  \caption{Evaluation of the \textbf{Mask2Former} architecture trained on Cityscapes and \textbf{validated on Cityscapes} at different sparsities. Results include the three preprocessing types (luminance, color-opponency, and single-color). Metrics reported are mean Intersection over Union (mIoU), mean Accuracy (mAcc), and average Accuracy (aAcc), averaged over three seeds. \textbf{Threshold-based sparsity:} For a given \textit{threshold}, all values within the range $[-\textit{threshold}, +\textit{threshold}]$ are set to zero. }   
  \begin{tabular}{@{}lc rrr@{}}
    \toprule
    Preprocessing &  Sparsity &\multicolumn{1}{c}{mIoU}&\multicolumn{1}{c}{mAcc}&\multicolumn{1}{c}{aAcc}  \\
    \midrule
    - & - & 77.43~$\pm$~0.35 & 86.83~$\pm$~0.19 & 96.10~$\pm$~0.05 \\
    \midrule
    \midrule
    Luminance & \multirow{3}{*}{-} & 73.93~$\pm$~0.80 & 85.33~$\pm$~0.92 & 95.59~$\pm$~0.13 \\
    Color-opponency & & 73.47~$\pm$~0.47 & 84.86~$\pm$~0.57 & 95.54~$\pm$~0.11 \\
    Single-color & & 74.02~$\pm$~0.62 & 84.64~$\pm$~0.85 & 95.75~$\pm$~0.02 \\
    \midrule
    Luminance  & \multirow{3}{*}{0.005} & 73.86~$\pm$~0.45 & 84.84~$\pm$~0.25 & 95.59~$\pm$~0.12 \\
    Color-opponency& & 73.25~$\pm$~0.98 & 84.48~$\pm$~0.53 & 95.53~$\pm$~0.12 \\
    Single-color & & 73.37~$\pm$~1.45 & 84.16~$\pm$~1.34 & 95.67~$\pm$~0.12 \\
    \midrule
    Luminance  & \multirow{3}{*}{0.0075} & 73.15~$\pm$~0.79 & 84.43~$\pm$~0.60 & 95.56~$\pm$~0.02 \\
    Color-opponency& & 73.92~$\pm$~0.76 & 85.17~$\pm$~0.96 & 95.54~$\pm$~0.05 \\
    Single-color & & 74.28~$\pm$~0.37 & 84.95~$\pm$~0.19 & 95.76~$\pm$~0.04 \\
    \midrule
    Luminance  & \multirow{3}{*}{0.01} & 72.79~$\pm$~1.16 & 84.29~$\pm$~0.74 & 95.42~$\pm$~0.06 \\
    Color-opponency& & 73.67~$\pm$~0.93 & 84.82~$\pm$~0.69 & 95.51~$\pm$~0.09 \\
    Single-color & & 74.17~$\pm$~1.53 & 85.16~$\pm$~1.71 & 95.75~$\pm$~0.05 \\
    \midrule
    Luminance  & \multirow{3}{*}{0.02} & 70.48~$\pm$~1.21 & 82.67~$\pm$~1.11 & 95.00~$\pm$~0.11 \\
    Color-opponency& & 71.34~$\pm$~1.34 & 83.41~$\pm$~1.15 & 95.06~$\pm$~0.16 \\
    Single-color & & 72.53~$\pm$~1.71 & 83.65~$\pm$~1.71 & 95.47~$\pm$~0.11 \\

    \bottomrule
  \end{tabular}
  \label{tab:sparse_results_mask2former_cityscapes_threshold}
\end{table}

\begin{table}[tb]
  \centering
  \caption{Evaluation of the \textbf{Mask2Former} architecture trained on Cityscapes and \textbf{validated on Dark Zurich} at different sparsities. Results include the three preprocessing types (luminance, color-opponency, and single-color). Metrics reported are mean Intersection over Union (mIoU), mean Accuracy (mAcc), and average Accuracy (aAcc), averaged over three seeds. \textbf{Threshold-based sparsity:} For a given \textit{threshold}, all values within the range $[-\textit{threshold}, +\textit{threshold}]$ are set to zero. }   
  \begin{tabular}{@{}lc rrr@{}}
    \toprule
    Preprocessing &  Sparsity &\multicolumn{1}{c}{mIoU}&\multicolumn{1}{c}{mAcc}&\multicolumn{1}{c}{aAcc}  \\
    \midrule
    - & - & 18.16~$\pm$~1.14 & 28.88~$\pm$~2.72 & 47.34~$\pm$~4.13 \\
    \midrule
    \midrule
    Luminance & \multirow{3}{*}{-} & 26.21~$\pm$~0.18 & 40.56~$\pm$~2.23 & 59.16~$\pm$~0.58 \\
    Color-opponency & & 22.84~$\pm$~1.82 & 37.37~$\pm$~2.58 & 55.26~$\pm$~2.45 \\
    Single-color & & 24.92~$\pm$~1.09 & 39.64~$\pm$~2.37 & 57.09~$\pm$~3.52 \\
    \midrule
    Luminance  & \multirow{3}{*}{0.005} & 25.56~$\pm$~1.10 & 39.33~$\pm$~1.72 & 60.46~$\pm$~3.71 \\
    Color-opponency& & 22.45~$\pm$~2.06 & 36.84~$\pm$~1.03 & 54.82~$\pm$~1.03 \\
    Single-color & & 25.32~$\pm$~0.22 & 39.18~$\pm$~1.25 & 56.43~$\pm$~0.36 \\
    \midrule
    Luminance  & \multirow{3}{*}{0.0075} & 25.88~$\pm$~1.34 & 39.22~$\pm$~1.99 & 56.92~$\pm$~0.70 \\
    Color-opponency& & 24.34~$\pm$~1.79 & 38.30~$\pm$~2.84 & 56.38~$\pm$~2.89 \\
    Single-color & & 23.90~$\pm$~0.89 & 36.86~$\pm$~1.95 & 54.22~$\pm$~0.50 \\
    \midrule
    Luminance  & \multirow{3}{*}{0.01} & 25.04~$\pm$~1.17 & 37.91~$\pm$~0.78 & 58.82~$\pm$~1.17 \\
    Color-opponency& & 22.48~$\pm$~1.13 & 36.09~$\pm$~0.89 & 53.24~$\pm$~0.97 \\
    Single-color & & 24.95~$\pm$~1.17 & 39.48~$\pm$~2.67 & 56.30~$\pm$~1.91 \\
    \midrule
    Luminance  & \multirow{3}{*}{0.02} & 24.33~$\pm$~1.72 & 36.86~$\pm$~2.90 & 55.09~$\pm$~3.89 \\
    Color-opponency& & 22.41~$\pm$~2.04 & 34.63~$\pm$~2.16 & 51.91~$\pm$~1.28 \\
    Single-color & & 22.50~$\pm$~0.78 & 37.56~$\pm$~2.41 & 53.74~$\pm$~0.89 \\

    \bottomrule
  \end{tabular}
  \label{tab:sparse_results_mask2former_dark_zurich_threshold}
\end{table}

\begin{table}[tb]
  \centering
  \caption{Evaluation of the \textbf{Mask2Former} architecture trained on Cityscapes and \textbf{validated on ACDC Night} at different sparsities. Results include the three preprocessing types (luminance, color-opponency, and single-color). Metrics reported are mean Intersection over Union (mIoU), mean Accuracy (mAcc), and average Accuracy (aAcc), averaged over three seeds. \textbf{Threshold-based sparsity:} For a given \textit{threshold}, all values within the range $[-\textit{threshold}, +\textit{threshold}]$ are set to zero. }   
  \begin{tabular}{@{}lc rrr@{}}
    \toprule
    Preprocessing &  Sparsity &\multicolumn{1}{c}{mIoU}&\multicolumn{1}{c}{mAcc}&\multicolumn{1}{c}{aAcc}  \\
    \midrule
    - & - & 19.65~$\pm$~1.22 & 32.32~$\pm$~1.80 & 51.57~$\pm$~2.56 \\
    \midrule
    \midrule
    Luminance & \multirow{3}{*}{-} & 28.45~$\pm$~1.18 & 42.50~$\pm$~2.86 & 62.43~$\pm$~0.83 \\
    Color-opponency & & 26.08~$\pm$~2.72 & 41.45~$\pm$~3.98 & 58.44~$\pm$~3.31 \\
    Single-color & & 26.64~$\pm$~0.26 & 42.38~$\pm$~1.67 & 60.03~$\pm$~2.94 \\
    \midrule
    Luminance  & \multirow{3}{*}{0.005} & 27.26~$\pm$~2.00 & 41.39~$\pm$~2.12 & 63.40~$\pm$~3.12 \\
    Color-opponency& & 25.01~$\pm$~1.83 & 40.05~$\pm$~1.12 & 58.06~$\pm$~0.95 \\
    Single-color & & 27.42~$\pm$~2.70 & 42.04~$\pm$~0.50 & 59.62~$\pm$~1.07 \\
    \midrule
    Luminance  & \multirow{3}{*}{0.0075} & 27.84~$\pm$~1.28 & 42.09~$\pm$~1.63 & 60.77~$\pm$~1.16 \\
    Color-opponency& & 25.80~$\pm$~2.77 & 39.84~$\pm$~3.32 & 59.72~$\pm$~2.36 \\
    Single-color & & 24.83~$\pm$~1.18 & 37.92~$\pm$~2.13 & 56.89~$\pm$~0.28 \\
    \midrule
    Luminance  & \multirow{3}{*}{0.01} & 26.80~$\pm$~1.41 & 39.99~$\pm$~1.47 & 61.66~$\pm$~1.58 \\
    Color-opponency& & 23.64~$\pm$~1.71 & 37.38~$\pm$~0.53 & 56.51~$\pm$~1.26 \\
    Single-color & & 26.38~$\pm$~1.51 & 41.55~$\pm$~2.54 & 58.97~$\pm$~1.47 \\
    \midrule
    Luminance  & \multirow{3}{*}{0.02} & 26.61~$\pm$~1.75 & 39.62~$\pm$~3.21 & 57.69~$\pm$~3.98 \\
    Color-opponency& & 23.47~$\pm$~2.09 & 35.62~$\pm$~2.18 & 54.03~$\pm$~2.14 \\
    Single-color & & 23.32~$\pm$~1.19 & 39.07~$\pm$~2.14 & 56.21~$\pm$~2.56 \\
    \bottomrule
  \end{tabular}
  \label{tab:sparse_results_mask2former_acdc_night_threshold}
\end{table}

\begin{table}[tb]
  \centering
  \caption{Evaluation of the \textbf{Mask2Former} architecture trained on Cityscapes and \textbf{validated on ACDC Fog} at different sparsities. Results include the three preprocessing types (luminance, color-opponency, and single-color). Metrics reported are mean Intersection over Union (mIoU), mean Accuracy (mAcc), and average Accuracy (aAcc), averaged over three seeds. \textbf{Threshold-based sparsity:} For a given \textit{threshold}, all values within the range $[-\textit{threshold}, +\textit{threshold}]$ are set to zero. }   
  \begin{tabular}{@{}lc rrr@{}}
    \toprule
    Preprocessing &  Sparsity &\multicolumn{1}{c}{mIoU}&\multicolumn{1}{c}{mAcc}&\multicolumn{1}{c}{aAcc}  \\
    \midrule
    - & - & 64.70~$\pm$~3.04 & 79.77~$\pm$~3.06 & 91.08~$\pm$~1.46 \\
    \midrule
    \midrule
    Luminance & \multirow{3}{*}{-} & 62.16~$\pm$~1.18 & 75.22~$\pm$~0.88 & 91.30~$\pm$~1.05 \\
    Color-opponency & & 61.58~$\pm$~0.95 & 77.08~$\pm$~1.58 & 89.45~$\pm$~1.28 \\
    Single-color & & 63.77~$\pm$~3.25 & 77.78~$\pm$~2.56 & 90.43~$\pm$~1.33 \\
    \midrule
    Luminance  & \multirow{3}{*}{0.005} & 63.31~$\pm$~2.75 & 76.30~$\pm$~3.42 & 92.21~$\pm$~0.50 \\
    Color-opponency& & 64.27~$\pm$~0.40 & 78.17~$\pm$~0.43 & 90.86~$\pm$~1.43 \\
    Single-color & & 62.21~$\pm$~1.92 & 78.07~$\pm$~1.52 & 90.80~$\pm$~1.12 \\
    \midrule
    Luminance  & \multirow{3}{*}{0.0075} & 65.32~$\pm$~3.37 & 78.15~$\pm$~3.16 & 92.08~$\pm$~0.22 \\
    Color-opponency& & 64.42~$\pm$~1.15 & 76.75~$\pm$~0.59 & 91.90~$\pm$~0.88 \\
    Single-color & & 64.62~$\pm$~1.12 & 78.39~$\pm$~0.33 & 90.07~$\pm$~0.88 \\
    \midrule
    Luminance  & \multirow{3}{*}{0.01} & 63.72~$\pm$~4.78 & 75.37~$\pm$~3.69 & 91.60~$\pm$~1.17 \\
    Color-opponency& & 64.02~$\pm$~0.71 & 76.25~$\pm$~1.10 & 91.65~$\pm$~0.91 \\
    Single-color & & 61.83~$\pm$~3.69 & 76.87~$\pm$~3.10 & 90.96~$\pm$~1.93 \\
    \midrule
    Luminance  & \multirow{3}{*}{0.02} & 60.92~$\pm$~1.92 & 73.32~$\pm$~2.72 & 90.68~$\pm$~0.67 \\
    Color-opponency& & 57.47~$\pm$~3.76 & 70.63~$\pm$~3.23 & 89.55~$\pm$~0.54 \\
    Single-color & & 59.42~$\pm$~6.66 & 74.27~$\pm$~6.31 & 88.32~$\pm$~3.72 \\
    \bottomrule
  \end{tabular}
  \label{tab:sparse_results_mask2former_acdc_fog_threshold}
\end{table}

\begin{table}[tb]
  \centering
  \caption{Evaluation of the \textbf{Mask2Former} architecture trained on Cityscapes and \textbf{validated on ACDC Rain} at different sparsities. Results include the three preprocessing types (luminance, color-opponency, and single-color). Metrics reported are mean Intersection over Union (mIoU), mean Accuracy (mAcc), and average Accuracy (aAcc), averaged over three seeds. \textbf{Threshold-based sparsity:} For a given \textit{threshold}, all values within the range $[-\textit{threshold}, +\textit{threshold}]$ are set to zero. }   
  \begin{tabular}{@{}lc rrr@{}}
    \toprule
    Preprocessing &  Sparsity &\multicolumn{1}{c}{mIoU}&\multicolumn{1}{c}{mAcc}&\multicolumn{1}{c}{aAcc}  \\
    \midrule
    - & - & 49.91~$\pm$~1.32 & 71.13~$\pm$~1.71 & 87.97~$\pm$~0.38 \\
    \midrule
    \midrule
    Luminance & \multirow{3}{*}{-} & 47.20~$\pm$~0.13 & 64.00~$\pm$~2.66 & 87.95~$\pm$~1.50 \\
    Color-opponency & & 44.45~$\pm$~0.48 & 60.91~$\pm$~0.56 & 86.78~$\pm$~1.08 \\
    Single-color & & 44.86~$\pm$~3.38 & 62.10~$\pm$~5.34 & 86.84~$\pm$~3.21 \\
    \midrule
    Luminance  & \multirow{3}{*}{0.005} & 46.79~$\pm$~2.79 & 62.21~$\pm$~3.94 & 88.84~$\pm$~1.17 \\
    Color-opponency& & 46.55~$\pm$~1.20 & 63.55~$\pm$~2.77 & 87.43~$\pm$~0.41 \\
    Single-color & & 45.66~$\pm$~1.02 & 64.00~$\pm$~0.16 & 87.90~$\pm$~0.85 \\
    \midrule
    Luminance  & \multirow{3}{*}{0.0075} & 47.57~$\pm$~1.64 & 63.40~$\pm$~2.48 & 88.77~$\pm$~0.93 \\
    Color-opponency& & 49.19~$\pm$~0.32 & 65.45~$\pm$~0.92 & 89.33~$\pm$~0.50 \\
    Single-color & & 45.74~$\pm$~1.83 & 64.07~$\pm$~1.60 & 86.91~$\pm$~1.78 \\
    \midrule
    Luminance  & \multirow{3}{*}{0.01} & 48.35~$\pm$~1.56 & 63.42~$\pm$~2.48 & 89.20~$\pm$~0.89 \\
    Color-opponency& & 47.25~$\pm$~2.20 & 63.12~$\pm$~1.34 & 88.74~$\pm$~0.89 \\
    Single-color & & 46.41~$\pm$~1.10 & 63.68~$\pm$~2.03 & 88.10~$\pm$~0.65 \\
    \midrule
    Luminance  & \multirow{3}{*}{0.02} & 46.83~$\pm$~1.47 & 60.90~$\pm$~2.67 & 88.47~$\pm$~0.41 \\
    Color-opponency& & 43.88~$\pm$~0.19 & 59.78~$\pm$~0.37 & 85.79~$\pm$~1.07 \\
    Single-color & & 42.83~$\pm$~2.65 & 59.83~$\pm$~3.83 & 83.68~$\pm$~5.26 \\
    \bottomrule
  \end{tabular}
  \label{tab:sparse_results_mask2former_acdc_rain_threshold}
\end{table}

\begin{table}[tb]
  \centering
  \caption{Evaluation of the \textbf{Mask2Former} architecture trained on Cityscapes and \textbf{validated on ACDC Snow} at different sparsities. Results include the three preprocessing types (luminance, color-opponency, and single-color). Metrics reported are mean Intersection over Union (mIoU), mean Accuracy (mAcc), and average Accuracy (aAcc), averaged over three seeds. \textbf{Threshold-based sparsity:} For a given \textit{threshold}, all values within the range $[-\textit{threshold}, +\textit{threshold}]$ are set to zero. }   
  \begin{tabular}{@{}lc rrr@{}}
    \toprule
    Preprocessing &  Sparsity &\multicolumn{1}{c}{mIoU}&\multicolumn{1}{c}{mAcc}&\multicolumn{1}{c}{aAcc}  \\
    \midrule
    - & - & 48.20~$\pm$~2.09 & 62.94~$\pm$~2.16 & 81.98~$\pm$~1.71 \\
    \midrule
    \midrule
    Luminance & \multirow{3}{*}{-} & 49.87~$\pm$~2.06 & 60.95~$\pm$~0.88 & 85.47~$\pm$~2.31 \\
    Color-opponency & & 47.48~$\pm$~0.67 & 60.15~$\pm$~0.81 & 82.73~$\pm$~1.13 \\
    Single-color & & 46.97~$\pm$~3.98 & 61.26~$\pm$~5.43 & 81.86~$\pm$~3.87 \\
    \midrule
    Luminance  & \multirow{3}{*}{0.005} & 52.00~$\pm$~4.36 & 63.09~$\pm$~3.14 & 86.74~$\pm$~2.19 \\
    Color-opponency& & 48.66~$\pm$~2.65 & 61.25~$\pm$~2.51 & 83.34~$\pm$~2.61 \\
    Single-color & & 47.65~$\pm$~2.28 & 60.89~$\pm$~3.06 & 83.96~$\pm$~1.62 \\
    \midrule
    Luminance  & \multirow{3}{*}{0.0075} & 48.88~$\pm$~2.06 & 60.96~$\pm$~0.80 & 84.89~$\pm$~1.39 \\
    Color-opponency& & 49.53~$\pm$~4.98 & 62.36~$\pm$~4.37 & 86.06~$\pm$~2.00 \\
    Single-color & & 44.84~$\pm$~3.33 & 57.65~$\pm$~3.49 & 81.84~$\pm$~2.52 \\
    \midrule
    Luminance  & \multirow{3}{*}{0.01} & 48.78~$\pm$~0.77 & 59.25~$\pm$~0.25 & 85.88~$\pm$~1.97 \\
    Color-opponency& & 49.22~$\pm$~1.95 & 61.26~$\pm$~1.00 & 85.18~$\pm$~1.90 \\
    Single-color & & 48.07~$\pm$~1.12 & 61.08~$\pm$~0.41 & 82.96~$\pm$~1.48 \\
    \midrule
    Luminance  & \multirow{3}{*}{0.02} & 47.71~$\pm$~2.59 & 58.17~$\pm$~2.00 & 86.28~$\pm$~1.07 \\
    Color-opponency& & 46.56~$\pm$~3.10 & 58.30~$\pm$~2.13 & 83.41~$\pm$~2.29 \\
    Single-color & & 43.20~$\pm$~3.16 & 56.25~$\pm$~4.12 & 78.76~$\pm$~3.18 \\
    \bottomrule
  \end{tabular}
  \label{tab:sparse_results_mask2former_acdc_snow_threshold}
\end{table}

\begin{table}[tb]
  \centering
  \caption{Evaluation of the \textbf{Mask2Former} architecture trained on Cityscapes and \textbf{validated on ACDC Mean} at different sparsities. Results include the three preprocessing types (luminance, color-opponency, and single-color). Metrics reported are mean Intersection over Union (mIoU), mean Accuracy (mAcc), and average Accuracy (aAcc), averaged over three seeds. \textbf{Threshold-based sparsity:} For a given \textit{threshold}, all values within the range $[-\textit{threshold}, +\textit{threshold}]$ are set to zero. }   
  \begin{tabular}{@{}lc rrr@{}}
    \toprule
    Preprocessing &  Sparsity &\multicolumn{1}{c}{mIoU}&\multicolumn{1}{c}{mAcc}&\multicolumn{1}{c}{aAcc}  \\
    \midrule
    - & - & 42.05~$\pm$~1.41 & 60.13~$\pm$~2.21 & 77.94~$\pm$~1.35\\
    \midrule
    \midrule
    Luminance & \multirow{3}{*}{-} & 46.83~$\pm$~1.10 & 59.72~$\pm$~2.32 & 81.61~$\pm$~1.38 \\
     Color-opponency &  & 44.93~$\pm$~0.66 & 58.84~$\pm$~0.86 & 79.18~$\pm$~0.62 \\
    Single-color &  & 45.23~$\pm$~1.74 & 59.85~$\pm$~2.53 & 79.63~$\pm$~2.76 \\
    \midrule
    Luminance & \multirow{3}{*}{0.005} & 47.21~$\pm$~2.28 & 60.01~$\pm$~1.66 & 82.64~$\pm$~1.75 \\
    Color-opponency & & 45.43~$\pm$~1.53 & 59.11~$\pm$~1.40 & 79.75~$\pm$~1.26 \\
    Single-color & & 45.64~$\pm$~2.38 & 60.73~$\pm$~1.41 & 80.40~$\pm$~0.87 \\
    \midrule
    Luminance & \multirow{3}{*}{0.0075} & 47.33~$\pm$~1.61 & 60.29~$\pm$~1.44 & 81.46~$\pm$~0.77 \\
    Color-opponency & & 46.73~$\pm$~1.66 & 59.47~$\pm$~1.12 & 81.58~$\pm$~0.62 \\
    Single-color & & 44.05~$\pm$~1.70 & 57.93~$\pm$~2.15 & 78.75~$\pm$~1.29 \\
    \midrule
    Luminance & \multirow{3}{*}{0.01} & 46.25~$\pm$~1.85 & 58.72~$\pm$~0.26 & 81.93~$\pm$~1.27 \\
    Color-opponency & & 45.34~$\pm$~2.50 & 58.05~$\pm$~1.33 & 80.33~$\pm$~1.02 \\
    Single-color & & 45.42~$\pm$~0.70 & 60.61~$\pm$~1.68 & 80.08~$\pm$~1.20 \\
    \midrule
    Luminance & \multirow{3}{*}{0.02} & 45.68~$\pm$~1.11 & 56.75~$\pm$~1.12 & 80.60~$\pm$~1.29 \\
    Color-opponency & & 42.46~$\pm$~1.76 & 54.13~$\pm$~1.27 & 78.00~$\pm$~1.03 \\
    Single-color & & 41.89~$\pm$~2.70 & 56.23~$\pm$~3.78 & 76.58~$\pm$~3.19 \\
    \bottomrule
  \end{tabular}
  \label{tab:sparse_results_mask2former_acdc_full_threshold}
\end{table}
\clearpage

%% file: tables/sparsity_threshold_zero_percentage.tex
\begin{table}[h]
  \centering
  \caption{Zero-pixel percentages in the test-set across \textbf{Cityscapes, Dark Zurich, all ACDC subsets, and ACDC Mean} at different sparsity levels. Results are reported for the three preprocessing types (luminance, color-opponency, and single-color). \textbf{Threshold-based sparsity:} For a given \textit{threshold}, all values within the range $[-\textit{threshold}, +\textit{threshold}]$ are set to zero.}
  \begin{tabular}{@{}lcrrrr@{}}
    \toprule
    Dataset & Preprocessing &\multicolumn{1}{c}{0.005}   &\multicolumn{1}{c}{0.0075}  &\multicolumn{1}{c}{0.01}  &\multicolumn{1}{c}{0.02}  \\
    \midrule
    \multirow{3}{*}{Cityscapes} & Luminance  & 20.61$\pm$3.76 & 29.37$\pm$5.01 & 37.10$\pm$5.96 & 58.43$\pm$7.26 \\
     & Color-opponency & 44.20$\pm$4.51 & 57.38$\pm$5.05 & 66.02$\pm$5.03 & 81.51$\pm$4.05 \\
     & Single-color & 18.33$\pm$3.18 & 26.31$\pm$4.22 & 33.64$\pm$5.13 & 55.62$\pm$6.90 \\
    \midrule
    \multirow{3}{*}{Dark Zurich} & Luminance  & 27.01$\pm$7.70 & 37.64$\pm$8.56 & 46.95$\pm$8.86 & 72.71$\pm$7.17 \\
     & Color-opponency & 66.44$\pm$3.88 & 74.62$\pm$3.55 & 80.41$\pm$3.30 & 90.45$\pm$2.50 \\
     & Single-color & 26.16$\pm$7.49 & 36.55$\pm$8.36 & 45.70$\pm$8.71 & 71.43$\pm$7.20 \\
    \midrule
    \multirow{3}{*}{ACDC Night} & Luminance  & 26.40$\pm$6.88 & 36.95$\pm$7.64 & 46.22$\pm$7.93 & 72.03$\pm$6.61 \\
     & Color-opponency & 66.14$\pm$3.55 & 74.34$\pm$3.24 & 80.12$\pm$3.00 & 90.21$\pm$2.33 \\
     & Single-color & 25.55$\pm$6.67 & 35.85$\pm$7.42 & 44.97$\pm$7.76 & 70.76$\pm$6.63 \\
    \midrule
    \multirow{3}{*}{ACDC Fog} & Luminance  & 33.83$\pm$2.97 & 43.90$\pm$3.82 & 51.03$\pm$4.63 & 67.22$\pm$5.91 \\
     & Color-opponency & 69.62$\pm$1.80 & 77.61$\pm$1.58 & 82.07$\pm$1.65 & 88.68$\pm$1.96 \\
     & Single-color & 32.28$\pm$2.88 & 42.39$\pm$3.70 & 49.64$\pm$4.52 & 66.49$\pm$5.85 \\
    \midrule
    \multirow{3}{*}{ACDC Rain} & Luminance  & 33.12$\pm$4.21 & 42.00$\pm$4.88 & 47.96$\pm$5.28 & 61.67$\pm$5.79 \\
     & Color-opponency & 68.59$\pm$2.45 & 76.50$\pm$2.15 & 80.72$\pm$2.03 & 86.75$\pm$1.98 \\
     & Single-color & 31.57$\pm$4.13 & 40.64$\pm$4.79 & 46.75$\pm$5.21 & 61.08$\pm$5.76 \\
    \midrule
    \multirow{3}{*}{ACDC Snow} & Luminance  & 27.00$\pm$4.63 & 34.64$\pm$6.03 & 39.71$\pm$6.93 & 51.25$\pm$8.19 \\
     & Color-opponency & 67.72$\pm$2.58 & 74.71$\pm$2.43 & 78.38$\pm$2.50 & 83.37$\pm$2.72 \\
     & Single-color & 25.79$\pm$4.56 & 33.58$\pm$5.91 & 38.81$\pm$6.83 & 50.87$\pm$8.14 \\
     \midrule
     \multirow{3}{*}{ACDC Mean} & Luminance  & 30.03$\pm$5.97 &  39.34$\pm$6.87 & 46.23$\pm$7.55 & 63.17$\pm$10.23 \\
     & Color-opponency & 67.99$\pm$2.96 & 75.77$\pm$2.77 & 80.32$\pm$2.70 & 87.30$\pm$3.42 \\
     & Single-color & 28.75$\pm$5.72 & 38.08$\pm$6.65 & 45.04$\pm$7.36 & 62.43$\pm$9.98 \\
    \bottomrule
  \end{tabular}
  \label{tab:zero_percentages_threshold}
\end{table}
\clearpage

%% file: tables/training_from_scratch.tex
\begin{table}[h]
  \centering
  \caption{Part 1 of the comparison between \textbf{SegFormer} trained from scratch on Cityscapes and \textbf{SegFormer} fine-tuned on Cityscapes using pretrained weights \textbf{validated on Cityscapes, Dark Zurich, all ACDC subsets, and ACDC Mean}. Results include the three preprocessing types (luminance, color-opponency, and single-color. Metrics reported are mean Intersection over Union (mIoU), mean Accuracy (mAcc), and average Accuracy (aAcc), averaged over three seeds.}
  \begin{tabular}{@{}lccrrr@{}}
  \toprule
    Dataset & Pretrained & Preprocessing & \multicolumn{1}{c}{mIoU} & \multicolumn{1}{c}{mAcc} & \multicolumn{1}{c}{aAcc} \\
    \midrule
    \multirow{8}{*}{Cityscapes} & \multirow{4}{*}{no} & - &   56.13~$\pm$~0.51 ~& 65.37~$\pm$~0.26 ~& 91.74~$\pm$~0.47 \\
    & &  Luminance & 51.63~$\pm$~1.23 ~& 61.67~$\pm$~1.92 ~& 90.35~$\pm$~0.51 \\
    & &  Color-opponency & 55.45~$\pm$~1.04 ~& 65.49~$\pm$~1.78 ~& 91.70~$\pm$~0.27 \\
    & & Single-color & 54.83~$\pm$~0.34 ~& 65.73~$\pm$~0.68 ~& 91.12~$\pm$~0.33 \\
    \cmidrule(lr){2-6}
    & \multirow{4}{*}{yes} & - &   73.86~$\pm$~0.29 & 82.58~$\pm$~0.59 & 95.36~$\pm$~0.03 \\
    & &  Luminance & 68.73~$\pm$~0.20 & 77.97~$\pm$~0.38 & 94.30~$\pm$~0.10 \\
    & &  Color-opponency & 71.26~$\pm$~0.16 & 80.22~$\pm$~0.58 & 94.80~$\pm$~0.05 \\
    & & Single-color & 71.59~$\pm$~0.67 & 80.56~$\pm$~0.59 & 94.79~$\pm$~0.09 \\
    \midrule
    \multirow{8}{*}{Dark Zurich} &\multirow{4}{*}{no} &  - &   7.46~$\pm$~0.71 ~& 17.35~$\pm$~2.13 ~& 32.52~$\pm$~1.05 \\
    & & Luminance & 13.52~$\pm$~2.92 ~& 26.22~$\pm$~3.31 ~& 41.30~$\pm$~8.80 \\
    & & Color-opponency & 11.80~$\pm$~0.50 ~& 22.54~$\pm$~0.93 ~& 39.86~$\pm$~2.05 \\
    & & Single-color & 10.90~$\pm$~0.78 ~& 23.43~$\pm$~1.07 ~& 36.88~$\pm$~2.83 \\
    \cmidrule(lr){2-6}
    & \multirow{4}{*}{yes} & - &   13.95~$\pm$~0.79 & 26.98~$\pm$~1.19 & 47.96~$\pm$~0.57 \\
    & &  Luminance & 19.99~$\pm$~0.31 & 33.67~$\pm$~1.31 & 54.02~$\pm$~1.98 \\
    & &  Color-opponency & 17.50~$\pm$~0.82 & 31.33~$\pm$~0.12 & 52.46~$\pm$~1.01 \\
    & & Single-color & 18.76~$\pm$~0.13 & 33.11~$\pm$~0.19 & 53.88~$\pm$~1.79 \\
    \midrule
    \multirow{8}{*}{ACDC Night}& \multirow{4}{*}{no}  & - &   8.69~$\pm$~0.88 ~& 17.99~$\pm$~1.93 ~& 36.26~$\pm$~2.60 \\
    & & Luminance & 15.12~$\pm$~2.91 ~& 26.96~$\pm$~3.07 ~& 46.91~$\pm$~8.34 \\
    & & Color-opponency & 13.13~$\pm$~0.34 ~& 23.73~$\pm$~1.33 ~& 43.88~$\pm$~1.55 \\
    & & Single-color & 12.41~$\pm$~1.03 ~& 24.97~$\pm$~1.21 ~& 40.25~$\pm$~2.88 \\
    \cmidrule(lr){2-6}
    & \multirow{4}{*}{yes} & - &   15.76~$\pm$~1.09 & 28.11~$\pm$~1.16 & 50.68~$\pm$~1.11 \\
    & &  Luminance & 22.84~$\pm$~0.50 & 35.40~$\pm$~1.27 & 58.25~$\pm$~1.50 \\
    & &  Color-opponency & 18.40~$\pm$~0.51 & 31.41~$\pm$~1.63 & 55.52~$\pm$~0.89 \\
    & & Single-color & 19.78~$\pm$~0.13 & 32.70~$\pm$~0.48 & 56.70~$\pm$~0.86 \\
    \bottomrule
  \end{tabular}
  \label{tab:training_from_scratch}
\end{table}

\begin{table}[tb]
  \centering
  \caption{Part 2 of the comparison between \textbf{SegFormer} trained from scratch on Cityscapes and \textbf{SegFormer} fine-tuned on Cityscapes using pretrained weights \textbf{validated on Cityscapes, Dark Zurich, all ACDC subsets, and ACDC Mean}. Results include the three preprocessing types (luminance, color-opponency, and single-color. Metrics reported are mean Intersection over Union (mIoU), mean Accuracy (mAcc), and average Accuracy (aAcc), averaged over three seeds.}
  \begin{tabular}{@{}lccrrr@{}}
  \toprule
    Dataset & Pretrained & Preprocessing & \multicolumn{1}{c}{mIoU} & \multicolumn{1}{c}{mAcc} & \multicolumn{1}{c}{aAcc} \\
    \midrule
    \multirow{8}{*}{ACDC Fog}& \multirow{4}{*}{no}  & - &   35.85~$\pm$~2.14 ~& 50.32~$\pm$~4.23 ~& 80.83~$\pm$~0.74 \\
    & & Luminance & 29.12~$\pm$~2.35 ~& 43.65~$\pm$~2.42 ~& 71.85~$\pm$~3.36 \\
    & & Color-opponency & 31.04~$\pm$~2.11 ~& 44.62~$\pm$~1.78 ~& 72.64~$\pm$~2.18 \\
    & & Single-color & 30.73~$\pm$~1.74 ~& 46.13~$\pm$~1.84 ~& 75.45~$\pm$~3.18 \\
    \cmidrule(lr){2-6}
    & \multirow{4}{*}{yes} & - &   63.67~$\pm$~0.94 & 74.53~$\pm$~0.83 & 92.79~$\pm$~0.19 \\
    & &  Luminance & 52.70~$\pm$~3.44 & 63.26~$\pm$~2.82 & 87.70~$\pm$~2.11 \\
    & &  Color-opponency & 57.79~$\pm$~1.80 & 67.87~$\pm$~2.60 & 90.20~$\pm$~0.80 \\
    & & Single-color & 54.55~$\pm$~0.14 & 66.95~$\pm$~0.67 & 90.42~$\pm$~0.54 \\
    \midrule
    \multirow{8}{*}{ACDC Rain}& \multirow{4}{*}{no}  & - &  26.49~$\pm$~2.98 ~& 39.56~$\pm$~3.29 ~& 75.76~$\pm$~3.35 \\
    & & Luminance & 21.63~$\pm$~1.44 ~& 36.90~$\pm$~2.97 ~& 63.55~$\pm$~3.78 \\
    & & Color-opponency & 23.81~$\pm$~1.99 ~& 38.07~$\pm$~1.79 ~& 65.92~$\pm$~2.66 \\
    & & Single-color & 23.92~$\pm$~0.66 ~& 40.78~$\pm$~2.17 ~& 68.58~$\pm$~3.87 \\
    \cmidrule(lr){2-6}
    & \multirow{4}{*}{yes} & - &   45.50~$\pm$~0.87 & 64.84~$\pm$~1.90 & 87.16~$\pm$~0.69 \\
    & &  Luminance & 38.12~$\pm$~0.91 & 50.77~$\pm$~1.20 & 82.60~$\pm$~2.43 \\
    & &  Color-opponency & 43.38~$\pm$~2.29 & 57.70~$\pm$~5.15 & 86.78~$\pm$~1.07 \\
    & & Single-color & 42.22~$\pm$~1.25 & 58.07~$\pm$~1.27 & 85.91~$\pm$~1.78 \\
    \midrule
    \multirow{8}{*}{ACDC Snow}& \multirow{4}{*}{no}  & - &  22.25~$\pm$~1.13 ~& 31.11~$\pm$~1.46 ~& 68.17~$\pm$~2.73 \\
    & & Luminance & 17.65~$\pm$~0.57 ~& 28.74~$\pm$~1.18 ~& 53.30~$\pm$~1.89 \\
    & & Color-opponency & 17.51~$\pm$~0.67 ~& 26.84~$\pm$~0.75 ~& 56.00~$\pm$~1.99 \\
    & & Single-color & 18.86~$\pm$~1.73 ~& 31.16~$\pm$~3.80 ~& 60.63~$\pm$~6.07 \\
    \cmidrule(lr){2-6}
    & \multirow{4}{*}{yes} & - &   46.92~$\pm$~0.51 & 57.57~$\pm$~0.80 & 86.05~$\pm$~0.19 \\
    & &  Luminance & 38.94~$\pm$~1.05 & 47.85~$\pm$~1.28 & 78.59~$\pm$~2.57 \\
    & &  Color-opponency & 42.60~$\pm$~0.63 & 50.74~$\pm$~1.28 & 83.58~$\pm$~1.17 \\
    & & Single-color & 41.89~$\pm$~2.11 & 50.58~$\pm$~1.07 & 83.20~$\pm$~2.76 \\
    \midrule
    \multirow{8}{*}{ACDC Mean}& \multirow{4}{*}{no}  & - &  23.32~$\pm$~1.64 & 34.74~$\pm$~2.53 & 65.26~$\pm$~1.85 \\
    & & Luminance & 20.88~$\pm$~1.54 & 34.06~$\pm$~0.94 & 58.90~$\pm$~3.36 \\
    & & Color-opponency & 21.38~$\pm$~1.26 & 33.32~$\pm$~0.95 & 59.61~$\pm$~1.53 \\
    & & Single-color & 21.48~$\pm$~0.69 & 35.76~$\pm$~1.64 & 61.23~$\pm$~2.94 \\
    \cmidrule(lr){2-6}
    & \multirow{4}{*}{yes} & - &   41.11~$\pm$~0.89 & 55.22~$\pm$~1.13 & 78.94~$\pm$~0.39\\
    & &  Luminance & 38.00~$\pm$~0.58 & 48.82~$\pm$~1.28 & 76.64~$\pm$~1.65 \\
    & &  Color-opponency & 38.87~$\pm$~0.36 & 50.58~$\pm$~2.39 & 78.83~$\pm$~0.96 \\
    & & Single-color &  38.74~$\pm$~0.98 & 51.29~$\pm$~0.14 & 78.88~$\pm$~1.46 \\
    \bottomrule
  \end{tabular}
  \label{tab:training_from_scratch_2}
\end{table}
\clearpage

%% file: tex_for_figures/appendix_ablation_grayscale_in_nighttime.tex
\begin{figure}[tb]
  \centering
  \includegraphics[width=\linewidth]{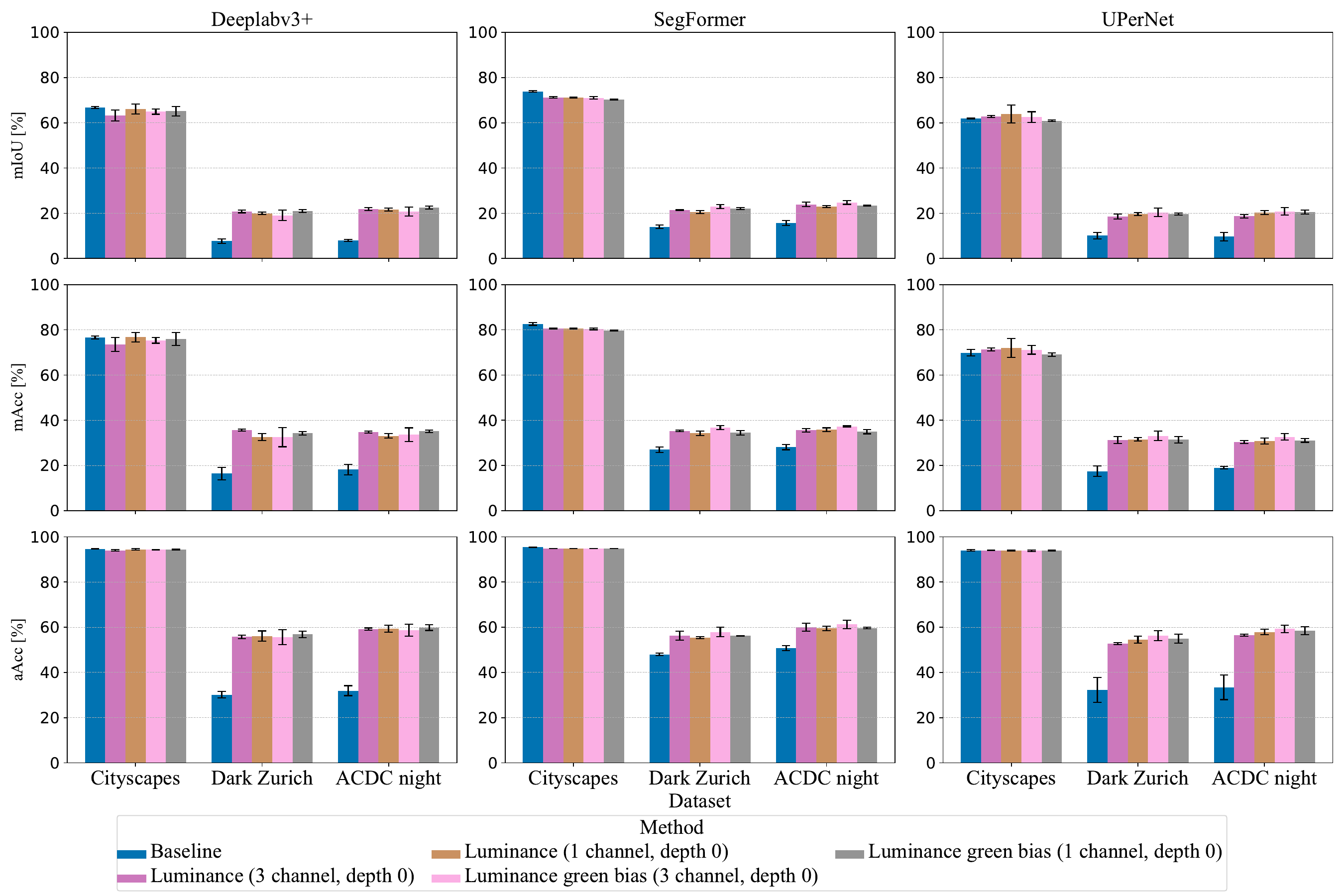}
  \caption{mIoU, mAcc and aAcc results of DeepLabv3+ (ResNetV1c-50), SegFormer, and InterImage architectures on Cityscapes, Dark Zurich and ACDC Night datasets for both luminance and luminance green bias variants. All architectures were trained with a depth of 0.}
  \label{fig:Validation_results_grayscale_in_nighttime_situations_appendix}
\end{figure}

%% file: tables/grayscale_evaluation.tex
\begin{table}[tb]
  \centering
  \caption{Evaluation of DeepLabv3+, SegFormer, and UPerNet architectures trained on Cityscapes and \textbf{validated on Cityscapes}. Results include two luminance variants (luminance and luminance green bias) each with depth=0. Metrics reported are mean Intersection over Union (mIoU), mean Accuracy (mAcc), and average Accuracy (aAcc), averaged over three seeds.}
  \begin{tabular}{@{}lccrrr@{}}
    \toprule
    Architecture & Preprocessing & Channel &\multicolumn{1}{c}{mIoU}&\multicolumn{1}{c}{mAcc}&\multicolumn{1}{c}{aAcc}  \\
    \midrule
    \multirow{6}{*}{DeepLabv3+}& - & 3 & 66.68~$\pm$~0.44 & 76.58~$\pm$~0.60 & 94.68~$\pm$~0.05 \\
    \cmidrule(lr){2-6}
     & Luminance green bias & \multirow{2}{*}{1} & 65.14~$\pm$~2.13 & 75.94~$\pm$~2.88 & 94.42~$\pm$~0.16 \\
     & Luminance &  & 66.02~$\pm$~2.18 & 76.76~$\pm$~2.09 & 94.47~$\pm$~0.27 \\
     \cmidrule(lr){2-6}
     & Luminance green bias & \multirow{2}{*}{3} & 64.97~$\pm$~1.19 & 75.34~$\pm$~1.28 & 94.31~$\pm$~0.13 \\
     & Luminance &  & 63.25~$\pm$~2.45 & 73.53~$\pm$~3.16 & 94.03~$\pm$~0.40 \\
    \midrule
    \multirow{6}{*}{SegFormer}& - & 3 & 73.86~$\pm$~0.29 & 82.58~$\pm$~0.59 & 95.36~$\pm$~0.03 \\
    \cmidrule(lr){2-6}
    & Luminance green bias & \multirow{2}{*}{1} & 70.25~$\pm$~0.16 & 79.57~$\pm$~0.21 & 94.75~$\pm$~0.02 \\
    & Luminance &  & 71.13~$\pm$~0.23 & 80.50~$\pm$~0.22 & 94.81~$\pm$~0.06 \\
    \cmidrule(lr){2-6}
    & Luminance green bias & \multirow{2}{*}{3} & 71.03~$\pm$~0.56 & 80.37~$\pm$~0.36 & 94.76~$\pm$~0.03 \\
    & Luminance &  & 71.19~$\pm$~0.32 & 80.56~$\pm$~0.18 & 94.82~$\pm$~0.06 \\
    \midrule
    \multirow{6}{*}{UPerNet}& - & 3 & 64.64~$\pm$~1.93 & 72.63~$\pm$~1.25 & 94.42~$\pm$~0.17 \\
    \cmidrule(lr){2-6}
     & Luminance green bias & \multirow{2}{*}{1}  & 60.90~$\pm$~0.30 & 69.00~$\pm$~0.71 & 93.93~$\pm$~0.17 \\
    & Luminance &  & 63.87~$\pm$~4.03 & 71.94~$\pm$~4.24 & 93.89~$\pm$~0.20 \\
    \cmidrule(lr){2-6}
    & Luminance green bias & \multirow{2}{*}{3} & 62.51~$\pm$~2.38 & 71.15~$\pm$~1.94 & 93.89~$\pm$~0.31 \\
    & Luminance &  & 62.84~$\pm$~0.41 & 71.32~$\pm$~0.66 & 94.00~$\pm$~0.13 \\
    \bottomrule
  \end{tabular}
  \label{tab:going_beyond_nighttime_cityscapes}
\end{table}

\begin{table}[tb]
  \centering
  \caption{Evaluation of DeepLabv3+, SegFormer, and UPerNet architectures trained on Cityscapes and \textbf{validated on Dark Zurich}. Results include two luminance variants (luminance and luminance green bias) each with depth=0. Metrics reported are mean Intersection over Union (mIoU), mean Accuracy (mAcc), and average Accuracy (aAcc), averaged over three seeds.}
  \begin{tabular}{@{}lccrrr@{}}
    \toprule
    Architecture & Preprocessing & Channel &\multicolumn{1}{c}{mIoU}&\multicolumn{1}{c}{mAcc}&\multicolumn{1}{c}{aAcc}  \\
    \midrule
    \multirow{6}{*}{DeepLabv3+}& - & 3 & 7.64~$\pm$~1.00 & 16.40~$\pm$~2.74 & 30.13~$\pm$~1.41 \\
    \cmidrule(lr){2-6}
    & Luminance green bias & \multirow{2}{*}{1} & 20.89~$\pm$~0.64 & 34.28~$\pm$~0.80 & 56.87~$\pm$~1.45 \\
    & Luminance &  & 19.92~$\pm$~0.55 & 32.51~$\pm$~1.51 & 56.13~$\pm$~2.24 \\
    \cmidrule(lr){2-6}
    & Luminance green bias & \multirow{2}{*}{3} & 19.04~$\pm$~2.31 & 32.55~$\pm$~4.29 & 55.59~$\pm$~3.38 \\
    & Luminance &  & 20.79~$\pm$~0.61 & 35.69~$\pm$~0.49 & 55.78~$\pm$~0.81 \\
    \midrule
    \multirow{6}{*}{SegFormer}& - & 3 & 13.95~$\pm$~0.79 & 26.98~$\pm$~1.19 & 47.96~$\pm$~0.57 \\
    \cmidrule(lr){2-6}
    & Luminance green bias & \multirow{2}{*}{1} & 22.06~$\pm$~0.48 & 34.47~$\pm$~1.01 & 56.23~$\pm$~0.13 \\
    & Luminance &  & 20.55~$\pm$~0.73 & 34.22~$\pm$~0.90 & 55.46~$\pm$~0.42 \\
    \cmidrule(lr){2-6}
    & Luminance green bias & \multirow{2}{*}{3} & 22.98~$\pm$~0.82 & 36.78~$\pm$~0.94 & 57.91~$\pm$~2.01 \\
    & Luminance &  & 21.47~$\pm$~0.23 & 35.31~$\pm$~0.35 & 56.24~$\pm$~1.97 \\
    \midrule
    \multirow{6}{*}{UPerNet}& - & 3 & 11.42~$\pm$~0.58 & 21.92~$\pm$~2.60 & 42.86~$\pm$~6.74 \\
    \cmidrule(lr){2-6}
    & Luminance green bias & \multirow{2}{*}{1} & 19.66~$\pm$~0.51 & 31.35~$\pm$~1.41 & 54.94~$\pm$~1.97 \\
    & Luminance &  & 19.63~$\pm$~0.62 & 31.52~$\pm$~0.78 & 54.58~$\pm$~1.51 \\
    \cmidrule(lr){2-6}
    & Luminance green bias & \multirow{2}{*}{3} & 20.34~$\pm$~1.87 & 33.09~$\pm$~2.06 & 56.29~$\pm$~2.14 \\
    & Luminance &  & 18.61~$\pm$~1.09 & 31.30~$\pm$~1.55 & 52.80~$\pm$~0.41 \\
    \bottomrule
  \end{tabular}
  \label{tab:going_beyond_nighttime_dark_zurich}%
\end{table}%

\begin{table}[tb]
  \centering
  \caption{Evaluation of DeepLabv3+, SegFormer, and UPerNet architectures trained on Cityscapes and \textbf{validated on ACDC Night}. Results include two luminance variants (luminance and luminance green bias) each with depth=0. Metrics reported are mean Intersection over Union (mIoU), mean Accuracy (mAcc), and average Accuracy (aAcc), averaged over three seeds.}
  \begin{tabular}{@{}lccrrr@{}}
    \toprule
    Architecture & Preprocessing & Channel &\multicolumn{1}{c}{mIoU}&\multicolumn{1}{c}{mAcc}&\multicolumn{1}{c}{aAcc}  \\
    \midrule
    \multirow{6}{*}{DeepLabv3+}& - & 3 & 8.01~$\pm$~0.45 & 18.14~$\pm$~2.24 & 31.94~$\pm$~2.20 \\
    \cmidrule(lr){2-6}
    & Luminance green bias & \multirow{2}{*}{1} & 22.48~$\pm$~0.71 & 35.12~$\pm$~0.62 & 59.80~$\pm$~1.22 \\
    & Luminance &  & 21.62~$\pm$~0.58 & 33.10~$\pm$~0.98 & 59.36~$\pm$~1.59 \\
    \cmidrule(lr){2-6}
    & Luminance green bias & \multirow{2}{*}{3} & 20.74~$\pm$~1.91 & 33.59~$\pm$~3.05 & 58.69~$\pm$~2.61 \\
    & Luminance &  & 21.87~$\pm$~0.71 & 34.84~$\pm$~0.42 & 59.17~$\pm$~0.51 \\
    \midrule
    \multirow{6}{*}{SegFormer}& - & 3 & 15.76~$\pm$~1.09 & 28.11~$\pm$~1.16 & 50.68~$\pm$~1.11 \\
    \cmidrule(lr){2-6}
    & Luminance green bias & \multirow{2}{*}{1} & 23.43~$\pm$~0.21 & 34.88~$\pm$~0.89 & 59.67~$\pm$~0.31 \\
    & Luminance &  & 22.96~$\pm$~0.39 & 35.85~$\pm$~0.78 & 59.50~$\pm$~1.04 \\
    \cmidrule(lr){2-6}
    & Luminance green bias & \multirow{2}{*}{3} & 24.76~$\pm$~0.83 & 37.23~$\pm$~0.29 & 61.24~$\pm$~1.81 \\
    & Luminance &  & 23.93~$\pm$~1.04 & 35.57~$\pm$~0.79 & 59.96~$\pm$~1.76 \\
    \midrule
    \multirow{6}{*}{UPerNet}& - & 3 & 11.53~$\pm$~1.24 & 21.96~$\pm$~2.20 & 45.24~$\pm$~7.17 \\
    \cmidrule(lr){2-6}
    & Luminance green bias & \multirow{2}{*}{1} & 20.55~$\pm$~0.94 & 31.12~$\pm$~0.87 & 58.43~$\pm$~1.73 \\
    & Luminance &  & 20.31~$\pm$~0.87 & 30.84~$\pm$~1.35 & 57.88~$\pm$~1.21 \\
    \cmidrule(lr){2-6}
    & Luminance green bias & \multirow{2}{*}{3} & 20.81~$\pm$~1.67 & 32.66~$\pm$~1.51 & 59.25~$\pm$~1.66 \\
    & Luminance &  & 18.70~$\pm$~0.75 & 30.40~$\pm$~0.63 & 56.50~$\pm$~0.38 \\
    \bottomrule
  \end{tabular}
  \label{tab:going_beyond_nighttime_cityscapes_acdc_night}%
\end{table}%

\begin{table}[tb]
  \centering
  \caption{Evaluation of DeepLabv3+, SegFormer, and UPerNet architectures trained on Cityscapes and \textbf{validated on ACDC Fog}. Results include two luminance variants (luminance and luminance green bias) each with depth=0. Metrics reported are mean Intersection over Union (mIoU), mean Accuracy (mAcc), and average Accuracy (aAcc), averaged over three seeds.}
  \begin{tabular}{@{}lccrrr@{}}
    \toprule
    Architecture & Preprocessing & Channel &\multicolumn{1}{c}{mIoU}&\multicolumn{1}{c}{mAcc}&\multicolumn{1}{c}{aAcc}  \\
    \midrule
    \multirow{6}{*}{DeepLabv3+}& - & 3 & 39.20~$\pm$~4.35 & 54.93~$\pm$~6.11 & 76.52~$\pm$~6.46 \\
    \cmidrule(lr){2-6}
    & Luminance green bias & \multirow{2}{*}{1} & 51.88~$\pm$~1.48 & 63.19~$\pm$~2.59 & 90.24~$\pm$~0.44 \\
    & Luminance &  & 52.60~$\pm$~1.94 & 61.64~$\pm$~2.24 & 90.38~$\pm$~0.99 \\
    \cmidrule(lr){2-6}
    & Luminance green bias & \multirow{2}{*}{3} & 47.49~$\pm$~3.24 & 58.31~$\pm$~3.34 & 82.17~$\pm$~11.65 \\
    & Luminance &  & 49.84~$\pm$~2.31 & 60.77~$\pm$~2.83 & 89.09~$\pm$~0.65 \\
    \midrule
    \multirow{6}{*}{SegFormer}& - & 3 & 63.67~$\pm$~0.94 & 74.53~$\pm$~0.83 & 92.79~$\pm$~0.19 \\
    \cmidrule(lr){2-6}
    & Luminance green bias & \multirow{2}{*}{1} & 58.86~$\pm$~3.27 & 69.27~$\pm$~3.05 & 90.92~$\pm$~0.46 \\
    & Luminance &  & 58.12~$\pm$~1.54 & 68.04~$\pm$~2.14 & 90.75~$\pm$~1.35 \\
    \cmidrule(lr){2-6}
    & Luminance green bias & \multirow{2}{*}{3} & 59.27~$\pm$~0.85 & 69.52~$\pm$~0.87 & 91.05~$\pm$~0.32 \\
    & Luminance &  & 59.38~$\pm$~1.60 & 69.48~$\pm$~1.56 & 91.65~$\pm$~0.18 \\
    \midrule
    \multirow{6}{*}{UPerNet}& - & 3 & 45.44~$\pm$~1.69 & 55.56~$\pm$~3.78 & 86.25~$\pm$~2.41 \\
    \cmidrule(lr){2-6}
    & Luminance green bias & \multirow{2}{*}{1} & 48.98~$\pm$~1.06 & 56.59~$\pm$~1.48 & 88.49~$\pm$~1.77 \\
    & Luminance &  & 49.41~$\pm$~2.44 & 57.32~$\pm$~2.59 & 88.24~$\pm$~1.22 \\
    \cmidrule(lr){2-6}
    & Luminance green bias & \multirow{2}{*}{3} & 47.93~$\pm$~2.64 & 57.65~$\pm$~1.86 & 85.04~$\pm$~7.06 \\
    & Luminance &  & 47.19~$\pm$~2.61 & 56.00~$\pm$~3.29 & 87.50~$\pm$~2.71 \\
    \bottomrule
  \end{tabular}
  \label{tab:going_beyond_nighttime_acdc_fog}%
\end{table}%

\begin{table}[tb]
  \centering
  \caption{Evaluation of DeepLabv3+, SegFormer, and UPerNet architectures trained on Cityscapes and \textbf{validated on ACDC Rain}. Results include two luminance variants (luminance and luminance green bias) each with depth=0. Metrics reported are mean Intersection over Union (mIoU), mean Accuracy (mAcc), and average Accuracy (aAcc), averaged over three seeds.}
  \begin{tabular}{@{}lccrrr@{}}
    \toprule
    Architecture & Preprocessing & Channel &\multicolumn{1}{c}{mIoU}&\multicolumn{1}{c}{mAcc}&\multicolumn{1}{c}{aAcc}  \\
    \midrule
    \multirow{6}{*}{DeepLabv3+}& - & 3 & 33.31~$\pm$~2.15 & 48.13~$\pm$~4.57 & 77.43~$\pm$~3.46 \\
    \cmidrule(lr){2-6}
    & Luminance green bias & \multirow{2}{*}{1} & 38.60~$\pm$~0.79 & 52.72~$\pm$~3.32 & 85.89~$\pm$~1.25 \\
    & Luminance &  & 39.12~$\pm$~1.13 & 51.11~$\pm$~2.79 & 86.59~$\pm$~0.41 \\
    \cmidrule(lr){2-6}
    & Luminance green bias & \multirow{2}{*}{3} & 34.45~$\pm$~4.04 & 46.76~$\pm$~5.08 & 78.70~$\pm$~8.52 \\
    & Luminance &  & 37.73~$\pm$~1.37 & 49.50~$\pm$~2.35 & 84.04~$\pm$~1.29 \\
    \midrule
    \multirow{6}{*}{SegFormer}& - & 3 & 45.50~$\pm$~0.87 & 64.84~$\pm$~1.90 & 87.16~$\pm$~0.69 \\
    \cmidrule(lr){2-6}
    & Luminance green bias & \multirow{2}{*}{1} & 39.65~$\pm$~1.35 & 55.69~$\pm$~1.72 & 83.49~$\pm$~0.27 \\
    & Luminance &  & 40.01~$\pm$~1.64 & 54.91~$\pm$~1.16 & 84.00~$\pm$~2.29 \\
    \cmidrule(lr){2-6}
    & Luminance green bias & \multirow{2}{*}{3} & 39.15~$\pm$~0.82 & 54.00~$\pm$~1.87 & 83.51~$\pm$~0.10 \\
    & Luminance &  & 39.37~$\pm$~0.98 & 54.01~$\pm$~1.52 & 84.30~$\pm$~1.82 \\
    \midrule
    \multirow{6}{*}{UPerNet}& - & 3 & 35.35~$\pm$~2.12 & 48.51~$\pm$~2.46 & 81.44~$\pm$~4.98 \\
    \cmidrule(lr){2-6}
    & Luminance green bias & \multirow{2}{*}{1} & 33.20~$\pm$~1.41 & 44.11~$\pm$~1.38 & 80.17~$\pm$~1.62 \\
    & Luminance &  & 32.60~$\pm$~1.39 & 41.31~$\pm$~1.66 & 79.11~$\pm$~2.16 \\
    \cmidrule(lr){2-6}
    & Luminance green bias & \multirow{2}{*}{3} & 30.20~$\pm$~2.71 & 41.30~$\pm$~3.13 & 73.92~$\pm$~8.32 \\
    & Luminance &  & 32.54~$\pm$~0.68 & 42.37~$\pm$~0.50 & 80.38~$\pm$~2.28 \\
    \bottomrule
  \end{tabular}
  \label{tab:going_beyond_nighttime_acdc_rain}%
\end{table}%

\begin{table}[tb]
  \centering
  \caption{Evaluation of DeepLabv3+, SegFormer, and UPerNet architectures trained on Cityscapes and \textbf{validated on ACDC Snow}. Results include two luminance variants (luminance and luminance green bias) each with depth=0. Metrics reported are mean Intersection over Union (mIoU), mean Accuracy (mAcc), and average Accuracy (aAcc), averaged over three seeds.}
  \begin{tabular}{@{}lccrrr@{}}
    \toprule
    Architecture & Preprocessing & Channel &\multicolumn{1}{c}{mIoU}&\multicolumn{1}{c}{mAcc}&\multicolumn{1}{c}{aAcc}  \\
    \midrule
    \multirow{6}{*}{DeepLabv3+}& - & 3 & 24.18~$\pm$~3.60 & 36.05~$\pm$~4.01 & 64.03~$\pm$~10.17 \\
    \cmidrule(lr){2-6}
    & Luminance green bias &  \multirow{2}{*}{1} & 36.48~$\pm$~0.17 & 48.96~$\pm$~2.15 & 82.91~$\pm$~0.89 \\
    & Luminance & & 37.75~$\pm$~1.66 & 48.66~$\pm$~1.10 & 83.59~$\pm$~2.36 \\
    \cmidrule(lr){2-6}
    & Luminance green bias & \multirow{2}{*}{3} & 33.79~$\pm$~2.96 & 46.19~$\pm$~2.87 & 76.96~$\pm$~4.85 \\
    & Luminance &  & 36.34~$\pm$~1.03 & 48.43~$\pm$~1.64 & 81.65~$\pm$~1.95 \\
    \midrule
    \multirow{6}{*}{SegFormer}& - & 3 & 46.92~$\pm$~0.51 & 57.57~$\pm$~0.80 & 86.05~$\pm$~0.19 \\
    \cmidrule(lr){2-6}
    & Luminance green bias & \multirow{2}{*}{1} & 43.69~$\pm$~1.70 & 52.98~$\pm$~2.71 & 83.47~$\pm$~0.77 \\
    & Luminance &  & 43.52~$\pm$~0.08 & 51.70~$\pm$~1.53 & 83.37~$\pm$~2.21 \\
    \cmidrule(lr){2-6}
     & Luminance green bias & \multirow{2}{*}{3} & 43.35~$\pm$~1.38 & 52.09~$\pm$~1.97 & 83.17~$\pm$~1.50 \\
    & Luminance &  & 43.76~$\pm$~2.56 & 52.78~$\pm$~2.40 & 83.52~$\pm$~3.08 \\
    \midrule
    \multirow{6}{*}{UPerNet}& - & 3 & 26.90~$\pm$~1.92 & 40.71~$\pm$~2.42 & 68.91~$\pm$~3.87 \\
    \cmidrule(lr){2-6}
    & Luminance green bias & \multirow{2}{*}{1} & 33.47~$\pm$~0.81 & 44.41~$\pm$~1.70 & 77.91~$\pm$~2.64 \\
    & Luminance &  & 32.40~$\pm$~3.65 & 42.47~$\pm$~3.18 & 76.29~$\pm$~2.47 \\
    \cmidrule(lr){2-6}
    & Luminance green bias & \multirow{2}{*}{3} & 29.07~$\pm$~0.99 & 41.22~$\pm$~2.06 & 67.63~$\pm$~5.03 \\
    & Luminance &  & 30.19~$\pm$~1.33 & 40.96~$\pm$~1.53 & 76.14~$\pm$~1.86 \\
    \bottomrule
  \end{tabular}
  \label{tab:going_beyond_nighttime_acdc_snow}%
\end{table}%
\clearpage

%% file: tex_for_figures/allweathernet_results.tex
\begin{figure*}[tb]
  \centering
  \includegraphics[width=\linewidth]{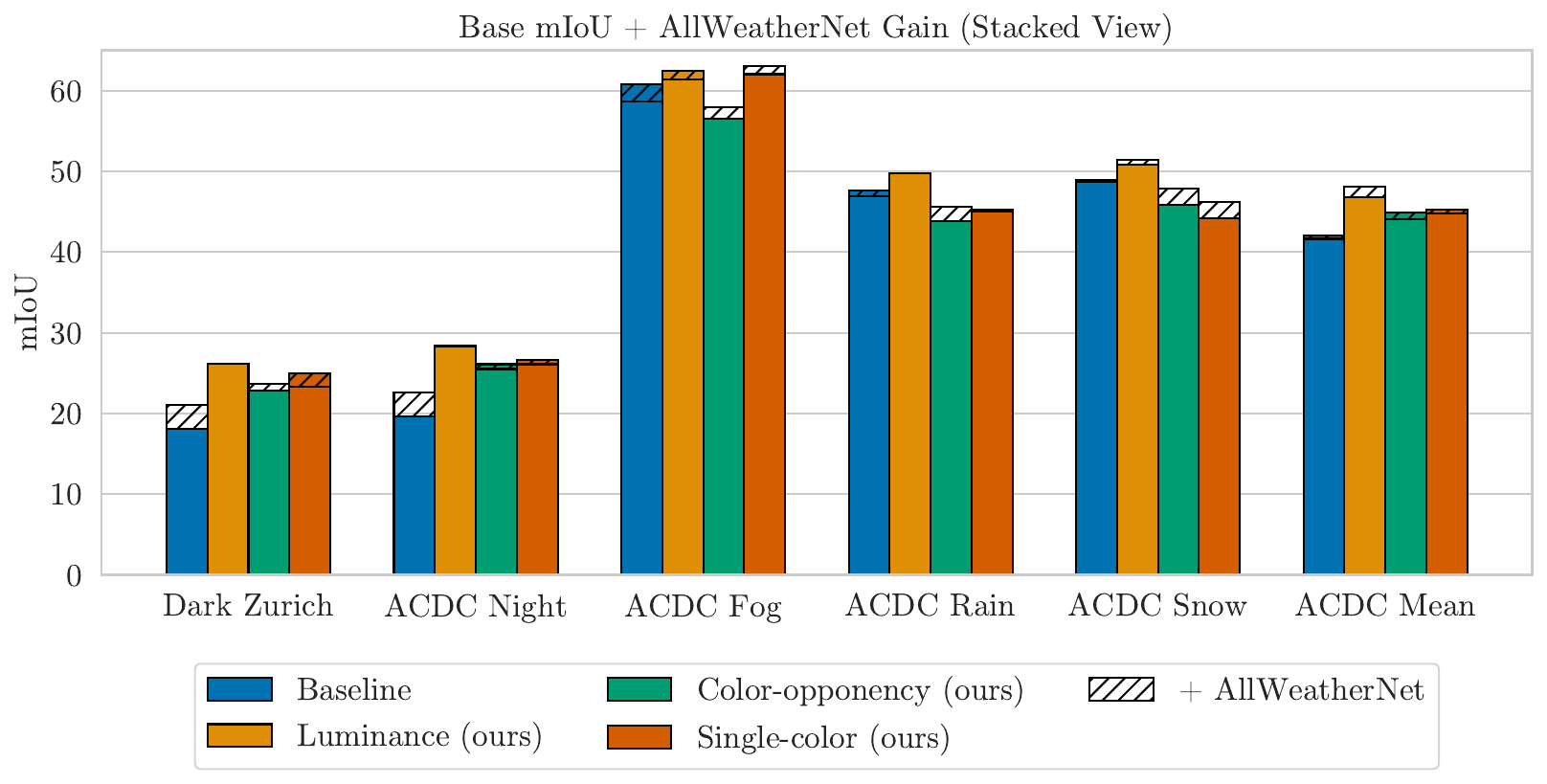}
  \caption{Evaluation of the \textbf{Mask2Former} architecture trained on Cityscapes and \textbf{validated on Dark Zurich, as well as on each ACDC split and the full ACDC dataset}, including the deviation to results validated on the enhanced datasets using AllWeatherNet.}
  \label{fig:allweathernet_results}
  \vspace{-0.5em}
\end{figure*}

%% file: tex_for_figures/appendix_ablation_depth_parameter_search.tex
\begin{figure}[tb]
  \centering
  \includegraphics[width=\linewidth]{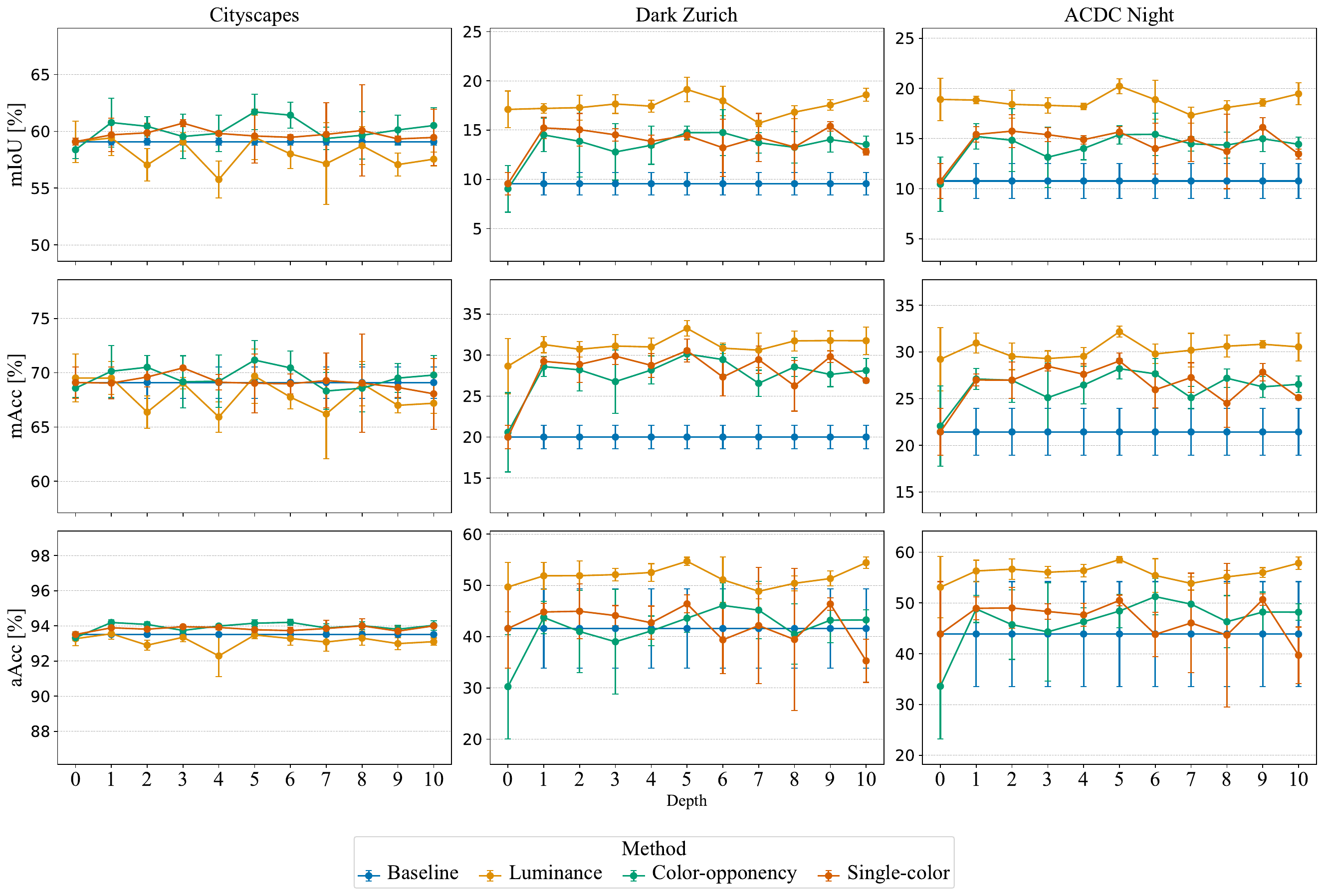}
  \caption{mIoU, mAcc and aAcc results for DeepLabv3+ (ResNetV1c-18) trained on Cityscapes, Dark Zurich, and ACDC Night datasets. Results are reported for each depth value from 0 to 10.}
  \label{fig:Validation_results_depth_parameter_search_appendix}
\end{figure}

%% file: tables/depth_all_results.tex
\begin{table*}[tb]
  \centering
  \caption{Evaluation of the DeepLabv3+ (ResNetV1c-18) architecture trained on Cityscapes and \textbf{validated on Cityscapes}. Results include two luminance variants (luminance and luminance green bias) each with depth=0 and three preprocessing types (luminance, color-opponency, and single-color) at depths 0 to 10. Metrics reported are mean Intersection over Union (mIoU), mean Accuracy (mAcc), and average Accuracy (aAcc), averaged over three seeds.}
  \begin{tabular}{@{}lccrrr@{}}
    \toprule
     Preprocessing & Dept & Channel &\multicolumn{1}{c}{mIoU}&\multicolumn{1}{c}{mAcc}&\multicolumn{1}{c}{aAcc}  \\
    \midrule
    - & 0 & 3 & 59.09~$\pm$~0.30 & 69.10~$\pm$~1.45 & 93.51~$\pm$~0.08 \\
    \midrule
    Luminance green bias &0&\multirow{2}{*}{1}  & 56.16~$\pm$~1.60 & 65.23~$\pm$~1.94 & 93.19~$\pm$~0.08 \\
    Luminance &0&  & 57.61~$\pm$~0.57 & 67.46~$\pm$~0.50 & 93.16~$\pm$~0.17 \\
    \midrule
    Luminance green bias &0& \multirow{12}{*}{3} & 57.51~$\pm$~0.40 & 67.82~$\pm$~0.47 & 93.09~$\pm$~0.10 \\
    \multirow{11}{*}{Luminance} &0&  & 59.08~$\pm$~1.82 & 69.52~$\pm$~2.22 & 93.27~$\pm$~0.39 \\
   &1& & 59.42~$\pm$~1.56 & 69.51~$\pm$~1.52 & 93.55~$\pm$~0.30 \\
   &2& & 57.04~$\pm$~1.43 & 66.37~$\pm$~1.48 & 92.90~$\pm$~0.29 \\
   &3& & 59.06~$\pm$~0.81 & 69.04~$\pm$~0.54 & 93.35~$\pm$~0.25 \\
   &4& & 55.76~$\pm$~1.61 & 65.92~$\pm$~1.41 & 92.29~$\pm$~1.17 \\
   &5& & 59.47~$\pm$~1.96 & 69.69~$\pm$~2.49 & 93.48~$\pm$~0.22 \\
   &6& & 58.01~$\pm$~1.30 & 67.75~$\pm$~1.08 & 93.28~$\pm$~0.37 \\
   &7& & 57.15~$\pm$~3.60 & 66.20~$\pm$~4.12 & 93.08~$\pm$~0.53 \\
   &8& & 58.73~$\pm$~1.67 & 69.00  ~$\pm$~2.04 & 93.30~$\pm$~0.39 \\
   &9& & 57.07~$\pm$~1.01 & 67.00  ~$\pm$~0.71 & 92.99~$\pm$~0.34 \\
   &10& & 57.55~$\pm$~0.61 & 67.19~$\pm$~0.94 & 93.10~$\pm$~0.21 \\
    \midrule
    \multirow{11}{*}{Color-opponency} &0&\multirow{11}{*}{3} & 58.38~$\pm$~0.76 & 68.58~$\pm$~0.83 & 93.33~$\pm$~0.16 \\
    &1& & 60.77~$\pm$~2.16 & 70.14~$\pm$~2.39 & 94.19~$\pm$~0.16 \\
    &2& & 60.44~$\pm$~0.86 & 70.51~$\pm$~1.10 & 94.08~$\pm$~0.16 \\
    &3& & 59.54~$\pm$~1.95 & 69.17~$\pm$~2.40 & 93.74~$\pm$~0.32 \\
    &4& & 59.82~$\pm$~1.60 & 69.22~$\pm$~2.42 & 93.99~$\pm$~0.11 \\
    &5& & 61.71~$\pm$~1.55 & 71.17~$\pm$~1.79 & 94.15~$\pm$~0.20 \\
    &6& & 61.42~$\pm$~1.13 & 70.45~$\pm$~1.57 & 94.20~$\pm$~0.16 \\
    &7& & 59.37~$\pm$~0.99 & 68.32~$\pm$~1.71 & 93.88~$\pm$~0.23 \\
    &8& & 59.65~$\pm$~2.08 & 68.58~$\pm$~2.19 & 94.01~$\pm$~0.22 \\
    &9& & 60.12~$\pm$~1.30 & 69.51~$\pm$~1.33 & 93.81~$\pm$~0.23 \\
    &10& & 60.51~$\pm$~1.58 & 69.79~$\pm$~1.80 & 94.03~$\pm$~0.27 \\
    \midrule
    \multirow{11}{*}{Single-color}& 0 & \multirow{11}{*}{3} & 59.09~$\pm$~0.30 & 69.10~$\pm$~1.45 & 93.51~$\pm$~0.08 \\
    &1& & 59.69~$\pm$~1.46 & 69.05~$\pm$~1.47 & 93.89~$\pm$~0.28 \\
    &2& & 59.87~$\pm$~0.83 & 69.61~$\pm$~0.90 & 93.81~$\pm$~0.01 \\
    &3& & 60.72~$\pm$~0.32 & 70.45~$\pm$~0.18 & 93.95~$\pm$~0.11 \\
    &4& & 59.81~$\pm$~0.19 & 69.13~$\pm$~0.73 & 93.91~$\pm$~0.13 \\
    &5& & 59.59~$\pm$~2.37 & 69.03~$\pm$~2.71 & 93.78~$\pm$~0.32 \\
    &6& & 59.47~$\pm$~0.24 & 68.99~$\pm$~0.92 & 93.73~$\pm$~0.19 \\
    &7& & 59.73~$\pm$~2.78 & 69.28~$\pm$~2.53 & 93.85~$\pm$~0.48 \\
    &8& & 60.07~$\pm$~4.01 & 69.05~$\pm$~4.54 & 94.00~$\pm$~0.39 \\
    &9& & 59.33~$\pm$~0.26 & 68.65~$\pm$~0.91 & 93.70~$\pm$~0.27 \\
    &10& & 59.46~$\pm$~2.47 & 68.06~$\pm$~3.26 & 94.00~$\pm$~0.09 \\
    \bottomrule
    \end{tabular}
  \label{tab:depth_search_validation_on_cityscapes}%
\end{table*}%

\begin{table*}[tb]
  \centering
  \caption{Evaluation of the DeepLabv3+ (ResNetV1c-18) architecture trained on Cityscapes and \textbf{validated on Dark Zurich}. Results include two luminance variants (luminance and luminance green bias) each with depth=0 and three preprocessing types (luminance, color-opponency, and single-color) at depths 0 to 10. Metrics reported are mean Intersection over Union (mIoU), mean Accuracy (mAcc), and average Accuracy (aAcc), averaged over three seeds.}
  \begin{tabular}{@{}lccccrrr@{}}
    \toprule
    Preprocessing & Dept & Channel  &\multicolumn{1}{c}{mIoU}&\multicolumn{1}{c}{mAcc}&\multicolumn{1}{c}{aAcc}  \\
    \midrule
    - &0& 3 & 9.56~$\pm$~1.13 & 20.00~$\pm$~1.43 & 41.60~$\pm$~7.74 \\
    \midrule
    Luminance green bias &0&\multirow{2}{*}{1}  & 19.14~$\pm$~2.64 & 32.88~$\pm$~2.45 & 54.83~$\pm$~2.73 \\
    Luminance &0&  & 17.64~$\pm$~0.63 & 31.89~$\pm$~0.45 & 53.17~$\pm$~0.84 \\
    \midrule
    Luminance green bias &0& \multirow{12}{*}{3} & 16.82~$\pm$~0.63 & 30.74~$\pm$~0.76 & 54.19~$\pm$~2.45 \\
   \multirow{11}{*}{Luminance} &0&  & 17.10~$\pm$~1.87 & 28.66~$\pm$~3.39 & 49.69~$\pm$~4.81 \\
   &1&  & 17.20~$\pm$~0.49 & 31.29~$\pm$~1.00 & 51.87~$\pm$~2.61 \\
   &2& & 17.28~$\pm$~1.26 & 30.72~$\pm$~0.95 & 51.90~$\pm$~2.88 \\
   &3& & 17.65~$\pm$~0.96 & 31.11~$\pm$~1.38 & 52.10~$\pm$~1.20 \\
   &4& & 17.43~$\pm$~0.60 & 31.00~$\pm$~1.06 & 52.52~$\pm$~1.72 \\
   &5& & 19.12~$\pm$~1.23 & 33.27~$\pm$~0.95 & 54.71~$\pm$~0.79 \\
   &6 & & 17.97~$\pm$~1.50 & 30.85~$\pm$~0.65 & 51.10~$\pm$~4.42 \\
   &7& & 15.70~$\pm$~0.33 & 30.61~$\pm$~2.10 & 48.85~$\pm$~1.45 \\
   &8& & 16.82~$\pm$~0.65 & 31.74~$\pm$~1.21 & 50.41~$\pm$~1.43 \\
   &9& & 17.55~$\pm$~0.53 & 31.78~$\pm$~1.19 & 51.34~$\pm$~1.47 \\
   &10 & & 18.58~$\pm$~0.67 & 31.76~$\pm$~1.69 & 54.43~$\pm$~1.16 \\
    \midrule
    \multirow{11}{*}{Color-opponency} &0&\multirow{11}{*}{3}  & 9.04~$\pm$~2.37 & 20.58~$\pm$~4.84 & 30.26~$\pm$~10.16 \\
    &1& & 14.51~$\pm$~1.71 & 28.60~$\pm$~1.21 & 43.74~$\pm$~3.12 \\
    &2& & 13.86~$\pm$~3.62 & 28.20~$\pm$~2.55 & 41.01~$\pm$~8.00 \\
    &3& & 12.78~$\pm$~2.87 & 26.77~$\pm$~3.87 & 39.01~$\pm$~10.23 \\
    &4& & 13.46~$\pm$~1.93 & 28.18~$\pm$~1.70 & 41.12~$\pm$~2.91 \\
    &5& & 14.74~$\pm$~0.67 & 30.14~$\pm$~0.69 & 43.61~$\pm$~2.77 \\
    &6& & 14.75~$\pm$~2.33 & 29.46~$\pm$~1.85 & 46.09~$\pm$~4.39 \\
    &7& & 13.72~$\pm$~1.04 & 26.56~$\pm$~1.60 & 45.18~$\pm$~5.57 \\
    &8& & 13.25~$\pm$~1.59 & 28.56~$\pm$~1.16 & 40.50~$\pm$~5.87 \\
    &9& & 14.04~$\pm$~1.28 & 27.64~$\pm$~1.49 & 43.21~$\pm$~4.35 \\
    &10& & 13.54~$\pm$~0.84 & 28.12~$\pm$~1.48 & 43.27~$\pm$~1.95 \\
    \midrule
    \multirow{11}{*}{Single-color} &0& \multirow{11}{*}{3} & 9.56~$\pm$~1.13 & 20.00~$\pm$~1.43 & 41.60~$\pm$~7.74 \\
    &1&  & 15.22~$\pm$~1.10 & 29.24~$\pm$~0.53 & 44.82~$\pm$~1.69 \\
    &2 & & 15.04~$\pm$~1.65 & 28.87~$\pm$~2.18 & 44.96~$\pm$~5.36 \\
    & 3 & & 14.52~$\pm$~0.62 & 29.88~$\pm$~1.00 & 44.12~$\pm$~1.95 \\
    &4& & 13.86~$\pm$~0.61 & 28.75~$\pm$~1.45 & 42.72~$\pm$~3.25 \\
    &5& & 14.46~$\pm$~0.46 & 30.55~$\pm$~1.40 & 46.40~$\pm$~1.72 \\
    &6& & 13.21~$\pm$~2.91 & 27.33~$\pm$~2.33 & 39.39~$\pm$~6.55 \\
    &7& & 14.26~$\pm$~2.43 & 29.45~$\pm$~1.66 & 42.17~$\pm$~11.35 \\
    &8& & 13.28~$\pm$~3.56 & 26.25~$\pm$~3.08 & 39.45~$\pm$~13.88 \\
    &9& & 15.39~$\pm$~0.48 & 29.81~$\pm$~0.69 & 46.37~$\pm$~1.24 \\
    &10& & 12.79~$\pm$~0.31 & 26.90~$\pm$~0.17 & 35.29~$\pm$~4.19 \\
    \bottomrule
    \end{tabular}
  \label{tab:depth_search_validation_on_dark_zurich}%
\end{table*}%

\begin{table*}[tb]
  \centering
  \caption{Evaluation of the DeepLabv3+ (ResNetV1c-18) architecture trained on Cityscapes and \textbf{validated on ACDC Night}. Results include two luminance variants (luminance and luminance green bias) each with depth=0 and three preprocessing types (luminance, color-opponency, and single-color) at depths 0 to 10. Metrics reported are mean Intersection over Union (mIoU), mean Accuracy (mAcc), and average Accuracy (aAcc), averaged over three seeds.}
  \begin{tabular}{@{}lccrrr@{}}
    \toprule
    Preprocessing & Dept & Channel &\multicolumn{1}{c}{mIoU}&\multicolumn{1}{c}{mAcc}&\multicolumn{1}{c}{aAcc}  \\
    \midrule
    - &0&3 & 10.77~$\pm$~1.72 & 21.44~$\pm$~2.50 & 43.89~$\pm$~10.31 \\
    \midrule
    Luminance green bias &0 &\multirow{2}{*}{1}  & 20.06~$\pm$~1.21 & 32.31~$\pm$~1.44 & 59.03~$\pm$~1.79 \\
    Luminance &0 & & 19.20~$\pm$~0.76 & 31.66~$\pm$~0.30 & 57.18~$\pm$~0.75 \\
    \midrule
    Luminance green bias &0 &\multirow{12}{*}{3} & 19.06~$\pm$~0.68 & 31.32~$\pm$~0.91 & 58.86~$\pm$~1.62 \\
   \multirow{11}{*}{Luminance}&0& & 18.90~$\pm$~2.11 & 29.22~$\pm$~3.41 & 53.10~$\pm$~6.04 \\
   &1& & 18.84~$\pm$~0.36 & 30.96~$\pm$~1.06 & 56.30~$\pm$~2.12 \\
   &2& & 18.41~$\pm$~1.39 & 29.53~$\pm$~1.43 & 56.66~$\pm$~1.98 \\
   &3& & 18.31~$\pm$~0.78 & 29.30~$\pm$~0.83 & 56.05~$\pm$~1.12 \\
   &4& & 18.20~$\pm$~0.31 & 29.54~$\pm$~0.94 & 56.36~$\pm$~1.25 \\
   &5& & 20.22~$\pm$~0.74 & 32.17~$\pm$~0.58 & 58.54~$\pm$~0.61 \\
   &6& & 18.88~$\pm$~1.95 & 29.79~$\pm$~1.04 & 55.41~$\pm$~3.31 \\
   &7& & 17.33~$\pm$~0.80 & 30.19~$\pm$~1.81 & 53.85~$\pm$~1.32 \\
   &8& & 18.10~$\pm$~0.67 & 30.62~$\pm$~1.17 & 55.13~$\pm$~1.29 \\
   &9& & 18.59~$\pm$~0.38 & 30.83~$\pm$~0.38 & 55.98~$\pm$~0.97 \\
   &10& & 19.46~$\pm$~1.10 & 30.55~$\pm$~1.49 & 57.84~$\pm$~1.21 \\
    \midrule
    \multirow{11}{*}{Color-opponency} &0&\multirow{11}{*}{3} & 10.45~$\pm$~2.70 & 22.07~$\pm$~4.30 & 33.60~$\pm$~10.38 \\
    &1& & 15.21~$\pm$~1.28 & 27.11~$\pm$~1.12 & 48.82~$\pm$~2.67 \\
    &2& & 14.83~$\pm$~3.13 & 26.97~$\pm$~2.36 & 45.72~$\pm$~6.83 \\
    &3& & 13.14~$\pm$~3.00 & 25.11~$\pm$~3.70 & 44.31~$\pm$~9.65 \\
    &4& & 14.01~$\pm$~1.13 & 26.46~$\pm$~2.00 & 46.30~$\pm$~2.72 \\
    &5& & 15.39~$\pm$~0.92 & 28.21~$\pm$~1.08 & 48.41~$\pm$~3.25 \\
    &6& & 15.42~$\pm$~2.11 & 27.66~$\pm$~1.64 & 51.25~$\pm$~3.56 \\
    &7& & 14.47~$\pm$~0.70 & 25.11~$\pm$~1.22 & 49.77~$\pm$~4.07 \\
    &8& & 14.35~$\pm$~1.29 & 27.19~$\pm$~0.99 & 46.31~$\pm$~5.14 \\
    &9& & 14.98~$\pm$~1.26 & 26.27~$\pm$~1.13 & 48.21~$\pm$~3.93 \\
    &10& & 14.43~$\pm$~0.71 & 26.54~$\pm$~0.88 & 48.19~$\pm$~1.63 \\
    \midrule
    \multirow{11}{*}{Single-color} &0&\multirow{11}{*}{3} & 10.77~$\pm$~1.72 & 21.44~$\pm$~2.50 & 43.89~$\pm$~10.31 \\
    &1& & 15.41~$\pm$~0.76 & 27.00~$\pm$~0.65 & 48.94~$\pm$~2.23 \\
    &2& & 15.74~$\pm$~1.62 & 26.99~$\pm$~1.96 & 49.02~$\pm$~4.01 \\
    &3& & 15.39~$\pm$~0.69 & 28.47~$\pm$~0.52 & 48.32~$\pm$~1.51 \\
    &4& & 14.89~$\pm$~0.42 & 27.61~$\pm$~1.15 & 47.66~$\pm$~2.24 \\
    &5& & 15.66~$\pm$~0.52 & 29.04~$\pm$~ 0.85 & 50.50~$\pm$~1.05 \\
    &6& & 14.00~$\pm$~ 2.53 & 25.96~$\pm$~1.95 & 43.81~$\pm$~4.35 \\
    &7& & 14.95~$\pm$~2.25 & 27.26~$\pm$~ 1.59 & 46.06~$\pm$~9.81 \\
    &8& & 13.72~$\pm$~3.72 & 24.51~$\pm$~ 2.59 & 43.67~$\pm$~14.13 \\
    &9& & 16.11~$\pm$~0.99 & 27.84~$\pm$~ 0.94 & 50.67~$\pm$~1.09 \\
    &10& & 13.46~$\pm$~0.50 & 25.12~$\pm$~0.25 & 39.71~$\pm$~5.56 \\
    \bottomrule
    \end{tabular}
  \label{tab:depth_search_validation_on_acdc_night}%
\end{table*}%

\begin{table*}[tb]
  \centering
    \caption{Evaluation of the DeepLabv3+ (ResNetV1c-18) architecture trained on Cityscapes and \textbf{validated on ACDC Fog}. Results include two luminance variants (luminance and luminance green bias) each with depth=0 and three preprocessing types (luminance, color-opponency, and single-color) at depths 0 to 10. Metrics reported are mean Intersection over Union (mIoU), mean Accuracy (mAcc), and average Accuracy (aAcc), averaged over three seeds.}
  \begin{tabular}{@{}lccccrrr@{}}
    \toprule
    Preprocessing & Dept & Channel &\multicolumn{1}{c}{mIoU}&\multicolumn{1}{c}{mAcc}&\multicolumn{1}{c}{aAcc}  \\
    \midrule
    - & 0 & 3& 36.09~$\pm$~1.90 & 52.31~$\pm$~0.63 & 85.22~$\pm$~1.18 \\
    \midrule
    Luminance green bias & 0 & \multirow{2}{*}{1} & 41.96~$\pm$~0.41 & 55.29~$\pm$~1.20 & 87.71~$\pm$~0.72 \\
    Luminance & 0 &  & 43.54~$\pm$~1.06 & 57.60~$\pm$~0.62 & 86.98~$\pm$~0.95 \\
    \midrule
    Luminance green bias & 0 &\multirow{12}{*}{3}& 44.04~$\pm$~0.96 & 57.29~$\pm$~0.77 & 88.61~$\pm$~0.41 \\
   \multirow{11}{*}{Luminance} & 0 & & 47.16~$\pm$~1.49 & 58.81~$\pm$~1.76 & 89.10~$\pm$~0.63 \\
    & 1 &  & 43.20~$\pm$~1.69 & 55.57~$\pm$~2.13 & 83.89~$\pm$~2.61 \\
   & 2 & & 43.04~$\pm$~2.15 & 53.12~$\pm$~1.72 & 86.68~$\pm$~2.39 \\
   & 3 &  & 43.78~$\pm$~2.20 & 54.87~$\pm$~1.85 & 85.95~$\pm$~3.52 \\
   & 4 & & 41.52~$\pm$~0.87 & 53.32~$\pm$~1.60 & 85.94~$\pm$~0.74 \\
   & 5 & & 44.88~$\pm$~0.25 & 55.51~$\pm$~0.88 & 85.51~$\pm$~3.25 \\
   & 6 & & 42.43~$\pm$~2.08 & 54.32~$\pm$~1.04 & 84.28~$\pm$~1.86 \\
   & 7 & & 41.20~$\pm$~3.38 & 52.92~$\pm$~3.02 & 85.75~$\pm$~2.68 \\
   & 8 & & 42.45~$\pm$~1.98 & 54.99~$\pm$~2.39 & 85.58~$\pm$~2.35 \\
   & 9 & & 41.33~$\pm$~1.83 & 54.39~$\pm$~2.29 & 85.14~$\pm$~3.26 \\
   & 10 & & 42.59~$\pm$~0.49 & 54.00~$\pm$~1.84 & 84.05~$\pm$~3.59 \\
    \midrule
    \multirow{11}{*}{Color-opponency} & 0 & \multirow{11}{*}{3} & 34.43~$\pm$~2.10 & 50.56~$\pm$~4.23 & 76.67~$\pm$~3.61 \\
    & 1 & & 40.80~$\pm$~2.48 & 54.30~$\pm$~2.50 & 82.07~$\pm$~1.28 \\
    & 2 & & 41.71~$\pm$~1.27 & 56.30~$\pm$~1.01 & 84.18~$\pm$~0.45 \\
    & 3 &  & 36.89~$\pm$~3.73 & 52.37~$\pm$~1.74 & 80.73~$\pm$~6.77 \\
    & 4 & & 39.01~$\pm$~0.74 & 54.24~$\pm$~2.31 & 85.57~$\pm$~0.70 \\
    & 5 & & 41.63~$\pm$~2.64 & 56.22~$\pm$~1.86 & 82.30~$\pm$~5.48 \\
    & 6 & & 42.71~$\pm$~2.44 & 55.97~$\pm$~2.68 & 85.15~$\pm$~1.95 \\
    & 7 & & 40.47~$\pm$~0.78 & 53.47~$\pm$~1.99 & 84.05~$\pm$~3.48 \\
    & 8 & & 42.93~$\pm$~1.84 & 55.64~$\pm$~2.42 & 86.27~$\pm$~1.55 \\
    & 9 & & 42.20~$\pm$~1.07 & 54.26~$\pm$~1.37 & 85.88~$\pm$~0.92 \\
    & 10 & & 39.94~$\pm$~2.62 & 53.81~$\pm$~1.98 & 79.93~$\pm$~4.77 \\
    \midrule
    \multirow{11}{*}{Single-color} & 0 & \multirow{11}{*}{3}& 36.09~$\pm$~1.90 & 52.31~$\pm$~0.63 & 85.22~$\pm$~1.18 \\
    & 1 &  & 38.54~$\pm$~2.56 & 52.75~$\pm$~2.51 & 82.22~$\pm$~3.08 \\
    & 2 & & 40.94~$\pm$~1.34 & 54.53~$\pm$~1.79 & 85.11~$\pm$~1.70 \\
    & 3 &  & 42.14~$\pm$~1.16 & 56.17~$\pm$~0.48 & 84.39~$\pm$~3.38 \\
    & 4 & & 36.79~$\pm$~4.44 & 53.48~$\pm$~2.72 & 75.15~$\pm$~12.15 \\
    & 5 & & 38.75~$\pm$~2.51 & 53.52~$\pm$~1.32 & 84.34~$\pm$~0.19 \\
    & 6 & & 39.40~$\pm$~2.13 & 53.21~$\pm$~2.12 & 84.24~$\pm$~2.91 \\
    & 7 & & 39.15~$\pm$~1.52 & 53.02~$\pm$~1.82 & 83.32~$\pm$~4.15 \\
    & 8 & & 39.48~$\pm$~1.80 & 54.44~$\pm$~2.51 & 83.51~$\pm$~5.62 \\
    & 9 & & 39.67~$\pm$~1.81 & 52.67~$\pm$~1.11 & 79.16~$\pm$~7.70 \\
    & 10 & & 38.22~$\pm$~1.63 & 52.79~$\pm$~1.09 & 84.42~$\pm$~0.79 \\
    \bottomrule
    \end{tabular}
  \label{tab:depth_search_validation_on_acdc_fog}%
\end{table*}%

\begin{table*}[tb]
  \centering
  \caption{Evaluation of the DeepLabv3+ (ResNetV1c-18) architecture trained on Cityscapes and \textbf{validated on ACDC Rain}. Results include two luminance variants (luminance and luminance green bias) each with depth=0 and three preprocessing types (luminance, color-opponency, and single-color) at depths 0 to 10. Metrics reported are mean Intersection over Union (mIoU), mean Accuracy (mAcc), and average Accuracy (aAcc), averaged over three seeds.}
  \begin{tabular}{@{}lccrrr@{}}
    \toprule
    Preprocessing & Dept & Channel &\multicolumn{1}{c}{mIoU}&\multicolumn{1}{c}{mAcc}&\multicolumn{1}{c}{aAcc}  \\
    \midrule
    - & 0 &3 & 32.22~$\pm$~0.29 & 46.63~$\pm$~0.91 & 82.86~$\pm$~0.98 \\
    \midrule
    Luminance green bias & 0 & \multirow{2}{*}{1} & 33.52~$\pm$~1.81 & 46.71~$\pm$~2.28 & 83.98~$\pm$~1.17 \\
    Luminance & 0 &  & 35.06~$\pm$~1.17 & 50.05~$\pm$~0.57 & 84.26~$\pm$~1.24 \\
    \midrule
    Luminance green bias & 0 &\multirow{12}{*}{3} & 34.55~$\pm$~2.19 & 47.50~$\pm$~2.62 & 84.25~$\pm$~1.55 \\
   \multirow{11}{*}{Luminance} & 0 & & 37.13~$\pm$~1.23 & 50.35~$\pm$~0.68 & 85.47~$\pm$~1.32 \\
    & 1 &  & 33.68~$\pm$~0.71 & 47.46~$\pm$~2.47 & 79.31~$\pm$~2.43 \\
   & 2 & & 32.74~$\pm$~1.77 & 43.15~$\pm$~2.03 & 82.98~$\pm$~2.50 \\
   & 3 &  & 32.43~$\pm$~2.13 & 43.65~$\pm$~2.58 & 78.91~$\pm$~4.43 \\
   & 4 & & 32.88~$\pm$~0.80 & 45.12~$\pm$~1.97 & 80.59~$\pm$~2.58 \\
   & 5 & & 33.89~$\pm$~1.08 & 45.96~$\pm$~1.55 & 81.10~$\pm$~4.44 \\
   & 6 & & 33.14~$\pm$~1.67 & 45.54~$\pm$~1.42 & 79.44~$\pm$~3.39 \\
   & 7 & & 32.10~$\pm$~2.10 & 43.55~$\pm$~1.80 & 82.05~$\pm$~3.17 \\
   & 8 & & 32.70~$\pm$~1.02 & 45.16~$\pm$~1.84 & 81.14~$\pm$~3.68 \\
   & 9 & & 33.39~$\pm$~0.72 & 46.41~$\pm$~2.41 & 82.90~$\pm$~1.48 \\
   & 10 & & 33.68~$\pm$~1.81 & 46.06~$\pm$~2.69 & 79.71~$\pm$~4.93 \\
    \midrule
    \multirow{11}{*}{Color-opponency} & 0 & \multirow{11}{*}{3} & 28.18~$\pm$~2.47 & 42.07~$\pm$~7.19 & 76.14~$\pm$~2.89 \\
    & 1 & & 32.90~$\pm$~1.54 & 46.46~$\pm$~0.51 & 79.26~$\pm$~1.16 \\
    & 2 & & 34.62~$\pm$~1.09 & 48.45~$\pm$~0.93 & 81.64~$\pm$~1.62 \\
    & 3 &  & 31.64~$\pm$~2.92 & 45.95~$\pm$~1.61 & 80.14~$\pm$~4.86 \\
    & 4 & & 33.94~$\pm$~1.03 & 48.64~$\pm$~2.10 & 84.31~$\pm$~1.26 \\
    & 5 & & 34.44~$\pm$~1.27 & 48.87~$\pm$~2.13 & 79.89~$\pm$~4.85 \\
    & 6 & & 36.32~$\pm$~0.83 & 49.84~$\pm$~2.24 & 83.91~$\pm$~0.99 \\
    & 7 & & 34.01~$\pm$~0.21 & 45.53~$\pm$~1.99 & 81.36~$\pm$~4.55 \\
    & 8 & & 36.09~$\pm$~0.55 & 48.47~$\pm$~1.99 & 84.77~$\pm$~1.33 \\
    & 9 & & 35.55~$\pm$~0.65 & 47.42~$\pm$~1.62 & 84.21~$\pm$~0.81 \\
    & 10 & & 33.28~$\pm$~0.53 & 47.01~$\pm$~0.92 & 78.06~$\pm$~4.17 \\
    \midrule
    \multirow{11}{*}{Single-color} & 0 &\multirow{11}{*}{3} & 32.22~$\pm$~0.29 & 46.63~$\pm$~0.91 & 82.86~$\pm$~0.98 \\
    & 1 &  & 32.60~$\pm$~1.80 & 46.00~$\pm$~2.26 & 79.68~$\pm$~3.54 \\
    & 2 & & 33.80~$\pm$~0.58 & 46.25~$\pm$~1.02 & 82.00~$\pm$~1.62 \\
    & 3 &  & 34.50~$\pm$~1.30 & 48.92~$\pm$~0.63 & 80.43~$\pm$~4.44 \\
    & 4 & & 32.07~$\pm$~3.11 & 46.41~$\pm$~1.16 & 76.73~$\pm$~9.03 \\
    & 5 & & 33.73~$\pm$~1.07 & 46.91~$\pm$~1.56 & 83.23~$\pm$~1.10 \\
    & 6 & & 33.08~$\pm$~0.61 & 46.39~$\pm$~1.33 & 80.71~$\pm$~2.92 \\
    & 7 & & 32.80~$\pm$~1.76 & 46.38~$\pm$~2.19 & 80.02~$\pm$~5.51 \\
    & 8 & & 33.95~$\pm$~2.12 & 47.19~$\pm$~1.76 & 82.40~$\pm$~4.75 \\
    & 9 & & 34.22~$\pm$~1.54 & 46.64~$\pm$~0.36 & 78.91~$\pm$~6.71 \\
    & 10 & & 32.85~$\pm$~1.05 & 45.74~$\pm$~1.17 & 81.52~$\pm$~1.96 \\
    \bottomrule
  \end{tabular}
  \label{tab:depth_search_validation_on_acdc_rain}%
\end{table*}

\begin{table*}[tb]
  \centering
    \caption{Evaluation of the DeepLabv3+ (ResNetV1c-18) architecture trained on Cityscapes and \textbf{validated on ACDC Snow}. Results include two luminance variants (luminance and luminance green bias) each with depth=0 and three preprocessing types (luminance, color-opponency, and single-color) at depths 0 to 10. Metrics reported are mean Intersection over Union (mIoU), mean Accuracy (mAcc), and average Accuracy (aAcc), averaged over three seeds.}
  \begin{tabular}{@{}lccrrr@{}}
    \toprule
   
    Preprocessing & Dept & Channel  &\multicolumn{1}{c}{mIoU}&\multicolumn{1}{c}{mAcc}&\multicolumn{1}{c}{aAcc}  \\
    \midrule
    - &0&3 & 25.31~$\pm$~0.53 & 36.81~$\pm$~0.60 & 75.33~$\pm$~2.91 \\
    \midrule
    Luminance green bias &0&\multirow{2}{*}{1}  & 31.54~$\pm$~0.36 & 43.64~$\pm$~0.09 & 81.44~$\pm$~1.08 \\
    Luminance &0 & & 31.56~$\pm$~0.60 & 44.72~$\pm$~0.82 & 80.86~$\pm$~0.72 \\
    \midrule
    Luminance green bias &0& \multirow{12}{*}{3}& 32.55~$\pm$~0.11 & 45.14~$\pm$~0.47 & 82.10~$\pm$~1.31 \\
   \multirow{11}{*}{Luminance} &0 & & 34.69~$\pm$~0.60 & 46.32~$\pm$~0.35 & 82.79~$\pm$~0.40 \\ 
   &1&  & 30.58~$\pm$~0.82 & 43.28~$\pm$~0.62 & 72.61~$\pm$~4.31 \\
   &2& & 29.30~$\pm$~2.12 & 40.75~$\pm$~1.39 & 77.44~$\pm$~3.88 \\
   &3& & 29.23~$\pm$~3.15 & 40.77~$\pm$~3.03 & 72.82~$\pm$~6.43 \\
   &4& & 28.63~$\pm$~1.54 & 40.93~$\pm$~1.73 & 74.42~$\pm$~3.66 \\
   &5& & 31.90~$\pm$~1.43 & 43.43~$\pm$~0.29 & 76.17~$\pm$~6.20 \\
   &6& & 29.18~$\pm$~2.41 & 41.75~$\pm$~0.81 & 72.36~$\pm$~1.93 \\
   &7& & 29.44~$\pm$~3.08 & 41.69~$\pm$~2.60 & 76.50~$\pm$~5.30 \\
   &8 & & 29.43~$\pm$~0.90 & 42.34~$\pm$~1.92 & 75.03~$\pm$~4.62 \\
   &9& & 29.32~$\pm$~1.27 & 41.94~$\pm$~1.54 & 76.11~$\pm$~3.76 \\
   &10& & 29.77~$\pm$~1.88 & 41.68~$\pm$~2.55 & 73.92~$\pm$~5.43 \\
    \midrule
    \multirow{11}{*}{Color-opponency} &0&\multirow{10}{*}{3}  & 22.46~$\pm$~1.06 & 35.19~$\pm$~3.89 & 65.61~$\pm$~2.84 \\
    &1& & 26.62~$\pm$~0.66 & 39.28~$\pm$~1.73 & 70.58~$\pm$~1.09 \\
    &2& & 28.66~$\pm$~1.70 & 40.61~$\pm$~1.06 & 76.41~$\pm$~2.28 \\
    &3& & 24.86~$\pm$~2.34 & 38.31~$\pm$~1.58 & 71.12~$\pm$~7.20 \\
    &4 & & 27.61~$\pm$~0.31 & 40.70~$\pm$~1.26 & 77.29~$\pm$~1.42 \\
    &5& & 28.09~$\pm$~3.09& 40.58~$\pm$~1.50 & 72.42~$\pm$~9.02  \\
    &6& & 29.85~$\pm$~0.92 & 41.54~$\pm$~1.19 & 77.17~$\pm$~1.43 \\
    &7& & 28.40~$\pm$~0.66 & 40.09~$\pm$~0.38 & 74.51~$\pm$~3.50 \\
    &8 & & 28.64~$\pm$~1.38 & 40.95~$\pm$~1.74 & 77.20~$\pm$~2.76 \\
    &9 & & 29.18~$\pm$~1.76 & 39.63~$\pm$~2.26 & 77.72~$\pm$~1.02 \\
    &10& & 26.51~$\pm$~1.04 & 39.37~$\pm$~0.65 & 70.29~$\pm$~ 4.21 \\
    \midrule
    \multirow{11}{*}{Single-color} &0&\multirow{10}{*}{3} & 25.31~$\pm$~0.53 & 36.81~$\pm$~0.60 & 75.33~$\pm$~2.91 \\
    &1& & 26.58~$\pm$~2.20 & 39.67~$\pm$~1.61 & 70.46~$\pm$~4.66 \\
    &2& & 26.78~$\pm$~0.23 & 39.35~$\pm$~1.02 & 73.29~$\pm$~1.97 \\
    &3& & 28.21~$\pm$~1.17 & 41.04~$\pm$~0.46 & 74.18~$\pm$~4.77 \\
    &4& & 24.76~$\pm$~3.37 & 39.13~$\pm$~3.24 & 66.67~$\pm$~10.46 \\
    &5& & 26.96~$\pm$~0.86 & 39.44~$\pm$~0.89 & 75.04~$\pm$~2.27 \\
    &6& & 26.86~$\pm$~1.86 & 39.18~$\pm$~1.14 & 72.96~$\pm$~5.91 \\
    &7& & 26.70~$\pm$~0.81 & 38.91~$\pm$~1.09 & 74.27~$\pm$~4.66 \\
    &8 & & 28.11~$\pm$~1.86 & 40.46~$\pm$~1.87 & 75.74~$\pm$~6.17 \\
    &9 & & 26.98~$\pm$~2.74 & 38.54~$\pm$~1.63 & 70.97~$\pm$~8.25 \\
    &10& & 26.04~$\pm$~0.20 & 38.44~$\pm$~0.38 & 75.14~$\pm$~1.05 \\
    \bottomrule
    \end{tabular}
  \label{tab:depth_search_validation_on_acdc_snow}
\end{table*}%
\clearpage

%% file: tex_for_figures/cifar_results_vanilla.tex
\begin{figure*}[tb]
  \centering
  \includegraphics[width=\linewidth]{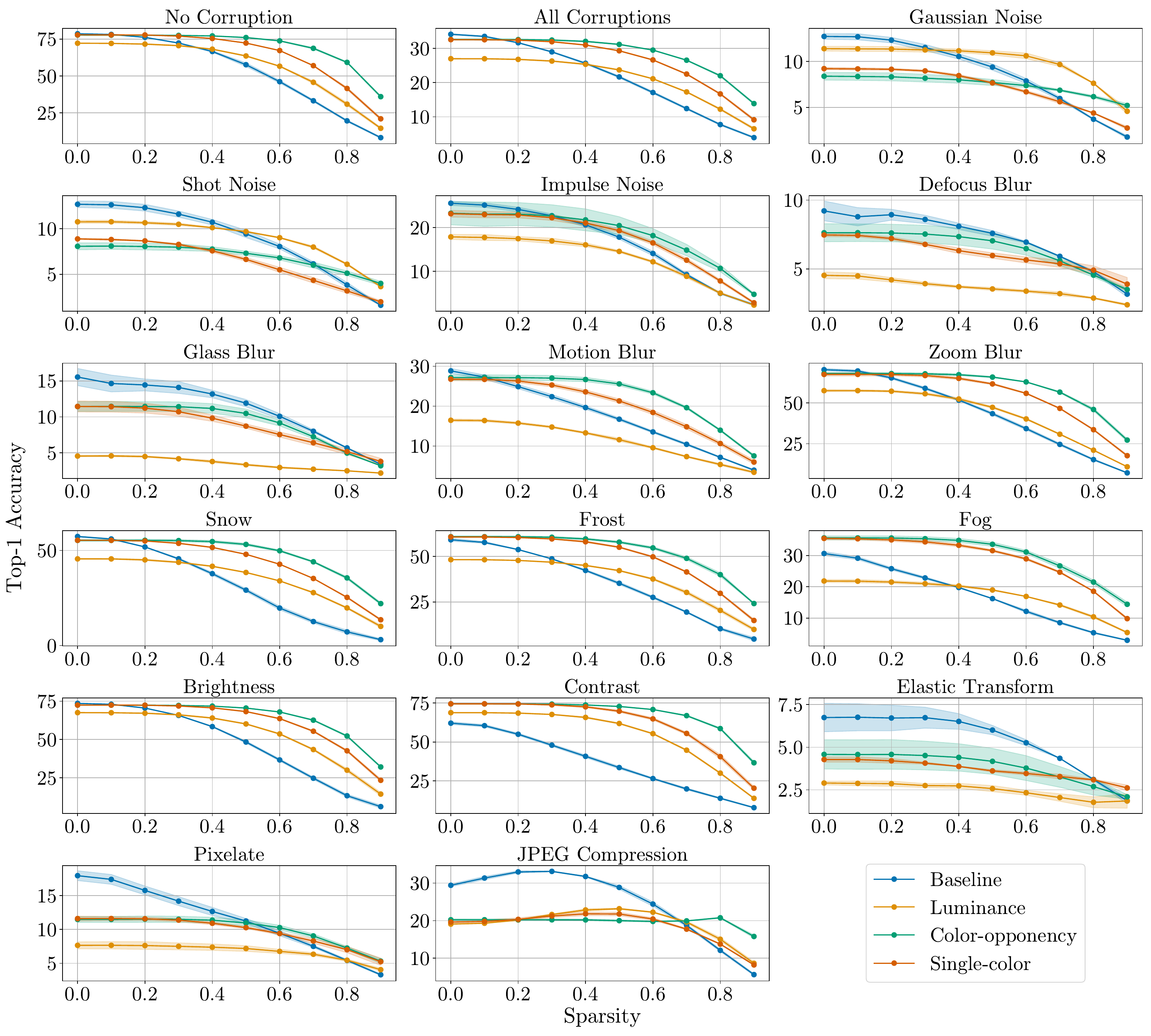}
  \caption{Evaluation of the ConvNeXt-Tiny architecture trained on CIFAR-100 and \textbf{validated on CIFAR-100-C}. Results include the three preprocessing types (luminance, color-opponency, and single-color). The reported metric is Top-1 Accuracy. All results are averaged over the severities from 1 to 5.}
  \label{fig:Validation results-cifar_vanilla}
  \vspace{-0.5em}
\end{figure*}

%% file: tex_for_figures/cifar_results_mixup.tex
\begin{figure*}[tb]
  \centering
  \includegraphics[width=\linewidth]{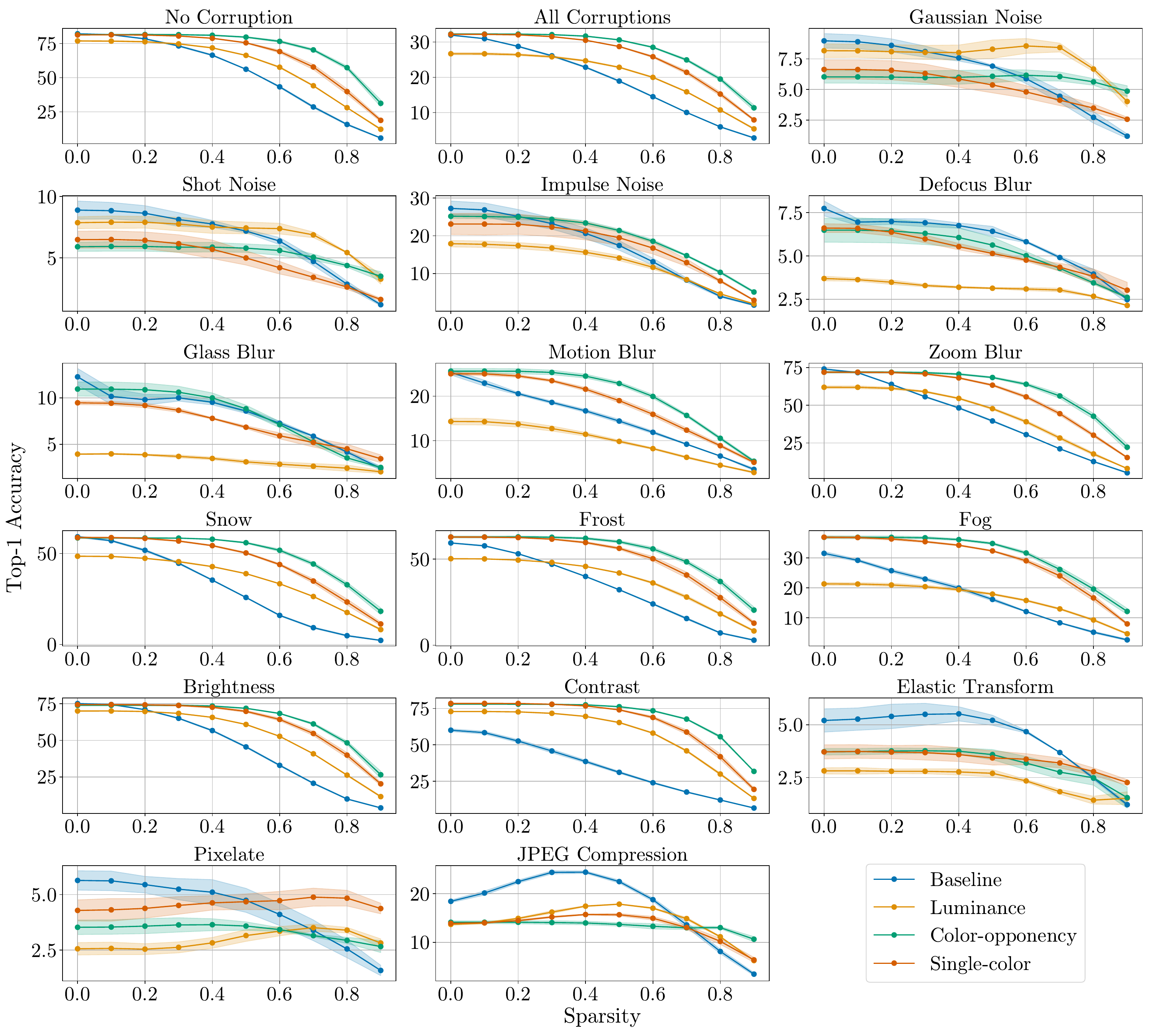}
  \caption{Evaluation of the ConvNeXt-Tiny architecture trained on CIFAR-100 with Cutmix and \textbf{validated on CIFAR-100-C}. Results include the three preprocessing types (luminance, color-opponency, and single-color). The reported metric is Top-1 Accuracy. All results are averaged over the severities from 1 to 5.}
  \label{fig:Validation results-cifar_mixup}
  \vspace{-0.5em}
\end{figure*}

%% file: tex_for_figures/extended_teaser_new.tex
\begin{figure}[h]
  \centering
  \includegraphics[width=\textwidth,height=0.62\textheight,keepaspectratio]{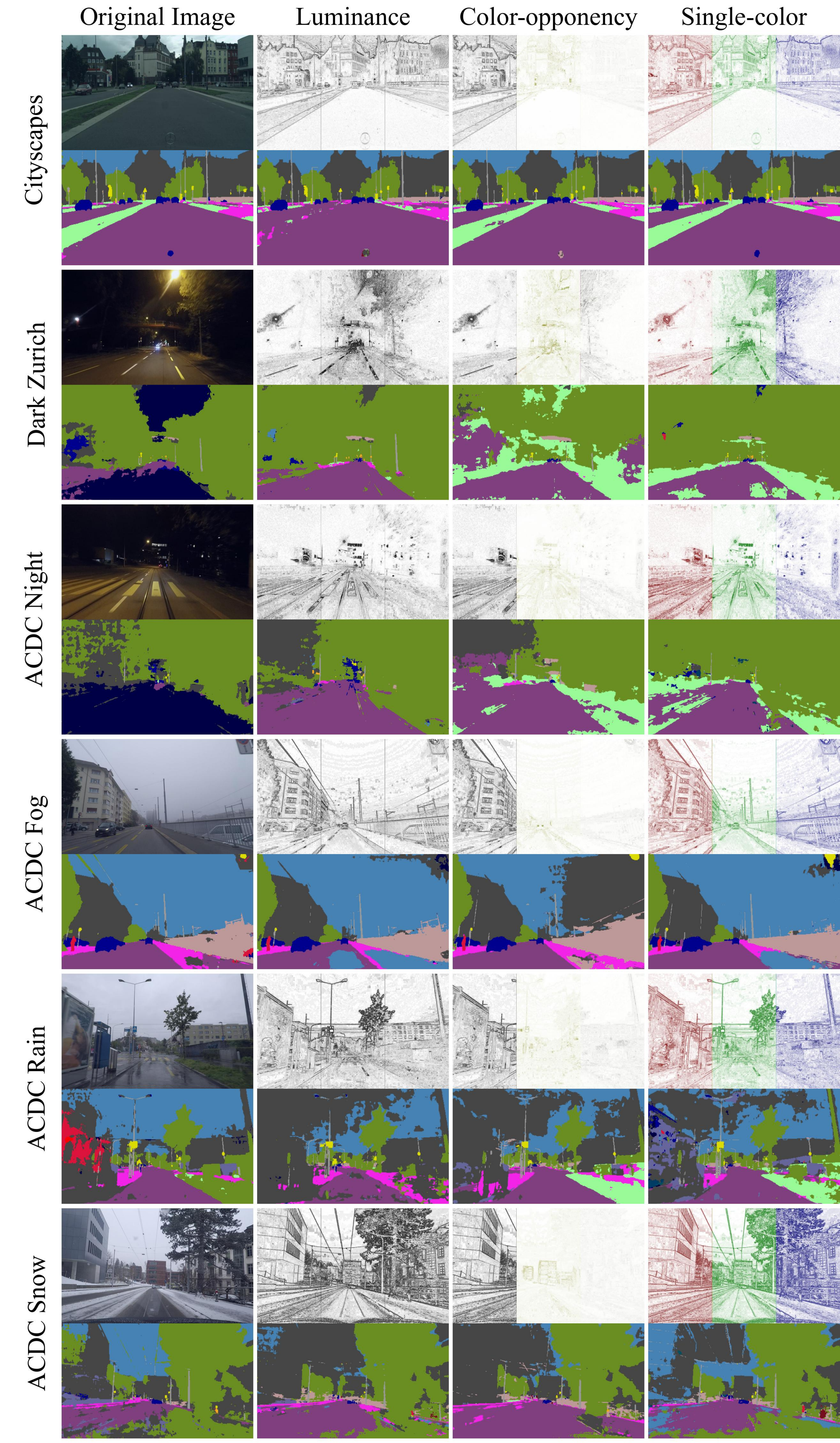}
  \caption{More examples of results (as in  \cref{fig:teaser_new}) from the in-distribution dataset (Cityscapes~\cite{Cityscapes_dataset}) and real-world robustness datasets (Dark Zurich~\cite{Dark_zurich_dataset} and ACDC~\cite{ACDC_dataset}). 
    Each pair of rows shows the original RGB input followed by the preprocessed variants explored in this work: the top rows show luminance, color-opponency, and single-color (three channels). The bottom rows visualize the semantic segmentation masks predicted using UPerNet trained with each preprocessing variant.}
    \label{fig:extended_teaser}
\end{figure}

%% file: tex_for_figures/color_opponency_depth0_new.tex
\begin{figure*}[tb]
  \centering
  \includegraphics[width=\textwidth,height=0.8\textheight,keepaspectratio]{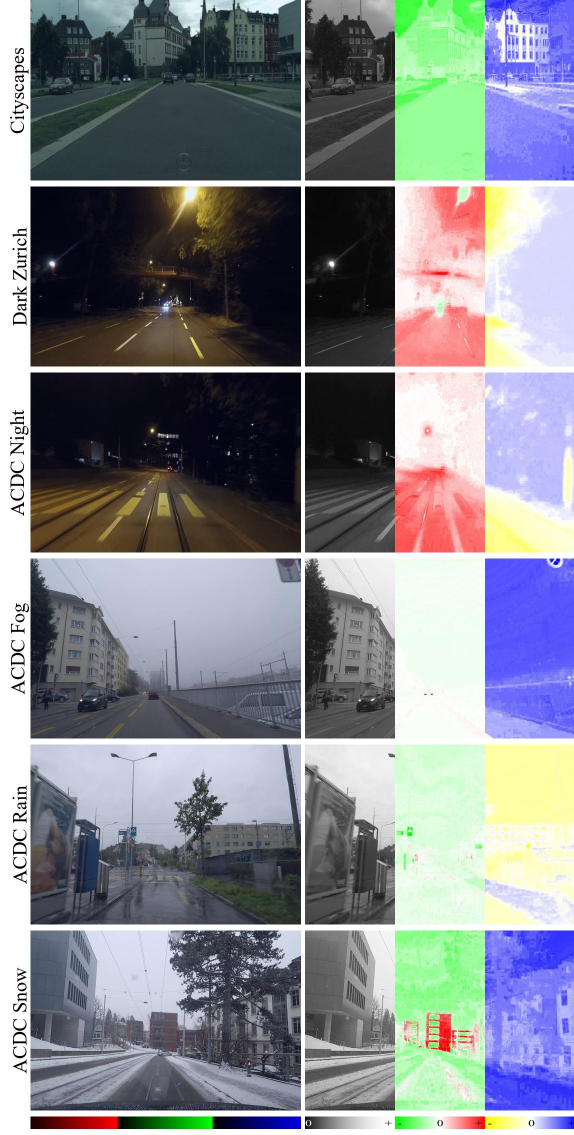}
  \caption{Comparison of original input images (left column) and their color-opponency reparameterized versions with depth=0 (right column) across Cityscapes~\cite{Cityscapes_dataset} and the real-world robustness datasets Dark Zurich~\cite{Dark_zurich_dataset} and ACDC~\cite{ACDC_dataset}.}
    \label{fig:color-opponency_visualization_appendix}
\end{figure*}
\clearpage